\documentclass{article}

\PassOptionsToPackage{numbers, compress}{natbib}
     \usepackage[final]{neurips_data_2022}

\usepackage[utf8]{inputenc} % allow utf-8 input
\usepackage[T1]{fontenc}    % use 8-bit T1 fonts
\usepackage{hyperref}       % hypers
\usepackage{url}            % simple URL typesetting
\usepackage{booktabs}       % professional-quality tables
\usepackage{amsfonts}       % blackboard math symbols
\usepackage{nicefrac}       % compact symbols for 1/2, etc.
\usepackage{microtype}      % microtypography
\usepackage[table]{xcolor}         % colors

\usepackage{pgf} %set color gradient depending on number value
\usepackage{xstring}
\usepackage{collcell} %color tabular cell
\usepackage{bm} %bold math symbols
\usepackage[draft]{minted} % code highlighting
\usepackage{amsmath}
\usepackage{multirow}
\usepackage{multicol}
\usepackage{comment}
\usepackage{mathtools}
\usepackage{subfig}
\usepackage{adjustbox}

\makeatletter
\def\PYGdefault@reset{\let\PYGdefault@it=\relax \let\PYGdefault@bf=\relax%
    \let\PYGdefault@ul=\relax \let\PYGdefault@tc=\relax%
    \let\PYGdefault@bc=\relax \let\PYGdefault@ff=\relax}
\def\PYGdefault@tok#1{\csname PYGdefault@tok@#1\endcsname}
\def\PYGdefault@toks#1+{\ifx\relax#1\empty\else%
    \PYGdefault@tok{#1}\expandafter\PYGdefault@toks\fi}
\def\PYGdefault@do#1{\PYGdefault@bc{\PYGdefault@tc{\PYGdefault@ul{%
    \PYGdefault@it{\PYGdefault@bf{\PYGdefault@ff{#1}}}}}}}
\def\PYGdefault#1#2{\PYGdefault@reset\PYGdefault@toks#1+\relax+\PYGdefault@do{#2}}

\@namedef{PYGdefault@tok@w}{\def\PYGdefault@tc##1{\textcolor[rgb]{0.73,0.73,0.73}{##1}}}
\@namedef{PYGdefault@tok@c}{\let\PYGdefault@it=\textit\def\PYGdefault@tc##1{\textcolor[rgb]{0.25,0.50,0.50}{##1}}}
\@namedef{PYGdefault@tok@cp}{\def\PYGdefault@tc##1{\textcolor[rgb]{0.74,0.48,0.00}{##1}}}
\@namedef{PYGdefault@tok@k}{\let\PYGdefault@bf=\textbf\def\PYGdefault@tc##1{\textcolor[rgb]{0.00,0.50,0.00}{##1}}}
\@namedef{PYGdefault@tok@kp}{\def\PYGdefault@tc##1{\textcolor[rgb]{0.00,0.50,0.00}{##1}}}
\@namedef{PYGdefault@tok@kt}{\def\PYGdefault@tc##1{\textcolor[rgb]{0.69,0.00,0.25}{##1}}}
\@namedef{PYGdefault@tok@o}{\def\PYGdefault@tc##1{\textcolor[rgb]{0.40,0.40,0.40}{##1}}}
\@namedef{PYGdefault@tok@ow}{\let\PYGdefault@bf=\textbf\def\PYGdefault@tc##1{\textcolor[rgb]{0.67,0.13,1.00}{##1}}}
\@namedef{PYGdefault@tok@nb}{\def\PYGdefault@tc##1{\textcolor[rgb]{0.00,0.50,0.00}{##1}}}
\@namedef{PYGdefault@tok@nf}{\def\PYGdefault@tc##1{\textcolor[rgb]{0.00,0.00,1.00}{##1}}}
\@namedef{PYGdefault@tok@nc}{\let\PYGdefault@bf=\textbf\def\PYGdefault@tc##1{\textcolor[rgb]{0.00,0.00,1.00}{##1}}}
\@namedef{PYGdefault@tok@nn}{\let\PYGdefault@bf=\textbf\def\PYGdefault@tc##1{\textcolor[rgb]{0.00,0.00,1.00}{##1}}}
\@namedef{PYGdefault@tok@ne}{\let\PYGdefault@bf=\textbf\def\PYGdefault@tc##1{\textcolor[rgb]{0.82,0.25,0.23}{##1}}}
\@namedef{PYGdefault@tok@nv}{\def\PYGdefault@tc##1{\textcolor[rgb]{0.10,0.09,0.49}{##1}}}
\@namedef{PYGdefault@tok@no}{\def\PYGdefault@tc##1{\textcolor[rgb]{0.53,0.00,0.00}{##1}}}
\@namedef{PYGdefault@tok@nl}{\def\PYGdefault@tc##1{\textcolor[rgb]{0.63,0.63,0.00}{##1}}}
\@namedef{PYGdefault@tok@ni}{\let\PYGdefault@bf=\textbf\def\PYGdefault@tc##1{\textcolor[rgb]{0.60,0.60,0.60}{##1}}}
\@namedef{PYGdefault@tok@na}{\def\PYGdefault@tc##1{\textcolor[rgb]{0.49,0.56,0.16}{##1}}}
\@namedef{PYGdefault@tok@nt}{\let\PYGdefault@bf=\textbf\def\PYGdefault@tc##1{\textcolor[rgb]{0.00,0.50,0.00}{##1}}}
\@namedef{PYGdefault@tok@nd}{\def\PYGdefault@tc##1{\textcolor[rgb]{0.67,0.13,1.00}{##1}}}
\@namedef{PYGdefault@tok@s}{\def\PYGdefault@tc##1{\textcolor[rgb]{0.73,0.13,0.13}{##1}}}
\@namedef{PYGdefault@tok@sd}{\let\PYGdefault@it=\textit\def\PYGdefault@tc##1{\textcolor[rgb]{0.73,0.13,0.13}{##1}}}
\@namedef{PYGdefault@tok@si}{\let\PYGdefault@bf=\textbf\def\PYGdefault@tc##1{\textcolor[rgb]{0.73,0.40,0.53}{##1}}}
\@namedef{PYGdefault@tok@se}{\let\PYGdefault@bf=\textbf\def\PYGdefault@tc##1{\textcolor[rgb]{0.73,0.40,0.13}{##1}}}
\@namedef{PYGdefault@tok@sr}{\def\PYGdefault@tc##1{\textcolor[rgb]{0.73,0.40,0.53}{##1}}}
\@namedef{PYGdefault@tok@ss}{\def\PYGdefault@tc##1{\textcolor[rgb]{0.10,0.09,0.49}{##1}}}
\@namedef{PYGdefault@tok@sx}{\def\PYGdefault@tc##1{\textcolor[rgb]{0.00,0.50,0.00}{##1}}}
\@namedef{PYGdefault@tok@m}{\def\PYGdefault@tc##1{\textcolor[rgb]{0.40,0.40,0.40}{##1}}}
\@namedef{PYGdefault@tok@gh}{\let\PYGdefault@bf=\textbf\def\PYGdefault@tc##1{\textcolor[rgb]{0.00,0.00,0.50}{##1}}}
\@namedef{PYGdefault@tok@gu}{\let\PYGdefault@bf=\textbf\def\PYGdefault@tc##1{\textcolor[rgb]{0.50,0.00,0.50}{##1}}}
\@namedef{PYGdefault@tok@gd}{\def\PYGdefault@tc##1{\textcolor[rgb]{0.63,0.00,0.00}{##1}}}
\@namedef{PYGdefault@tok@gi}{\def\PYGdefault@tc##1{\textcolor[rgb]{0.00,0.63,0.00}{##1}}}
\@namedef{PYGdefault@tok@gr}{\def\PYGdefault@tc##1{\textcolor[rgb]{1.00,0.00,0.00}{##1}}}
\@namedef{PYGdefault@tok@ge}{\let\PYGdefault@it=\textit}
\@namedef{PYGdefault@tok@gs}{\let\PYGdefault@bf=\textbf}
\@namedef{PYGdefault@tok@gp}{\let\PYGdefault@bf=\textbf\def\PYGdefault@tc##1{\textcolor[rgb]{0.00,0.00,0.50}{##1}}}
\@namedef{PYGdefault@tok@go}{\def\PYGdefault@tc##1{\textcolor[rgb]{0.53,0.53,0.53}{##1}}}
\@namedef{PYGdefault@tok@gt}{\def\PYGdefault@tc##1{\textcolor[rgb]{0.00,0.27,0.87}{##1}}}
\@namedef{PYGdefault@tok@err}{\def\PYGdefault@bc##1{{\setlength{\fboxsep}{\string -\fboxrule}\fcolorbox[rgb]{1.00,0.00,0.00}{1,1,1}{\strut ##1}}}}
\@namedef{PYGdefault@tok@kc}{\let\PYGdefault@bf=\textbf\def\PYGdefault@tc##1{\textcolor[rgb]{0.00,0.50,0.00}{##1}}}
\@namedef{PYGdefault@tok@kd}{\let\PYGdefault@bf=\textbf\def\PYGdefault@tc##1{\textcolor[rgb]{0.00,0.50,0.00}{##1}}}
\@namedef{PYGdefault@tok@kn}{\let\PYGdefault@bf=\textbf\def\PYGdefault@tc##1{\textcolor[rgb]{0.00,0.50,0.00}{##1}}}
\@namedef{PYGdefault@tok@kr}{\let\PYGdefault@bf=\textbf\def\PYGdefault@tc##1{\textcolor[rgb]{0.00,0.50,0.00}{##1}}}
\@namedef{PYGdefault@tok@bp}{\def\PYGdefault@tc##1{\textcolor[rgb]{0.00,0.50,0.00}{##1}}}
\@namedef{PYGdefault@tok@fm}{\def\PYGdefault@tc##1{\textcolor[rgb]{0.00,0.00,1.00}{##1}}}
\@namedef{PYGdefault@tok@vc}{\def\PYGdefault@tc##1{\textcolor[rgb]{0.10,0.09,0.49}{##1}}}
\@namedef{PYGdefault@tok@vg}{\def\PYGdefault@tc##1{\textcolor[rgb]{0.10,0.09,0.49}{##1}}}
\@namedef{PYGdefault@tok@vi}{\def\PYGdefault@tc##1{\textcolor[rgb]{0.10,0.09,0.49}{##1}}}
\@namedef{PYGdefault@tok@vm}{\def\PYGdefault@tc##1{\textcolor[rgb]{0.10,0.09,0.49}{##1}}}
\@namedef{PYGdefault@tok@sa}{\def\PYGdefault@tc##1{\textcolor[rgb]{0.73,0.13,0.13}{##1}}}
\@namedef{PYGdefault@tok@sb}{\def\PYGdefault@tc##1{\textcolor[rgb]{0.73,0.13,0.13}{##1}}}
\@namedef{PYGdefault@tok@sc}{\def\PYGdefault@tc##1{\textcolor[rgb]{0.73,0.13,0.13}{##1}}}
\@namedef{PYGdefault@tok@dl}{\def\PYGdefault@tc##1{\textcolor[rgb]{0.73,0.13,0.13}{##1}}}
\@namedef{PYGdefault@tok@s2}{\def\PYGdefault@tc##1{\textcolor[rgb]{0.73,0.13,0.13}{##1}}}
\@namedef{PYGdefault@tok@sh}{\def\PYGdefault@tc##1{\textcolor[rgb]{0.73,0.13,0.13}{##1}}}
\@namedef{PYGdefault@tok@s1}{\def\PYGdefault@tc##1{\textcolor[rgb]{0.73,0.13,0.13}{##1}}}
\@namedef{PYGdefault@tok@mb}{\def\PYGdefault@tc##1{\textcolor[rgb]{0.40,0.40,0.40}{##1}}}
\@namedef{PYGdefault@tok@mf}{\def\PYGdefault@tc##1{\textcolor[rgb]{0.40,0.40,0.40}{##1}}}
\@namedef{PYGdefault@tok@mh}{\def\PYGdefault@tc##1{\textcolor[rgb]{0.40,0.40,0.40}{##1}}}
\@namedef{PYGdefault@tok@mi}{\def\PYGdefault@tc##1{\textcolor[rgb]{0.40,0.40,0.40}{##1}}}
\@namedef{PYGdefault@tok@il}{\def\PYGdefault@tc##1{\textcolor[rgb]{0.40,0.40,0.40}{##1}}}
\@namedef{PYGdefault@tok@mo}{\def\PYGdefault@tc##1{\textcolor[rgb]{0.40,0.40,0.40}{##1}}}
\@namedef{PYGdefault@tok@ch}{\let\PYGdefault@it=\textit\def\PYGdefault@tc##1{\textcolor[rgb]{0.25,0.50,0.50}{##1}}}
\@namedef{PYGdefault@tok@cm}{\let\PYGdefault@it=\textit\def\PYGdefault@tc##1{\textcolor[rgb]{0.25,0.50,0.50}{##1}}}
\@namedef{PYGdefault@tok@cpf}{\let\PYGdefault@it=\textit\def\PYGdefault@tc##1{\textcolor[rgb]{0.25,0.50,0.50}{##1}}}
\@namedef{PYGdefault@tok@c1}{\let\PYGdefault@it=\textit\def\PYGdefault@tc##1{\textcolor[rgb]{0.25,0.50,0.50}{##1}}}
\@namedef{PYGdefault@tok@cs}{\let\PYGdefault@it=\textit\def\PYGdefault@tc##1{\textcolor[rgb]{0.25,0.50,0.50}{##1}}}

\makeatother

\makeatletter
\def\PYG@reset{\let\PYG@it=\relax \let\PYG@bf=\relax%
    \let\PYG@ul=\relax \let\PYG@tc=\relax%
    \let\PYG@bc=\relax \let\PYG@ff=\relax}
\def\PYG@tok#1{\csname PYG@tok@#1\endcsname}
\def\PYG@toks#1+{\ifx\relax#1\empty\else%
    \PYG@tok{#1}\expandafter\PYG@toks\fi}
\def\PYG@do#1{\PYG@bc{\PYG@tc{\PYG@ul{%
    \PYG@it{\PYG@bf{\PYG@ff{#1}}}}}}}
\def\PYG#1#2{\PYG@reset\PYG@toks#1+\relax+\PYG@do{#2}}

\@namedef{PYG@tok@w}{\def\PYG@tc##1{\textcolor[rgb]{0.73,0.73,0.73}{##1}}}
\@namedef{PYG@tok@c}{\let\PYG@it=\textit\def\PYG@tc##1{\textcolor[rgb]{0.25,0.50,0.50}{##1}}}
\@namedef{PYG@tok@cp}{\def\PYG@tc##1{\textcolor[rgb]{0.74,0.48,0.00}{##1}}}
\@namedef{PYG@tok@k}{\let\PYG@bf=\textbf\def\PYG@tc##1{\textcolor[rgb]{0.00,0.50,0.00}{##1}}}
\@namedef{PYG@tok@kp}{\def\PYG@tc##1{\textcolor[rgb]{0.00,0.50,0.00}{##1}}}
\@namedef{PYG@tok@kt}{\def\PYG@tc##1{\textcolor[rgb]{0.69,0.00,0.25}{##1}}}
\@namedef{PYG@tok@o}{\def\PYG@tc##1{\textcolor[rgb]{0.40,0.40,0.40}{##1}}}
\@namedef{PYG@tok@ow}{\let\PYG@bf=\textbf\def\PYG@tc##1{\textcolor[rgb]{0.67,0.13,1.00}{##1}}}
\@namedef{PYG@tok@nb}{\def\PYG@tc##1{\textcolor[rgb]{0.00,0.50,0.00}{##1}}}
\@namedef{PYG@tok@nf}{\def\PYG@tc##1{\textcolor[rgb]{0.00,0.00,1.00}{##1}}}
\@namedef{PYG@tok@nc}{\let\PYG@bf=\textbf\def\PYG@tc##1{\textcolor[rgb]{0.00,0.00,1.00}{##1}}}
\@namedef{PYG@tok@nn}{\let\PYG@bf=\textbf\def\PYG@tc##1{\textcolor[rgb]{0.00,0.00,1.00}{##1}}}
\@namedef{PYG@tok@ne}{\let\PYG@bf=\textbf\def\PYG@tc##1{\textcolor[rgb]{0.82,0.25,0.23}{##1}}}
\@namedef{PYG@tok@nv}{\def\PYG@tc##1{\textcolor[rgb]{0.10,0.09,0.49}{##1}}}
\@namedef{PYG@tok@no}{\def\PYG@tc##1{\textcolor[rgb]{0.53,0.00,0.00}{##1}}}
\@namedef{PYG@tok@nl}{\def\PYG@tc##1{\textcolor[rgb]{0.63,0.63,0.00}{##1}}}
\@namedef{PYG@tok@ni}{\let\PYG@bf=\textbf\def\PYG@tc##1{\textcolor[rgb]{0.60,0.60,0.60}{##1}}}
\@namedef{PYG@tok@na}{\def\PYG@tc##1{\textcolor[rgb]{0.49,0.56,0.16}{##1}}}
\@namedef{PYG@tok@nt}{\let\PYG@bf=\textbf\def\PYG@tc##1{\textcolor[rgb]{0.00,0.50,0.00}{##1}}}
\@namedef{PYG@tok@nd}{\def\PYG@tc##1{\textcolor[rgb]{0.67,0.13,1.00}{##1}}}
\@namedef{PYG@tok@s}{\def\PYG@tc##1{\textcolor[rgb]{0.73,0.13,0.13}{##1}}}
\@namedef{PYG@tok@sd}{\let\PYG@it=\textit\def\PYG@tc##1{\textcolor[rgb]{0.73,0.13,0.13}{##1}}}
\@namedef{PYG@tok@si}{\let\PYG@bf=\textbf\def\PYG@tc##1{\textcolor[rgb]{0.73,0.40,0.53}{##1}}}
\@namedef{PYG@tok@se}{\let\PYG@bf=\textbf\def\PYG@tc##1{\textcolor[rgb]{0.73,0.40,0.13}{##1}}}
\@namedef{PYG@tok@sr}{\def\PYG@tc##1{\textcolor[rgb]{0.73,0.40,0.53}{##1}}}
\@namedef{PYG@tok@ss}{\def\PYG@tc##1{\textcolor[rgb]{0.10,0.09,0.49}{##1}}}
\@namedef{PYG@tok@sx}{\def\PYG@tc##1{\textcolor[rgb]{0.00,0.50,0.00}{##1}}}
\@namedef{PYG@tok@m}{\def\PYG@tc##1{\textcolor[rgb]{0.40,0.40,0.40}{##1}}}
\@namedef{PYG@tok@gh}{\let\PYG@bf=\textbf\def\PYG@tc##1{\textcolor[rgb]{0.00,0.00,0.50}{##1}}}
\@namedef{PYG@tok@gu}{\let\PYG@bf=\textbf\def\PYG@tc##1{\textcolor[rgb]{0.50,0.00,0.50}{##1}}}
\@namedef{PYG@tok@gd}{\def\PYG@tc##1{\textcolor[rgb]{0.63,0.00,0.00}{##1}}}
\@namedef{PYG@tok@gi}{\def\PYG@tc##1{\textcolor[rgb]{0.00,0.63,0.00}{##1}}}
\@namedef{PYG@tok@gr}{\def\PYG@tc##1{\textcolor[rgb]{1.00,0.00,0.00}{##1}}}
\@namedef{PYG@tok@ge}{\let\PYG@it=\textit}
\@namedef{PYG@tok@gs}{\let\PYG@bf=\textbf}
\@namedef{PYG@tok@gp}{\let\PYG@bf=\textbf\def\PYG@tc##1{\textcolor[rgb]{0.00,0.00,0.50}{##1}}}
\@namedef{PYG@tok@go}{\def\PYG@tc##1{\textcolor[rgb]{0.53,0.53,0.53}{##1}}}
\@namedef{PYG@tok@gt}{\def\PYG@tc##1{\textcolor[rgb]{0.00,0.27,0.87}{##1}}}
\@namedef{PYG@tok@err}{\def\PYG@bc##1{{\setlength{\fboxsep}{\string -\fboxrule}\fcolorbox[rgb]{1.00,0.00,0.00}{1,1,1}{\strut ##1}}}}
\@namedef{PYG@tok@kc}{\let\PYG@bf=\textbf\def\PYG@tc##1{\textcolor[rgb]{0.00,0.50,0.00}{##1}}}
\@namedef{PYG@tok@kd}{\let\PYG@bf=\textbf\def\PYG@tc##1{\textcolor[rgb]{0.00,0.50,0.00}{##1}}}
\@namedef{PYG@tok@kn}{\let\PYG@bf=\textbf\def\PYG@tc##1{\textcolor[rgb]{0.00,0.50,0.00}{##1}}}
\@namedef{PYG@tok@kr}{\let\PYG@bf=\textbf\def\PYG@tc##1{\textcolor[rgb]{0.00,0.50,0.00}{##1}}}
\@namedef{PYG@tok@bp}{\def\PYG@tc##1{\textcolor[rgb]{0.00,0.50,0.00}{##1}}}
\@namedef{PYG@tok@fm}{\def\PYG@tc##1{\textcolor[rgb]{0.00,0.00,1.00}{##1}}}
\@namedef{PYG@tok@vc}{\def\PYG@tc##1{\textcolor[rgb]{0.10,0.09,0.49}{##1}}}
\@namedef{PYG@tok@vg}{\def\PYG@tc##1{\textcolor[rgb]{0.10,0.09,0.49}{##1}}}
\@namedef{PYG@tok@vi}{\def\PYG@tc##1{\textcolor[rgb]{0.10,0.09,0.49}{##1}}}
\@namedef{PYG@tok@vm}{\def\PYG@tc##1{\textcolor[rgb]{0.10,0.09,0.49}{##1}}}
\@namedef{PYG@tok@sa}{\def\PYG@tc##1{\textcolor[rgb]{0.73,0.13,0.13}{##1}}}
\@namedef{PYG@tok@sb}{\def\PYG@tc##1{\textcolor[rgb]{0.73,0.13,0.13}{##1}}}
\@namedef{PYG@tok@sc}{\def\PYG@tc##1{\textcolor[rgb]{0.73,0.13,0.13}{##1}}}
\@namedef{PYG@tok@dl}{\def\PYG@tc##1{\textcolor[rgb]{0.73,0.13,0.13}{##1}}}
\@namedef{PYG@tok@s2}{\def\PYG@tc##1{\textcolor[rgb]{0.73,0.13,0.13}{##1}}}
\@namedef{PYG@tok@sh}{\def\PYG@tc##1{\textcolor[rgb]{0.73,0.13,0.13}{##1}}}
\@namedef{PYG@tok@s1}{\def\PYG@tc##1{\textcolor[rgb]{0.73,0.13,0.13}{##1}}}
\@namedef{PYG@tok@mb}{\def\PYG@tc##1{\textcolor[rgb]{0.40,0.40,0.40}{##1}}}
\@namedef{PYG@tok@mf}{\def\PYG@tc##1{\textcolor[rgb]{0.40,0.40,0.40}{##1}}}
\@namedef{PYG@tok@mh}{\def\PYG@tc##1{\textcolor[rgb]{0.40,0.40,0.40}{##1}}}
\@namedef{PYG@tok@mi}{\def\PYG@tc##1{\textcolor[rgb]{0.40,0.40,0.40}{##1}}}
\@namedef{PYG@tok@il}{\def\PYG@tc##1{\textcolor[rgb]{0.40,0.40,0.40}{##1}}}
\@namedef{PYG@tok@mo}{\def\PYG@tc##1{\textcolor[rgb]{0.40,0.40,0.40}{##1}}}
\@namedef{PYG@tok@ch}{\let\PYG@it=\textit\def\PYG@tc##1{\textcolor[rgb]{0.25,0.50,0.50}{##1}}}
\@namedef{PYG@tok@cm}{\let\PYG@it=\textit\def\PYG@tc##1{\textcolor[rgb]{0.25,0.50,0.50}{##1}}}
\@namedef{PYG@tok@cpf}{\let\PYG@it=\textit\def\PYG@tc##1{\textcolor[rgb]{0.25,0.50,0.50}{##1}}}
\@namedef{PYG@tok@c1}{\let\PYG@it=\textit\def\PYG@tc##1{\textcolor[rgb]{0.25,0.50,0.50}{##1}}}
\@namedef{PYG@tok@cs}{\let\PYG@it=\textit\def\PYG@tc##1{\textcolor[rgb]{0.25,0.50,0.50}{##1}}}

\def\PYGZus{\char`\_}

\def\PYGZsh{\char`\#}

\def\PYGZhy{\char`\-}
\def\PYGZsq{\char`\'}

\makeatother
\title{Pythae: Unifying Generative Autoencoders in Python A Benchmarking Use Case}

\author{%
    Clément Chadebec\\
  Université Paris Cité, INRIA, Inserm, SU\\
  Centre de Recherche des Cordeliers \thanks{15 Rue de l'École de Médecine, 75006 Paris}\\
  \texttt{clement.chadebec@inria.fr} \\
  \And
  Louis J. Vincent\\
  Implicity \thanks{ \url{https://www.implicity.com} - Implicity Paris, France. }\\
  Université Paris Cité, INRIA, Inserm, SU \\
  Centre de Recherche des Cordeliers \footnotemark[1]\\
   \texttt{louis.vincent@inria.fr} \\
   \AND
   Stéphanie Allassonnière \\
   Université Paris Cité, INRIA, Inserm, SU \\
   Centre de Recherche des Cordeliers \footnotemark[1]\\
   \texttt{stephanie.allassonniere@inria.fr}
}

\begin{document}

\maketitle
\begin{abstract}
    %The use of amortized inference in the context of generative models has been a key advance which has led to multiple variational autoencoder methods who have proven to be useful in a variety of applications.
    
    In recent years, deep generative models have attracted increasing interest due to their capacity to model complex distributions. Among those models, variational autoencoders have gained popularity as they have proven both to be computationally efficient and yield impressive results in multiple fields. Following this breakthrough, extensive research has been done in order to improve the original publication, resulting in a variety of different VAE models in response to different tasks. In this paper we present \textbf{Pythae}, a versatile \textit{open-source} Python library providing both a \textit{unified implementation} and a dedicated framework allowing \textit{straightforward}, \emph{reproducible} and \textit{reliable} use of generative autoencoder models. As an example of application, we propose to use this library to perform a case study benchmark where we present and compare 19 generative autoencoder models representative of some of the main improvements on downstream tasks such as image reconstruction, generation, classification, clustering and interpolation. The open-source library can be found at \url{https://github.com/clementchadebec/benchmark_VAE}.
    %plug and play, easy to use 
    
\begin{comment}
    %Why VAEs are important
  Generative models are nice (but costly ?)
  
  Building up on this breakthrough, various papers have proposed une amelioration quelconque
  
  (However these techniques can sometimes be inaccessible and hard to compare / comprehend ?)
  
\end{comment}
  
\end{abstract}

\section{Introduction}

%Expand on abstract

%$\bullet$ Learning models capable of representing complex distributions is a big challenge in ML

%Generative models are nice (but costly ?)
Over the past few years, 
%deep generative models such as Generative Adversarial Networks (GAN) \citep{goodfellow_generative_2014} or 
generative models have proven to be a promising approach for modelling datasets with complex inherent distributions such a natural images. Among those, Variational AutoEncoders (VAE) \citep{kingma_auto-encoding_2014,rezende_stochastic_2014} have gained popularity due to their computational efficiency and scalability, leading to many applications such as speech modelling \citep{blaauw_modeling_2016}, clustering \citep{dilokthanakul_deep_2017,yang_deep_2019}, data augmentation \citep{chadebec_data_2021} or image generation \citep{razavi_generating_2019}.
Similarly to autoencoders, these models encourage good reconstruction of an observed input data from a latent representation, but they further assume latent vectors to be random variables involved in the generation process of the observed data.
This imposes a latent structure wherein latent variables are driven to follow a prior distribution that can then be used to generate new data. 
%All of these aspects make VAEs a computationally efficient and highly scalable class of models useful for many tasks such as speech modelling \citep{blaauw_modeling_2016}, clustering \citep{dilokthanakul_deep_2017,yang_deep_2019}, data augmentation \citep{chadebec_data_2021} or image generation \citep{razavi_generating_2019}.
Since this breakthrough, various contributions have been made to enrich the original VAE scheme through new generating strategies \citep{dilokthanakul_deep_2017,tomczak_vae_2018,ghosh_variational_2020,bauer_resampled_2019,pang_learning_2020}, reconstruction objectives \citep{larsen_autoencoding_2015,snell2017vae} and more adapted latent representations \citep{higgins_beta-vae_2017,kim2018factorvae,oord2018vae,arvanitidis_latent_2018,chadebec_data_2021} to cite a few. 
A drawback of VAEs is that due to the intractability of the log-likelihood objective function, VAEs have to resort to optimizing a lower bound on the true objective as a proxy, which has been mentionned as a major limitation of the model \citep{burda_importance_2016,alemi_deep_2016,higgins_beta-vae_2017,cremer_inference_2018,zhang_advances_2018}. Hence, extensive research has been proposed to improve this bound through richer distributions \citep{salimans_markov_2015,rezende_variational_2015,kingma_improved_2016,caterini_hamiltonian_2018}.
More recently, it has been shown that autoencoders can be turned into generative approaches through latent density estimation \cite{ghosh_variational_2020}, extending the concept of \emph{Generative AutoEncoders} (GAE) to a more general class of autoencoder models.

Nonetheless, most of this research has been done in parallel across disjoint sub-fields of research and to the best of our knowledge little to no work has been done on homogenising and integrating these distinct methods in a common framework. Moreover, for many of the aforementioned publications, implementations may not be available or maintained, therefore requiring time-consuming re-implementation. This induces a strong bottleneck for research to move forward in this field and makes reproducibility challenging, which calls for the need of a unified generative autoencoder framework.  To address this issue we introduce \textbf{Pythae} (\textbf{Pyth}on \textbf{A}uto\textbf{E}ncoder), a versatile open source Python library for generative autoencoders providing unified implementations of common methods, along with a reproducible framework allowing for easy model training, data generation and experiment tracking. We then propose to illustrate the usefulness of the proposed library on a benchmark case study of 19 generative autoencoder methods on classical image datasets. We consider five different downstream tasks: image reconstruction and generation, latent vector classification and clustering, and image interpolation on three well known imaging datasets.

\section{Variational autoencoders}

    In this section, we recall the original VAE setting and present some of the main improvements that were proposed to enhance the model.

    \subsection{Background}
%    - We consider that observed entities $x$ are spanned from $z$
    Given $x \in \mathbb{R}^D$, a set of observed variables deriving from an unknown distribution $p(x)$, a VAE assumes that there exists $z \in \mathbb{R}^d$ such that $z$ is a latent representation of $x$. The generation process of $x$ thus decomposes as
    \begin{equation}\label{eq: gen process}
        p_\theta(x) = \int_\mathcal{Z} p_\theta(x | z) p_z(z) dz\,,
    \end{equation}
    where $p_z$ is the \textit{prior distribution} on the latent space $\mathbb{R}^d$. The distribution $p_\theta(x | z)$ is referred to as the \emph{decoder} and is modelled with a simple parametric distribution whose parameters are given by a neural network. Since the true posterior $p_{\theta}(z|x)$ is most of the time intractable due to the integral in Eq.~\eqref{eq: gen process} recourse to Variational Inference \citep{jordan_introduction_1999} is needed and a variational distribution $q_{\phi}(z|x)$ which we refer to as the \emph{encoder} is introduced. The approximate posterior $q_\phi$ is again taken as a simple parametric distribution whose parameters are also modelled by a neural network. This allows to define an unbiased estimate $\widehat{p}_{\theta}$ of the marginal distribution $p_{\theta}(x)$ using importance sampling with $q_{\phi}(z|x)$ \emph{i.e.} 
    $\widehat{p}_{\theta}(x) =  \frac{p_{\theta}(x|z)p_z(z)}{q_{\phi}(z|x)}$ and $\mathbb{E}_{z \sim q_\phi}\big[\widehat{p}_{\theta} \big] = p_{\theta}$. Applying Jensen's inequality leads to a lower bound on the likelihood given in Eq.~\eqref{eq: gen process}:
    \begin{equation}\label{eq: elbo decomposition}
        \begin{aligned}
           \log \underbrace{\mathbb{E}_{z \sim q_\phi}\big[\widehat{p}_{\theta}(x) \big]}_{p_{\theta}(x)} &\geq \mathbb{E}_{z \sim q_\phi}\big[\log \widehat{p}_{\theta}(x) \big]=\underbrace{\mathbb{E}_{z \sim q_\phi}[\log p_\theta(x | z)]}_\text{reconstruction} - \underbrace{\mathcal{D}_{KL}\big[ q_\phi(z|x) || p_z(z) \big]}_\text{regularisation}\,,
        \end{aligned}
    \end{equation}
where $\mathcal{D}_{KL}(p||q)$ is the Kullback-Leibler divergence between distributions $p$ and $q$. This bound is referred to as the Evidence Lower Bound (ELBO) \cite{kingma_auto-encoding_2014} and is used as the training objective to maximize in the traditional VAE scheme. It can be interpreted as a two terms objective \citep{ghosh_variational_2020} where the \emph{reconstruction} loss forces the output of the decoder to be close to the original input $x$, while the \emph{regularisation} loss forces the posterior distribution $q_\phi(z | x)$ outputted by the encoder to be close to the prior distribution $p_z(z)$. Under standard VAE assumption, the prior distribution is a multivariate standard Gaussian $p_z = \mathcal{N}(0,I_d)$, the approximate posterior is set to $q_\phi(z|x) =\mathcal{N}\big(z \big\vert \mu(x),\Sigma(x)\big)$ where  $\big(\mu(x),\Sigma(x)\big)$ are outputs of the encoder network. 

    \subsection{Improvements upon the classical VAE method}\label{sec: enhancing the model}
    
    Building on the breakthrough of VAEs, several papers have proposed improvements to the model. In this section we present 4 axes which we consider to be representative of the major advancements made on VAEs, as well as classical models characterising the main improvements within each of these axes.
    %We choose 4 main axes, blabalbla segway to library because we choose them to be REPRESENTATIVE of GAEs

    \paragraph{Improving the prior}\label{par: better prior} It has been shown that the role of the prior distribution $p_z$ is crucial in the good performance of the VAE \citep{hoffman_elbo_2016} and choosing a family of overly simplistic priors can lead to over-regularization \citep{connor2021variational} and poor reconstruction performance \citep{dai_diagnosing_2018}. In particular, it was shown that the prior maximizing the ELBO objective is the \emph{aggregated posterior} $q(z) = \frac{1}{N}\sum_{i=1}^N q_{\phi}(z|x_i)$ \citep{tomczak_vae_2018}. However, it should be noted that a perfect fit between the prior and the aggregated posterior is not necessarily desired since it has been shown in \citep{bauer_resampled_2019-1,tomczak_vae_2018} that it may lead to over-fitting as it essentially amounts to the model memorising the training set. Hence, multi-modal priors \citep{nalisnick_approximate_2016,dilokthanakul_deep_2017,tomczak_vae_2018} were proposed, followed by hierarchical latent variable models \citep{sonderby_ladder_2016,klushyn_learning_2019} and prior learning based approaches \citep{chen_variational_2016,aneja_ncp-vae_2020} to address the poor expressiveness of the prior distribution and model richer generative distributions. 
    Considering a specific geometry of the latent space also led to alternative priors taking into account geometrical aspects of the latent space \citep{davidson_hyperspherical_2018,falorsi_explorations_2018,mathieu_continuous_2019,arvanitidis_latent_2018,chen_metrics_2018,shao_riemannian_2018,kalatzis_variational_2020,chadebec_data_2021}. Another interesting approach proposed for instance in \citep{oord2018vae,ghosh_variational_2020} consists in using density estimation post training with another distribution or normalising flows \citep{rezende_variational_2015} on the learned latent codes. 
    
    % \begin{itemize}
    %     \item \textbf{VAE with a VampPrior} \citet{tomczak_vae_2018} propose to replace the traditional prior distribution with a mixture of variational posteriors anchored around selected \textit{pseudo-input} points that are learned to be representative of the posterior distribution
    
    %      \item \textbf{Hyperspherical VAE} \citet{davidson_hyperspherical_2018} propose an alternative uniform prior set on a hypersphere to deal with datasets that do not have a defined center point, allowing latent clusters of data to evenly spread over the latent space.
    
    %     \item \textbf{Vector Quantized VAE} \citet{oord2018vae} adapt the VAE setting to a case where the latent space is a discrete embedding space and the prior is learned rather than imposed.
    
    % \end{itemize}

    \textbf{Towards a better lower bound} Another major axis of improvement of the VAE model has been to tighten the gap between the ELBO objective and the true log probability \cite{burda_importance_2016, alemi_deep_2016, higgins_beta-vae_2017, cremer_inference_2018, zhang_advances_2018}. 
    The ELBO objective can indeed be written as the difference between the true log probability and a KL divergence between the approximate posterior and the true posterior
    \begin{equation}\label{eq: elbo2}
    \mathcal{L}_{\text{ELBO}}(x) = \log p_{\theta}(x) - \mathcal{D}_{KL}\big[ q_\phi(z | x) || p(z|x) \big]\,.
    \end{equation}
    Hence, if one wants to make the ELBO gap tighter, particular attention should be paid to the choice in the approximate posterior $q_{\phi}(z|x)$. In the original model, $q_{\phi}(z|x)$ is chosen as a simple distribution for tractability of the ELBO in Eq.~\eqref{eq: elbo decomposition}. 
    However, several approaches have been proposed to extend the choice of $q_{\phi}$ to a wider class of distributions using MCMC sampling \citep{salimans_markov_2015} or normalising flows \citep{rezende_variational_2015}. For instance, \citet{kingma_improved_2016} improve upon the works of \cite{rezende_variational_2015} with an inverse auto-regressive normalising flow (IAF), a new type of normalizing flow that better scales to high-dimensional latent spaces. With this objective in mind a Hamiltonian VAE aimed at targeting the true posterior during training with a Hamiltonian Monte Carlo \citep{neal_hamiltonian_2005} inspired scheme was proposed \citep{caterini_hamiltonian_2018} and extended to Riemannian latent spaces in \citep{chadebec_data_2021}.

    \paragraph{Encouraging disentanglement} 
    Although there is no clear consensus upon the definition of disentanglement, it is commonly referred to as the independence between features in a representation \citep{mathieu2019disentangling}. This is a desirable behaviour for VAEs, as it is argued that disentangled features may be more representative and interpretable \citep{higgins_beta-vae_2017}.
    %One key interest of VAEs is to learn a {more representative and interpretable} latent representation of the input data. 
    In that regard, several approaches have been proposed encouraging a better disentanglement of the features in the latent space. \citet{higgins_beta-vae_2017} first argue that increasing the weight of the KL divergence term in the ELBO loss enforces a higher disentanglement of the latent features as the posterior probability is forced to match a multivariate normal standard Gaussian. 
    Following this idea, \citep{burgess2018vae} propose to achieve disentanglement by gradually increasing the proximity between the posterior and the prior \citep{burgess2018vae}. Other methods challenge the view that disentanglement can be achieved by simply forcing the posterior to match the prior, or raise the point that in this case disentanglement is achieved at the cost of a bad reconstruction. From these observations, new approaches arise such as \citep{kim2018factorvae} who augment the VAE objective with a penalty that encourages factorial representation of the marginal distributions, or  \citep{chen2019vae} that enforce a
    penalty on the total correlation favouring disentanglement. 
    
    % \begin{itemize}
    %     \item \textbf{$\beta$-VAE}
    
    % \citet{higgins_beta-vae_2017} argue that increasing the weight of the KL divergence term in the ELBO loss enforces a higher disentanglement of the latent features as the posterior probability is forced to match a multivariate normal standard gaussian.
    
    %     \item \textbf{Disentangled $\beta$-VAE}
    
    % \citet{burgess2018vae} propose a modification of $\beta$-VAE that gradually increases the proximity between the posterior and the prior.
    
    %     \item \textbf{$\beta$-TC-VAE}
    
    % \citet{chen2019vae} extend on the ideas of the previous papers by adding multiple hyperparameters.
    
    %     \item \textbf{Factor VAE}
    
    % \citet{kim2018factorvae} augment the VAE objective with a penalty that encourages factorial representation of the marginal distributions, enforcing a stronger disentangling of the latent space.
    
    % \end{itemize}
    
    \textbf{Amending the distance between distributions} It can be stressed that the reconstruction term $\mathbb{E}_{z \sim q_\phi(z | x)}[\log p_\theta(x | z)]$ in eq.~\eqref{eq: elbo decomposition} has a crucial role in the reconstruction and that its choice should be dependent of the application. For instance, methods using a discriminator \citep{larsen_autoencoding_2015} or using a deterministic differentiable loss function  \citep{snell2017vae} acting as a distance between the input data and its reconstruction were also proposed. 
    The second term in the ELBO measures the distance between the approximate posterior and the prior distribution through the KL divergence and it has however been argued that other distances between probability distributions could be used instead. Hence, approaches using a GAN to distinguish samples from the posterior from samples from the prior distribution \citep{makhzani2016vae} or methods based on optimal transport have also been proposed \citep{tolstikhin2018wasserstein,zhao2018vae}.

\section{The Pythae library}
\paragraph{Why Pythae ?} To the best of our knowledge, although some well referenced libraries grouping different Variational Auto-Encoder methods exist (\emph{e.g.} \cite{Subramanian2020}), there exists no framework providing both adaptable and easy-to-use unified implementations of state-of-the-art Generative AutoEncoder (GAE) methods. 
%This induces a strong brake for research to move forward in this field since new comers often need to compare to existing baseline methods whose implementations might no longer be maintained, hard to access and re-use or completely unavailable. 
This induces both a strong brake for reproducible research and democratisation of the models since implementations might be difficult to adapt to other use-cases, no longer maintained, or completely unavailable.

\paragraph{Project vision}  Starting from this observation, we created \textbf{Pythae}, an open-source python library inspired from \citep{pedregosa_scikit-learn_2011,wolf-etal-2020-transformers} providing unified implementations of generative autoencoding methods, allowing for easy use and training of GAE models. Pythae is designed with the following points in mind:
\begin{itemize}
    \item \textbf{Usable by all} Pythae makes GAE models accessible to all - beginners to experts. This means beginners can run \emph{ready-to-use} models with a few lines of code, while more advanced users can easily access and adapt different methods to their specific use-cases, with custom encoder/decoder definition. Indeed, the library was designed to be flexible enough to allow users to use existing implementations on their own data, with custom model hyper-parameters, training configurations and network architectures. 
    
    The library has an online documentation\footnote{The full documentation can be found at \url{https://pythae.readthedocs.io/en/latest/}.} and is also explained and illustrated through tutorials available either on a local machine or on the \emph{Google Colab} platform \citep{Bisong2019}.

    \item \textbf{Unified implementation} The brick-like structure of Pythae allows for seamless but efficient interchange between models, sampling techniques, network architectures, model hyper-parameters and training schemes. Pythae is unit-tested ensuring code quality and continuous development with a code coverage of 98\% as of release 0.6. The library is made available on \emph{pip} and \emph{conda} allowing an easy integration. Its development is performed through releases that ensure stable and robust implementations. 
    
    \item \textbf{A reproducible research environment} Pythae is open to all and as such encourages transparent and reproducible research, as illustrated in the next section. With a variety of different interchangeable models gathered in a common library, it can be used as a sandbox for research and applications. Moreover, the library also integrates an easy-to-use experiment tracking tool (\textit{wandb}) \citep{wandb} allowing to monitor runs launched with Pythae and compare them through a graphic interface, and an online model sharing tool, the HuggingFace Hub, allowing to share models with peers.
    
    \item \textbf{Evolving and driven by the community} Pythae's design is intended to evolve with the addition of new models to enrich the existing model base. Furthermore, peers can contribute by reviewing and submitting models to enrich the library, a few of which have already been added at the time of this publication. 
    
    %\item \textbf{Open to and driven by the community}
    %\itme \textbf{An evolving reproducible research environment}
    %\item \textbf{Evolving and open to the community} Pythae is intended to evolve with the addition of new models to enrich the existing model base. Furthermore, Pythae is open-source and open to contributions to enrich the library, a few of which have already been made by peers.
    
    %beyond providing unified implementations, we believe that Pythae can also provide an open model testing environment driven by the community where people can contribute by adding their own models, and by doing so encourage reproducible research and the gathering of proposed methods at a same location. 
    %Pythae is open-source and open to contributions to enrich the library, a few of which have already been made by peers.
\end{itemize}

%This provides \emph{off-the-shelf} ready-to-use implementations directly applicable to any use case. This library is distributed under the Apache2.0 licence. Pythae is unit-tested ensuring code quality and continuous development with a code coverage of 97\% as of release 0.1. To make it as accessible as possible, the library has an online documentation\footnote{The full documentation can be found at \url{https://pythae.readthedocs.io/en/latest/}.} and is also explained and illustrated through tutorials available either on a local machine or on the \emph{Google Colab} platform \citep{Bisong2019}. This library is intended to evolve with the addition of new models through time to enrich the existing model base. Furthermore, beyond providing unified implementations, we believe that Pythae can also provide an open model testing environment driven by the community where people can contribute by adding their own models, and by doing so encourage reproducible research and the gathering of proposed methods at a same location. 

\paragraph{Code structure} Pythae was thought for easy model training and data generation, while striving for simplicity with a quick and user-friendly model selection and configuration. The backbone of the library is the module \textit{pythae.models} in which all the autoencoder models are implemented. Each model implementation is accompanied with a configuration \emph{dataclass} containing any hyper-parameters relative to the model and allowing easing configuration loading and saving from \emph{json} files. 

\begin{figure}[ht]
    \centering
    \includegraphics[width=\linewidth]{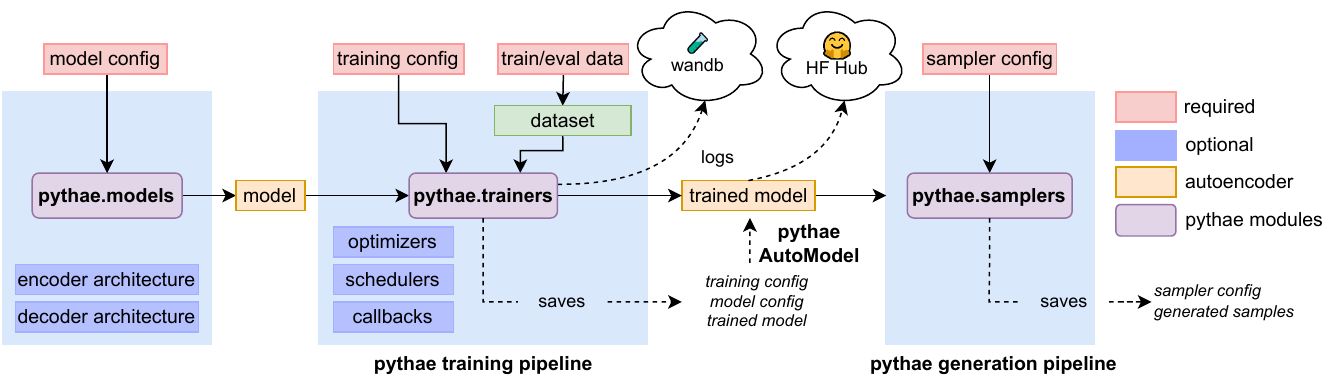}
    \caption{Pythae library diagram}
    \label{fig:pythae_diagram}
\end{figure}

All the models are implemented using a common API allowing for a seamless integration with \emph{pythae.trainers} (for training) and \emph{pythae.samplers} (for generation) along with a simplified usage as illustrated in Fig.~\ref{fig:pythae_diagram}. In particular, Pythae provides pipelines allowing to train an autoencoder model or to generate new data with only a few lines of code, as shown in Appendix.~\ref{appA}.

It mainly relies on the Pytorch \citep{paszke_automatic_2017} framework and in its basic usage only essential hyper-parameter configurations and data (arrays or tensors) are needed to launch a model training or generation. More advanced options allowing further flexibility such as defining custom encoder and decoder neural-networks are also available and can be found in the documentation and tutorials. It can adapt to various types of data through the use of different already-implemented or user provided encoder and decoder neural-network architectures. In addition, Pythae also provides several ways to generate new data through different popular sampling methods in the \emph{pythae.samplers} module. We detail some aspects of the library in Appendix.~\ref{appA}. 

% This library is intended to evolve with the addition of new models through time to enrich the existing model base. Furthermore, beyond providing unified implementations, we believe that Pythae can also provide an open benchmarking environment driven by the community where people can contribute by adding their own models and by doing so encourage reproducible research.

% Fig.~\ref{fig:pythae_diagram} provides a diagram presenting the library and the different modules articulated together. In particular, Pythae provides pipelines allowing to train an AutoEncoder model or to generate new data in only a few lines of code. In its basic usage only configurations and data are needed to launch a model training or generation. More advanced options allowing further flexibility such as defining custom encoder and decoder neural nets are also available and can found in the documentation and tutorials. We detail some aspects of the library in Appendix.~A. 

% Pythae is
% \begin{itemize}
%     \item Open-access 
%     Althought most of the main methods have already been implemented people are encouraged to contribute
    
%     Insert a table / list with all the VAEs

%     \item Easy to use
%     Aimed at people with or without experience with VAEs - (with our without a mathematical background)
    
%     \item Versatile
%     Can adapt to most data types, interchangeable
%     -> Insert schema d'utilisation
    
%     \item Aim of Pythae: reproducible research (hyper-parameters tuning), analysis of VAE models with experiment monitoring (wandb), unified implementations allowing reliable comparison. 
    
%     \item Inspired from scikit-learn \citep{pedregosa_scikit-learn_2011} and transformers lib \citep{wolf-etal-2020-transformers}
    
%     We will have to add the link to wandb exp \& report as url.
% \end{itemize}

% \subsection{An example of usage of Pythae}

% Any model can be configured and trained with 3 straightforward steps

\section{Case study benchmark}

By nature of its structured framework, Pythae allows for easy comparison between models on any chosen task. As an illustrative purpose, we propose a case study where we use Pythae to perform a straightforward benchmark comparison of models implemented in Pythae on a selection of well-known elementary tasks. 
%We choose to assess each model's capacity 
%The aim is not to determine which model performs best, but rather to identify general trends betweens groups of GAEs.
The aim of these tasks is to underline general trends within groups of GAEs, based on common behaviours, as well as judge the versatility of the models. However, this benchmark should not be considered as a means to rank models on these tasks, as performances depend on sometimes complex hyper-parameter tuning and training, which we consider to be outside of the scope of this benchmarking use case. The scripts used for the benchmark are provided in supplementary materials.
%Tables ref et ref are meant as illustrative support for results in figure ref.
%We chose to assess the quality of image reconstruction and generation, image interpolation and

\subsection{Benchmark setting}
In this section, we present the setting of the benchmark. Comprehensive results for all the experiments are available through the monitoring tool \citep{wandb} used in Pythae to allow complete transparency.

\paragraph{The data}
To perform the different tasks presented in this paper, 3 classical and widely used image datasets are considered: MNIST \citep{lecun_mnist_1998}, CIFAR10 \citep{krizhevsky2009learning} and CELEBA \citep{liu2015faceattributes}. These datasets are publicly available, widely used for generative model related papers and have well known associated metrics in the literature. Each dataset is split into a train set, a validation set and a test set. For MNIST and CIFAR10 the validation set is composed of the last 10k images extracted from the official train set and the test set corresponds to the official one. For CELEBA, we use the official train/val/test split. %For all the tasks presented in this benchmark, these splits remains unchanged.

\paragraph{The models}
We propose to compare 19 generative autoencoder models representative of the improvements proposed in the literature and presented in Sec.~\ref{sec: enhancing the model}. 
Descriptions and explanations of each implemented model can be found in Appendix.~\ref{appD}.
We use as baseline an Autoencoder (\textbf{AE}) and a Variational Autoencoder (\textbf{VAE}). To assess the influence of a \emph{more expressive prior}, we propose using a VAE with VAMP prior (\textbf{VAMP}) \citep{tomczak_vae_2018} and regularised autoencoders with either a gradient penalty (\textbf{RAE-GP}) or a L2 penalty on the weights of the decoder (\textbf{RAE-L2}) that use \emph{ex-post} density estimation \citep{ghosh_variational_2020}. 
To represent models trying to reach a \emph{better lower bound}, we choose a Importance Weighted Autoencoder (\textbf{IWAE}) \citep{burda_importance_2016} and VAEs adding either simple linear normalising flows (\textbf{VAE-lin-NF}) \citep{rezende_variational_2015} or using IAF (\textbf{VAE-IAF}) \citep{kingma_improved_2016}. 
For \emph{disentanglement-based models}, we select a $\mathbf{\beta}$\textbf{-VAE} \citep{higgins_beta-vae_2017}, a \textbf{FactorVAE} \citep{kim2018factorvae} and a $\beta$\textbf{-TC VAE} \citep{chen2019vae}. 
To stress the \emph{influence of the distance} used between distributions we add a Wasserstein Autoencoder (\textbf{WAE})\citep{tolstikhin2018wasserstein} and an \textbf{InfoVAE} \citep{zhao2018vae} with either Inverse Multi-Quadratic (IMQ) or a Radial Basis Function kernel (RBF) together with an Adversarial Autoencoder (\textbf{AAE})\citep{makhzani2016vae}, a \textbf{VAEGAN}\citep{larsen_autoencoding_2015} and a VAE using structural similarity  metric for reconstruction (\textbf{MSSSIM-VAE}) \citep{snell2017vae}. 
Finally, we add a \textbf{VQVAE} \citep{oord2018vae} since having a discrete latent space has shown to yield promising results. 
Models implemented in Pythae requiring too much training time or more intricate  hyper-parameter tuning were excluded from the benchmarks. 

In the following, we will distinguish \emph{AE-based} (autoencoder-based) methods (AE, RAE, WAE and VQVAE) from the other \emph{variational-based} methods.

\paragraph{Training paradigm} Each of the aforementioned models is equipped with the same neural network architecture for both the encoder and decoder leading to a comparable number of parameters \footnote{Some models may actually have additional parameters in their intrinsic structure \emph{e.g.} a VQVAE learns a dictionary of embeddings, a VAMP learns the pseudo-inputs, a VAE-IAF learns auto-regressive flows. Nonetheless, since we work on images, the number of parameters remains in the same order of magnitude.}. For each task, 10 different configurations are considered for each model, allowing a simple exploration of the models' hyper-parameters, leading to 10 trained models for each dataset and each neural network type (ConvNet or ResNet) leading to a total of 1140 models\footnote{The training setting (curves, configs ...) can be found at \url{https://wandb.ai/benchmark_team/trainings} while detailed experimental set-up is available in Appendix.~\ref{appC}.}. It is important to note that the hyper-parameter exploration is not exhaustive and models sensitive to hyper-parameter tuning may have better performances with a more extensive parameter search. The sets of hyper-parameters explored are detailed in Appendix.~\ref{appD} for each model.

\subsection{Experiments}
In this section, we present the main results observed on 5 downstream tasks.

\subsubsection{Fixed latent dimension}

In this first part, latent dimensions are set to 16, 256 and 64 for the MNIST, CIFAR10 and CELEBA datasets respectively, as we observed those latent dimensions to lead to good performances. 
See results per model and across the 10 configurations specified in Appendix.~\ref{appD} to assess the influence of the parameters on the tasks. 

\paragraph{Task 1: Image reconstruction}

For each model, reconstruction error is evaluated by selecting the configuration minimising the Mean Square Error (MSE) between the input and the output of the model on the validation set, while results are shown on the test set\footnote{See the whole results at \url{https://wandb.ai/benchmark_team/reconstructions}}.
We show in Table.~\ref{tab:reconstuction} the MSE and Frechet Inception Distance\footnote{We used the implementation of \url{https://github.com/bioinf-jku/TTUR}} (FID) \citep{heusel_gans_2017} of the reconstructions from this model on the test set. 
It is important to note that using the MSE as a metric places models using different reconstruction losses (VAEGAN and MSSSIM-VAE) at a disadvantage. 

As expected, the autoencoder-based models seem to perform best for the reconstruction task. Nonetheless, this experiment also shows the interest of adding regularisation to the autoencoder since improvements over the AE (RAE-GP, RAE-L2) achieve better performance than the regular AE.
Moreover, $\beta$-VAE type models demonstrate their versatility as small enough $\beta$ values can lead to less regularisation, therefore favouring a better reconstruction.

\paragraph{Task 2: Image generation} 
% Multiple sampler choices are chosen to generate new data:
% \begin{enumerate}
%     \item A simple distribution chosen as $\mathcal{N}(0,I_d)$ for variational approaches ($\mathcal{N}$).
%     \item fitting a 10 components mixture of Gaussian in the latent space of the post training as proposed in \citep{ghosh_variational_2020} (GMM).
%     \item fitting a normalising flow taken as a Masked Autoregressive Flow (MAF) with two-layer MADE \citep{germain2015made}.
%     \item fitting a VAE in a similar fashion as \citep{dai_diagnosing_2018}.
% \end{enumerate}
We consider an image generation task with the trained models. In this experiment, we also explore different ways of sampling new data, either 1) using a simple distribution chosen as $\mathcal{N}(0,I_d)$ and corresponding to the standard prior for variational approaches ($\mathcal{N}$); 2) fitting a 10 components mixture of Gaussian in the latent space post training as proposed in \citep{ghosh_variational_2020} (GMM), 3) fitting a normalising flow taken as a Masked Autoregressive Flow (MAF) \citep{papamakarios2017masked} or 4) fitting a VAE in a similar fashion as \citep{dai_diagnosing_2018}. 
For the MAF, two-layer MADE \citep{germain2015made} are used. For each sampler, we select the models achieving the lowest FID on the validation set and compute the Inception Score\footnote{We used the implementation of \url{https://github.com/openai/improved-gan}} (IS) \citep{salimans2016improved} and the FID on the test set\footnote{See the whole results at \url{https://wandb.ai/benchmark_team/generations}}. 
It should be noted that although the use of IS and FID has been criticised \citep{barratt2018note, shmelkov_how_2018, chong2020effectively, morozov2020self, jung2021internalized}, we still choose to use those metrics for clarity's sake as they are within the most commonly used metrics for image generation on generative models.
The main results are shown in Table.~\ref{tab:generation} for the normal and GMM sampler (see Appendix.~\ref{appD} for the other sampling schemes). 

One of the key findings of this experiment is that performing \emph{ex-post} density (therefore not using the standard Gaussian prior) for the variational approach tends to almost always lead to better generation metrics even when a simple 10-components mixture of Gaussian is used. Interestingly, we note that when a more advanced density estimation model such as a MAF is used, results appear equivalent to those of the GMM (see Appendix.~\ref{appD}). This may be due to the simplicity of the database we used and in consequence of the distribution of the latent codes that can be approximated well enough with a GMM. It should nonetheless be noted that the number of components in the GMM remains a key parameter which was set to the number of classes for MNIST and CIFAR10 since it is known, however too high a value may lead to overfitting while a low one may lead to worse results. %For this task the adversarial approach proposed in the VAEGAN clearly surpasses peers on MNIST and CELEBA with impressive results when compared to other models. Nonetheless, fitting such a model may be challenging due to the adversarial approach and results can be considerably affected if parameters are not chosen correctly (see Appendix.~D).

\paragraph{Task 3: Classification} 
To measure the meaningfulness of the learned latent representations we perform a simple classification task with a single layer classifier as proposed in \citep{coates2011analysis}. The rationale behind this is that if a GAE succeeds in learning a disentangled latent representation a simple linear classifier should perform well \citep{berthelot2018understanding}. A single layer classifier is trained in a supervised manner on the latent embeddings of MNIST and CIFAR10. The train/val/test split used is the same as for the autoencoder training. For each model configuration, we perform 20 runs of the classifier on the latent embeddings and define the best hyper-parameter configuration as the one achieving the highest mean accuracy on these 20 runs on the validation set. We report the mean accuracy on the test set across the 20 runs for the selected configuration in Table.~\ref{tab:classif_cluster} (left)\footnote{See the whole results at \url{https://wandb.ai/benchmark_team/classifications}}. 

As expected, models explicitly encouraging disentanglement in the latent space such as the $\beta$-VAE and $\beta$-TC VAE achieve better classification when compared to a standard VAE. Nonetheless, AE-based models seem again the best suited for such a task since variational approaches tend to enforce a continuous space, consequently bringing latent representations of different classes closer to each other. As a general observation, we can state that models with a more flexible prior achieve better results on this task.

%The complete results are available at \url{}

\paragraph{Task 4: Clustering} As a complement to the previous task, performing clustering directly in the latent space of the trained autoencoders can give insights on the quality of the latent representation. Indeed, a well defined latent space will maintain the separation of the classes inherent to the datasets, leading to easy and stable $k$-means performances.
To do so, we propose to fit 100 separate runs of the $k$-means algorithm and we show the mean accuracy obtained on the train embeddings in Table.~\ref{tab:classif_cluster} (right)\footnote{See the whole results at \url{https://wandb.ai/benchmark_team/clustering}}. This experiment allows us to explore and measure the \emph{clusterability} of the generated latent spaces \citep{berthelot2018understanding}. To measure accuracy we assign the label of the most prevalent class to each cluster. 

The conclusions of this experiment are slightly different from the previous one since models targeting disentanglement seem to be equalled by the original VAE.
Interestingly, adversarial approaches and other alternatives to the standard VAE KL regularisation method seem to achieve the best results.

\paragraph{Task 5: Interpolation} Finally, we propose to assess the ability of the model to perform meaningful interpolations. For this task, we consider a starting and ending image in the test set of MNIST and CIFAR10 and perform a linear interpolation in the generated latent spaces between the two encoded images. We show in Appendix.~\ref{appB} the decoding along the interpolation curves. For this task, no metric was found relevant since the notion of "good" interpolation can be disputable. Nonetheless, the obtained interpolated images can be reconstructed and qualitatively evaluated. 

For this task, variational approaches were found to obtain better results as the inherent structure the posterior distribution imposes in the latent space results in a "smoother" transition from one image to another  when compared to autoencoders that mainly superpose images, especially in higher dimensional latent spaces.

\subsubsection{Varying latent dimension}

\begin{figure}[ht]
    \centering
    \includegraphics[width=\linewidth]{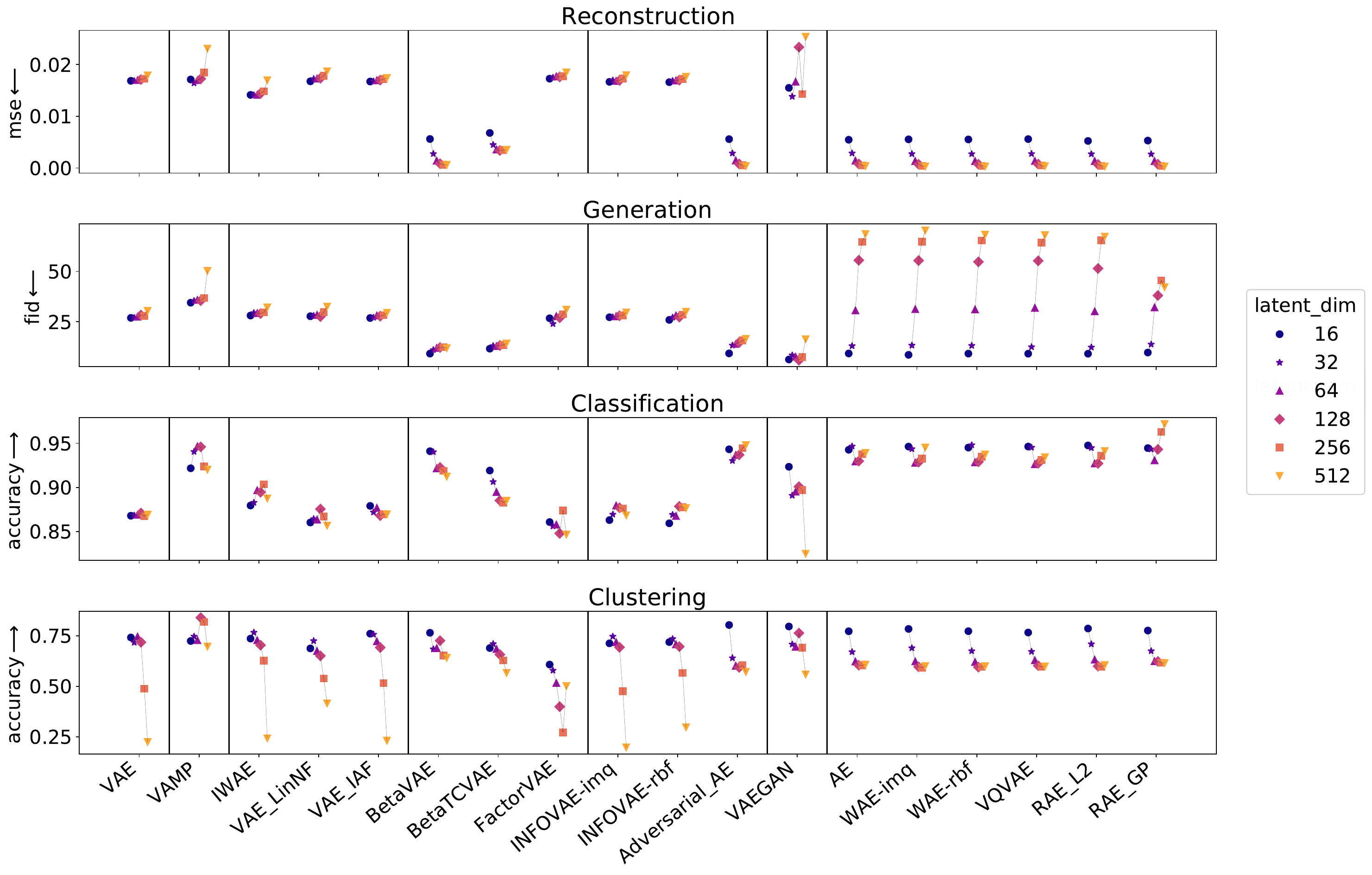}
    \caption{\emph{From top to bottom:} Evolution of the reconstruction MSE, generation FID, classification accuracy and clustering accuracy with respect to the latent space dimension on the MNIST dataset.}
    \label{fig:benchmarking_results}
\end{figure}

An important parameter of autoencoder models which is too often neglected  in the literature is the dimension of the latent space. We now propose to keep the same configurations as previously but re-evaluate Tasks 1 to 5 with the latent space varying in the range $[16,32,64,128,256,512]$. Results are shown in Fig.~\ref{fig:benchmarking_results} for MNIST and a ConvNet\footnote{MSSSIM-VAE was removed from this plot for visualisation purposes.} (see Appendix.~\ref{appD} for CIFAR, ResNet and interpolations). For the generation task, we select the sampler with lowest FID on the validation set.  

%In the previous experiments this parameter was kept fixed, and we propose in this section to discuss the influence of this parameter on Tasks 1 to 4 for all models. For this experiment we keep the same configurations as previously, but only consider the MNIST and CIFAR10 datasets and consider latent dimensions of 16, 32, 64, 128, 256 and 512. 
%Again, we consider a set of 10 hyper-parameter configurations for each autoencoder (unchanged from the previous section) with 2 different neural networks architectures (ConvNet and ResNet). The models are trained with the same setting as explained previously, leading to 4800 trained models. 
\paragraph{Assessing the influence of the latent dimension} A clear difference in behaviour is exhibited between variational-based and AE-based methods.
For each given task, AEs share a common trend with respect to the evolution of the latent dimension: a common optimal latent dimension within the range $[16,32,64,128,256,512]$  is found for each task, but differs drastically among different tasks (\emph{e.g.} 512 for reconstruction, 16 for generation, either 16 or 512 for classification and 512 for clustering with the MNIST dataset). This suggests the existence of a common intra-group optimal latent space dimension for a given task. 
In addition, we observe that $\beta$-VAE type methods (with the right hyper-parameter choice) can exhibit similar behaviours to AE models. 
% with regards to reconstruction and generation
The same observation can be made for variational-based methods, where it is interesting to note that although lower performances are achieved, the apparent optimal latent dimension varies less with respect to the choice of the task. Therefore, a latent dimension of 16 to 32 appears to be the optimal choice for all 4 Tasks on the MNIST dataset, and 32 to 128 on the CIFAR10 dataset. % AE MORE SENSIBLE TO LATENT DIM CHOIE For AE based methods, the choice of the latent dimension seems to have a measurable impact on its performance, and the optimal choice varies drastically from one dataset to another.
It should be noted that unsupervised tasks such as clustering of the latent representation of the CIFAR10 dataset are hard and models are expected to perform poorly, leading to less interpretable results.

%It should also be noted that these 2 groups have different behaviours in terms of stability to the latent dimension choice: while GAEs are only sensitive to the latent dimension on the clustering task, AEs are sensitive to the latent dimension for reconstruction, generation and interpolation.

\begin{table}[p]
    \centering
    \tiny
    \caption{Mean Squared Error ($10^{-3}$) and FID (lower is better) computed with 10k samples on the test set. For each model, the best configuration is the one achieving the lowest MSE on the validation set.}
    \label{tab:reconstuction}
    \begin{tabular}{ccccccc|cccccc}
    \toprule
        \multirow{3}{*}{Model} & \multicolumn{6}{c|}{ConvNet}& \multicolumn{6}{c}{ResNet} \\
        %\cmidrule(r){4-7}
        &\multicolumn{2}{c}{MNIST (16)} &  \multicolumn{2}{c}{CIFAR10 (256)} & \multicolumn{2}{c|}{CELEBA (64)} & \multicolumn{2}{c}{MNIST (16)} & \multicolumn{2}{c}{CIFAR10 (256)} & \multicolumn{2}{c}{CELEBA (64)} \\
         & MSE $\downarrow$  & FID $\downarrow $ & MSE $\downarrow$ & FID $\downarrow$  & MSE $\downarrow$  & FID $\downarrow$ & MSE $\downarrow$ & FID $\downarrow$ & MSE $\downarrow$ & FID $\downarrow$ & MSE $\downarrow$  & FID $\downarrow$  \\
        \midrule
        VAE & 16.85 & 30.71 & 16.24 & 218.66 & 9.83 & 49.22 & 17.24 & 36.06 & 16.33 & 176.63 & 10.59 & 58.75 \\
        \midrule
        VAMP  & 24.17 & 44.95 & 17.45 & 221.40 & 10.81 & 51.64 & 17.11 & 37.58 & 16.87 & 177.03 & 11.50 & 60.89 \\
        \midrule
        IWAE & 14.14 & 34.28 & 16.19 & 237.14 & 9.47 & 50.00 & 15.79 & 38.74 & 16.02 & 183.37 & 10.14 & 60.18 \\
        VAE-lin-NF & 16.75 & 31.14 & 16.57 & 221.39 & 9.90 & 49.84 & 17.23 & 36.74 & 16.59 & 177.08 & 10.68 & 58.73 \\
        VAE-IAF & 16.71 & 30.64 & 16.33 & 223.65 & 9.87 & 50.05 & 17.05 & 35.98 & 16.39 & 177.05 & 10.63 & 58.41 \\
        \midrule
        $\beta$-VAE & 5.61 & 10.55 & 3.60 & 50.55 & 7.28 & \textbf{46.96} &  5.87 & 15.81 & 2.40 & 55.67 & 7.78 & 51.59 \\
        %Dis $\beta$-VAE & 53.34 & 195.71 & 20.69 & 290.17 & 14.95 & 93.39 & 39.73 & 98.60 & 12.95 & 189.97 & 12.45 & 80.38 \\
        $\beta$-TC VAE & 6.78 & 14.11 & 5.06 & 53.49 & 7.65 & 50.82  & 7.12 & 18.44 & 4.05 & 66.89 & 8.08 & 52.70 \\
        Factor VAE & 17.27 & 30.39 & 16.41 & 224.3 & 10.16 & 53.61 & 18.13 & 37.97 & 16.55 & 176.8 & 10.93 & 59.46  \\
        \midrule
        InfoVAE-IMQ & 16.65 & 30.62 & 16.19 & 216.44 & 9.81 & 50.51 & 17.17 & 37.33 & 16.32 & 173.79 & 10.63 & 58.04 \\
        InfoVAE-RBF  & 16.59 & 30.63 & 16.23 & 217.52 & 9.85 & 50.14 & 17.01 & 37.04 & 16.32 & 175.37 & 10.64 & 58.68\\
        AAE & 5.59 & 10.87 & 2.60 & 40.66 & 7.25 & 50.22 & 5.98 & 17.01 & \textbf{2.33} & 55.93 & 7.76 & 50.97\\
        \midrule
        MSSSIM-VAE  & 32.60 & 37.91 & 39.42 & 276.70 & 35.60 & 124.52 & 33.67 & 40.25 & 39.61 & 254.34 & 35.43 & 119.92 \\
        VAEGAN & 15.49 & \textbf{5.54} & 31.40 & 289.35 & 8.91 & 86.58 & 23.25 & \textbf{11.35} & 30.22 & 300.07 & 9.32 & 86.32 \\
        \midrule
        AE & 5.47 & 11.61 & 2.82 & 41.98 & 7.03 & 51.08 & 6.13 & 13.74 & 2.34 & 55.43 & 7.74 & 50.54 \\
        WAE-IMQ  & 5.55 & 11.29 & 2.81 & 41.79& 7.04 & 52.11 & 5.78 & 16.21 &  2.34 & 56.55 & 7.74 & 50.50 \\
        WAE-RBF & 5.53 & 11.34 & 2.82 & 42.21 & 7.03 & 51.43 & 5.80 & 16.14 & 2.34 & 56.00 & 7.74 & 51.38 \\
        VQVAE & 5.59 & 11.02 & 2.84 & 44.60 & 7.06 & 52.27 & 6.00 & 15.27 & 2.34 & \textbf{55.84} & \textbf{7.73} & \textbf{50.29} \\
        RAE-L2 & \textbf{5.24} & 15.37 & \textbf{2.25} & 49.28 & \textbf{6.90} & 53.98 & \textbf{5.76} & 17.27 &  2.35 & 57.85 & 7.74 & 51.07 \\
        RAE-GP   & 5.31 & 12.08 & 2.81 & \textbf{41.15} & 7.06 & 51.85  & 5.83 & 15.69 & 2.34 & 56.71 & 7.76 & 51.36 \\
    \bottomrule
    \end{tabular}
\end{table}

\begin{table}[p]
    \centering
    \tiny
    \caption{\emph{Left:} Mean test accuracy of a single layer classifier on the embedding obtained in the latent spaces of each model average on 20 runs. \emph{Right:} Mean accuracy of 100 $k$-means fitted on the training embeddings coming from the autoencoders.}
    \label{tab:classif_cluster}
    \begin{tabular}{ccc|cc||cc|cc}
    \toprule
    \multirow{4}{*}{Model} & \multicolumn{4}{c||}{Classification} & \multicolumn{4}{c}{Clustering} \\
    \cmidrule{2-9}
    & \multicolumn{2}{c|}{ConvNet} & \multicolumn{2}{c||}{ResNet} & \multicolumn{2}{c|}{ConvNet} & \multicolumn{2}{c}{ResNet}\\
        %\cmidrule(r){4-7}
        &MNIST &  CIFAR10& MNIST& CIFAR10 & MNIST &  CIFAR10& MNIST& CIFAR10\\
        \midrule
        VAE & 86.75 (0.05) & 32.61 (0.03) & 86.80 (0.03) & 32.37 (0.03) & 69.71 (2.01) & 17.18 (0.68) & 74.21 (0.97) & 18.12 (0.74) \\
        \midrule
        VAMP & 92.17 (0.02) & 33.46 (0.17) & 92.58 (0.04) & 33.03 (0.22)  & 67.26 (1.25) & 24.03 (0.24) & 72.48 (0.96) & 23.35 (0.11) \\
        \midrule
        IWAE & 87.96 (0.04) & 31.86 (0.04) & 88.18 (0.03) & 32.26 (0.04) & 63.93 (1.73) & 19.55 (0.67) & 73.66 (2.30) & 18.44 (0.86)\\
        VAE-lin-NF & 86.04 (0.04) & 31.57 (0.02) & 85.85 (0.05) & 31.74 (0.03) & 65.48 (2.76) & 17.09 (0.64) & 68.80 (3.65) & 18.74 (0.68) \\
        VAE-IAF & 88.32 (0.02) & 33.52 (0.02) & 87.91 (0.02) & 32.41 (0.02) & 75.31 (1.69) & 17.81 (0.73) & 76.11 (2.15) & 18.42 (0.66)\\
        \midrule        %Dis $\beta$-VAE & 35.52 (0.08) & 28.25 (0.04) & 58.02 (0.07) & 31.93 (0.07) & 12.74 (0.13) & 10.77 (0.09) & 11.51 (0.06) & 10.84 (0.09) \\
        $\beta$-TC VAE & 90.96 (0.02) & 45.40 (0.05) & 91.91 (0.02) & \textbf{42.17 (0.07)} & 65.68 (0.91) & 24.14 (0.65) & 68.98 (2.67) & \textbf{25.57 (0.61)} \\
        Factor VAE & 86.08 (0.06) & 31.38 (0.04) & 83.44 (0.05) & 31.76 (0.04) & 51.02 (1.73) & 15.77 (0.60) & 60.79 (2.06) & 17.56 (0.68) \\
        \midrule
        InfoVAE-IMQ & 86.33 (0.04) & 32.48 (0.02) & 86.31 (0.06) & 32.10 (0.05)& 68.17 (2.34) & 16.65 (0.80) & 71.31 (2.62) & 18.10 (0.79)\\
        InfoVAE-RBF & 85.94 (0.03) & 32.50 (0.03) & 86.12 (0.04) & 31.67 (0.03) & 66.02 (1.14) & 16.22 (0.69) & 71.93 (1.91) & 18.61 (0.67) \\
        AAE & 93.28 (0.03) & 43.93 (0.07) & 94.31 (0.03) & 40.62 (0.11) & 74.19 (3.22) & 24.72 (0.75) & \textbf{80.41 (2.09)} & 24.76 (0.53) \\
        \midrule
        MSSSIM-VAE & 78.30 (0.03) & 20.26 (0.06) & 76.54 (0.03) & 20.24 (0.04) & 49.33 (1.32) & 11.70 (0.19) & 48.58 (1.37) & 11.70 (0.17) \\
        VAEGAN  & 92.34 (0.02) & 26.56 (0.04) & 90.31 (0.03) & 29.90 (0.03)& \textbf{77.29 (1.19)} & 17.20 (0.45) & 79.67 (0.90) & 22.23 (0.44) \\
        \midrule
        AE & 93.81 (0.02) & 42.15 (0.07) & 94.26 (0.03) & 40.47 (0.13) & 73.55 (0.60) & 23.19 (0.52) & 77.30 (0.84) & 23.18 (0.37) \\
        WAE-IMQ & 93.60 (0.02) & \textbf{45.89 (0.07)} & 94.62 (0.03) & 41.35 (0.03) & 72.33 (2.92) & 23.81 (0.61) & 78.46 (3.48) & 25.09 (0.82) \\
        WAE-RBF  & 93.72 (0.02) & 43.38 (0.08) & 94.51 (0.02) & 40.63 (0.08)& 74.20 (1.94) & 23.70 (0.71) & 77.33 (1.92) & 24.66 (0.63) \\
        VQVAE & 93.45 (0.02) & 42.89 (0.07) & \textbf{94.63 (0.04)} & 40.40 (0.09)  & 72.61 (0.40) & 23.85 (0.48) & 76.68 (2.36) & 23.68 (0.37) \\
        RAE-L2 & 94.75 (0.01) & 42.76 (0.08) & 94.43 (0.03) & 40.22 (0.05) & 74.07 (0.36) & 23.77 (0.54) & 78.66 (0.29) & 24.84 (0.73) \\
        RAE-GP & \textbf{94.10 (0.02)} & 43.66 (0.07) & 94.45 (0.02) & 40.93 (0.14) & 72.88 (0.52) & \textbf{24.84 (0.53)} & 77.66 (1.29) & 23.86 (0.32)\\
    \bottomrule
    \end{tabular}
\end{table}

\newcommand{\gc}{\cellcolor{gray!25}}
\newcommand{\CC}[1]{\cellcolor{blue!#1}}

\begin{table}[ht]
    \centering
    \tiny
    \caption{Inception Score (higher is better) and FID (lower is better) computed with 10k samples on the test set. For each model and sampler we report the results obtained by the model achieving the lowest FID score on the validation set.}
    \label{tab:generation}
    \begin{tabular}{c|c|cccccc|cccccc}
    \toprule
        \multirow{3}{*}{Model} & \multirow{3}{*}{Sampler} & \multicolumn{6}{c|}{ConvNet}& \multicolumn{6}{c}{ResNet} \\
        %\cmidrule(r){4-7}
        &&\multicolumn{2}{c}{MNIST} & \multicolumn{2}{c}{CIFAR10}& \multicolumn{2}{c|}{CELEBA} & \multicolumn{2}{c}{MNIST}& \multicolumn{2}{c}{CIFAR10}& \multicolumn{2}{c}{CELEBA} \\
        & & FID $\downarrow$  & IS $\uparrow$  & FID & IS  & FID  & IS $\uparrow$ & FID $\downarrow$  & IS $\uparrow$  & FID & IS  & FID  & IS  \\
        \midrule
        \multirow{2}{*}{VAE} & $\mathcal{N}$ & 28.5 & 2.1 & 241.0 & 2.2 & 54.8 & 1.9 & 31.3 & 2.0 & 181.7 & 2.5 & 66.6 & 1.6  \\
                            & \gc GMM & \gc 26.9 & \gc 2.1 & \gc 235.9 & \gc 2.3 & \gc 52.4 & \gc 1.9 & \gc 32.3 & \gc 2.1 & \gc 179.7 & \gc 2.5 & \gc 63.0 & \gc 1.7  \\
                            %& VAE & 40.3 & 2.0 & 337.5 & 1.7 & 70.9 & 1.6 & 48.7 &  1.8 & 358.0 & 1.3 & 76.4 & 1.4 \\
                   %          & MAF & 26.8 & 2.1 & 239.5 & 2.2 & 52.5 & 2.0 & 31.0 &  2.1 & 181.5 & 2.5 & 62.9 & 1.7 \\
                            % & IAF Sampler \\
        \midrule
        \multirow{1}{*}{VAMP} & VAMP & 64.2 & 2.0 & 329.0 & 1.5 & 56.0 & 1.9 & 34.5 &  2.1 & 181.9 & 2.5 & 67.2 & 1.6  \\
        \midrule
        \multirow{2}{*}{IWAE} & $\mathcal{N}$ &29.0 & 2.1 & 245.3 & 2.1 & 55.7 & 1.9 & 32.4 &  2.0 & 191.2 & 2.4 & 67.6 & 1.6  \\
                             &\gc GMM & \gc 28.4 & \gc 2.1 & \gc 241.2 & \gc 2.1 & \gc 52.7 & \gc 1.9 & \gc 34.4 & \gc 2.1 & \gc 188.8 & \gc 2.4 & \gc 64.1 & \gc 1.7 \\
                             %& VAE & 42.4 & 2.0 & 346.6 & 1.5 & 74.3 & 1.5 & 50.1 &  1.9 & 364.8 & 1.2 & 76.4 & 1.4  \\
                   %          & MAF & 28.1 & 2.1 & 243.4 & 2.1 & 52.7 & 1.9 & 32.5 &  2.1 & 190.4 & 2.4 & 64.3 & 1.7  \\
                            % & IAF Sampler \\
        %\midrule
        \multirow{2}{*}{VAE-lin NF} & $\mathcal{N}$ &  29.3 & 2.1 & 240.3 & 2.1 & 56.5 & 1.9 & 32.5 &  2.0 & 185.5 & 2.4 & 67.1 & 1.6 \\
                             & \gc GMM & \gc 28.4 & \gc 2.1 & \gc 237.0 & \gc 2.2 & \gc 53.3 & \gc 1.9 & \gc 33.1 &  \gc 2.1 & \gc 183.1 & \gc 2.5 & \gc 62.8 & \gc 1.7 \\
                             %& VAE & 40.1 & 2.0 & 311.0 & 1.6 & 71.1 & 1.6 & 49.7 &  1.9 & 296.2 & 1.7 & 75.6 & 1.4  \\
                   %          & MAF & 27.7 & 2.1 & 239.1 & 2.1 & 53.4 & 2.0 & 32.4 &  2.0 & 184.2 & 2.5 & 62.7 & 1.7 \\
                            % & IAF Sampler \\
        %\midrule
        \multirow{2}{*}{VAE-IAF} & $\mathcal{N}$ & 27.5 & 2.1 & 236.0 & 2.2 & 55.4 & 1.9 & 30.6 &  2.0 & 183.6 & 2.5 & 66.2 & 1.6  \\
                             & \gc GMM & \gc 27.0 & \gc 2.1 & \gc 235.4 & \gc 2.2 & \gc 53.6 & \gc 1.9 & \gc 32.2 &  \gc 2.1 & \gc 180.8 & \gc 2.5 & \gc 62.7 & \gc 1.7  \\
                            % & VAE & 39.4 & 2.0 & 330.5 & 1.1 & 73.0 & 1.5 & 44.8 &  1.9 & 322.7 & 1.5 & 76.7 & 1.4  \\
                   %          & MAF & 26.9 & 2.1 & 236.8 & 2.2 & 53.6 & 1.9 & 30.6 &  2.1 & 182.5 & 2.5 & 63.0 & 1.7 \\
                            % & IAF Sampler \\
        \midrule
        \multirow{2}{*}{$\beta$-VAE} & $\mathcal{N}$ & 21.4 & 2.1 & 115.4 & 3.6 & 56.1 & 1.9 & 19.1 &  2.0 & 124.9 & 3.4 & 65.9 & 1.6  \\
                             & \gc GMM & \gc 9.2 & \gc 2.2 & \gc 92.2 & \gc 3.9 & \gc 51.7 & \gc 1.9 & \gc 11.4 & \gc 2.1 & \gc 112.6 & \gc 3.6 & \gc 59.3 & \gc 1.7 \\
                            %  & VAE & 14.0 & 2.2 & 139.6 & 3.6 & 55.0 & 1.9 & 20.3 &  2.1 & 152.5 & 3.5 & 61.5 & 1.7 \\
                   %          & MAF & 9.5 & 2.2 & 100.9 & 3.5 & 51.5 & 2.0 & 12.0 &  2.1 & 120.0 & 3.6 & 59.7 & 1.8 \\
                            % & IAF Sampler \\ 
        %\midrule
        %  \multirow{4}{*}{Dis $\beta$-VAE} & $\mathcal{N}(0,1)$ & 96.5 & 2.3 & 219.4 & 3.6 & 130.8 & 1.6 & 109.4 &  2.7 & 209.8 & 3.2 & 110.6 & 1.5 \\
        %                       & GMM &  192.7 & 2.3 & 300.8 & 1.8 & 94.0 & 1.5 & 98.7 &  2.1 & 202.3 & 2.6 & 83.2 & 1.6 \\
        %                       & $2$-s sampler & 236.1 & 1.5 & 371.2 & 1.0 & 167.9 & 1.0 & 250.8 &  1.1 & 349.0 & 1.2 & 161.0 & 1.2  \\
        %                       & MAF sampler & 191.8 & 2.2 & 300.6 & 1.8 & 94.3 & 1.5 & 98.3 &  2.1 & 200.6 & 2.6 & 82.6 & 1.7  \\
        % %                     % & IAF Sampler \\
        %  \midrule
        \multirow{2}{*}{$\beta$-TC VAE} & $\mathcal{N}$ & 21.3 & 2.1 & 116.6 & 2.8 & 55.7 & 1.8 & 20.7 &  2.0 & 125.8 & 3.4 & 65.9 & 1.6  \\
                             & \gc GMM & \gc 11.6 & \gc 2.2 & \gc 89.3 & \gc 4.1 & \gc 51.8 & \gc 1.9 & \gc 13.3 & \gc 2.1 & \gc \textbf{106.5} & \gc \textbf{3.7} & \gc 59.3 & \gc 1.7  \\
                            % & VAE & 18.4 & 2.2 & 127.9 & 4.2 & 59.7 & 1.8 & 28.3 &  2.0 & 164.0 & 3.3 & 66.4 & 1.5 \\
                   %          & MAF & 12.0 & 2.2 & 95.6 & 3.6 & 52.2 & 1.9 & 13.7 &  2.1 & 116.6 & 3.4 & 60.1 & 1.7  \\
                            % & IAF Sampler \\
        %\midrule
        \multirow{2}{*}{FactorVAE} & $\mathcal{N}$ & 27.0 & 2.1 & 236.5 & 2.2 & 53.8 & 1.9 & 31.0 &  2.0 & 185.4 & 2.5 & 66.4 & 1.7 \\
                             & \gc GMM & \gc 26.9 & \gc 2.1 & \gc 234.0 & \gc 2.2 & \gc 52.4 & \gc 2.0 & \gc 32.7 & \gc 2.1 & \gc 184.4 & \gc 2.5 & \gc 63.3 & \gc 1.7 \\
                            % & VAE & 41.2 & 1.9 & 338.3 & 1.5 & 75.0 & 1.5 & 54.7 &  1.8 & 316.2 & 1.3 & 77.7 & 1.4 \\
                   %          & MAF & 26.7 & 2.2 & 236.7 & 2.2 & 52.7 & 1.9 & 32.8 &  2.1 & 185.8 & 2.5 & 63.4 & 1.7 \\
                            % & IAF Sampler \\
        \midrule
        \multirow{2}{*}{InfoVAE - RBF} & $\mathcal{N}$ & 27.5 & 2.1 & 235.2 & 2.1 & 55.5 & 1.9 & 31.1 &  2.0 & 182.8 & 2.5 & 66.5 & 1.6  \\
                             & \gc GMM  & \gc 26.7  & \gc 2.1  & \gc 230.4  & \gc 2.2  & \gc 52.7  & \gc 1.9  & \gc 32.3  & \gc  2.1  & \gc 179.5  & \gc 2.5  & \gc 62.8  & \gc 1.7 \\
                            % & VAE & 39.7 & 2.0 & 327.2 & 1.5 & 73.7 & 1.5 & 50.6 &  1.9 & 363.4 & 1.2 & 75.8 & 1.4  \\
                   %          & MAF & 25.9 & 2.1 & 233.5 & 2.2 & 52.2 & 2.0 & 30.5 &  2.1 & 181.3 & 2.5 & 62.7 & 1.7  \\
                            % & IAF Sampler \\
        %\midrule
        \multirow{2}{*}{InfoVAE - IMQ} & $\mathcal{N}$ & 28.3 & 2.1 & 233.8 & 2.2 & 56.7 & 1.9 & 31.0 &  2.0 & 182.4 & 2.5 & 66.4 & 1.6  \\
                             & \gc GMM & \gc 27.7 & \gc 2.1 & \gc 231.9 & \gc 2.2 & \gc 53.7 & \gc 1.9 & \gc 32.8 & \gc 2.1 & \gc 180.7 & \gc 2.6 & \gc 62.3 & \gc 1.7 \\
                            % & VAE & 40.4 & 1.9 & 323.8 & 1.6 & 73.7 & 1.5 & 49.9 &  1.9 & 341.8 & 1.8 & 75.7 & 1.4  \\
                   %          & MAF & 27.2 & 2.1 & 232.3 & 2.1 & 53.8 & 2.0 & 30.6 &  2.1 & 182.5 & 2.5 & 62.6 & 1.7  \\
                            % & IAF Sampler \\
        %\midrule
        \multirow{2}{*}{AAE} & $\mathcal{N}$ & 16.8 & 2.2 & 139.9 & 2.6 & 59.9 & 1.8 & 19.1 &  2.1 & 164.9 & 2.4 & 64.8 & 1.7  \\
                              & \gc GMM  & \gc 9.3  & \gc 2.2  & \gc 92.1  & \gc 3.8  & \gc 53.9  & \gc 2.0  & \gc 11.1  & \gc  2.1  & \gc 118.5  & \gc 3.5  & \gc 58.7  & \gc 1.8 \\
                            % & VAE & 13.4 & 2.2 & 144.0 & 3.4 & 58.2 & 1.8 & 15.1 &  2.1 & 145.2 & 3.6 & 59.0 & 1.7 \\
                   %          & MAF & 9.3 & 2.2 & 101.1 & 3.2 & 53.8 & 2.0 & 11.9 &  2.1 & 133.6 & 3.1 & 59.2 & 1.8  \\
                            % & IAF Sampler \\
        \midrule
        \multirow{2}{*}{MSSSIM-VAE} & $\mathcal{N}$ & 26.7 & 2.2 & 279.9 & 1.7 & 124.3 & 1.3 & 28.0 &  2.1 & 254.2 & 1.7 & 119.0 & 1.3 \\
                              & \gc GMM  & \gc 27.2  & \gc 2.2  & \gc 279.7  & \gc 1.7  & \gc 124.3  & \gc 1.3  & \gc 28.8  & \gc  2.1  & \gc 253.1  & \gc 1.7  & \gc 119.2  & \gc 1.3 \\
                   %          & VAE & 51.2 & 1.9 & 355.5 & 1.1 & 137.9 & 1.2 & 51.6 &  1.9 & 372.1 & 1.1 & 136.5 & 1.2  \\
                    %         & MAF & 26.9 & 2.2 & 279.8 & 1.7 & 124.0 & 1.3 & 27.5 &  2.1 & 254.1 & 1.7 & 119.5 & 1.3  \\
                            % & IAF Sampler \\
        %\midrule
        \multirow{2}{*}{VAEGAN} & $\mathcal{N}$ & 8.7 & 2.2 & 199.5 & 2.2 & 39.7 & 1.9 & 12.8 &  2.2 & 198.7 & 2.2 & 122.8 & 2.0  \\
                              & \gc GMM  & \gc \textbf{6.3}  & \gc \textbf{2.2}  & \gc 197.5  & \gc 2.1  & \gc \textbf{35.6}  & \gc 1.8  & \gc \textbf{6.5}  & \gc  2.2  & \gc 188.2  & \gc 2.6  & \gc 84.3  & \gc 1.7  \\
                            % & VAE & 11.2 & 2.1 & 310.9 & 2.0 & 54.5 & 1.6 & 9.2 &  2.1 & 272.7 & 2.0 & 88.8 & 1.6  \\
                   %          & MAF & 6.9 & 2.3 & 199.0 & 2.1 & 36.7 & 1.8 & 6.6 & \textbf{ 2.2} & 191.9 & 2.5 & 84.8 & 1.7 \\
                            % & IAF Sampler \\
        \midrule
        \multirow{2}{*}{AE}  & $\mathcal{N}$ & 26.7 & 2.1 & 201.3 & 2.1 & 327.7 & 1.0 & 221.8 &  1.3 & 210.1 & 2.1 & 275.0 & 2.9  \\
                               & \gc GMM  & \gc 9.3  & \gc 2.2  & \gc 97.3  & \gc 3.6  & \gc 55.4  & \gc 2.0  & \gc 11.0  & \gc  2.1  & \gc 120.7  & \gc 3.4  & \gc \textbf{57.4}  & \gc 1.8 \\
                    %2         & MAF &  9.9 & 2.2 & 108.3 & 3.1 & 55.7 & 2.0 & 12.0 &  2.1 & 136.5 & 3.0 & 58.3 & 1.8 \\
                            % & IAF Sampler \\
        %\midrule
        \multirow{2}{*}{WAE - RBF} & $\mathcal{N}$ & 21.2 & 2.2 & 175.1 & 2.0 & 332.6 & 1.0 & 21.2 &  2.1 & 170.2 & 2.3 & 69.4 & 1.6  \\
                              & \gc GMM  & \gc 9.2  & \gc 2.2  & \gc 97.1  & \gc 3.6  & \gc 55.0  & \gc 2.0  & \gc 11.2  & \gc  2.1  & \gc 120.3  & \gc 3.4  & \gc 58.3  & \gc 1.7 \\
                   %          & MAF & 9.8 & 2.2 & 108.2 & 3.1 & 56.0 & 2.0 & 11.8 &  2.2 & 135.3 & 3.0 & 58.3 & 1.8  \\
        %\midrule
        \multirow{2}{*}{WAE - IMQ} & $\mathcal{N}$ & 18.9 & 2.2 & 164.4 & 2.2 & 64.6 & 1.7 & 20.3 &  2.1 & 150.7 & 2.5 & 67.1 & 1.6 \\
                              & \gc GMM  & \gc 8.6  & \gc 2.2  & \gc 96.5  & \gc 3.6  & \gc 51.7  & \gc 2.0  & \gc 11.2  & \gc  2.1  & \gc 119.0  & \gc 3.5  & \gc 57.7  & \gc 1.8 \\
                   %          & MAF &  9.5 & 2.2 & 107.8 & 3.1 & 51.6 & 2.0 & 11.8 &  2.1 & 130.2 & 3.0 & 58.7 & 1.7  \\
        %\midrule
        \multirow{1}{*}{VQVAE} & $\mathcal{N}$ & 28.2 & 2.0 & 152.2 & 2.0 & 306.9 & 1.0 & 170.7 &  1.6 & 195.7 & 1.9 & 140.3 & \textbf{2.2}  \\
          & \gc  GMM   & \gc 9.1   & \gc 2.2   & \gc 95.2   & \gc 3.7   & \gc 51.6   & \gc 2.0   & \gc 10.7   & \gc  2.1   & \gc  120.1   & \gc 3.4   & \gc 57.9   & \gc 1.8  \\
                   %          & MAF & 9.6 & 2.2 & 104.7 & 3.2 & 52.3 & 1.9 & 11.7 &  2.2 & 136.8 & 3.0 & 57.9 & 1.8 \\
        %\midrule
        \multirow{1}{*}{RAE-L2} & $\mathcal{N}$ & 25.0 & 2.0 & 156.1 & 2.6 & 86.1 & 2.8 & 63.3 &  2.2 & 170.9 & 2.2 & 168.7 & 3.1  \\
                              & \gc GMM  & \gc 9.1  & \gc 2.2  & \gc \textbf{85.3}  & \gc \textbf{3.9}  & \gc 55.2  & \gc 1.9  & \gc 11.5  & \gc  2.1  & \gc 122.5  & \gc 3.4  & \gc 58.3  & \gc 1.8 \\
                   %          & MAF & 9.5 & 2.2 & 93.4 & 3.5 & 55.2 & 2.0 & 12.3 &  2.2 & 136.6 & 3.0 & 59.1 & 1.7  \\
                            % & IAF Sampler \\
        %\midrule
        \multirow{1}{*}{RAE - GP} & $\mathcal{N}$ &  27.1 & 2.1 & 196.8 & 2.1 & 86.1 & \textbf{
2.4} & 61.5 &  2.2 & 229.1 & 2.0 & 201.9 & 3.1  \\
                               & \gc GMM   & \gc 9.7   & \gc  2.2   & \gc 96.3   & \gc  3.7   & \gc  52.5   & \gc  1.9   & \gc 11.4   & \gc 2.1  & \gc 123.3   & \gc 3.4   & \gc 59.0  & \gc 1.8   \\
                   %          & MAF & 9.7 & 2.2 & 106.3 & 3.2 & 52.5 & 1.9 & 12.2 &  2.2 & 139.4 & 3.0 & 59.5 & 1.8  \\

    \bottomrule
    \end{tabular}
\end{table}

\section{Conclusion}

In this paper, we introduce \textbf{Pythae}, a new open-source Python library unifying common and state-of-the-art Generative AutoEncoder (GAE) implementations, allowing reliable and reproducible model training, data generation and experiment tracking. This library was designed as an open model testing environment driven by the community, wherein peers are encouraged to contribute by adding their own models, and by doing so favour reproducible research and accessibility to ready-to-use GAE models. As an illustration of the capabilities of Pythae, we perform a benchmarking of 19 generative autoencoder models on 5 downstream tasks (image reconstruction, generation, classification, clustering and interpolation) leading to some interesting findings on the general behaviours of generative autoencoder models. We hope that the library will continue to be adopted by the community and expand thanks to the increasing number of contributions.
%First, this benchmark shows that as far as generation is concerned, performing \emph{ex-post} density estimation in the latent space of the autoencoders to balance the potential poor expressiveness of the prior distribution tends to almost always improve in terms of the FID and IS metrics. Interestingly, results were either equivalent or better when fitting a simple mixture of Gaussian than considering more fancy normalizing flows models or another VAE to estimate the actual distribution of the latent code. As for reconstruction, it appears that best performing models are autoencoder based models with some regularization. 

%\subsection{Image generation}

\newcommand{\cb}{\cellcolor{blue!25}}

% Define min mid and max values for color gradient
\newcommand*{\MinNumber}{1.0}%
\newcommand*{\MidNumber}{50.0} %
\newcommand*{\MaxNumber}{100.0}%

%Apply the gradient macro for 3 colors black, red, blue (as an example)
\newcommand{\ApplyGradientBlack}[1]{%
    \IfDecimal{#1}{%
      \pgfmathsetmacro{\PercentColor}{100.0*(#1-1)/(3 - 1)}
      \textcolor{black!\PercentColor}{#1}
    }{\textbf{#1}}% else it's not a decimal
}
\newcommand{\ApplyGradientFID}[1]{%
    \IfDecimal{#1}{%
      \pgfmathsetmacro{\PercentColor}{100.0*(#1-0)/(100 - 0)}
      \textcolor{black!\PercentColor}{#1}
    }{\textbf{#1}}% else it's not a decimal
}
\newcommand{\ApplyGradientRed}[1]{%
    \IfDecimal{#1}{%
      \pgfmathsetmacro{\PercentColor}{100.0*(#1-\MinNumber)/(\MaxNumber-\MinNumber)}
      \textcolor{red!\PercentColor}{#1}
    }{\textbf{#1}}% else it's not a decimal
}
\newcommand{\ApplyGradientBlue}[1]{%
    \IfDecimal{#1}{%
      \pgfmathsetmacro{\PercentColor}{100.0*(#1-0)/(300-0)}
      \textcolor{blue!\PercentColor}{#1}
    }{\textbf{#1}}% else it's not a decimal
}

%create new tabular column type based on color
\newcolumntype{K}{>{\collectcell\ApplyGradientBlack}m{0.3cm}<{\endcollectcell}}
\newcolumntype{R}{>{\collectcell\ApplyGradientRed}m{0.3cm}<{\endcollectcell}}
\newcolumntype{B}{>{\collectcell\ApplyGradientBlue}m{0.3cm}<{\endcollectcell}}
\newcolumntype{F}{>{\collectcell\ApplyGradientBlackFID}m{0.5cm}<{\endcollectcell}}

\begin{ack}
The research leading to these results has received funding from the French government under management of Agence Nationale de la Recherche as part of the ``Investissements d'avenir'' program, reference ANR-19-P3IA-0001 (PRAIRIE 3IA Institute) and reference ANR-10-IAIHU-06 (Agence Nationale de la Recherche-10-IA Institut Hospitalo-Universitaire-6). 
This work was granted access to the HPC
resources of IDRIS under the allocation AD011013517 made by
GENCI (Grand Equipement National de Calcul Intensif).
\end{ack}
\clearpage
\begin{small}
\bibliographystyle{plainnat}
\bibliography{references}

\begin{thebibliography}{72}
\providecommand{\natexlab}[1]{#1}
\providecommand{\url}[1]{\texttt{#1}}
\expandafter\ifx\csname urlstyle\endcsname\relax
  \providecommand{\doi}[1]{doi: #1}\else
  \providecommand{\doi}{doi: \begingroup \urlstyle{rm}\Url}\fi

\bibitem[Alemi et~al.(2017)Alemi, Fischer, Dillon, and Murphy]{alemi_deep_2016}
Alexander~A Alemi, Ian Fischer, Joshua~V Dillon, and Kevin Murphy.
\newblock Deep variational information bottleneck.
\newblock In \emph{International Conference on Learning Representations}, 2017.

\bibitem[Aneja et~al.(2020)Aneja, Schwing, Kautz, and
  Vahdat]{aneja_ncp-vae_2020}
Jyoti Aneja, Alexander Schwing, Jan Kautz, and Arash Vahdat.
\newblock {NCP}-{VAE}: Variational autoencoders with noise contrastive priors.
\newblock \emph{{arXiv}:2010.02917 [cs, stat]}, 2020.

\bibitem[Arvanitidis et~al.(2018)Arvanitidis, Hansen, and
  Hauberg]{arvanitidis_latent_2018}
Georgios Arvanitidis, Lars~Kai Hansen, and S{\"o}ren Hauberg.
\newblock Latent space oddity: On the curvature of deep generative models.
\newblock In \emph{6th International Conference on Learning Representations,
  ICLR 2018}, 2018.

\bibitem[Barratt and Sharma(2018)]{barratt2018note}
Shane Barratt and Rishi Sharma.
\newblock A note on the inception score.
\newblock \emph{arXiv preprint arXiv:1801.01973}, 2018.

\bibitem[Bauer and Mnih(2019{\natexlab{a}})]{bauer_resampled_2019}
Matthias Bauer and Andriy Mnih.
\newblock Resampled priors for variational autoencoders.
\newblock In \emph{The 22nd International Conference on Artificial Intelligence
  and Statistics}, pages 66--75. PMLR, 2019{\natexlab{a}}.

\bibitem[Bauer and Mnih(2019{\natexlab{b}})]{bauer_resampled_2019-1}
Matthias Bauer and Andriy Mnih.
\newblock Resampled priors for variational autoencoders.
\newblock pages 66--75. {PMLR}, 2019{\natexlab{b}}.
\newblock ISBN 2640-3498.

\bibitem[Berthelot* et~al.(2019)Berthelot*, Raffel*, Roy, and
  Goodfellow]{berthelot2018understanding}
David Berthelot*, Colin Raffel*, Aurko Roy, and Ian Goodfellow.
\newblock Understanding and improving interpolation in autoencoders via an
  adversarial regularizer.
\newblock In \emph{International Conference on Learning Representations}, 2019.
\newblock URL \url{https://openreview.net/forum?id=S1fQSiCcYm}.

\bibitem[Biewald(2020)]{wandb}
Lukas Biewald.
\newblock Experiment tracking with weights and biases, 2020.
\newblock URL \url{https://www.wandb.com/}.
\newblock Software available from wandb.com.

\bibitem[Bisong(2019)]{Bisong2019}
Ekaba Bisong.
\newblock \emph{Google Colaboratory}, pages 59--64.
\newblock Apress, Berkeley, CA, 2019.
\newblock ISBN 978-1-4842-4470-8.
\newblock \doi{10.1007/978-1-4842-4470-8_7}.
\newblock URL \url{https://doi.org/10.1007/978-1-4842-4470-8_7}.

\bibitem[Blaauw and Bonada(2016)]{blaauw_modeling_2016}
Merlijn Blaauw and Jordi Bonada.
\newblock Modeling and transforming speech using variational autoencoders.
\newblock \emph{Morgan N, editor. Interspeech 2016; 2016 Sep 8-12; San
  Francisco, CA.[place unknown]: ISCA; 2016. p. 1770-4.}, 2016.
\newblock Publisher: International Speech Communication Association (ISCA).

\bibitem[Burda et~al.(2016)Burda, Grosse, and
  Salakhutdinov]{burda_importance_2016}
Yuri Burda, Roger Grosse, and Ruslan Salakhutdinov.
\newblock Importance weighted autoencoders.
\newblock \emph{{arXiv}:1509.00519 [cs, stat]}, 2016.

\bibitem[Burgess(2018)]{burgess2018vae}
Christopher P. et~al. Burgess.
\newblock Understanding disentangling in $\beta$-vae.
\newblock \emph{arXiv preprint arXiv:1804.03599}, 2018.

\bibitem[Caterini et~al.(2018)Caterini, Doucet, and
  Sejdinovic]{caterini_hamiltonian_2018}
Anthony~L Caterini, Arnaud Doucet, and Dino Sejdinovic.
\newblock Hamiltonian variational auto-encoder.
\newblock In \emph{Advances in Neural Information Processing Systems}, pages
  8167--8177, 2018.

\bibitem[Chadebec et~al.(2021)Chadebec, Thibeau-Sutre, Burgos, and
  Allassonnière]{chadebec_data_2021}
Clément Chadebec, Elina Thibeau-Sutre, Ninon Burgos, and Stéphanie
  Allassonnière.
\newblock Data {Augmentation} in {High} {Dimensional} {Low} {Sample} {Size}
  {Setting} {Using} a {Geometry}-{Based} {Variational} {Autoencoder}.
\newblock \emph{arXiv preprint arXiv:2105.00026}, 2021.

\bibitem[Chen et~al.(2018{\natexlab{a}})Chen, Klushyn, Kurle, Jiang, Bayer, and
  Smagt]{chen_metrics_2018}
Nutan Chen, Alexej Klushyn, Richard Kurle, Xueyan Jiang, Justin Bayer, and
  Patrick Smagt.
\newblock Metrics for deep generative models.
\newblock In \emph{International Conference on Artificial Intelligence and
  Statistics}, pages 1540--1550. PMLR, 2018{\natexlab{a}}.

\bibitem[Chen et~al.(2018{\natexlab{b}})Chen, Li, Grosse, and
  Duvenaud]{chen2019vae}
Ricky~TQ Chen, Xuechen Li, Roger~B Grosse, and David~K Duvenaud.
\newblock Isolating sources of disentanglement in variational autoencoders.
\newblock \emph{Advances in neural information processing systems}, 31,
  2018{\natexlab{b}}.

\bibitem[Chen et~al.(2016)Chen, Kingma, Salimans, Duan, Dhariwal, Schulman,
  Sutskever, and Abbeel]{chen_variational_2016}
Xi~Chen, Diederik~P Kingma, Tim Salimans, Yan Duan, Prafulla Dhariwal, John
  Schulman, Ilya Sutskever, and Pieter Abbeel.
\newblock Variational lossy autoencoder.
\newblock \emph{{arXiv} preprint {arXiv}:1611.02731}, 2016.

\bibitem[Chong and Forsyth(2020)]{chong2020effectively}
Min~Jin Chong and David Forsyth.
\newblock Effectively unbiased fid and inception score and where to find them.
\newblock In \emph{Proceedings of the IEEE/CVF conference on computer vision
  and pattern recognition}, pages 6070--6079, 2020.

\bibitem[Coates et~al.(2011)Coates, Ng, and Lee]{coates2011analysis}
Adam Coates, Andrew Ng, and Honglak Lee.
\newblock An analysis of single-layer networks in unsupervised feature
  learning.
\newblock In \emph{Proceedings of the fourteenth international conference on
  artificial intelligence and statistics}, pages 215--223. JMLR Workshop and
  Conference Proceedings, 2011.

\bibitem[Connor et~al.(2021)Connor, Canal, and Rozell]{connor2021variational}
Marissa Connor, Gregory Canal, and Christopher Rozell.
\newblock Variational autoencoder with learned latent structure.
\newblock In \emph{International Conference on Artificial Intelligence and
  Statistics}, pages 2359--2367. PMLR, 2021.

\bibitem[Cremer et~al.(2018)Cremer, Li, and Duvenaud]{cremer_inference_2018}
Chris Cremer, Xuechen Li, and David Duvenaud.
\newblock Inference suboptimality in variational autoencoders.
\newblock In \emph{International Conference on Machine Learning}, pages
  1078--1086. PMLR, 2018.

\bibitem[Dai and Wipf(2018)]{dai_diagnosing_2018}
Bin Dai and David Wipf.
\newblock Diagnosing and {Enhancing} {VAE} {Models}.
\newblock In \emph{International Conference on Learning Representations}, 2018.

\bibitem[Davidson et~al.(2018)Davidson, Falorsi, De~Cao, Kipf, and
  Tomczak]{davidson_hyperspherical_2018}
Tim~R Davidson, Luca Falorsi, Nicola De~Cao, Thomas Kipf, and Jakub~M Tomczak.
\newblock Hyperspherical variational auto-encoders.
\newblock In \emph{34th Conference on Uncertainty in Artificial Intelligence
  2018, UAI 2018}, pages 856--865. Association For Uncertainty in Artificial
  Intelligence (AUAI), 2018.

\bibitem[Dilokthanakul et~al.(2017)Dilokthanakul, Mediano, Garnelo, Lee,
  Salimbeni, Arulkumaran, and Shanahan]{dilokthanakul_deep_2017}
Nat Dilokthanakul, Pedro A.~M. Mediano, Marta Garnelo, Matthew C.~H. Lee, Hugh
  Salimbeni, Kai Arulkumaran, and Murray Shanahan.
\newblock Deep unsupervised clustering with gaussian mixture variational
  autoencoders.
\newblock \emph{{arXiv}:1611.02648 [cs, stat]}, 2017.

\bibitem[Falorsi et~al.(2018)Falorsi, de~Haan, Davidson, De~Cao, Weiler,
  Forré, and Cohen]{falorsi_explorations_2018}
Luca Falorsi, Pim de~Haan, Tim~R. Davidson, Nicola De~Cao, Maurice Weiler,
  Patrick Forré, and Taco~S. Cohen.
\newblock Explorations in homeomorphic variational auto-encoding.
\newblock \emph{{arXiv}:1807.04689 [cs, stat]}, 2018.

\bibitem[Germain et~al.(2015)Germain, Gregor, Murray, and
  Larochelle]{germain2015made}
Mathieu Germain, Karol Gregor, Iain Murray, and Hugo Larochelle.
\newblock Made: Masked autoencoder for distribution estimation.
\newblock In \emph{International Conference on Machine Learning}, pages
  881--889. PMLR, 2015.

\bibitem[Ghosh et~al.(2020)Ghosh, Sajjadi, Vergari, Black, and
  Schölkopf]{ghosh_variational_2020}
Partha Ghosh, Mehdi~SM Sajjadi, Antonio Vergari, Michael Black, and Bernhard
  Schölkopf.
\newblock From variational to deterministic autoencoders.
\newblock In \emph{8th {International} {Conference} on {Learning}
  {Representations}, {ICLR} 2020}, 2020.

\bibitem[Gretton et~al.(2012)Gretton, Borgwardt, Rasch, Sch{\"o}lkopf, and
  Smola]{gretton2012kernel}
Arthur Gretton, Karsten~M Borgwardt, Malte~J Rasch, Bernhard Sch{\"o}lkopf, and
  Alexander Smola.
\newblock A kernel two-sample test.
\newblock \emph{The Journal of Machine Learning Research}, 13\penalty0
  (1):\penalty0 723--773, 2012.

\bibitem[Heusel et~al.(2017)Heusel, Ramsauer, Unterthiner, Nessler, and
  Hochreiter]{heusel_gans_2017}
Martin Heusel, Hubert Ramsauer, Thomas Unterthiner, Bernhard Nessler, and Sepp
  Hochreiter.
\newblock Gans trained by a two time-scale update rule converge to a local nash
  equilibrium.
\newblock In \emph{Advances in Neural Information Processing Systems}, 2017.

\bibitem[Higgins et~al.(2017)Higgins, Matthey, Pal, Burgess, Glorot, Botvinick,
  Mohamed, and Lerchner]{higgins_beta-vae_2017}
Irina Higgins, Loic Matthey, Arka Pal, Christopher Burgess, Xavier Glorot,
  Matthew Botvinick, Shakir Mohamed, and Alexander Lerchner.
\newblock beta-{VAE}: Learning basic visual concepts with a constrained
  variational framework.
\newblock \emph{{ICLR}}, 2\penalty0 (5):\penalty0 6, 2017.

\bibitem[Hoffman and Johnson(2016)]{hoffman_elbo_2016}
Matthew~D Hoffman and Matthew~J Johnson.
\newblock Elbo surgery: yet another way to carve up the variational evidence
  lower bound.
\newblock In \emph{Workshop in Advances in Approximate Bayesian Inference,
  {NIPS}}, volume~1, page~2, 2016.

\bibitem[Jordan et~al.(1999)Jordan, Ghahramani, Jaakkola, and
  Saul]{jordan_introduction_1999}
Michael~I Jordan, Zoubin Ghahramani, Tommi~S Jaakkola, and Lawrence~K Saul.
\newblock An introduction to variational methods for graphical models.
\newblock \emph{Machine Learning}, 37\penalty0 (2):\penalty0 183--233, 1999.

\bibitem[Jung and Keuper(2021)]{jung2021internalized}
Steffen Jung and Margret Keuper.
\newblock Internalized biases in fr{\'e}chet inception distance.
\newblock In \emph{NeurIPS 2021 Workshop on Distribution Shifts: Connecting
  Methods and Applications}, 2021.

\bibitem[Kalatzis et~al.(2020)Kalatzis, Eklund, Arvanitidis, and
  Hauberg]{kalatzis_variational_2020}
Dimitrios Kalatzis, David Eklund, Georgios Arvanitidis, and Soren Hauberg.
\newblock Variational autoencoders with riemannian brownian motion priors.
\newblock In \emph{International Conference on Machine Learning}, pages
  5053--5066. PMLR, 2020.

\bibitem[Kim and Mnih(2018)]{kim2018factorvae}
Hyunjik Kim and Andriy Mnih.
\newblock Disentangling by factorising.
\newblock In \emph{International Conference on Machine Learning}, pages
  2649--2658. PMLR, 2018.

\bibitem[Kingma and Ba(2014)]{kingma_adam_2014}
Diederik~P Kingma and Jimmy Ba.
\newblock Adam: A method for stochastic optimization.
\newblock \emph{{arXiv} preprint {arXiv}:1412.6980}, 2014.

\bibitem[Kingma and Welling(2014)]{kingma_auto-encoding_2014}
Diederik~P. Kingma and Max Welling.
\newblock Auto-encoding variational bayes.
\newblock \emph{{arXiv}:1312.6114 [cs, stat]}, 2014.

\bibitem[Kingma et~al.(2016)Kingma, Salimans, Jozefowicz, Chen, Sutskever, and
  Welling]{kingma_improved_2016}
Durk~P. Kingma, Tim Salimans, Rafal Jozefowicz, Xi~Chen, Ilya Sutskever, and
  Max Welling.
\newblock Improved variational inference with inverse autoregressive flow.
\newblock \emph{Advances in neural information processing systems}, 29, 2016.

\bibitem[Klushyn et~al.(2019)Klushyn, Chen, Kurle, and
  Cseke]{klushyn_learning_2019}
Alexej Klushyn, Nutan Chen, Richard Kurle, and Botond Cseke.
\newblock Learning {Hierarchical} {Priors} in {VAEs}.
\newblock \emph{Advances in neural information processing systems}, page~10,
  2019.

\bibitem[Krizhevsky et~al.(2009)Krizhevsky, Hinton,
  et~al.]{krizhevsky2009learning}
Alex Krizhevsky, Geoffrey Hinton, et~al.
\newblock Learning multiple layers of features from tiny images.
\newblock 2009.

\bibitem[Larsen et~al.(2016)Larsen, S{\o}nderby, Larochelle, and
  Winther]{larsen_autoencoding_2015}
Anders Boesen~Lindbo Larsen, S{\o}ren~Kaae S{\o}nderby, Hugo Larochelle, and
  Ole Winther.
\newblock Autoencoding beyond pixels using a learned similarity metric.
\newblock In \emph{International conference on machine learning}, pages
  1558--1566. PMLR, 2016.

\bibitem[{LeCun}(1998)]{lecun_mnist_1998}
Yann {LeCun}.
\newblock The {MNIST} database of handwritten digits.
\newblock 1998.

\bibitem[Liu et~al.(2015)Liu, Luo, Wang, and Tang]{liu2015faceattributes}
Ziwei Liu, Ping Luo, Xiaogang Wang, and Xiaoou Tang.
\newblock Deep learning face attributes in the wild.
\newblock In \emph{Proceedings of International Conference on Computer Vision
  (ICCV)}, December 2015.

\bibitem[Makhzani(2016)]{makhzani2016vae}
Alireza et~al. Makhzani.
\newblock Adversarial autoencoders.
\newblock \emph{arXiv preprint arXiv:1511.05644}, 2016.

\bibitem[Mathieu et~al.(2019{\natexlab{a}})Mathieu, Le~Lan, Maddison, Tomioka,
  and Teh]{mathieu_continuous_2019}
Emile Mathieu, Charline Le~Lan, Chris~J Maddison, Ryota Tomioka, and Yee~Whye
  Teh.
\newblock Continuous hierarchical representations with poincaré variational
  auto-encoders.
\newblock In \emph{Advances in neural information processing systems}, pages
  12565--12576, 2019{\natexlab{a}}.

\bibitem[Mathieu et~al.(2019{\natexlab{b}})Mathieu, Rainforth, Siddharth, and
  Teh]{mathieu2019disentangling}
Emile Mathieu, Tom Rainforth, Nana Siddharth, and Yee~Whye Teh.
\newblock Disentangling disentanglement in variational autoencoders.
\newblock In \emph{International Conference on Machine Learning}, pages
  4402--4412. PMLR, 2019{\natexlab{b}}.

\bibitem[Morozov et~al.(2020)Morozov, Voynov, and Babenko]{morozov2020self}
Stanislav Morozov, Andrey Voynov, and Artem Babenko.
\newblock On self-supervised image representations for gan evaluation.
\newblock In \emph{International Conference on Learning Representations}, 2020.

\bibitem[Nalisnick et~al.(2016)Nalisnick, Hertel, and
  Smyth]{nalisnick_approximate_2016}
Eric Nalisnick, Lars Hertel, and Padhraic Smyth.
\newblock Approximate inference for deep latent gaussian mixtures.
\newblock In \emph{NIPS Workshop on Bayesian Deep Learning}, volume~2, page
  131, 2016.

\bibitem[Neal(2005)]{neal_hamiltonian_2005}
Radford~M Neal.
\newblock Hamiltonian importance sampling.
\newblock In \emph{talk presented at the Banff International Research Station
  ({BIRS}) workshop on Mathematical Issues in Molecular Dynamics}, 2005.

\bibitem[Pang et~al.(2020)Pang, Han, Nijkamp, Zhu, and Wu]{pang_learning_2020}
Bo~Pang, Tian Han, Erik Nijkamp, Song-Chun Zhu, and Ying~Nian Wu.
\newblock Learning latent space energy-based prior model.
\newblock \emph{Advances in Neural Information Processing Systems}, 33, 2020.

\bibitem[Papamakarios et~al.(2017)Papamakarios, Pavlakou, and
  Murray]{papamakarios2017masked}
George Papamakarios, Theo Pavlakou, and Iain Murray.
\newblock Masked autoregressive flow for density estimation.
\newblock \emph{Advances in neural information processing systems}, 30, 2017.

\bibitem[Paszke et~al.(2017)Paszke, Gross, Chintala, Chanan, Yang, {DeVito},
  Lin, Desmaison, Antiga, and Lerer]{paszke_automatic_2017}
Adam Paszke, Sam Gross, Soumith Chintala, Gregory Chanan, Edward Yang, Zachary
  {DeVito}, Zeming Lin, Alban Desmaison, Luca Antiga, and Adam Lerer.
\newblock Automatic differentiation in pytorch.
\newblock 2017.

\bibitem[Pedregosa et~al.(2011)Pedregosa, Varoquaux, Gramfort, Michel, Thirion,
  Grisel, Blondel, Prettenhofer, Weiss, Dubourg, Vanderplas, Passos,
  Cournapeau, Brucher, Perrot, and Duchesnay]{pedregosa_scikit-learn_2011}
F.~Pedregosa, G.~Varoquaux, A.~Gramfort, V.~Michel, B.~Thirion, O.~Grisel,
  M.~Blondel, P.~Prettenhofer, R.~Weiss, V.~Dubourg, J.~Vanderplas, A.~Passos,
  D.~Cournapeau, M.~Brucher, M.~Perrot, and E.~Duchesnay.
\newblock Scikit-learn: Machine learning in python.
\newblock \emph{Journal of Machine Learning Research}, 12:\penalty0 2825--2830,
  2011.

\bibitem[Razavi et~al.(2020)Razavi, Oord, and Vinyals]{razavi_generating_2019}
Ali Razavi, Aaron van~den Oord, and Oriol Vinyals.
\newblock Generating diverse high-fidelity images with vq-vae-2.
\newblock \emph{Advances in Neural Information Processing Systems}, 2020.

\bibitem[Rezende and Mohamed(2015)]{rezende_variational_2015}
Danilo Rezende and Shakir Mohamed.
\newblock Variational inference with normalizing flows.
\newblock In \emph{International Conference on Machine Learning}, pages
  1530--1538. PMLR, 2015.

\bibitem[Rezende et~al.(2014)Rezende, Mohamed, and
  Wierstra]{rezende_stochastic_2014}
Danilo~Jimenez Rezende, Shakir Mohamed, and Daan Wierstra.
\newblock Stochastic backpropagation and approximate inference in deep
  generative models.
\newblock In \emph{International conference on machine learning}, pages
  1278--1286. PMLR, 2014.

\bibitem[Salimans et~al.(2015)Salimans, Kingma, and
  Welling]{salimans_markov_2015}
Tim Salimans, Diederik Kingma, and Max Welling.
\newblock Markov chain monte carlo and variational inference: Bridging the gap.
\newblock In \emph{International Conference on Machine Learning}, pages
  1218--1226, 2015.

\bibitem[Salimans et~al.(2016)Salimans, Goodfellow, Zaremba, Cheung, Radford,
  and Chen]{salimans2016improved}
Tim Salimans, Ian Goodfellow, Wojciech Zaremba, Vicki Cheung, Alec Radford, and
  Xi~Chen.
\newblock Improved techniques for training gans.
\newblock \emph{Advances in neural information processing systems}, 29, 2016.

\bibitem[Shao et~al.(2018)Shao, Kumar, and Fletcher]{shao_riemannian_2018}
Hang Shao, Abhishek Kumar, and P.~Thomas Fletcher.
\newblock The riemannian geometry of deep generative models.
\newblock In \emph{2018 {IEEE}/{CVF} Conference on Computer Vision and Pattern
  Recognition Workshops ({CVPRW})}, pages 428--4288. {IEEE}, 2018.
\newblock ISBN 978-1-5386-6100-0.
\newblock \doi{10.1109/CVPRW.2018.00071}.

\bibitem[Shmelkov et~al.(2018)Shmelkov, Schmid, and Alahari]{shmelkov_how_2018}
Konstantin Shmelkov, Cordelia Schmid, and Karteek Alahari.
\newblock How good is my gan?
\newblock In \emph{Proceedings of the European Conference on Computer Vision
  (ECCV)}, pages 213--229, 2018.

\bibitem[Snell et~al.(2017)Snell, Ridgeway, Liao, Roads, Mozer, and
  Zemel]{snell2017vae}
Jake Snell, Karl Ridgeway, Renjie Liao, Brett~D Roads, Michael~C Mozer, and
  Richard~S Zemel.
\newblock Learning to generate images with perceptual similarity metrics.
\newblock In \emph{2017 IEEE International Conference on Image Processing
  (ICIP)}, pages 4277--4281. IEEE, 2017.

\bibitem[S{\o}nderby et~al.(2016)S{\o}nderby, Raiko, Maal{\o}e, S{\o}nderby,
  and Winther]{sonderby_ladder_2016}
Casper~Kaae S{\o}nderby, Tapani Raiko, Lars Maal{\o}e, S{\o}ren~Kaae
  S{\o}nderby, and Ole Winther.
\newblock Ladder variational autoencoder.
\newblock In \emph{29th Annual Conference on Neural Information Processing
  Systems (NIPS 2016)}, 2016.

\bibitem[Subramanian(2020)]{Subramanian2020}
A.K Subramanian.
\newblock Pytorch-vae.
\newblock \url{https://github.com/AntixK/PyTorch-VAE}, 2020.

\bibitem[Tolstikhin et~al.(2018)Tolstikhin, Bousquet, Gelly, and
  Sch{\"o}lkopf]{tolstikhin2018wasserstein}
I~Tolstikhin, O~Bousquet, S~Gelly, and B~Sch{\"o}lkopf.
\newblock Wasserstein auto-encoders.
\newblock In \emph{6th International Conference on Learning Representations
  (ICLR 2018)}, 2018.

\bibitem[Tomczak and Welling(2018)]{tomczak_vae_2018}
Jakub Tomczak and Max Welling.
\newblock Vae with a vampprior.
\newblock In \emph{International Conference on Artificial Intelligence and
  Statistics}, pages 1214--1223. PMLR, 2018.

\bibitem[Van Den~Oord et~al.(2017)Van Den~Oord, Vinyals, et~al.]{oord2018vae}
Aaron Van Den~Oord, Oriol Vinyals, et~al.
\newblock Neural discrete representation learning.
\newblock \emph{Advances in neural information processing systems}, 30, 2017.

\bibitem[Wang et~al.(2003)Wang, Simoncelli, and Bovik]{wang2003multiscale}
Zhou Wang, Eero~P Simoncelli, and Alan~C Bovik.
\newblock Multiscale structural similarity for image quality assessment.
\newblock In \emph{The Thrity-Seventh Asilomar Conference on Signals, Systems
  \& Computers, 2003}, volume~2, pages 1398--1402. Ieee, 2003.

\bibitem[Wang et~al.(2004)Wang, Bovik, Sheikh, and Simoncelli]{wang2004image}
Zhou Wang, Alan~C Bovik, Hamid~R Sheikh, and Eero~P Simoncelli.
\newblock Image quality assessment: from error visibility to structural
  similarity.
\newblock \emph{IEEE transactions on image processing}, 13\penalty0
  (4):\penalty0 600--612, 2004.

\bibitem[Wolf et~al.(2020)Wolf, Debut, Sanh, Chaumond, Delangue, Moi, Cistac,
  Rault, Louf, Funtowicz, Davison, Shleifer, von Platen, Ma, Jernite, Plu, Xu,
  Scao, Gugger, Drame, Lhoest, and Rush]{wolf-etal-2020-transformers}
Thomas Wolf, Lysandre Debut, Victor Sanh, Julien Chaumond, Clement Delangue,
  Anthony Moi, Pierric Cistac, Tim Rault, Rémi Louf, Morgan Funtowicz, Joe
  Davison, Sam Shleifer, Patrick von Platen, Clara Ma, Yacine Jernite, Julien
  Plu, Canwen Xu, Teven~Le Scao, Sylvain Gugger, Mariama Drame, Quentin Lhoest,
  and Alexander~M. Rush.
\newblock Transformers: State-of-the-art natural language processing.
\newblock In \emph{Proceedings of the 2020 Conference on Empirical Methods in
  Natural Language Processing: System Demonstrations}, pages 38--45, Online,
  October 2020. Association for Computational Linguistics.
\newblock URL \url{https://www.aclweb.org/anthology/2020.emnlp-demos.6}.

\bibitem[Yang et~al.(2019)Yang, Cheung, Li, and Fang]{yang_deep_2019}
Linxiao Yang, Ngai-Man Cheung, Jiaying Li, and Jun Fang.
\newblock Deep clustering by gaussian mixture variational autoencoders with
  graph embedding.
\newblock In \emph{Proceedings of the IEEE/CVF International Conference on
  Computer Vision}, pages 6440--6449, 2019.

\bibitem[Zhang et~al.(2018)Zhang, Bütepage, Kjellström, and
  Mandt]{zhang_advances_2018}
Cheng Zhang, Judith Bütepage, Hedvig Kjellström, and Stephan Mandt.
\newblock Advances in variational inference.
\newblock \emph{{IEEE} Transactions on Pattern Analysis and Machine
  Intelligence}, 41\penalty0 (8):\penalty0 2008--2026, 2018.

\bibitem[Zhao et~al.(2016)Zhao, Song, and Ermon]{zhao2018vae}
Shengjia Zhao, Jiaming Song, and Stefano Ermon.
\newblock Infovae: Information maximizing variational autoencoders.
\newblock \emph{arXiv preprint arXiv:1706.02262}, 2016.

\end{thebibliography}

\end{small}

\clearpage
\appendix

\section{Usage of Pythae}\label{appA}

In this section we illustrate through simple examples how to use \textbf{Pythae} pipelines. The library is documented\footnote{\url{https://pythae.readthedocs.io/en/latest/?badge=latest}} and also available on pypi\footnote{\url{https://pypi.org/project/pythae/}} allowing a wider use and easier integration in other codes. All of the implementations proposed in the library are adaptations of the official code when available and allowed by the licence. If not, the method is re-implemented. Table.~\ref{implemented-vaes} lists all the implemented models as of June 2022.

\begin{enumerate}
    \item \textbf{Training configuration} Before launching a model training, one must specify the training configuration that should be used. This can be done easily by instantiating a \textbf{BaseTrainerConfig} instance taking as input all the hyper-parameters related to the training (number of training epochs, learning rate to apply...). See the full documentation for additional arguments that can be passed to the \textbf{BaseTrainerConfig}.
    \begin{Verbatim}[commandchars=\\\{\},frame=lines,
        framesep=2mm,
        baselinestretch=1.2,
        fontsize=\footnotesize,
        linenos]
        \PYG{k+kn}{from} \PYG{n+nn}{pythae.trainers} \PYG{k+kn}{import} \PYG{n}{BaseTrainerConfig}
        \PYG{c+c1}{\PYGZsh{} Set up the model configuration}
        \PYG{n}{my\PYGZus{}training\PYGZus{}config} \PYG{o}{=} \PYG{n}{BaseTrainerConfig}\PYG{p}{(}
            \PYG{n}{output\PYGZus{}dir}\PYG{o}{=}\PYG{l+s+s1}{\PYGZsq{}my\PYGZus{}model\PYGZsq{}}\PYG{p}{,}
            \PYG{n}{num\PYGZus{}epochs}\PYG{o}{=}\PYG{l+m+mi}{50}\PYG{p}{,}
            \PYG{n}{learning\PYGZus{}rate}\PYG{o}{=}\PYG{l+m+mf}{1e\PYGZhy{}3}\PYG{p}{,}
            \PYG{n}{batch\PYGZus{}size}\PYG{o}{=}\PYG{l+m+mi}{200}\PYG{p}{)}
\end{Verbatim}
        % \begin{minted}[
        % frame=lines,
        % framesep=2mm,
        % baselinestretch=1.2,
        % fontsize=\footnotesize,
        % linenos
        % ]{python}
        % from pythae.trainers import BaseTrainerConfig
        % # Set up the model configuration
        % my_training_config = BaseTrainerConfig(
        %     output_dir='my_model',
        %     num_epochs=50,
        %     learning_rate=1e-3,
        %     batch_size=200)
        % \end{minted}
        \clearpage
    \item \textbf{Model configuration} Similarly to the TrainerConfig, the model can then be instantiated with the model configuration specifying any hyper-parameters relevant to the model. Note that each model has its own configuration with specific hyper-parameters. See the online documentation for more details.
    \begin{Verbatim}[commandchars=\\\{\},frame=lines,
        framesep=2mm,
        baselinestretch=1.2,
        fontsize=\footnotesize,
        linenos]
        \PYG{k+kn}{from} \PYG{n+nn}{pythae.models} \PYG{k+kn}{import} \PYG{n}{BetaVAE}\PYG{p}{,} \PYG{n}{BetaVAEConfig}
        \PYG{c+c1}{\PYGZsh{} Set up the model configuration}
        \PYG{n}{my\PYGZus{}vae\PYGZus{}config} \PYG{o}{=} \PYG{n}{BetaVAEConfig}\PYG{p}{(}
            \PYG{n}{input\PYGZus{}dim}\PYG{o}{=}\PYG{p}{(}\PYG{l+m+mi}{1}\PYG{p}{,} \PYG{l+m+mi}{28}\PYG{p}{,} \PYG{l+m+mi}{28}\PYG{p}{),}
            \PYG{n}{latent\PYGZus{}dim}\PYG{o}{=}\PYG{l+m+mi}{16}\PYG{p}{,}
            \PYG{n}{beta}\PYG{o}{=}\PYG{l+m+mi}{2}\PYG{p}{)}
        \PYG{c+c1}{\PYGZsh{} Build the model}
        \PYG{n}{my\PYGZus{}vae\PYGZus{}model} \PYG{o}{=} \PYG{n}{BetaVAE}\PYG{p}{(}\PYG{n}{model\PYGZus{}config}\PYG{o}{=}\PYG{n}{my\PYGZus{}vae\PYGZus{}config}\PYG{p}{)}
\end{Verbatim}
        % \begin{minted}[
        % frame=lines,
        % framesep=2mm,
        % baselinestretch=1.2,
        % fontsize=\footnotesize,
        % linenos
        % ]{python}
        % from pythae.models import BetaVAE, BetaVAEConfig
        % # Set up the model configuration
        % my_vae_config = BetaVAEConfig(
        %     input_dim=(1, 28, 28),
        %     latent_dim=16,
        %     beta=2)
        % # Build the model
        % my_vae_model = BetaVAE(model_config=my_vae_config)
        % \end{minted}
        
    \item \textbf{Training} A model training can then be launched by simply using the built-in training pipeline in which only the training/evaluation data need to be specified.
     \begin{Verbatim}[commandchars=\\\{\},frame=lines,
        framesep=2mm,
        baselinestretch=1.2,
        fontsize=\footnotesize,
        linenos]
        \PYG{k+kn}{from} \PYG{n+nn}{pythae.pipelines} \PYG{k+kn}{import} \PYG{n}{TrainingPipeline}
        \PYG{n}{pipeline} \PYG{o}{=} \PYG{n}{TrainingPipeline}\PYG{p}{(}
            \PYG{n}{training\PYGZus{}config}\PYG{o}{=}\PYG{n}{my\PYGZus{}training\PYGZus{}config}\PYG{p}{,}
            \PYG{n}{model}\PYG{o}{=}\PYG{n}{my\PYGZus{}vae\PYGZus{}model}\PYG{p}{)}
        \PYG{c+c1}{\PYGZsh{} Launch the Pipeline}
        \PYG{n}{pipeline}\PYG{p}{(}
            \PYG{n}{train\PYGZus{}data}\PYG{o}{=}\PYG{n}{your\PYGZus{}train\PYGZus{}data}\PYG{p}{,} \PYG{c+c1}{\PYGZsh{} arrays or tensors}
            \PYG{n}{eval\PYGZus{}data}\PYG{o}{=}\PYG{n}{your\PYGZus{}eval\PYGZus{}data}\PYG{p}{)} \PYG{c+c1}{\PYGZsh{} arrays or tensors}
\end{Verbatim}
        % \begin{minted}[
        % frame=lines,
        % framesep=2mm,
        % baselinestretch=1.2,
        % fontsize=\footnotesize,
        % linenos
        % ]{python}
        % from pythae.pipelines import TrainingPipeline
        % pipeline = TrainingPipeline(
        %     training_config=my_training_config, 
        %     model=my_vae_model)
        % # Launch the Pipeline
        % pipeline(
        %     train_data=your_train_data, # arrays or tensors
        %     eval_data=your_eval_data) # arrays or tensors
        
        % \end{minted}
        
    \item \textbf{Model reloading} 
    The weights and configuration of the trained model can be reloaded using the \textbf{AutoModel} instance proposed in Pythae.

    \begin{Verbatim}[commandchars=\\\{\},frame=lines,
        framesep=2mm,
        baselinestretch=1.2,
        fontsize=\footnotesize,
        linenos]
        \PYG{k+kn}{from} \PYG{n+nn}{pythae.models} \PYG{k+kn}{import} \PYG{n}{AutoModel}
        \PYG{n}{my\PYGZus{}trained\PYGZus{}vae} \PYG{o}{=} \PYG{n}{AutoModel}\PYG{o}{.}\PYG{n}{load\PYGZus{}from\PYGZus{}folder}\PYG{p}{(}\PYG{l+s+s1}{\PYGZsq{}path/to/trained\PYGZus{}model\PYGZsq{}}\PYG{p}{)}
\end{Verbatim}
    % \begin{minted}[
    %     frame=lines,
    %     framesep=2mm,
    %     baselinestretch=1.2,
    %     fontsize=\footnotesize,
    %     linenos
    %     ]{python}
    %     from pythae.models import AutoModel
    %     my_trained_vae = AutoModel.load_from_folder('path/to/trained_model')
    %     \end{minted}
        
    \item \textbf{Data generation} A data generation pipeline can be instantiated similarly to a model training. The pipeline can then be called with any relevant arguments such as the number of samples to generate or the training and evaluation data that may be needed to fit the sampler.

    \begin{Verbatim}[commandchars=\\\{\},frame=lines,
        framesep=2mm,
        baselinestretch=1.2,
        fontsize=\footnotesize,
        linenos]
        \PYG{k+kn}{from} \PYG{n+nn}{pythae.samplers} \PYG{k+kn}{import} \PYG{n}{GaussianMixtureSamplerConfig}
        \PYG{k+kn}{from} \PYG{n+nn}{pythae.pipelines} \PYG{k+kn}{import} \PYG{n}{GenerationPipeline}
        \PYG{c+c1}{\PYGZsh{} Define your sampler configuration}
        \PYG{n}{gmm\PYGZus{}sampler\PYGZus{}config} \PYG{o}{=} \PYG{n}{GaussianMixtureSamplerConfig}\PYG{p}{(}
            \PYG{n}{n\PYGZus{}components}\PYG{o}{=}\PYG{l+m+mi}{10}\PYG{p}{)}
        \PYG{c+c1}{\PYGZsh{} Build the pipeline}
        \PYG{n}{pipeline} \PYG{o}{=} \PYG{n}{GenerationPipeline}\PYG{p}{(}
            \PYG{n}{model}\PYG{o}{=}\PYG{n}{my\PYGZus{}trained\PYGZus{}vae}\PYG{p}{,}
            \PYG{n}{sampler\PYGZus{}config}\PYG{o}{=}\PYG{n}{gmm\PYGZus{}sampler\PYGZus{}config}\PYG{p}{)}
        \PYG{c+c1}{\PYGZsh{} Launch generation}
        \PYG{n}{generated\PYGZus{}samples} \PYG{o}{=} \PYG{n}{pipeline}\PYG{p}{(}
            \PYG{n}{num\PYGZus{}samples}\PYG{o}{=}\PYG{l+m+mi}{100}\PYG{p}{,}
            \PYG{n}{return\PYGZus{}gen}\PYG{o}{=}\PYG{k+kc}{True}\PYG{p}{,}
            \PYG{n}{train\PYGZus{}data}\PYG{o}{=}\PYG{n}{train\PYGZus{}data}\PYG{p}{,}
            \PYG{n}{eval\PYGZus{}data}\PYG{o}{=}\PYG{k+kc}{None}\PYG{p}{)}
\end{Verbatim}
    
    % \begin{minted}[
    %     frame=lines,
    %     framesep=2mm,
    %     baselinestretch=1.2,
    %     fontsize=\footnotesize,
    %     linenos
    %     ]{python}
    %     from pythae.samplers import GaussianMixtureSamplerConfig
    %     from pythae.pipelines import GenerationPipeline
    %     # Define your sampler configuration
    %     gmm_sampler_config = GaussianMixtureSamplerConfig(
    %         n_components=10)
    %     # Build the pipeline
    %     pipeline = GenerationPipeline(
    %         model=my_trained_vae,
    %         sampler_config=gmm_sampler_config)
    %     # Launch generation
    %     generated_samples = pipeline(
    %         num_samples=100,
    %         return_gen=True,
    %         train_data=train_data,
    %         eval_data=None)
    % \end{minted}
        
\end{enumerate}

 \begin{table}[ht]
 \small
      \caption{List of implemented VAEs}
      \label{implemented-vaes}
      \centering
      \begin{tabular}{lll}
        \toprule
        %\multicolumn{2}{c}{Part}                   \\
        %\cmidrule(r){1-2}
        Name     & Reference \\
        \midrule
        Variational Autoencoder (VAE) & \citet{kingma_auto-encoding_2014}\\
        Beta Variational Autoencoder (BetaVAE) & \citet{higgins_beta-vae_2017}\\
        {VAE with Linear Normalizing Flows} (VAE\_LinNF) & \citet{rezende_variational_2015}\\
        {VAE with Inverse Autoregressive Flows} (VAE\_IAF) & \citet{kingma_improved_2016}\\
        Disentangled $\beta$-VAE (DisentangledBetaVAE)	& \citet{higgins_beta-vae_2017}\\
        Disentangling by Factorising (FactorVAE) & \citet{kim2018factorvae} \\
        {Beta-TC-VAE (BetaTCVAE)}	 & \citet{chen2019vae}\\
        {Importance Weighted Autoencoder} (IWAE) & \citet{burda_importance_2016}\\
        {VAE with perceptual metric similarity (MSSSIM\_VAE)}	& \citet{snell2017vae} \\
        Wasserstein Autoencoder (WAE) & \citet{tolstikhin2018wasserstein}\\
        {Info Variational Autoencoder (INFOVAE\_MMD)}	& \citet{zhao2018vae}\\
        VAMP Autoencoder (VAMP)	& \citet{tomczak_vae_2018} \\
        {Hyperspherical VAE (SVAE)} & \citet{davidson_hyperspherical_2018}\\
        {Adversarial Autoencoder (Adversarial\_AE)} & \citet{makhzani2016vae}\\
        {Variational Autoencoder GAN (VAEGAN)} & \citet{larsen_autoencoding_2015}\\
        {Vector Quantized VAE (VQVAE)} & \citet{oord2018vae}\\
        Hamiltonian VAE (HVAE) & \citet{caterini_hamiltonian_2018}\\
        {Regularized AE with L2 decoder param (RAE\_L2)} &  \citet{ghosh_variational_2020}\\
        {Regularized AE with gradient penalty (RAE\_GP)} & \citet{ghosh_variational_2020} \\
        {Riemannian Hamiltonian VAE (RHVAE)} & \citet{chadebec_data_2021} \\
        \bottomrule
      \end{tabular}
    \end{table}
\paragraph{Maintenance plan:} We intend for this library to be maintained in the long term. In that view, the main author's contact details will remain available and up-to-date on the github repository, which will remain the main discussion channel. Additionally, we are currently considering adding back-up contributors that will also support this effort in the long-term. Since this library has already started to be a community effort with external contributors, we further hope that the community will also continue to help reviewing and updating the current implementations.
    
\paragraph{Original papers reproducibility}
We validate the implementations by reproducing some results presented in the original publications when the official code has been released or when enough details about the experimental section of the papers were available (we indeed noted that in many papers key elements for reproducibility were missing such as the data split considered, which criteria is used to select the model on which the metrics are computed, the hyper-parameters are not fully disclosed or the network architectures is unclear making reproduction very hard if not impossible in certain cases). This insists on the fact that the framework is flexible enough to reproduce results from publications. Finally, we have open-sourced the scripts, configurations and results on the repository at \url{https://github.com/clementchadebec/benchmark_VAE/tree/main/examples/scripts/reproducibility} and made the trained models available on the HuggingFace Hub (e.g. \url{https://huggingface.co/clementchadebec/reproduced_iwae}).
    
\clearpage

\section{Interpolations}\label{appB}

\begin{figure}[ht]
    \centering
    \captionsetup[subfigure]{position=above, labelformat = empty}
    \adjustbox{minipage=6em,raise=\dimexpr -5.\height}{\small VAE}
    \subfloat[CIFAR10 (32)]{\includegraphics[width=2.3in]{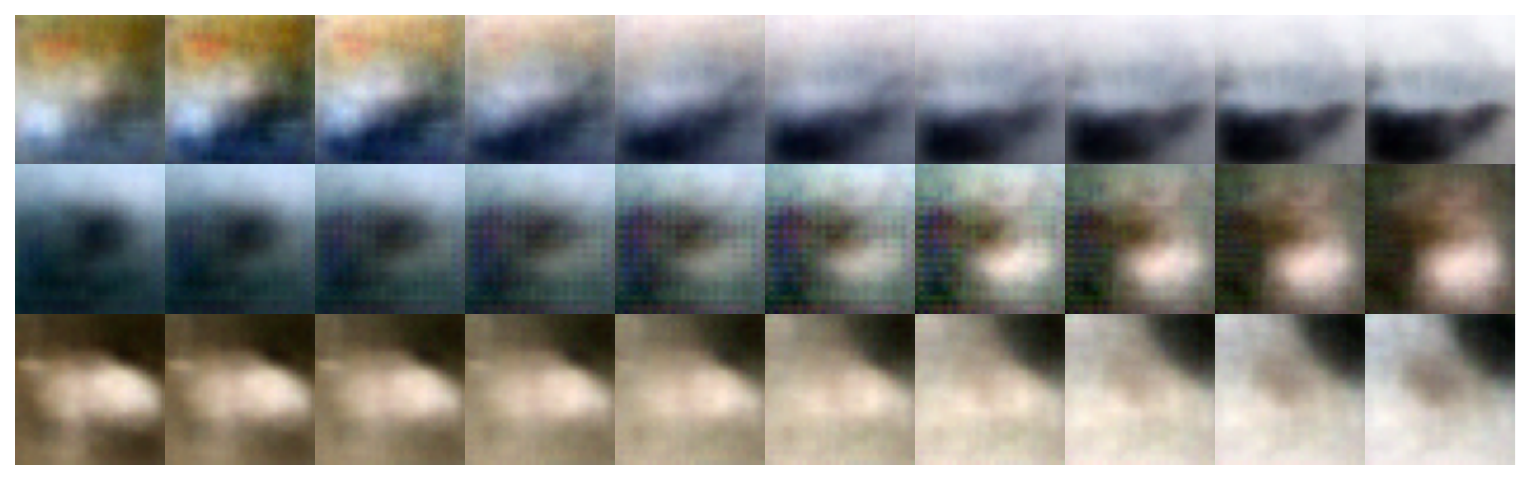}}
    \subfloat[CIFAR10 (256)]{\includegraphics[width=2.3in]{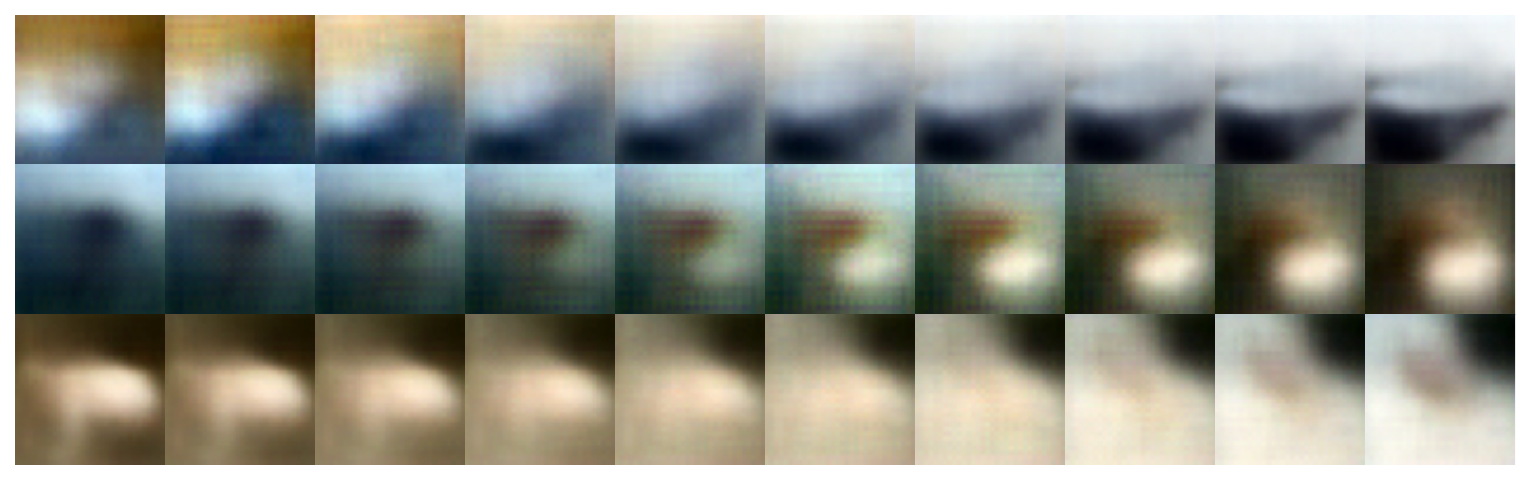}}\\\vspace{-1.2em}
    \adjustbox{minipage=6em,raise=\dimexpr -5.\height}{\small VAMP}
    \subfloat{\includegraphics[width=2.3in]{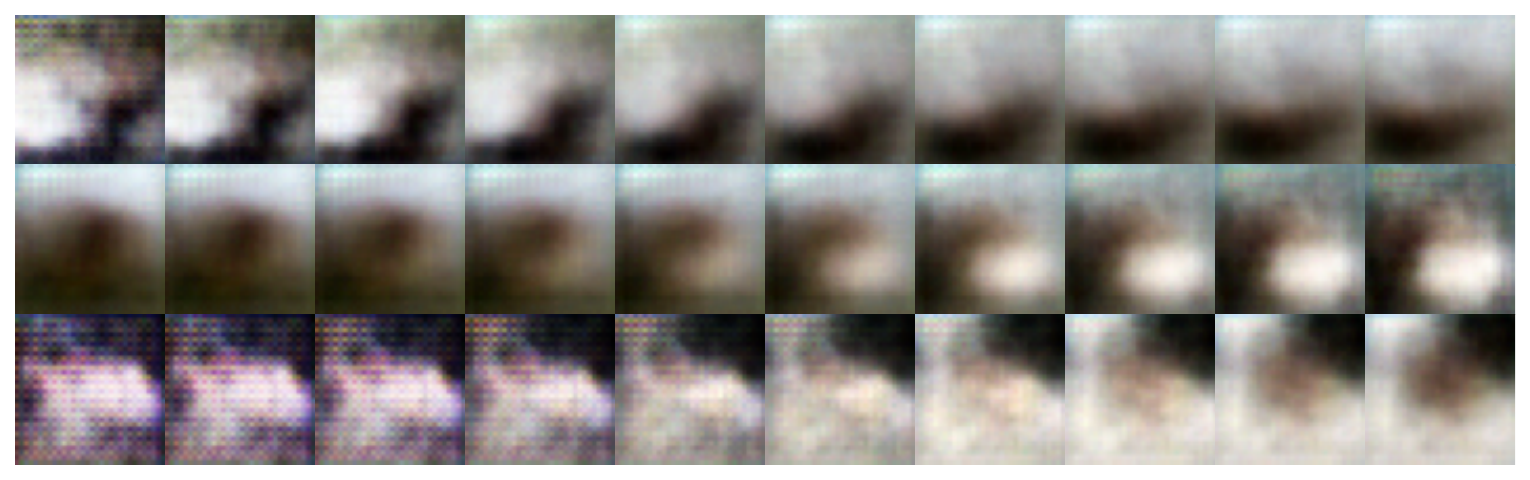}}
    \subfloat{\includegraphics[width=2.3in]{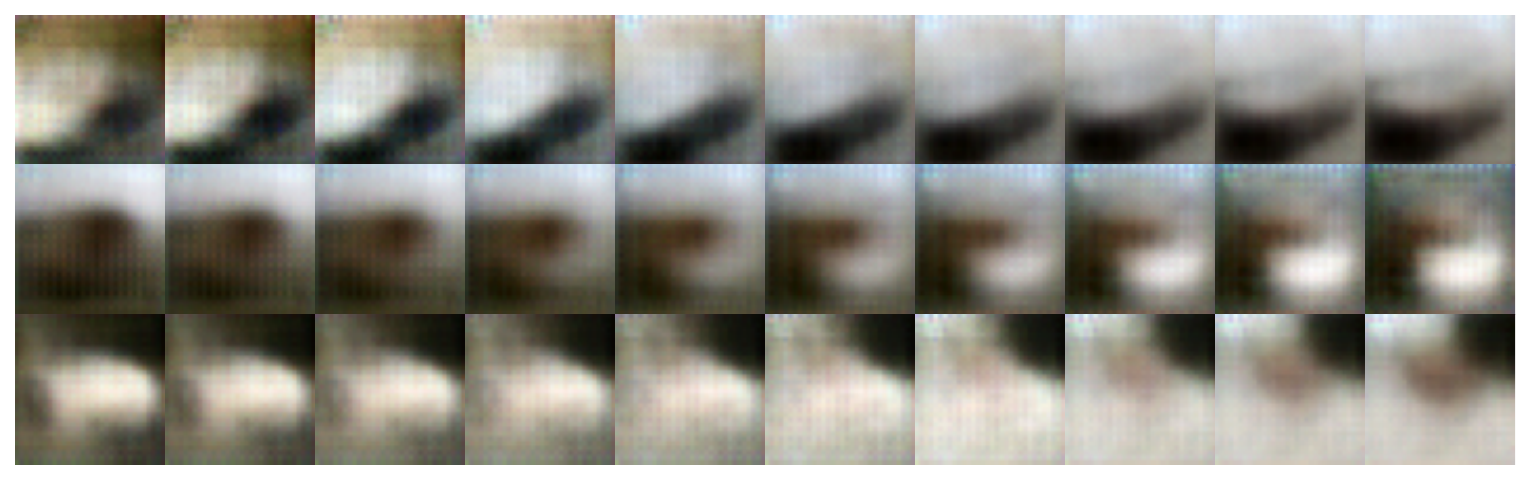}}\\\vspace{-1.2em}
    \adjustbox{minipage=6em,raise=\dimexpr -5.\height}{\small IWAE}
    \subfloat{\includegraphics[width=2.3in]{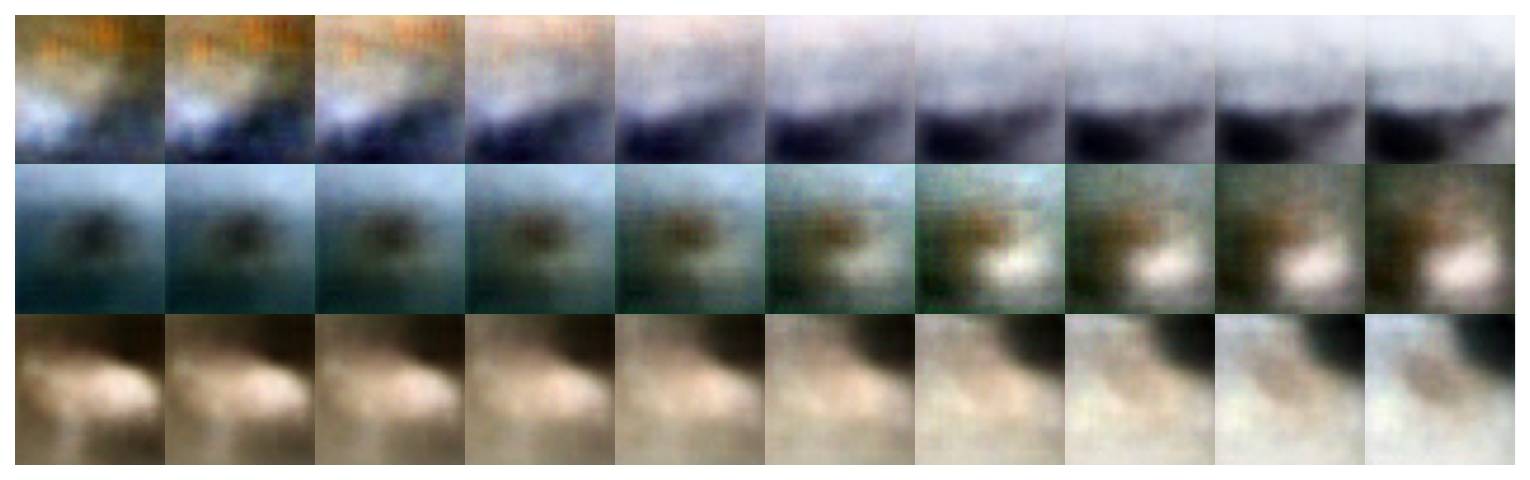}}
    \subfloat{\includegraphics[width=2.3in]{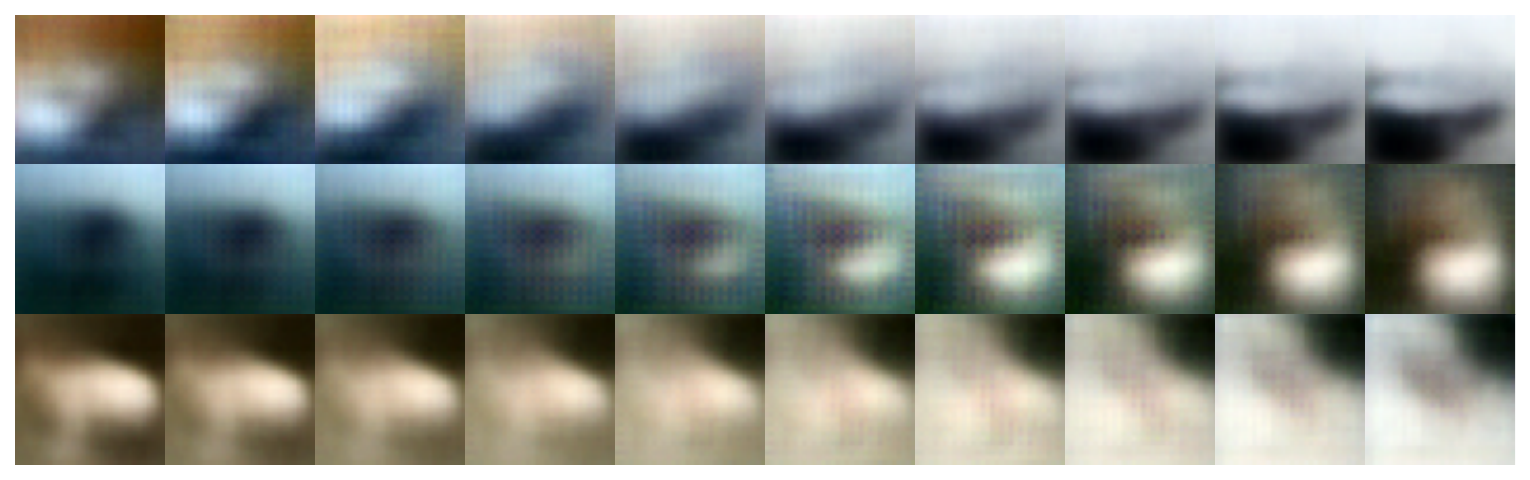}}\\\vspace{-1.2em}
    \adjustbox{minipage=6em,raise=\dimexpr -5.\height}{\small VAE-lin-NF}
    \subfloat{\includegraphics[width=2.3in]{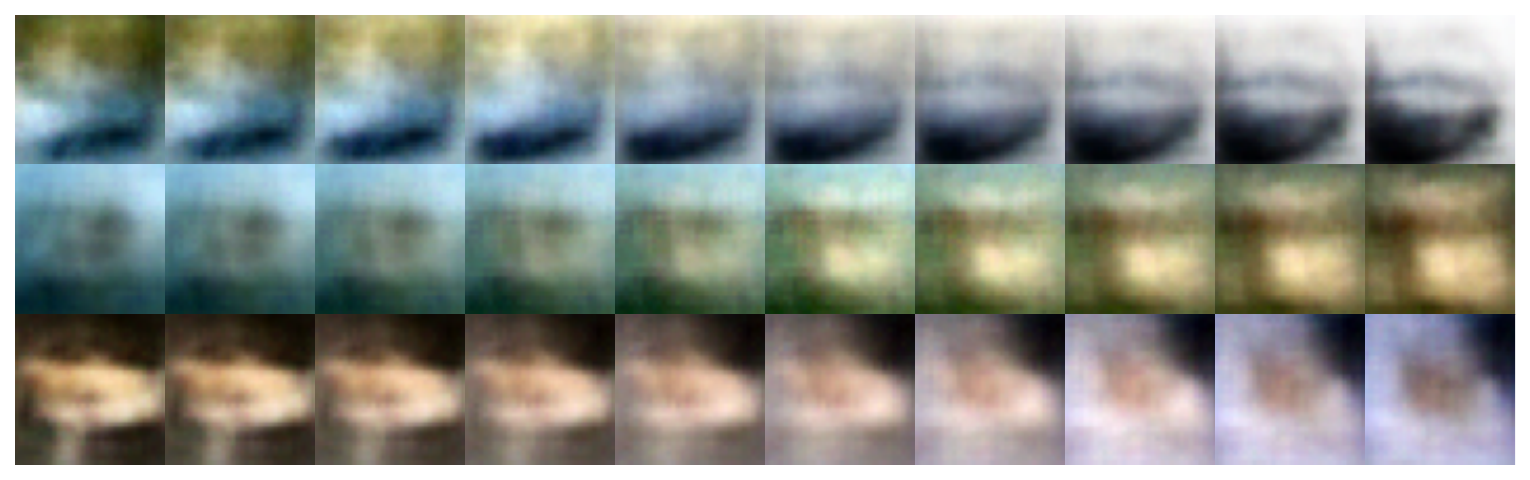}}
    \subfloat{\includegraphics[width=2.3in]{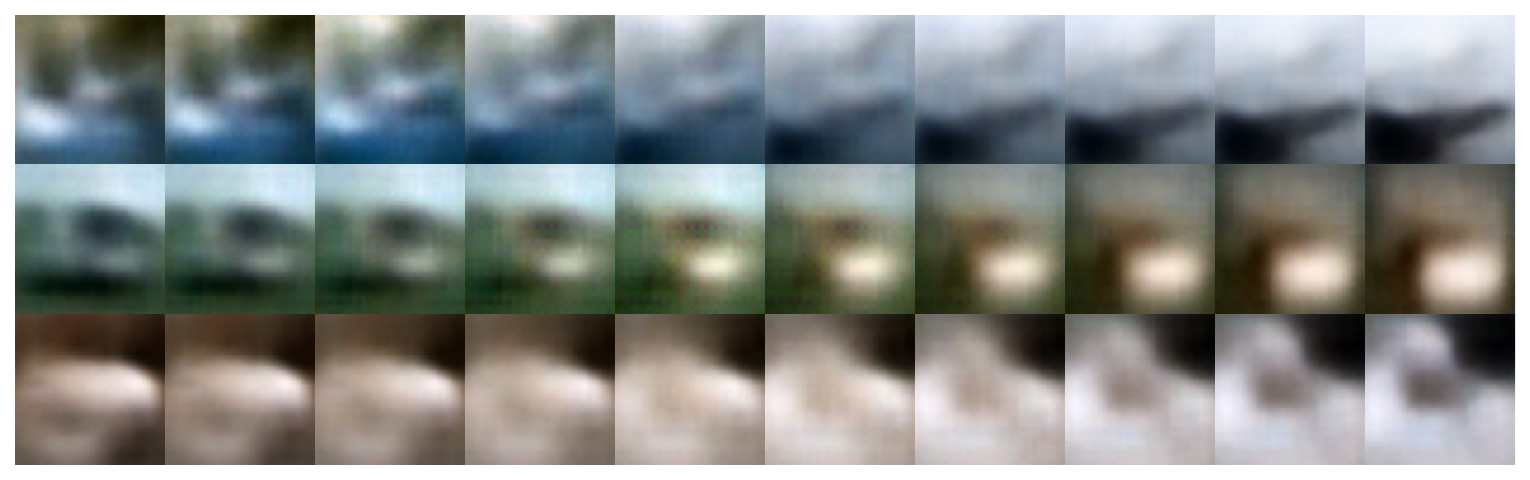}}\\\vspace{-1.2em}
     \adjustbox{minipage=6em,raise=\dimexpr -5.\height}{\small VAE-IAF}
    \subfloat{\includegraphics[width=2.3in]{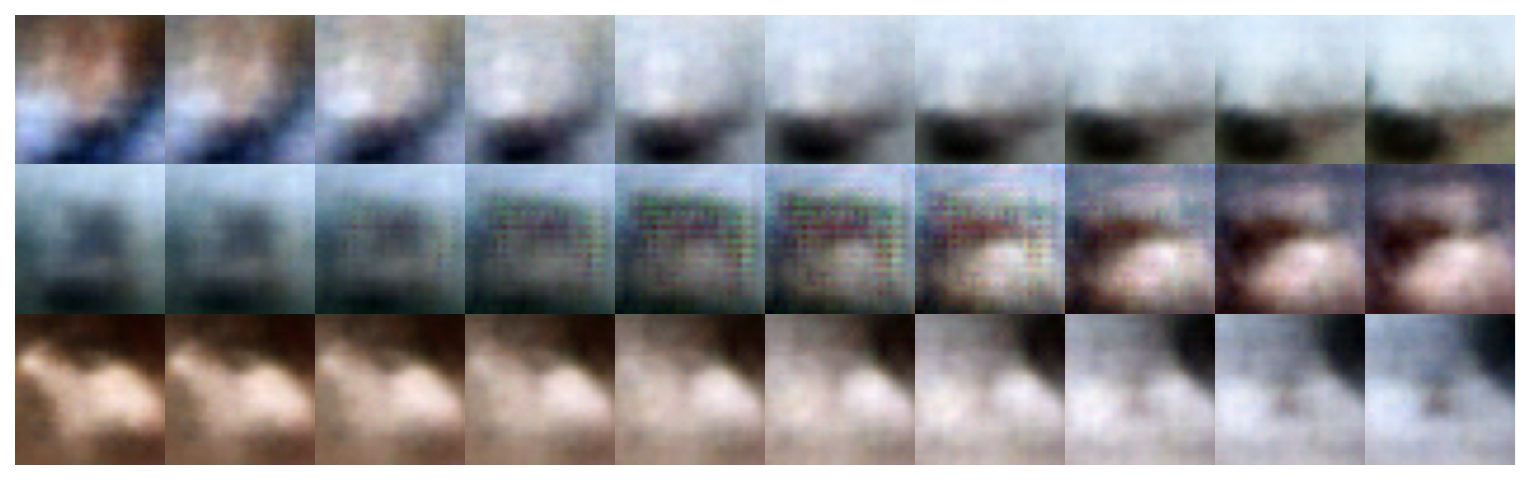}}
    \subfloat{\includegraphics[width=2.3in]{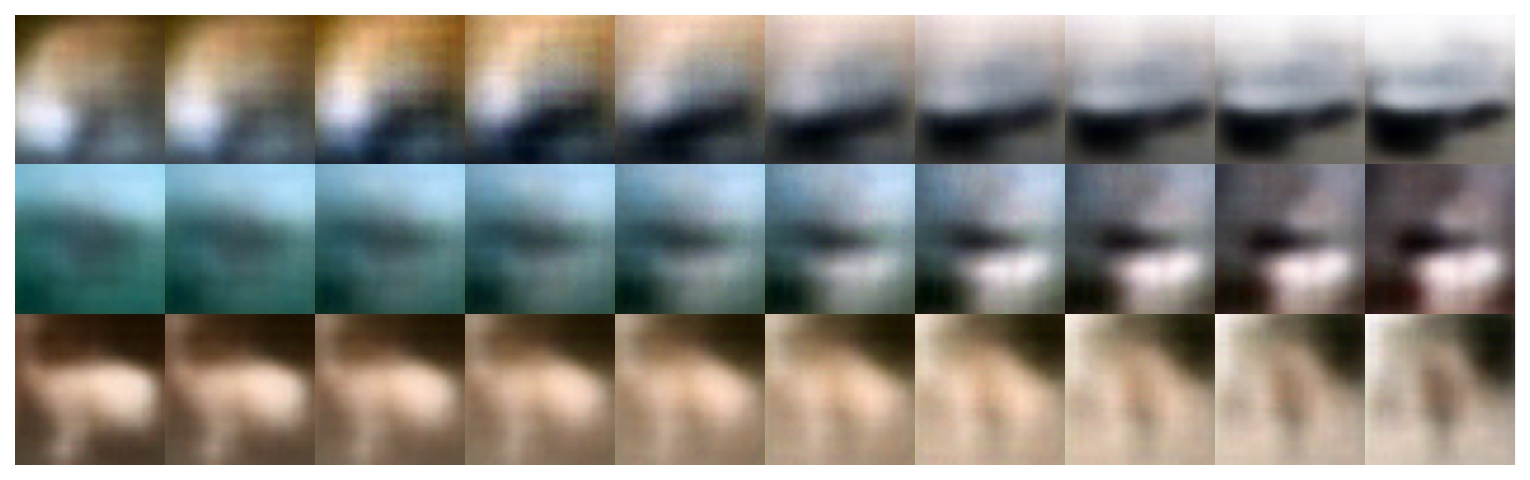}}\\\vspace{-1.2em}
    \adjustbox{minipage=6em,raise=\dimexpr -5.\height}{\small $\beta$-VAE}
    \subfloat{\includegraphics[width=2.3in]{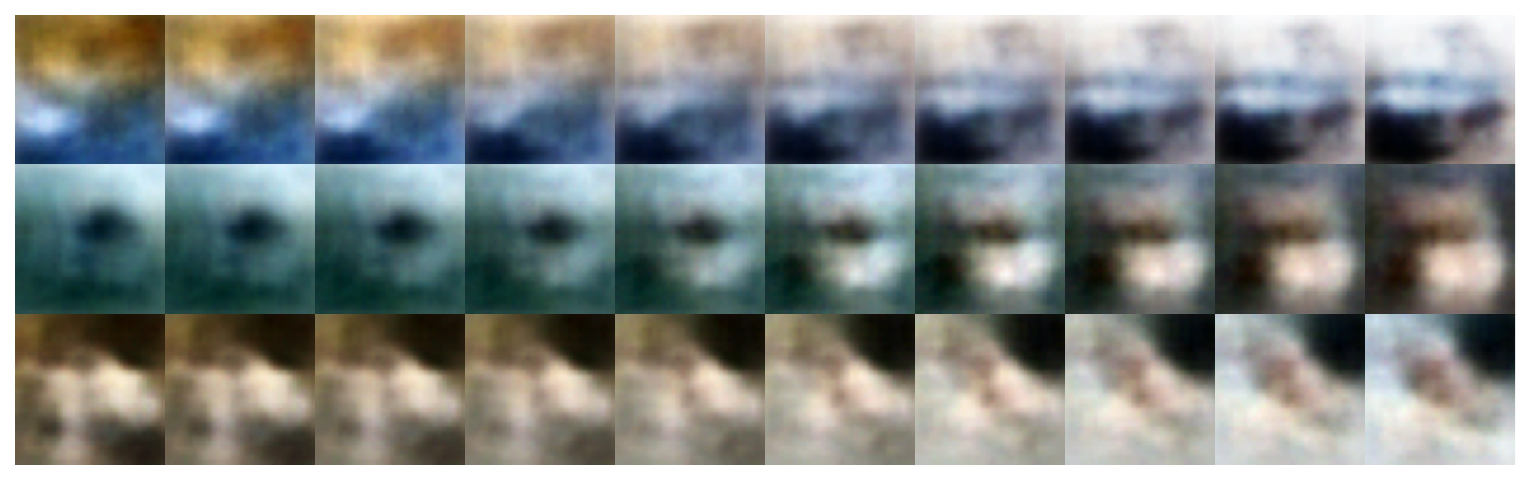}}
    \subfloat{\includegraphics[width=2.3in]{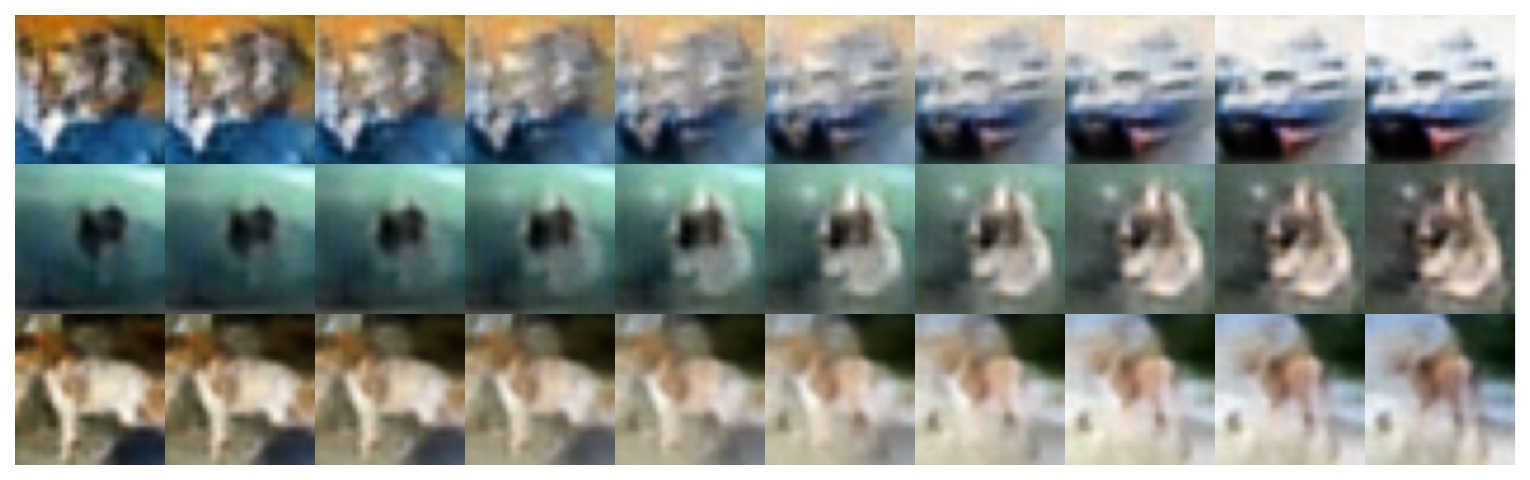}}\\\vspace{-1.2em}
    \adjustbox{minipage=6em,raise=\dimexpr -5.\height}{\small $\beta$-TC-VAE}
    \subfloat{\includegraphics[width=2.3in]{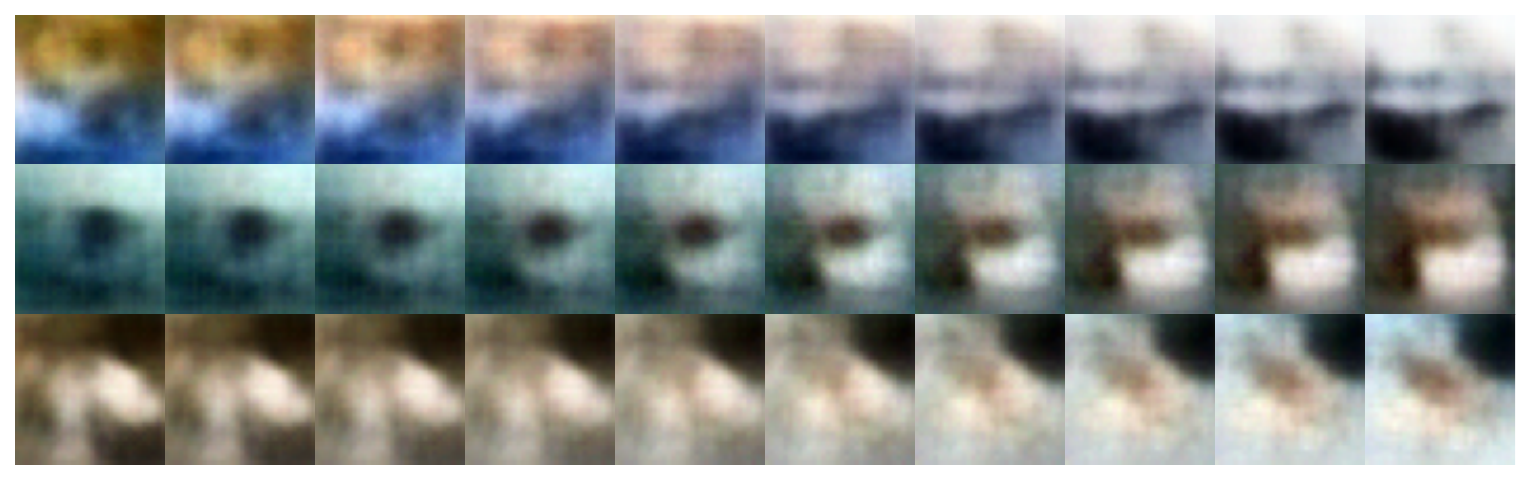}}
    \subfloat{\includegraphics[width=2.3in]{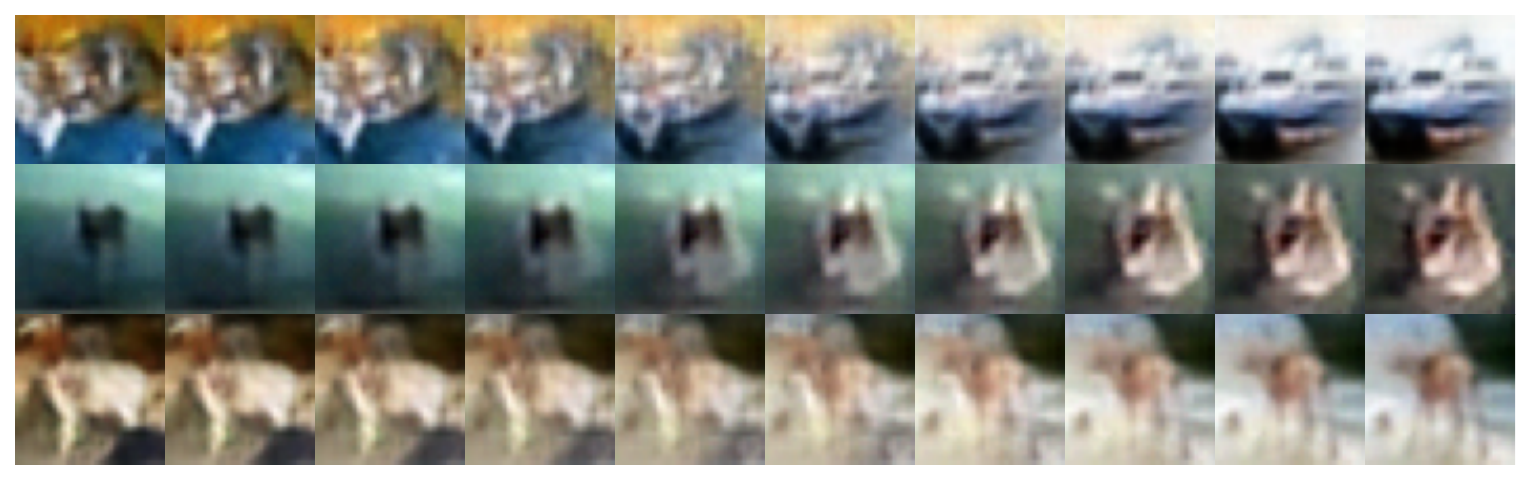}}\\\vspace{-1.2em}
    \adjustbox{minipage=6em,raise=\dimexpr -5.\height}{\small Factor-VAE}
    \subfloat{\includegraphics[width=2.3in]{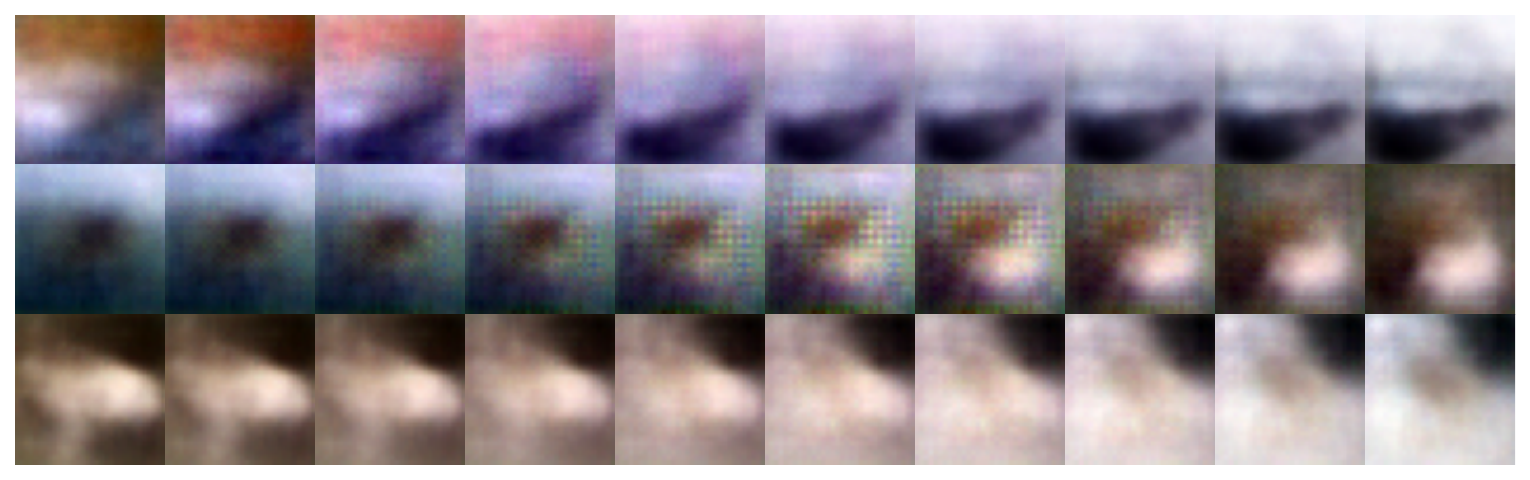}}
    \subfloat{\includegraphics[width=2.3in]{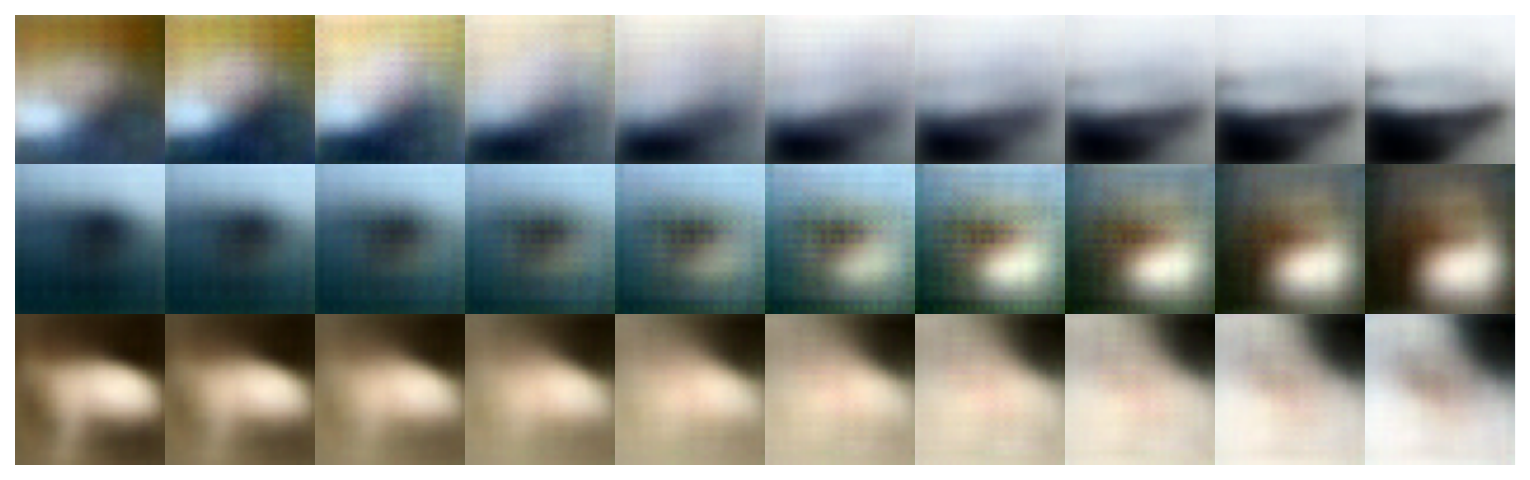}}\\\vspace{-1.2em}
    \adjustbox{minipage=6em,raise=\dimexpr -5.\height}{\small InfoVAE - IMQ}
    \subfloat{\includegraphics[width=2.3in]{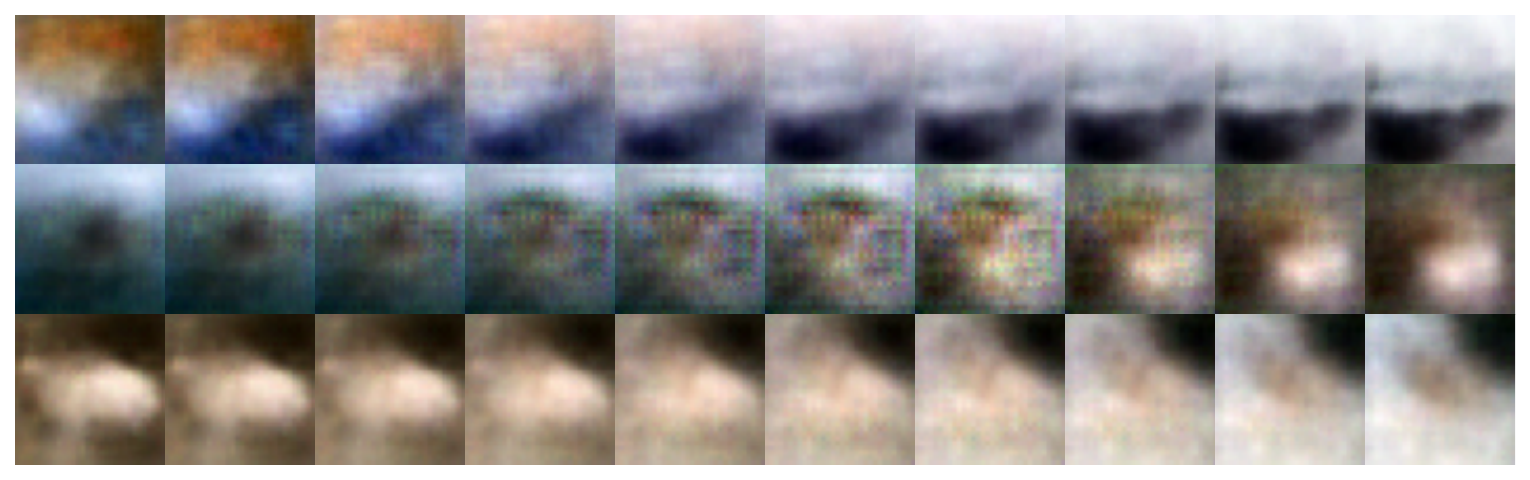}}
    \subfloat{\includegraphics[width=2.3in]{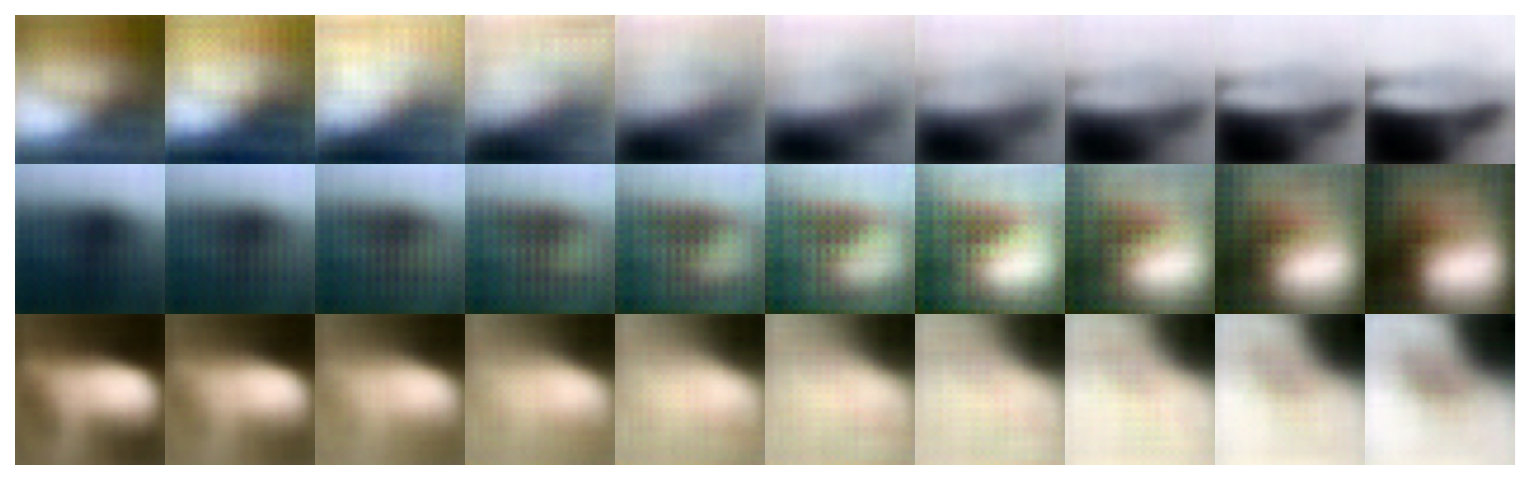}}\\\vspace{-1.2em}
    \adjustbox{minipage=6em,raise=\dimexpr -5.\height}{\small InfoVAE - RBF}
    \subfloat{\includegraphics[width=2.3in]{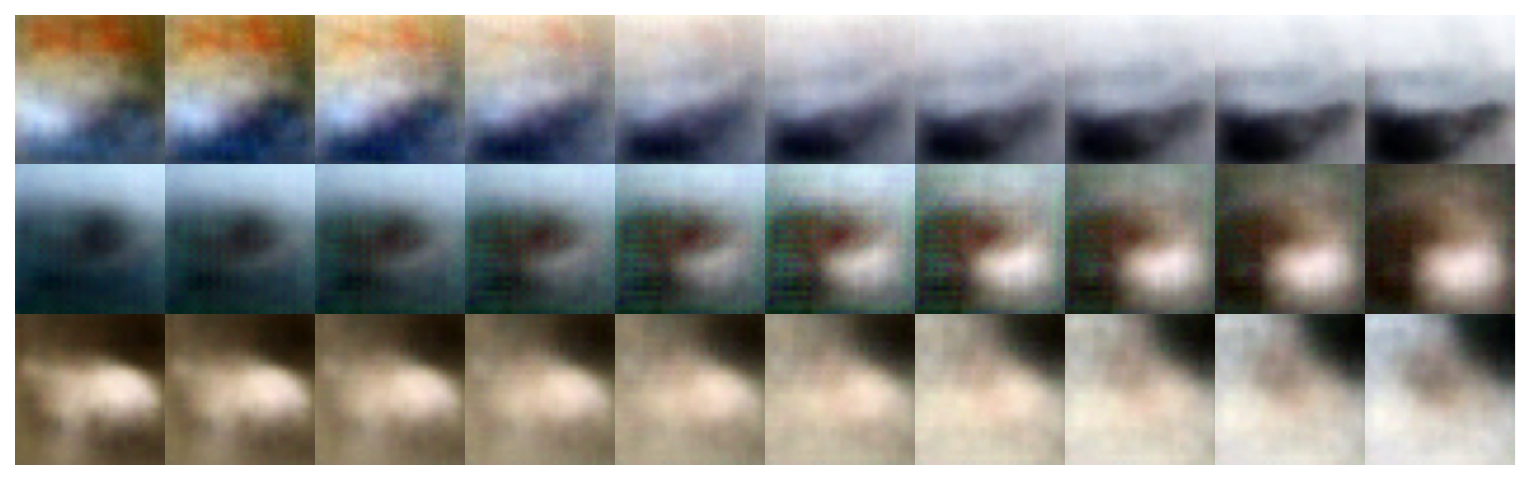}}
    \subfloat{\includegraphics[width=2.3in]{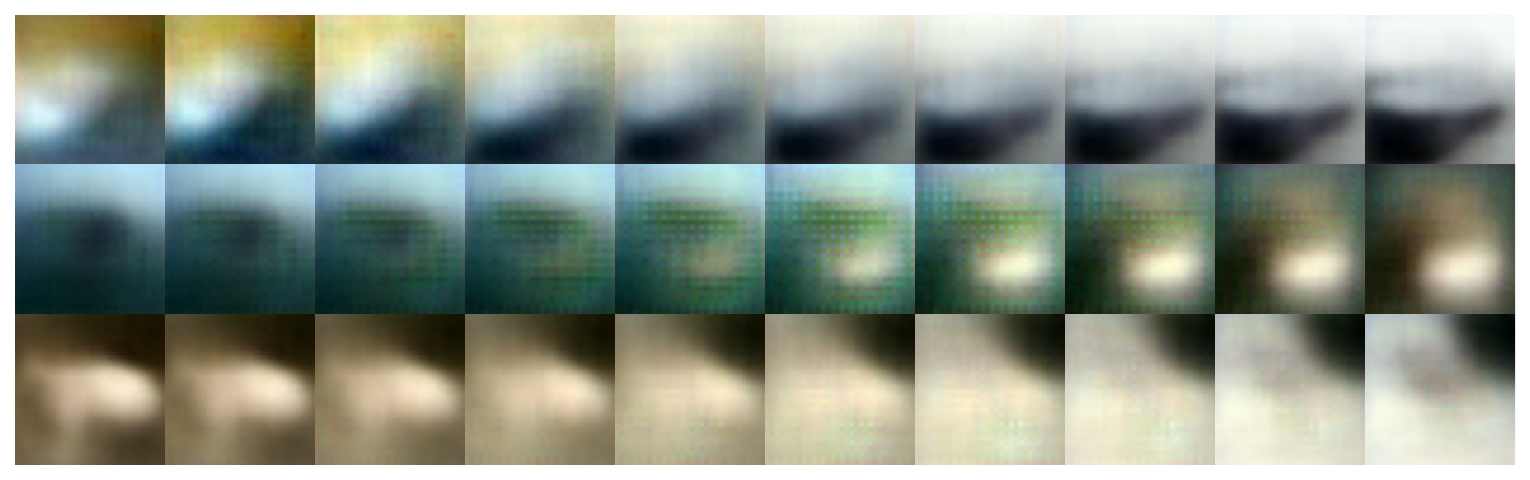}}
    \caption{Interpolations on CIFAR10 with the same starting and ending images for latent spaces of dimension 32 and 256. For each model we select the configuration achieving the lowest FID on the generation task on the validation set with a GMM sampler.}
    \label{fig:interpolations cifar 1}
    \end{figure}

In this section, we show the interpolations obtained on the three considered datasets. For each model, we select both a starting image and an ending image from the test set and perform a linear interpolation between the corresponding embeddings in the learned latent space. We then show the decoded trajectory all along the interpolation line. For this task, we use the model configuration that obtained the lowest FID on the validation set with a GMM sampler from the generation task.
We show the resulting interpolations for latent spaces of dimension 16 and 256 for MNIST, 32 and 256 for CIFAR10 and 64
for CELEBA.
As mentioned in the paper, for this complex task, variational approaches tend to outperform the AE-based methods. This is well illustrated on MNIST with a latent space of dimension 256 since all the AE-based approaches eventually superpose the starting and ending image, making the interpolation visually irrelevant. Impressively, the regularisation imposed by the variational approaches prevents such undesirable behaviours from occurring. 
This adds to the observation made in Sec.~4.2.2 of the paper where we note some robustness to the latent dimension for the variational methods. Nonetheless, as stated in the paper this regularisation can also degrade image reconstruction, leading to very blurry interpolations, as illustrated on Fig.~\ref{fig:interpolations cifar 1}. 

\begin{figure}[ht]
    \centering
    \captionsetup[subfigure]{position=above, labelformat = empty}
     \adjustbox{minipage=6em,raise=\dimexpr -5.\height}{\small VAE}
    \subfloat[MNIST (16)]{\includegraphics[width=2.3in]{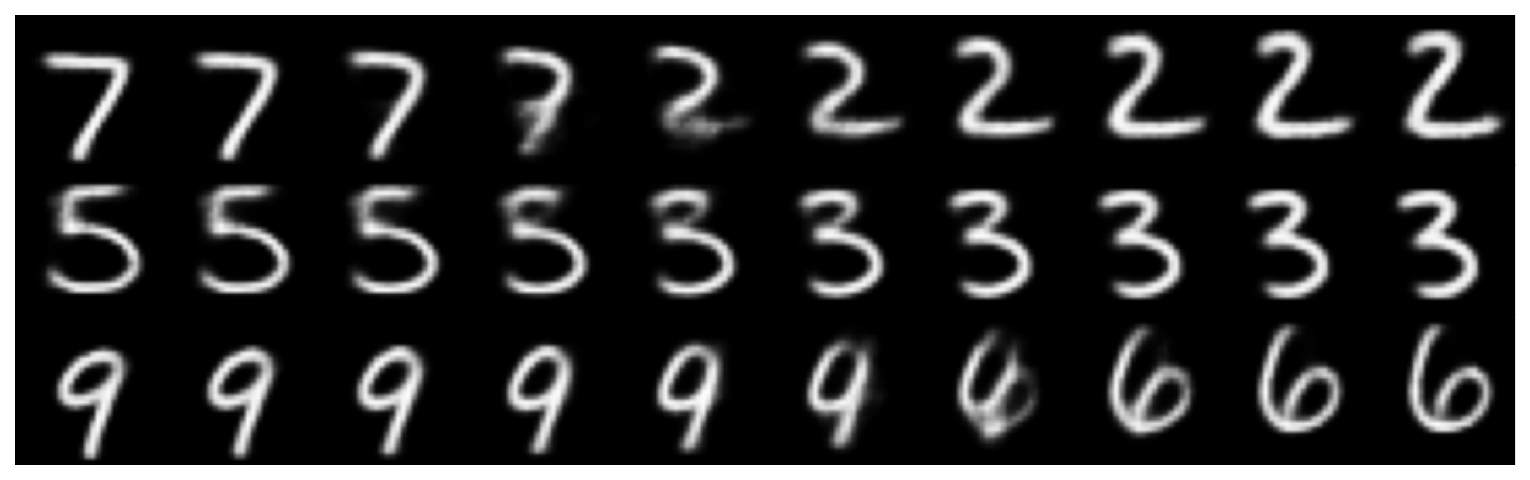}}
    \subfloat[MNIST (256)]{\includegraphics[width=2.3in]{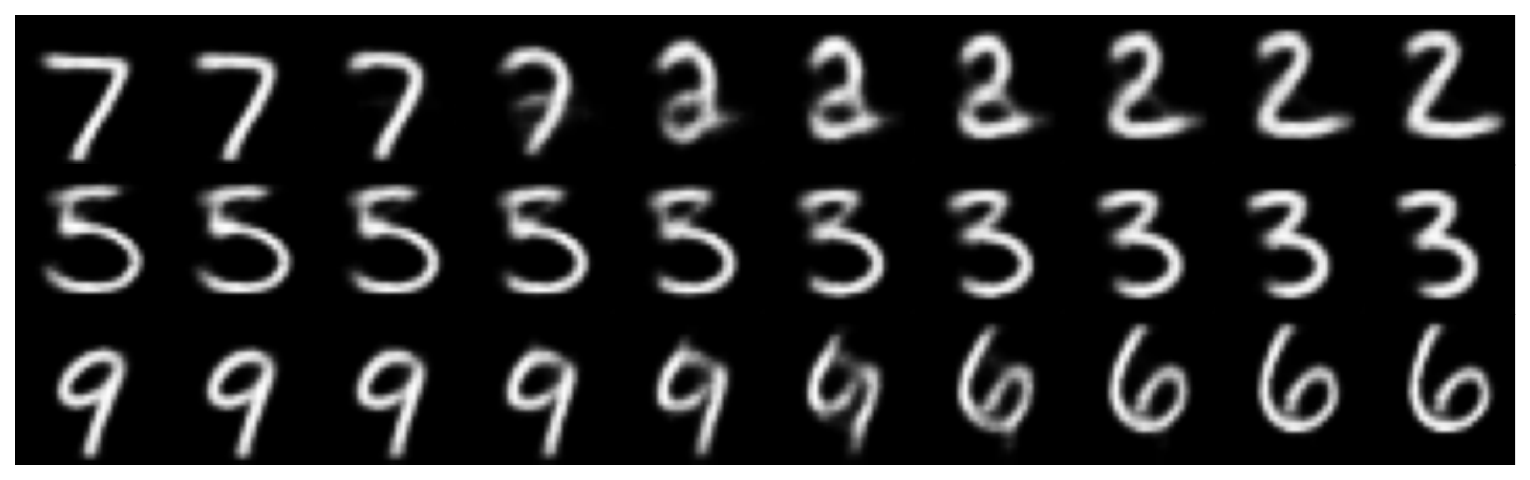}}\\\vspace{-1.2em}
     \adjustbox{minipage=6em,raise=\dimexpr -5.\height}{\small VAMP}
    \subfloat{\includegraphics[width=2.3in]{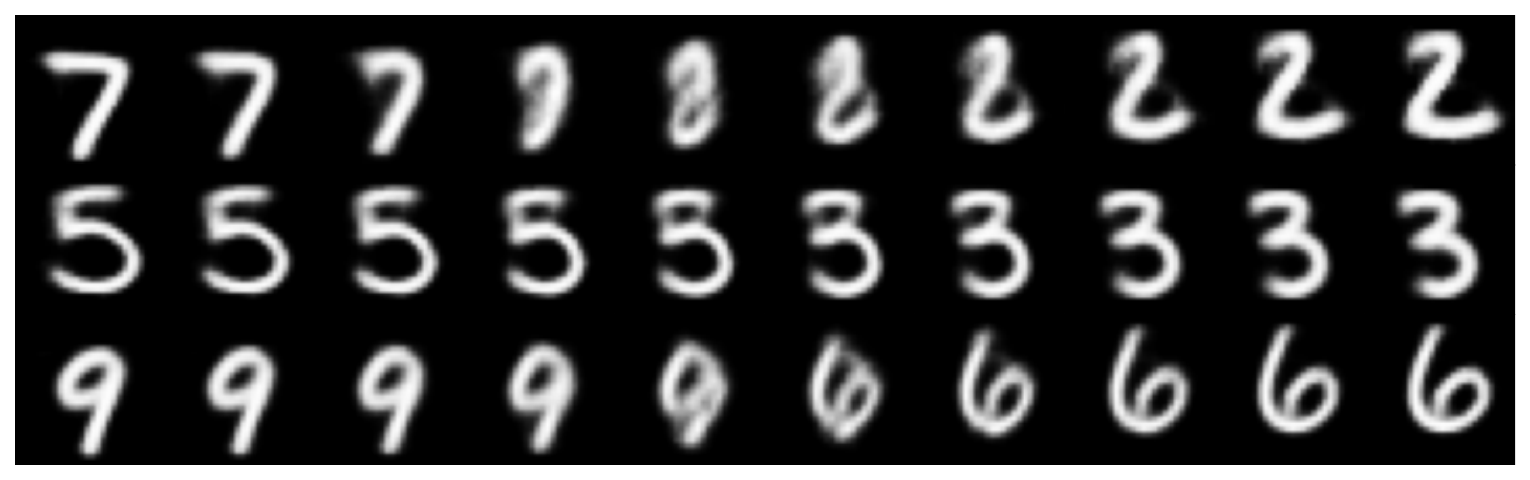}}
    \subfloat{\includegraphics[width=2.3in]{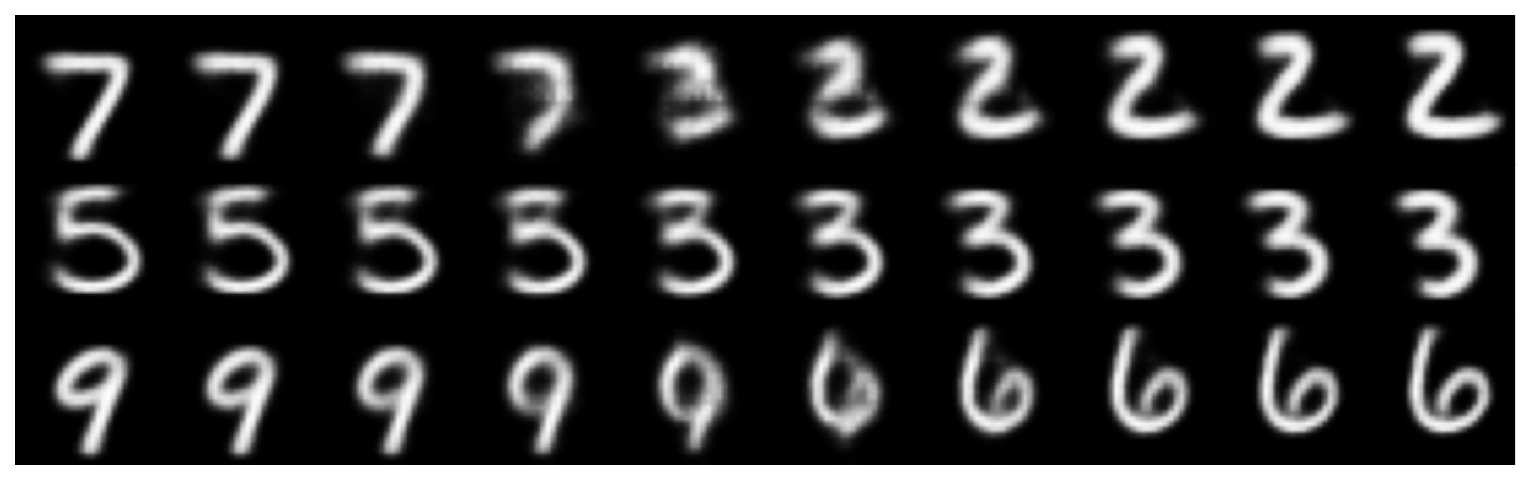}}\\\vspace{-1.2em}
    \adjustbox{minipage=6em,raise=\dimexpr -5.\height}{\small IWAE}
    \subfloat{\includegraphics[width=2.3in]{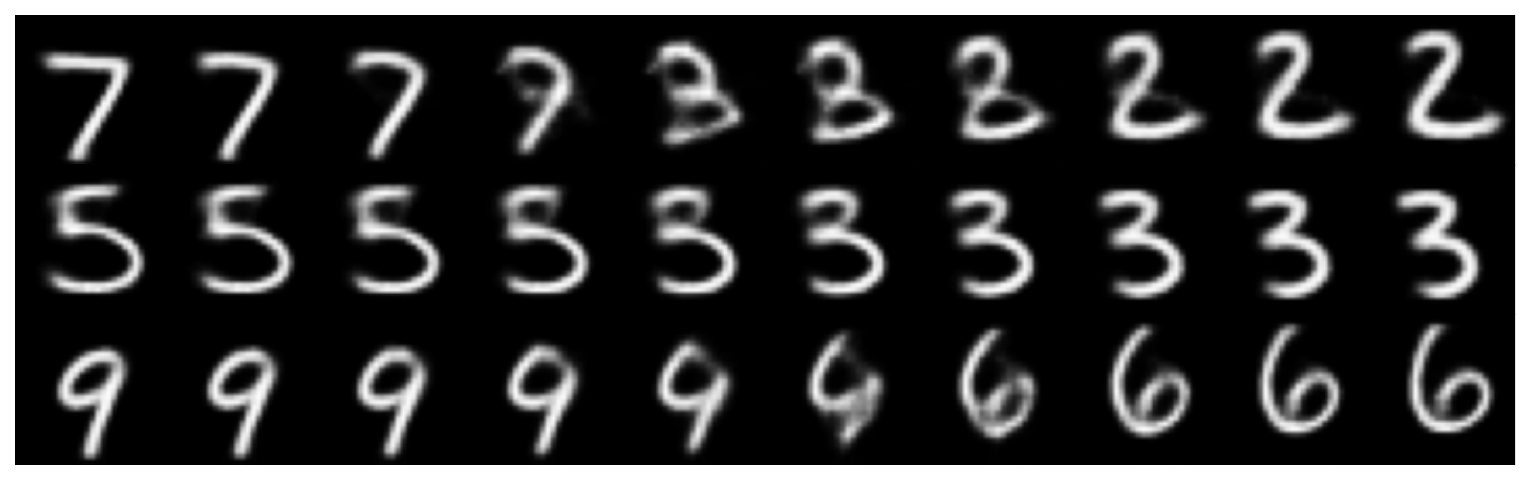}}
    \subfloat{\includegraphics[width=2.3in]{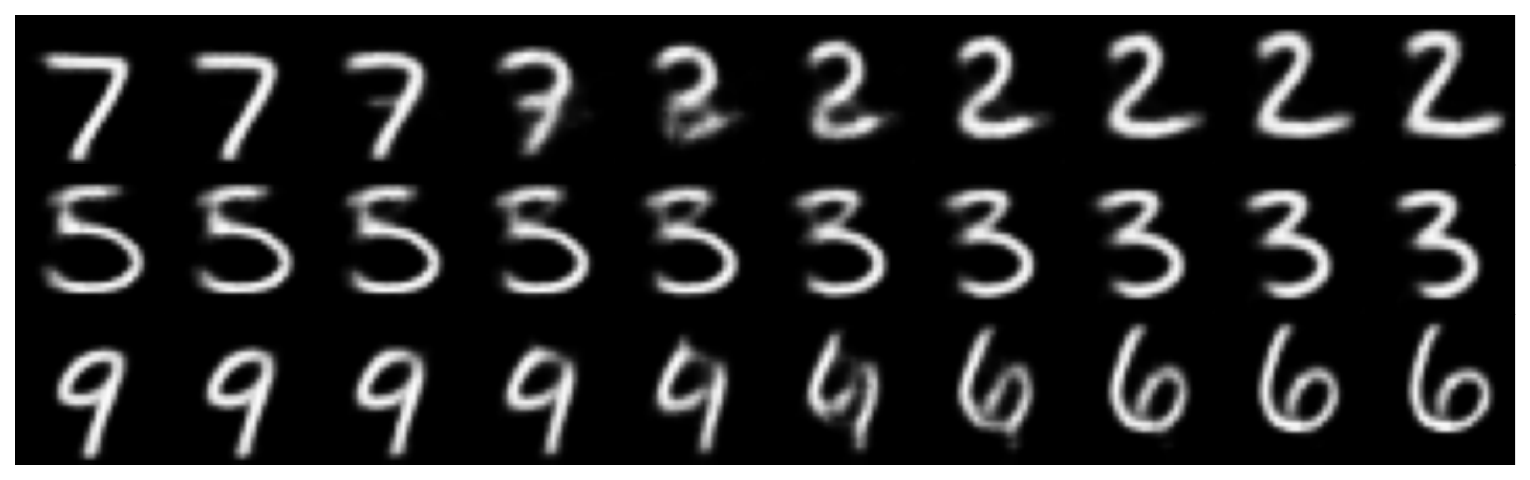}}\\\vspace{-1.2em}
    \adjustbox{minipage=6em,raise=\dimexpr -5.\height}{\small VAE-lin-NF}
    \subfloat{\includegraphics[width=2.3in]{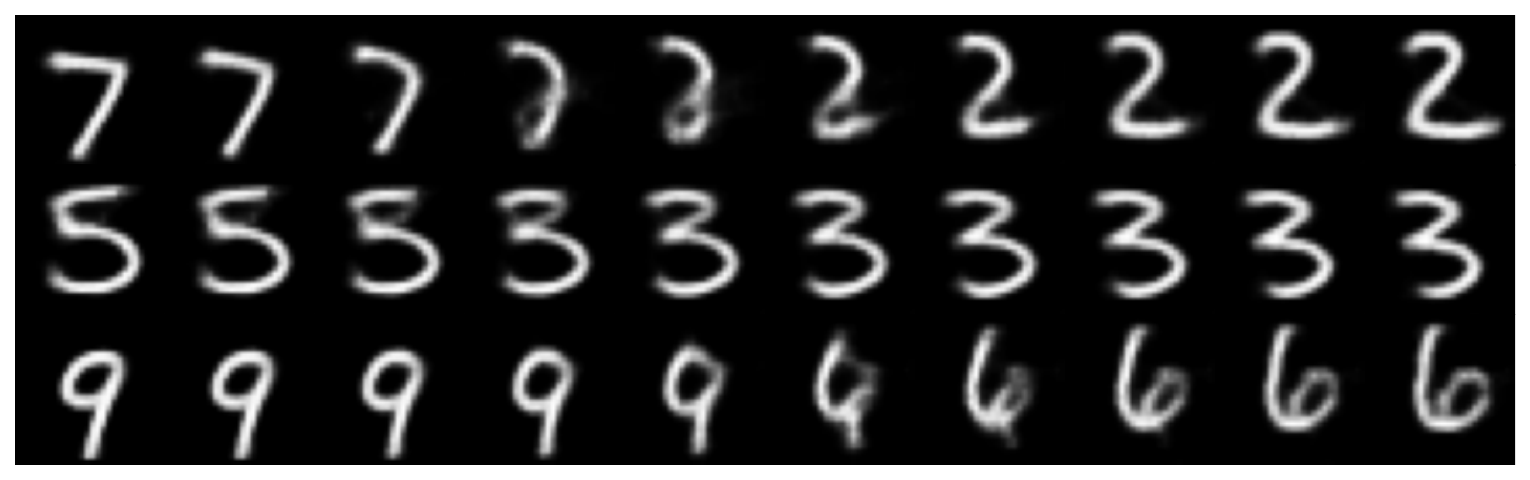}}
    \subfloat{\includegraphics[width=2.3in]{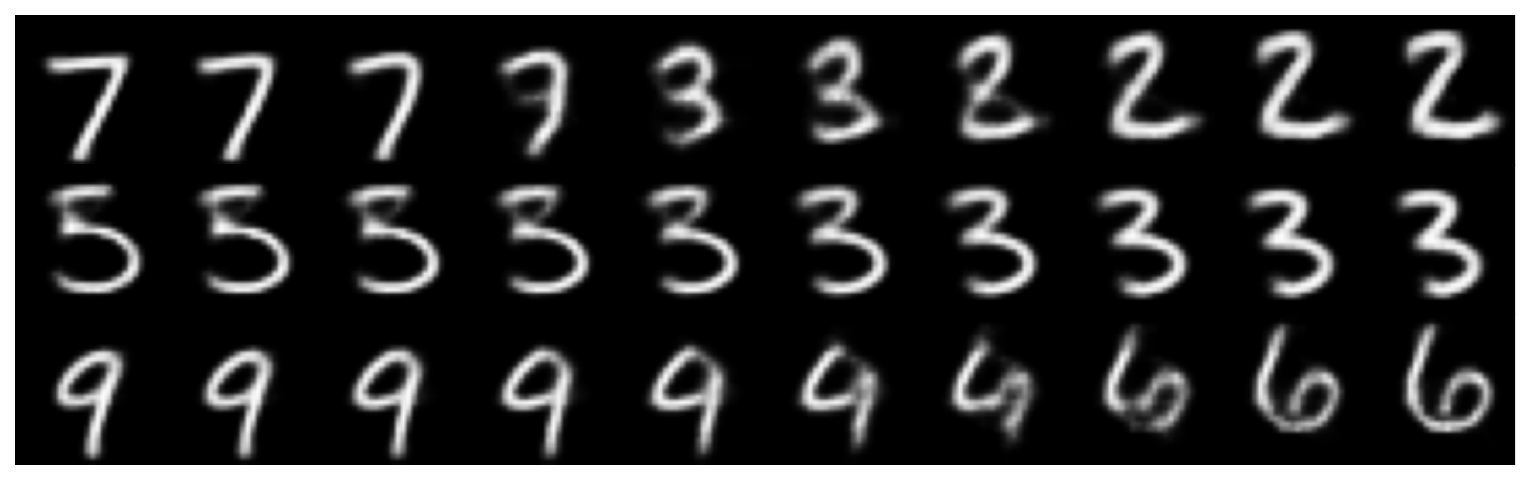}}\\\vspace{-1.2em}
    \adjustbox{minipage=6em,raise=\dimexpr -5.\height}{\small VAE-IAF}
    \subfloat{\includegraphics[width=2.3in]{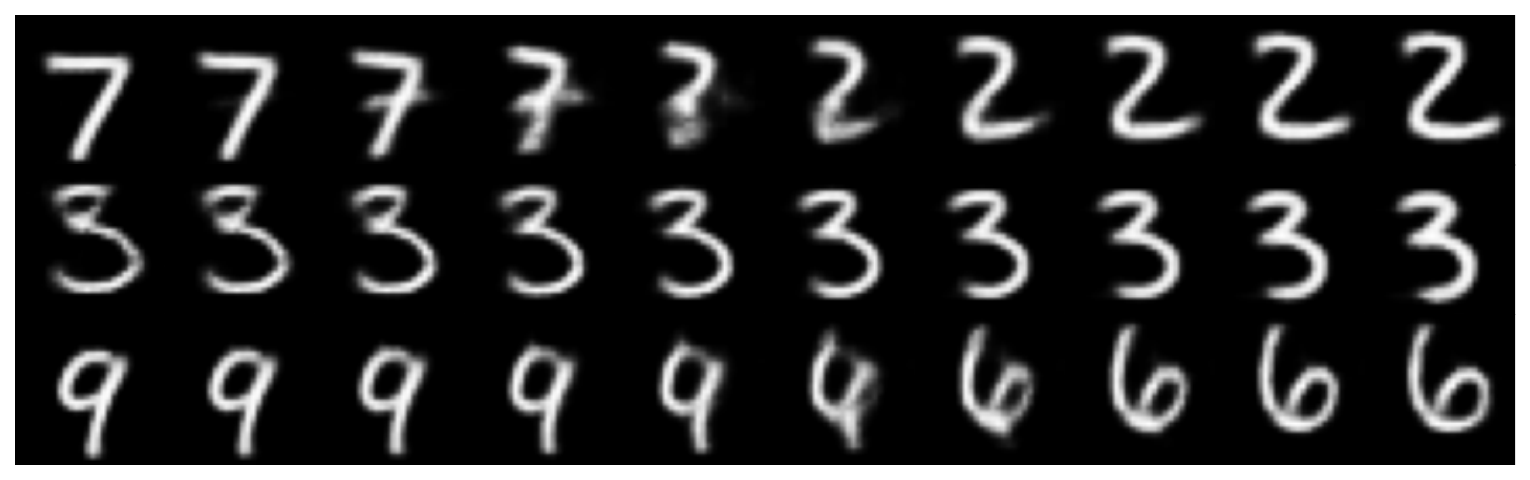}}
    \subfloat{\includegraphics[width=2.3in]{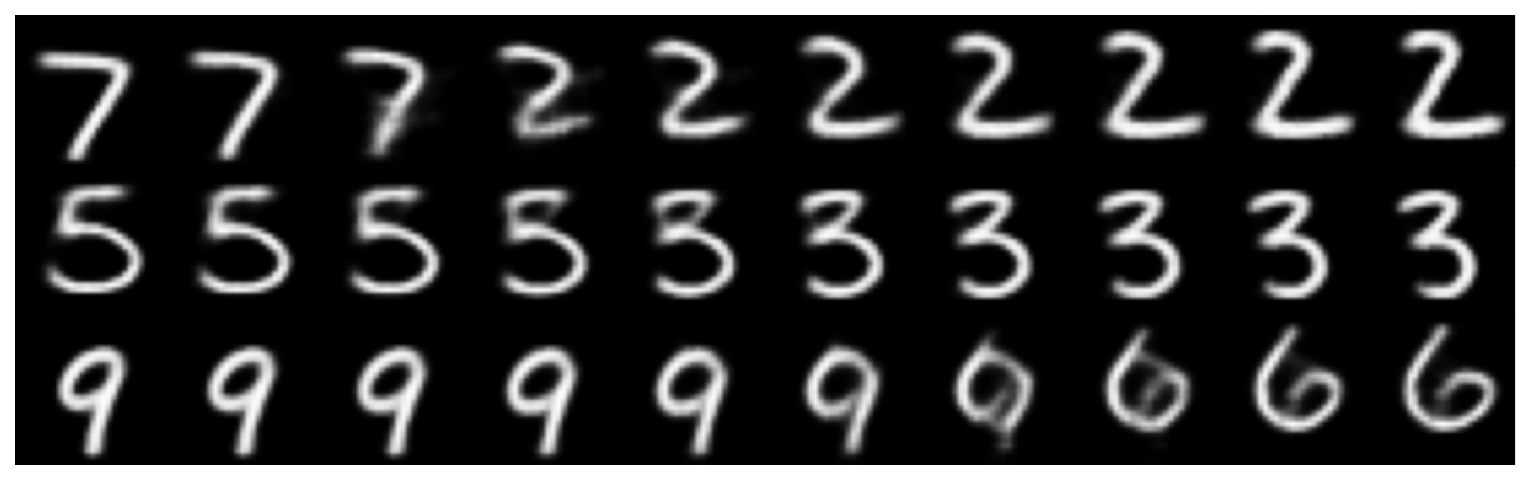}}\\\vspace{-1.2em}
     \adjustbox{minipage=6em,raise=\dimexpr -5.\height}{\small $\beta$-VAE}
    \subfloat{\includegraphics[width=2.3in]{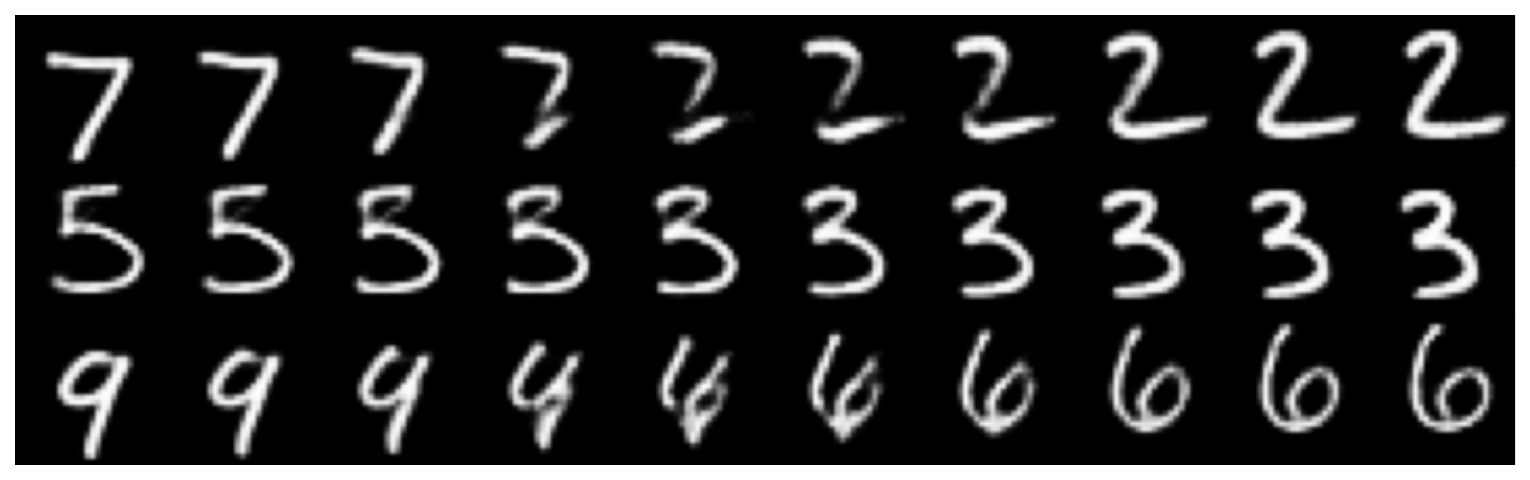}}
    \subfloat{\includegraphics[width=2.3in]{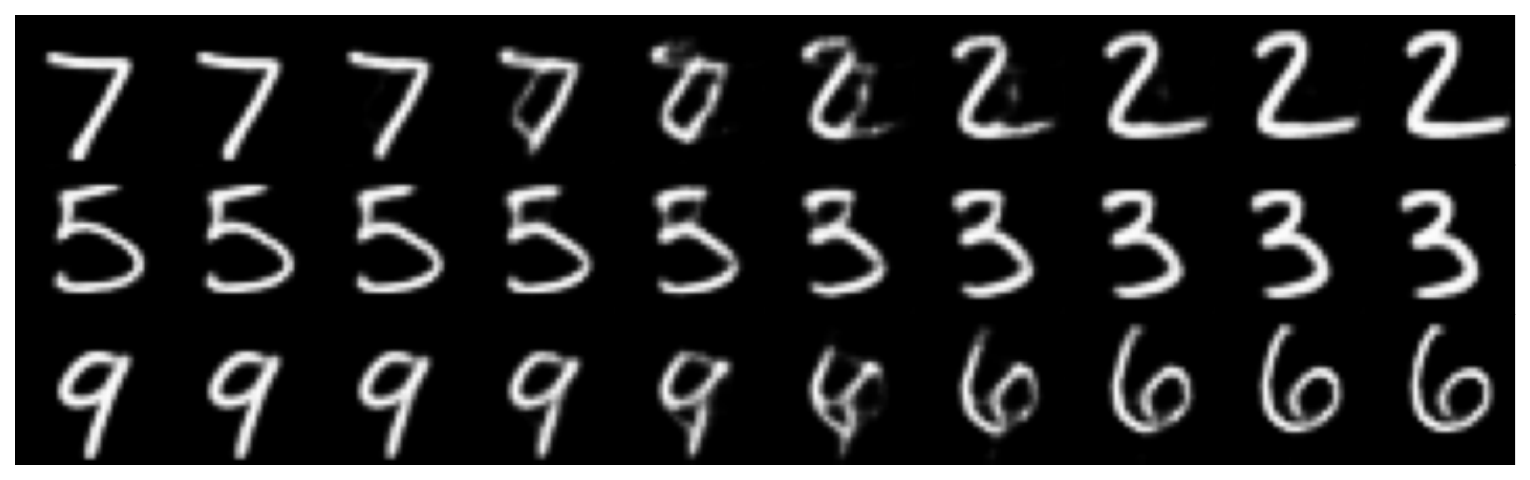}}\\\vspace{-1.2em}
    \adjustbox{minipage=6em,raise=\dimexpr -5.\height}{\small $\beta$-TC-VAE}
    \subfloat{\includegraphics[width=2.3in]{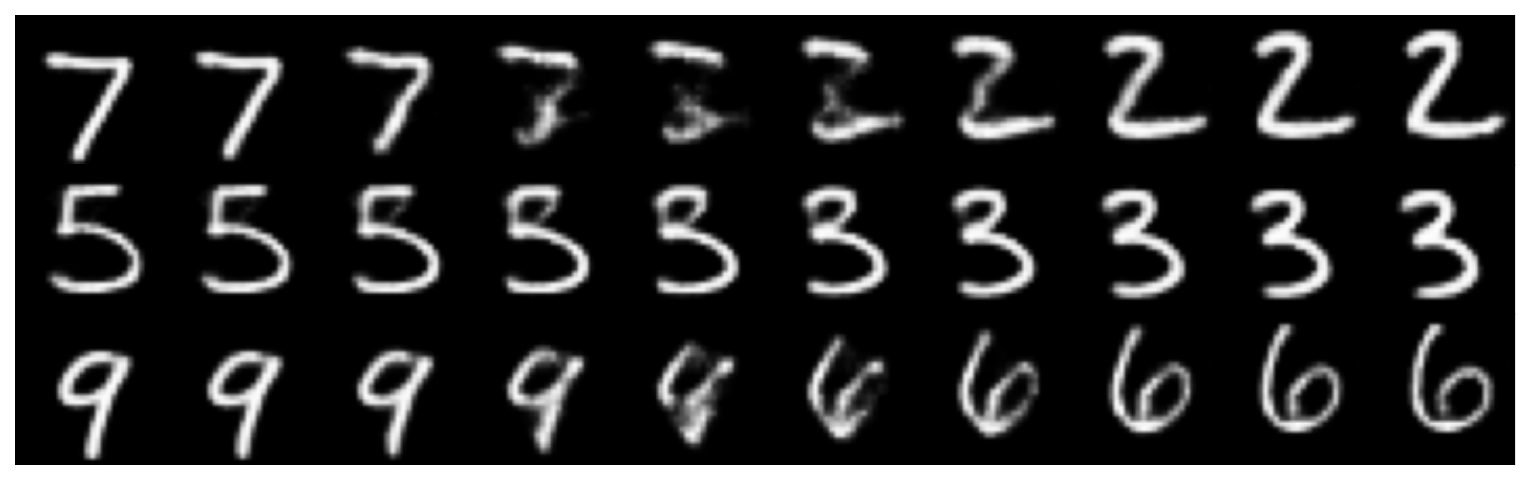}}
    \subfloat{\includegraphics[width=2.3in]{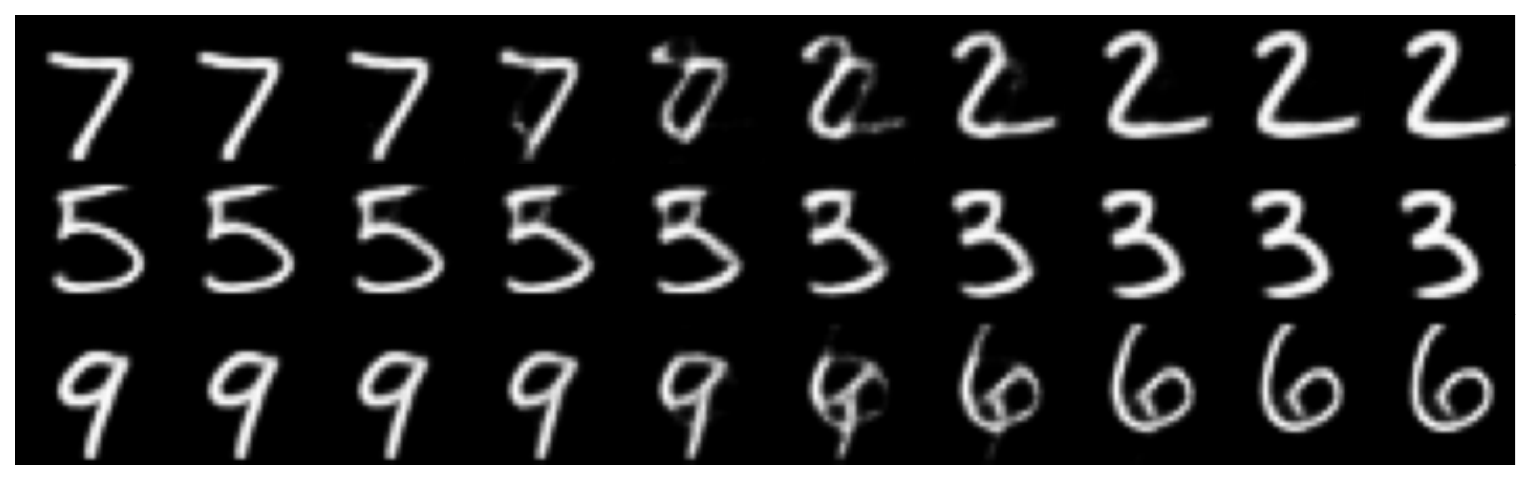}}\\\vspace{-1.2em}
    \adjustbox{minipage=6em,raise=\dimexpr -5.\height}{\small Factor-VAE}
    \subfloat{\includegraphics[width=2.3in]{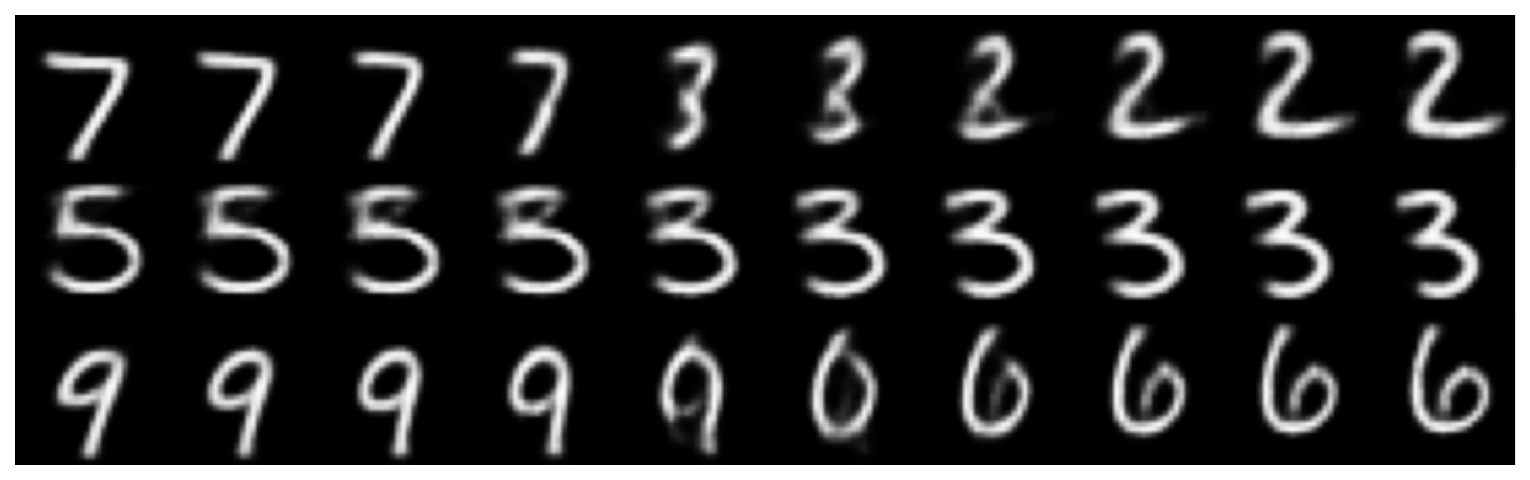}}
    \subfloat{\includegraphics[width=2.3in]{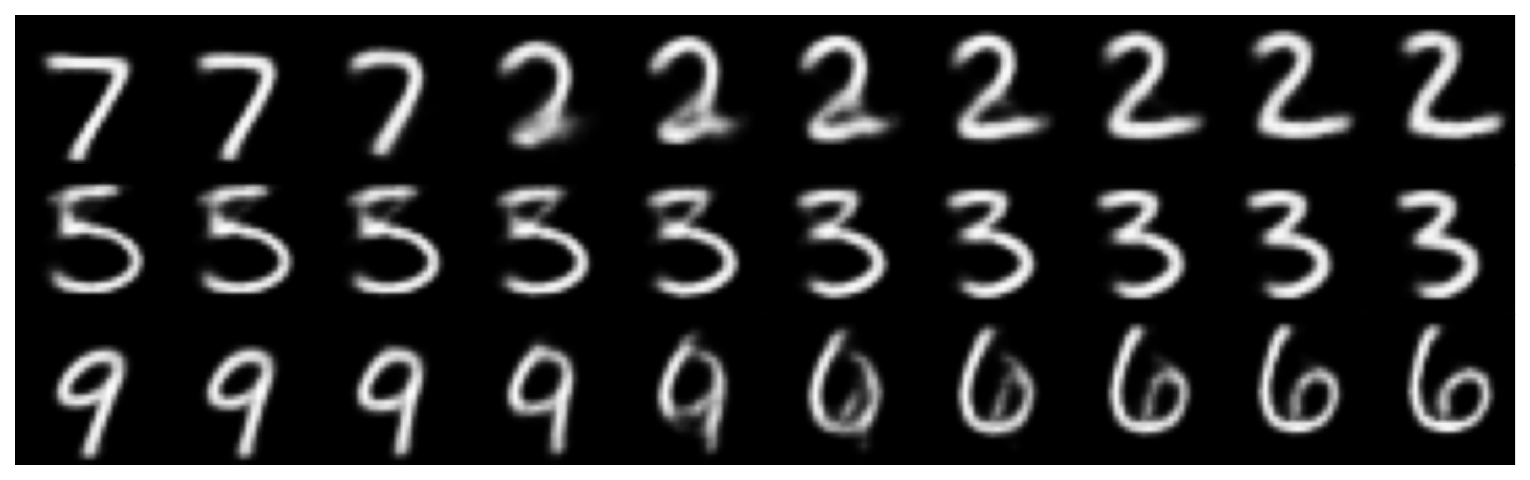}}\\\vspace{-1.2em}
    \adjustbox{minipage=6em,raise=\dimexpr -5.\height}{\small InfoVAE - IMQ}
    \subfloat{\includegraphics[width=2.3in]{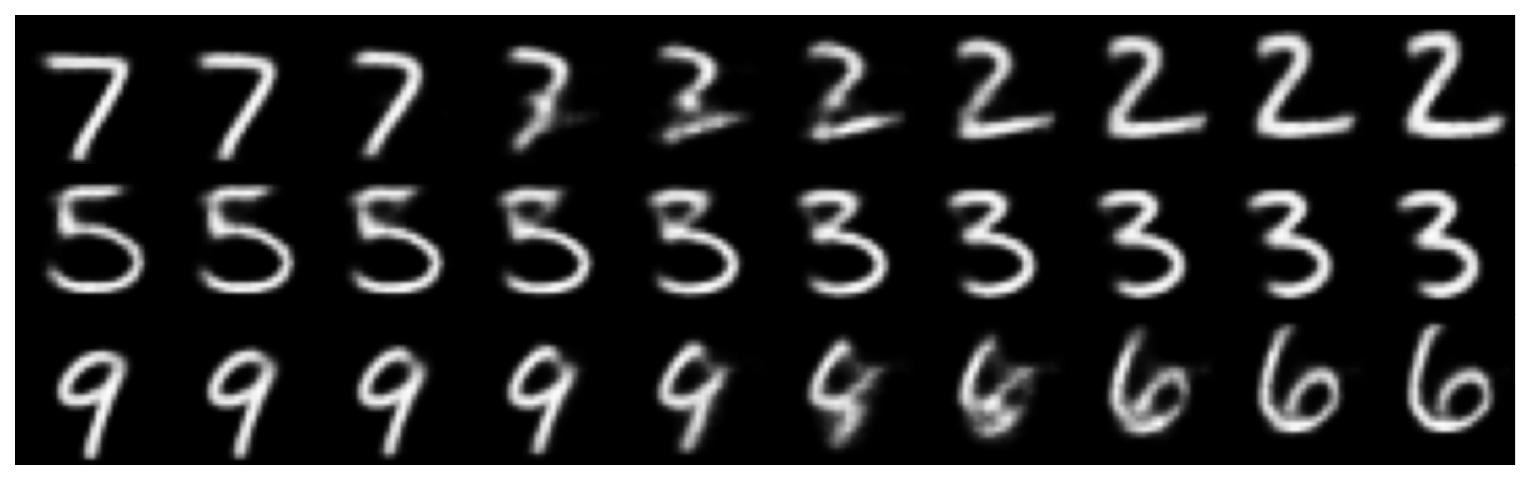}}
    \subfloat{\includegraphics[width=2.3in]{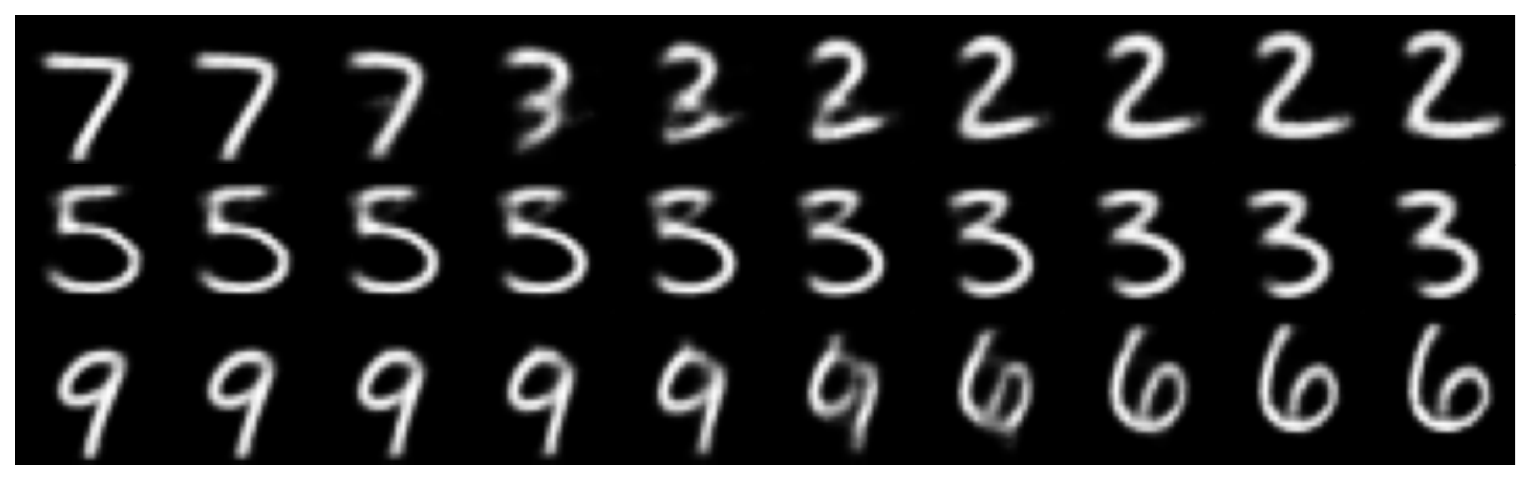}}\\\vspace{-1.2em}
    \adjustbox{minipage=6em,raise=\dimexpr -5.\height}{\small InfoVAE - RBF}
    \subfloat{\includegraphics[width=2.3in]{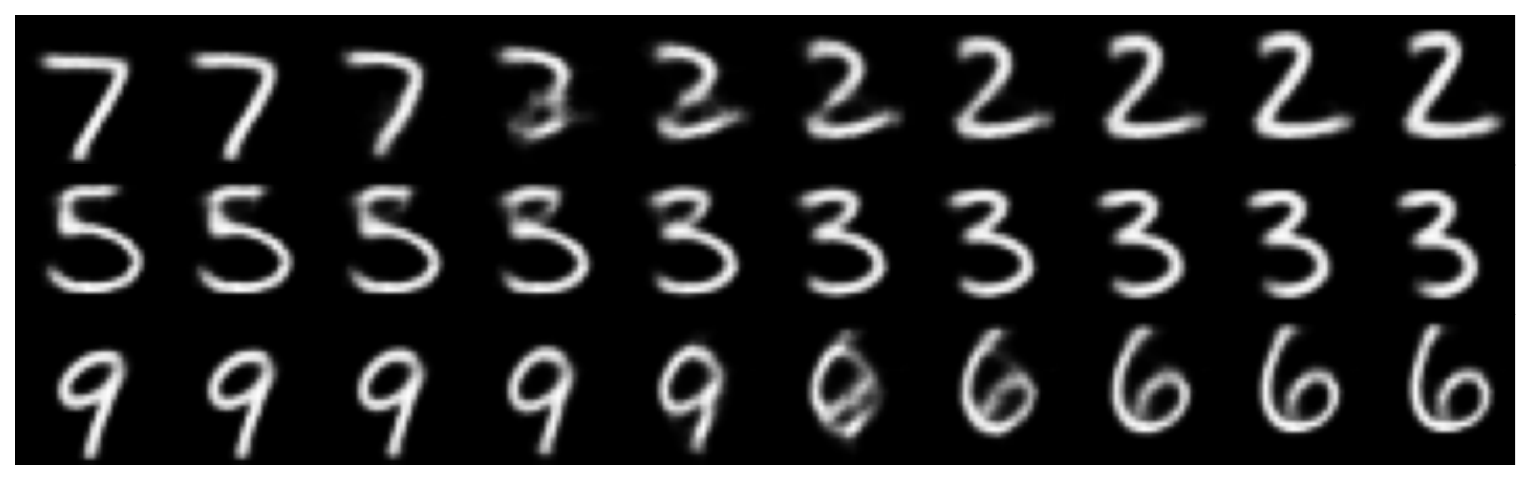}}
    \subfloat{\includegraphics[width=2.3in]{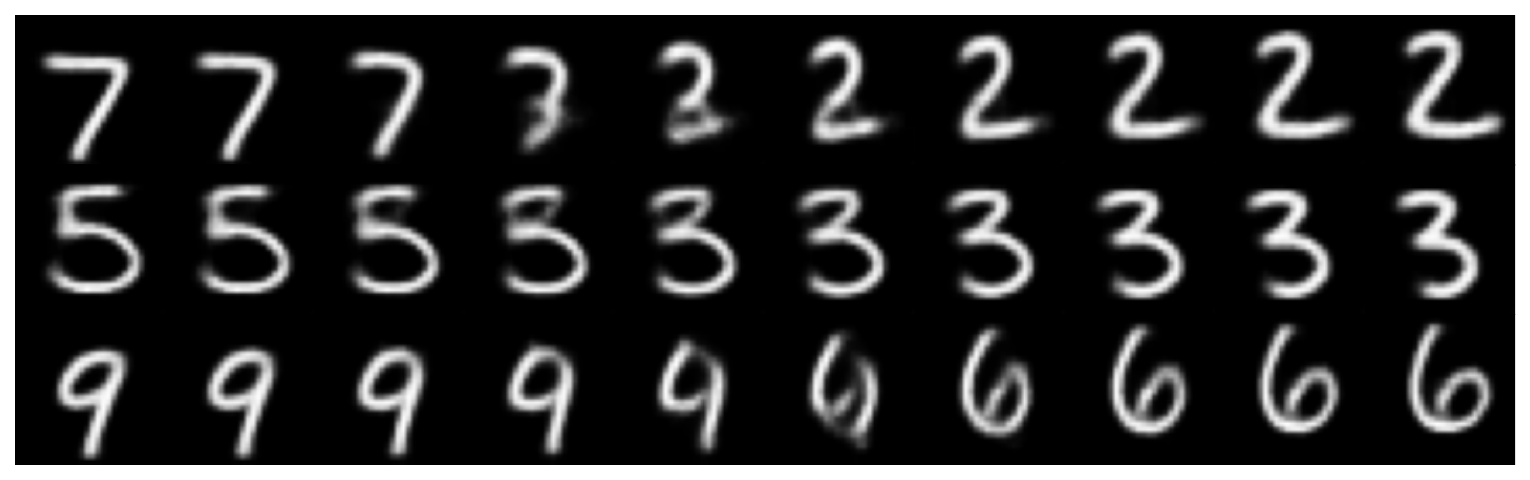}}
    \caption{Interpolations on MNIST with the same starting and ending images for latent spaces of dimension 16 and 256. For each model we select the configuration achieving the lowest FID on the generation task on the validation set with a GMM sampler.}
    \label{fig:interpolations mnist 1}
    \end{figure}

\begin{figure}[ht]
    \centering
    \captionsetup[subfigure]{position=above, labelformat = empty}
    \adjustbox{minipage=6em,raise=\dimexpr -5.\height}{\small AAE}
    \subfloat[MNIST (16)]{\includegraphics[width=2.3in]{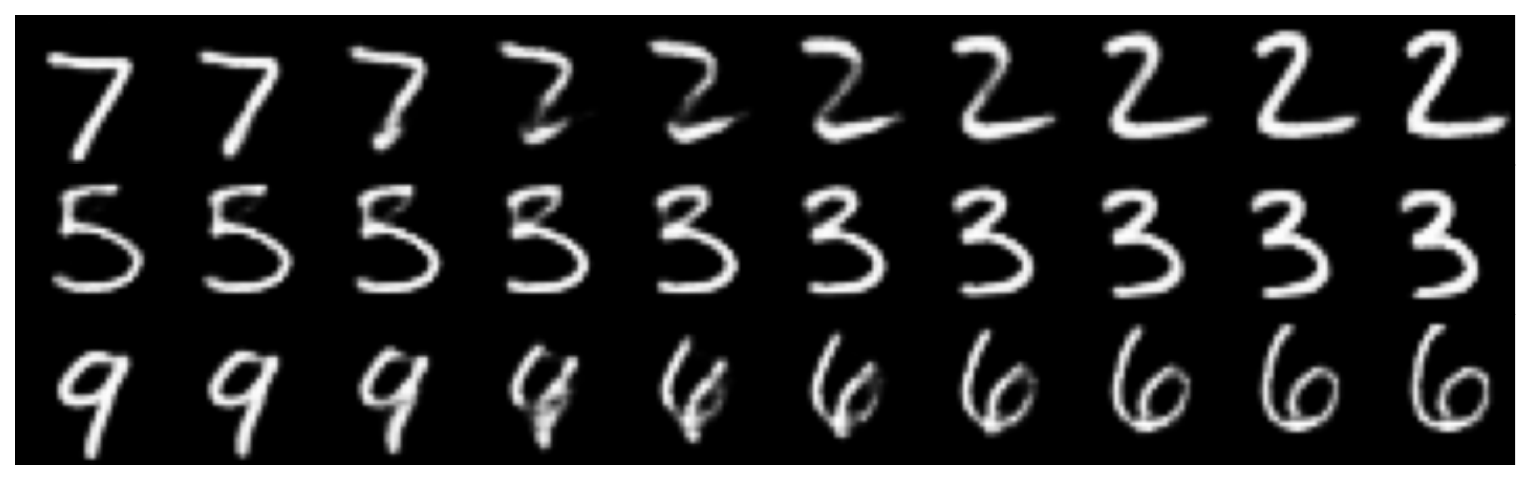}}
    \subfloat[MNIST (256)]{\includegraphics[width=2.3in]{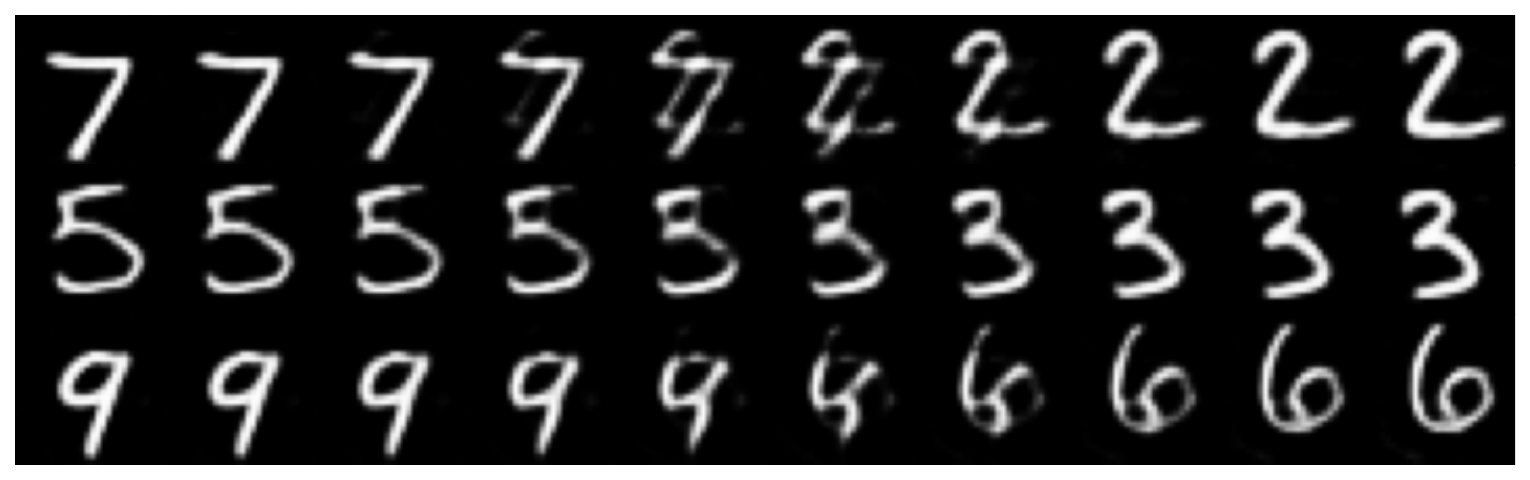}}\\\vspace{-1.2em}
    \adjustbox{minipage=6em,raise=\dimexpr -5.\height}{\small MSSSIM-VAE}
    \subfloat{\includegraphics[width=2.3in]{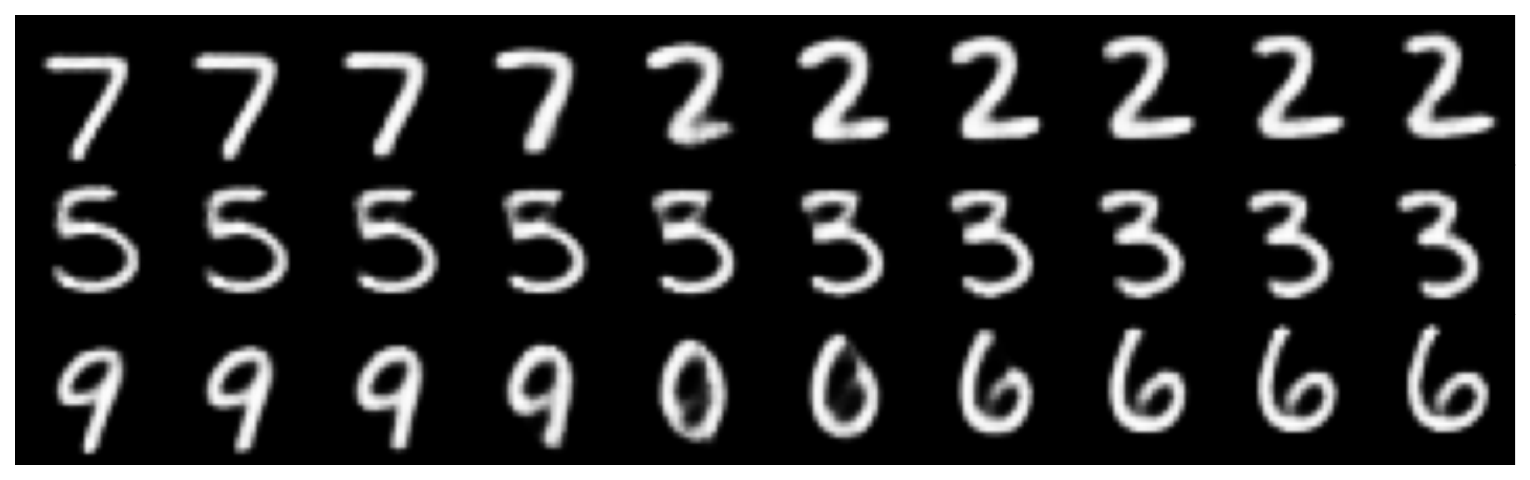}}
    \subfloat{\includegraphics[width=2.3in]{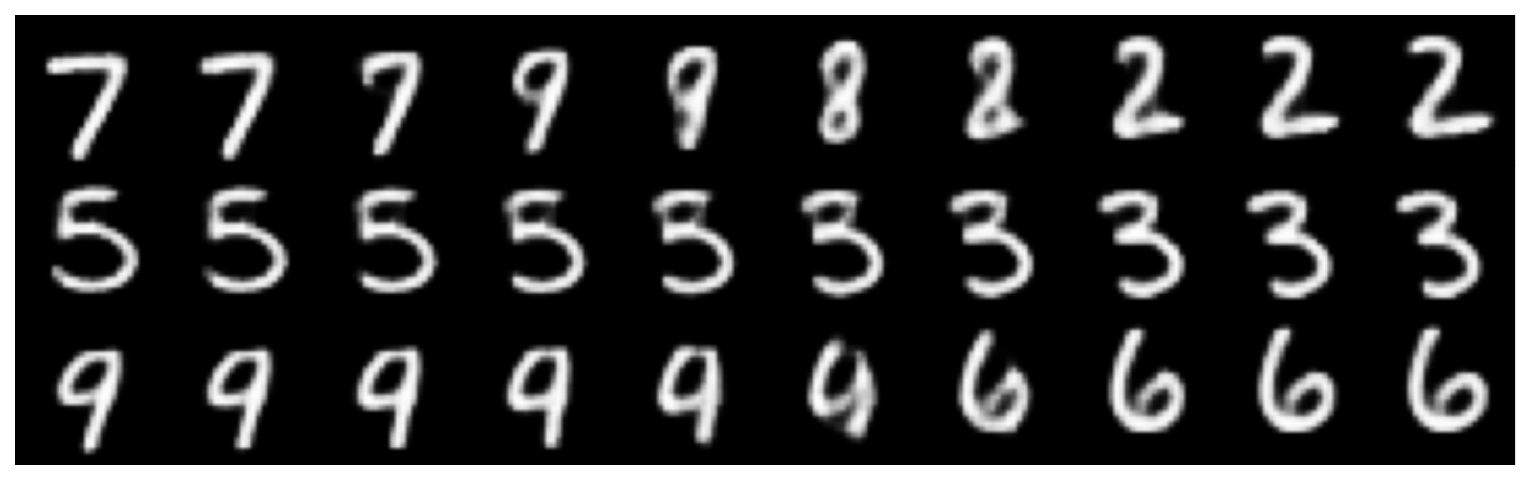}}\\\vspace{-1.2em}
    \adjustbox{minipage=6em,raise=\dimexpr -5.\height}{\small VAEGAN}
    \subfloat{\includegraphics[width=2.3in]{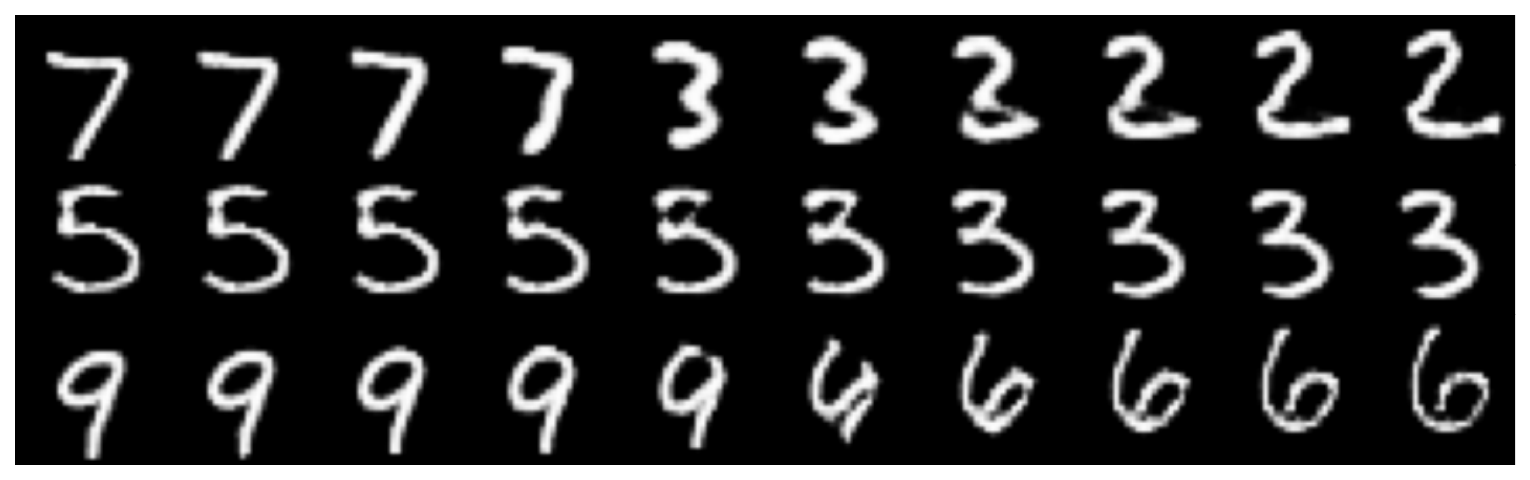}}
    \subfloat{\includegraphics[width=2.3in]{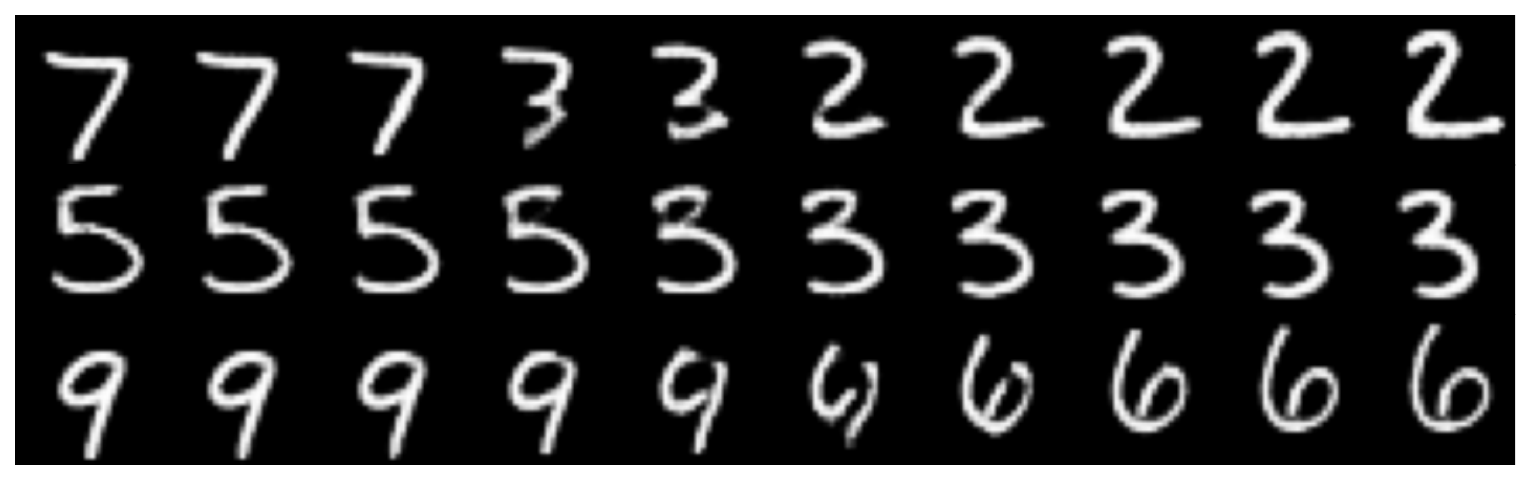}}\\\vspace{-1.2em}
    \adjustbox{minipage=6em,raise=\dimexpr -5.\height}{\small AE}
    \subfloat{\includegraphics[width=2.3in]{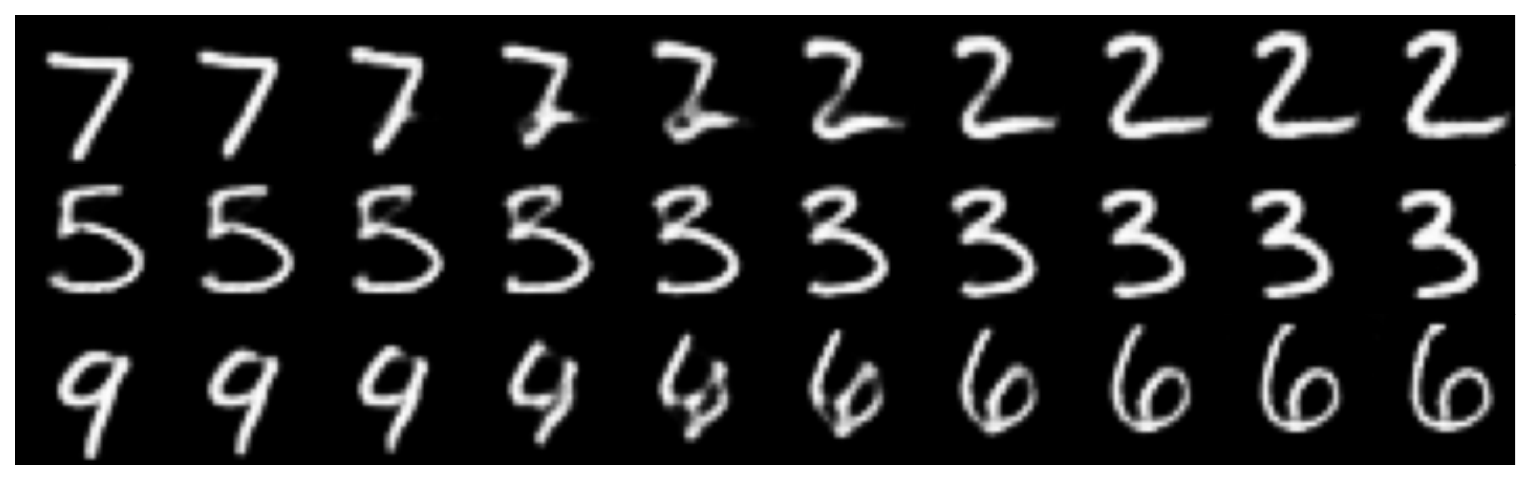}}
    \subfloat{\includegraphics[width=2.3in]{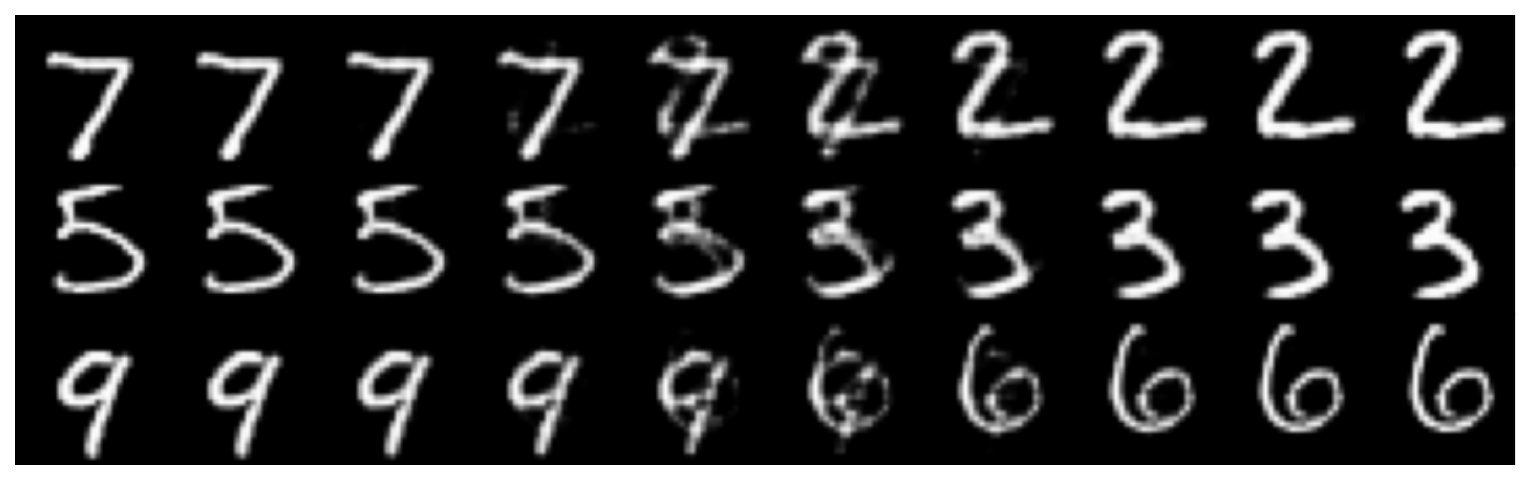}}\\\vspace{-1.2em}
    \adjustbox{minipage=6em,raise=\dimexpr -5.\height}{\small WAE-IMQ}
    \subfloat{\includegraphics[width=2.3in]{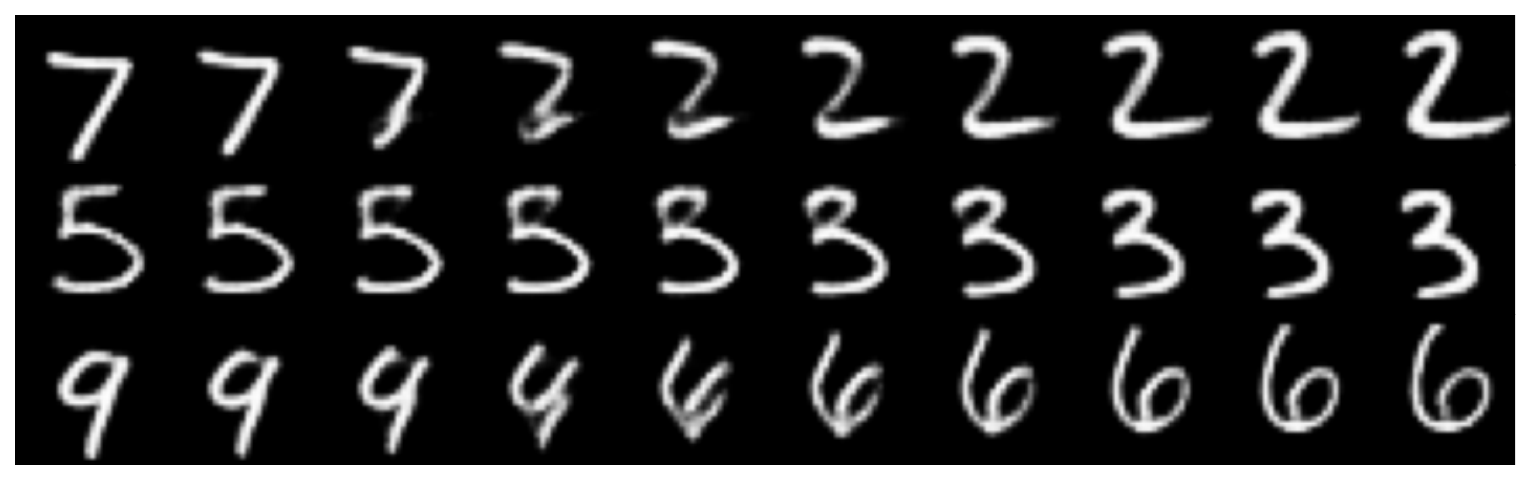}}
    \subfloat{\includegraphics[width=2.3in]{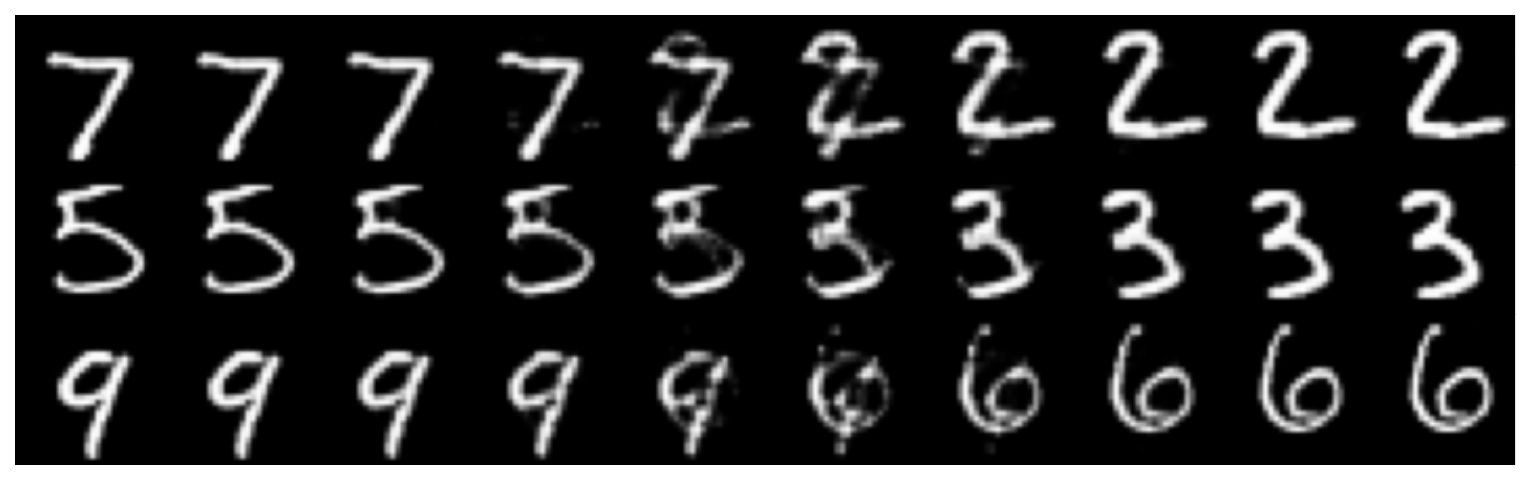}}\\\vspace{-1.2em}
    \adjustbox{minipage=6em,raise=\dimexpr -5.\height}{\small WAE-RBF}
    \subfloat{\includegraphics[width=2.3in]{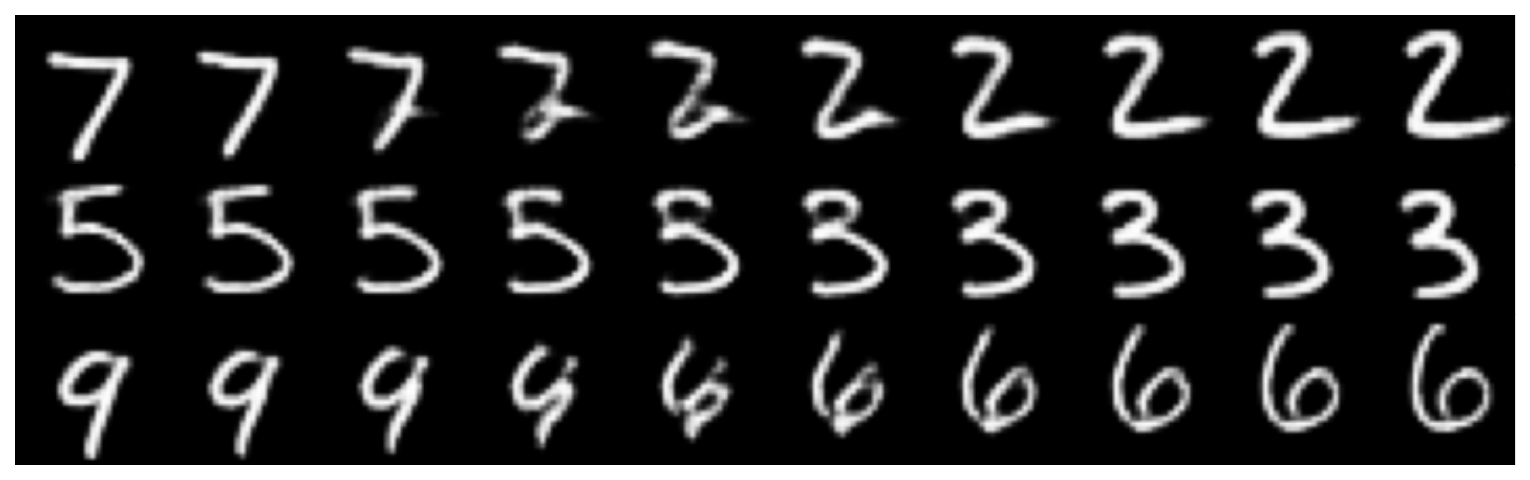}}
    \subfloat{\includegraphics[width=2.3in]{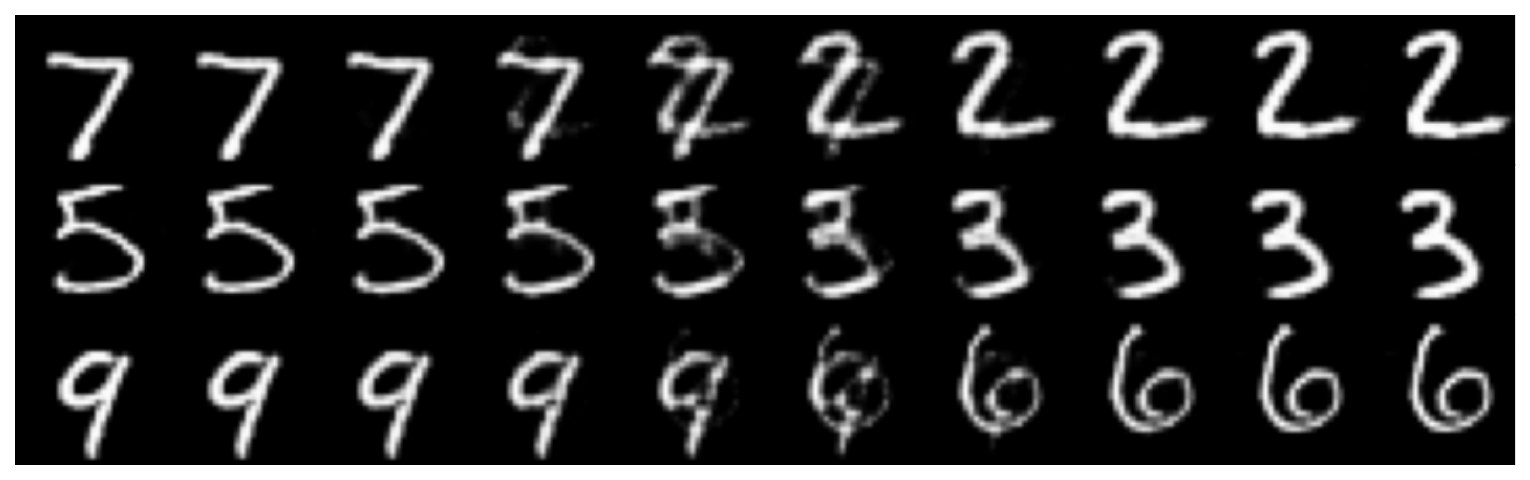}}\\\vspace{-1.2em}
    \adjustbox{minipage=6em,raise=\dimexpr -5.\height}{\small VQVAE}
    \subfloat{\includegraphics[width=2.3in]{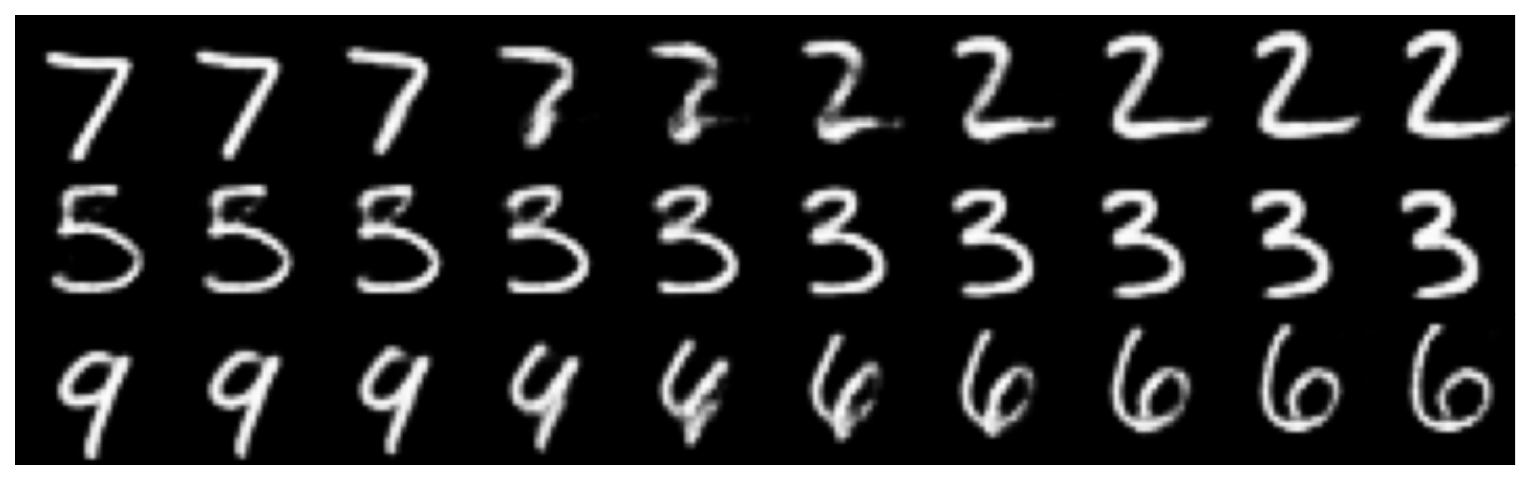}}
    \subfloat{\includegraphics[width=2.3in]{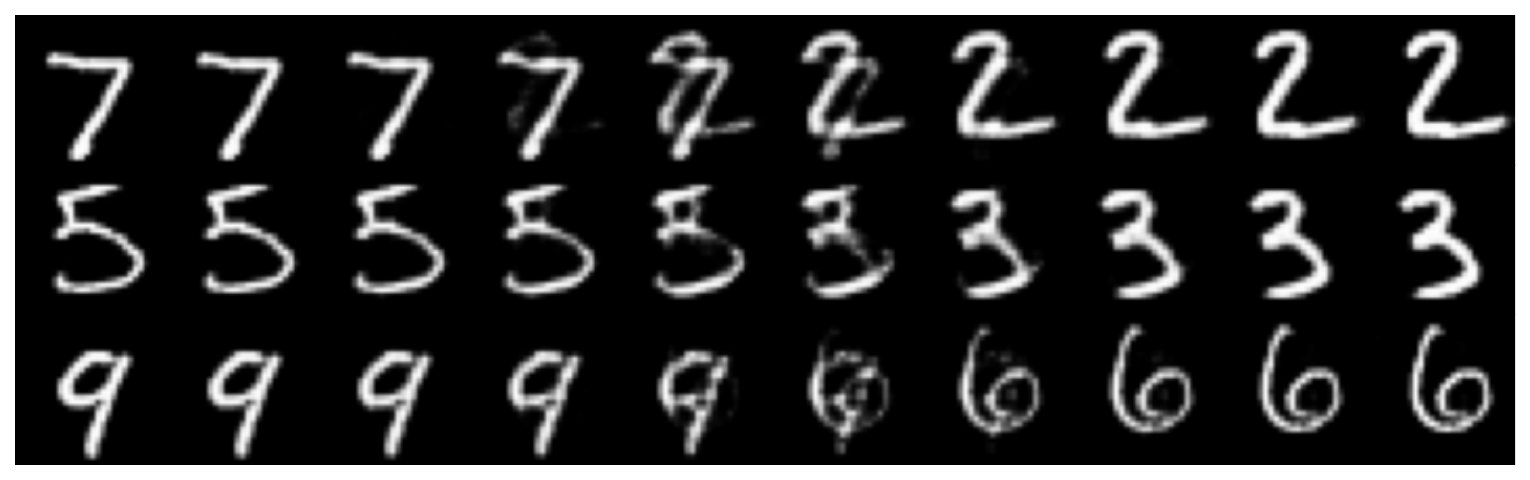}}\\\vspace{-1.2em}
    \adjustbox{minipage=6em,raise=\dimexpr -5.\height}{\small RAE-l2}
    \subfloat{\includegraphics[width=2.3in]{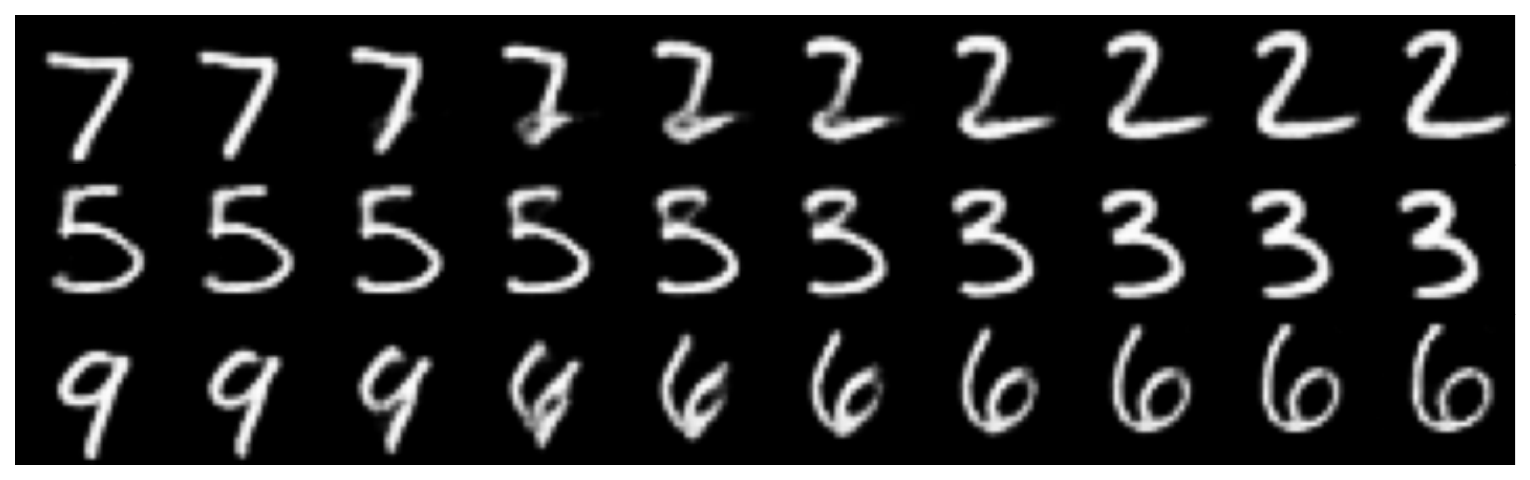}}
    \subfloat{\includegraphics[width=2.3in]{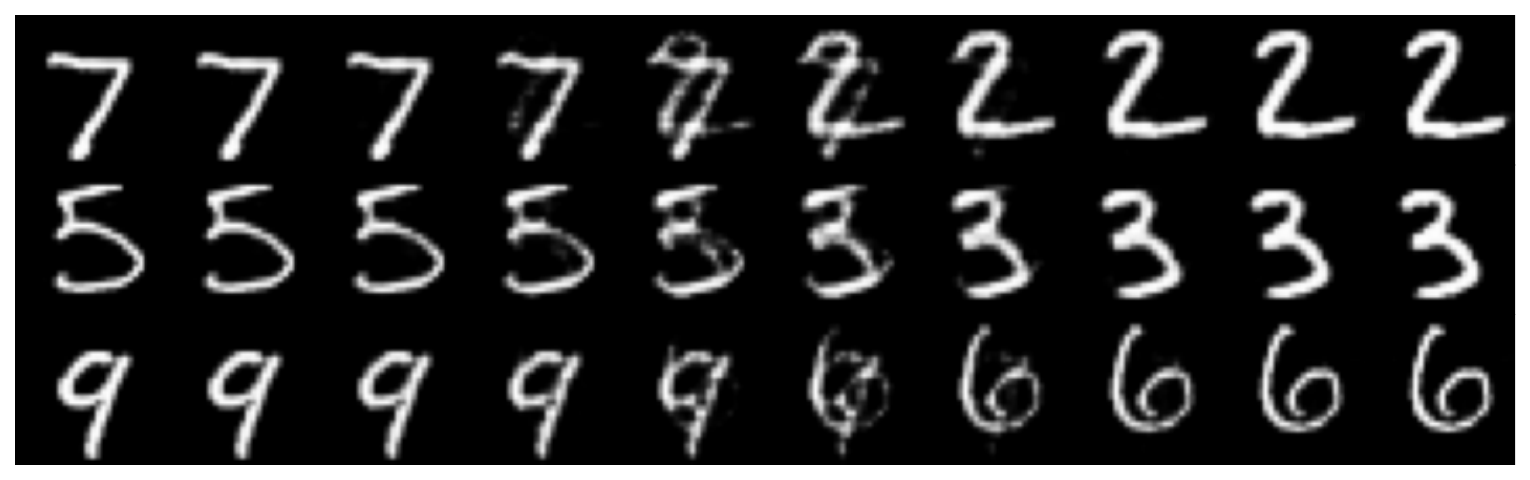}}\\\vspace{-1.2em}
    \adjustbox{minipage=6em,raise=\dimexpr -5.\height}{\small RAE-GP}
    \subfloat{\includegraphics[width=2.3in]{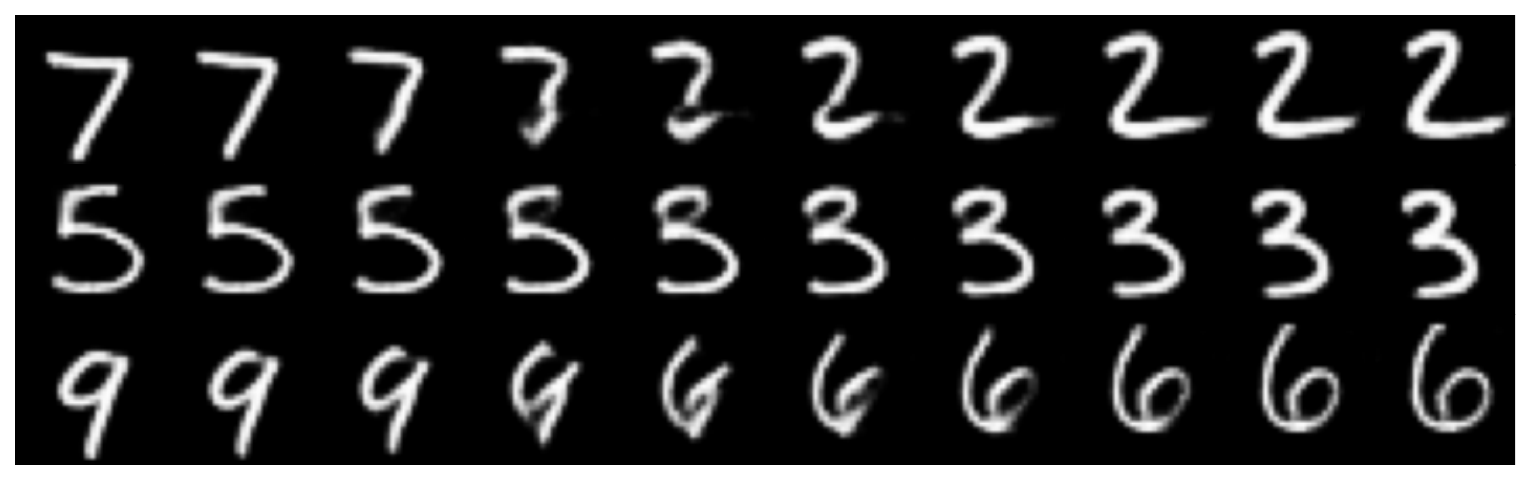}}
    \subfloat{\includegraphics[width=2.3in]{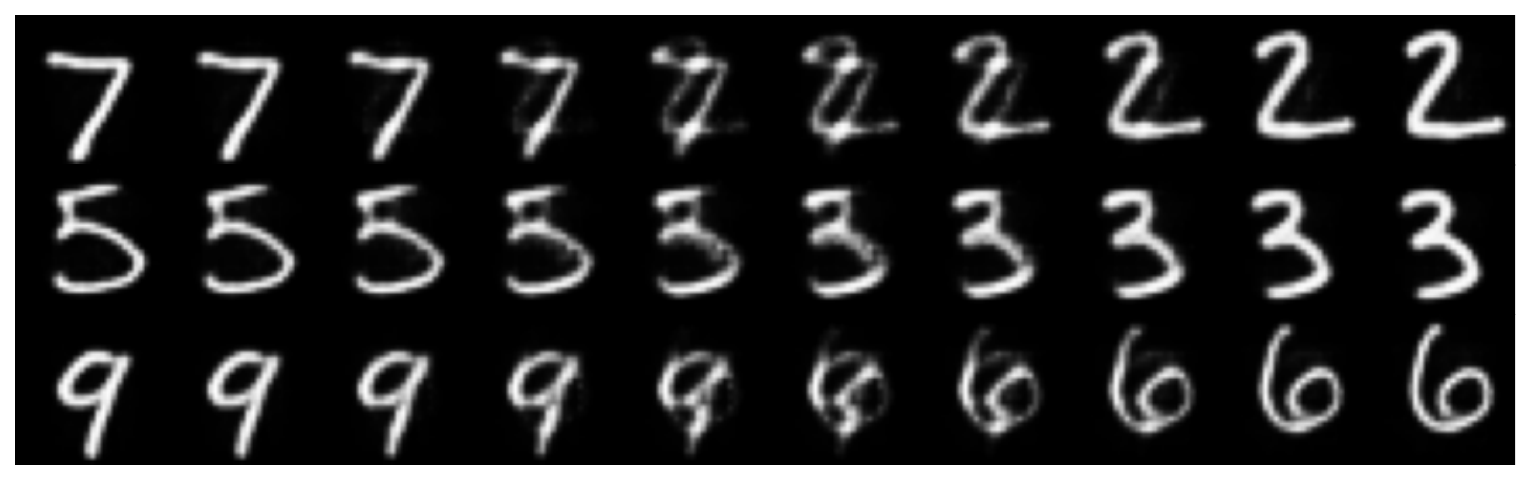}}
    \caption{Interpolations on MNIST with the same starting and ending images for latent spaces of dimension 16 and 256. For each model we select the configuration achieving the lowest FID on the generation task on the validation set with a GMM sampler.}
    \label{fig:interpolations mnist 2}
    \end{figure}

\begin{figure}[ht]
    \centering
    \captionsetup[subfigure]{position=above, labelformat = empty}
    \adjustbox{minipage=6em,raise=\dimexpr -5.\height}{\small AAE}
    \subfloat[CIFAR10 (32)]{\includegraphics[width=2.3in]{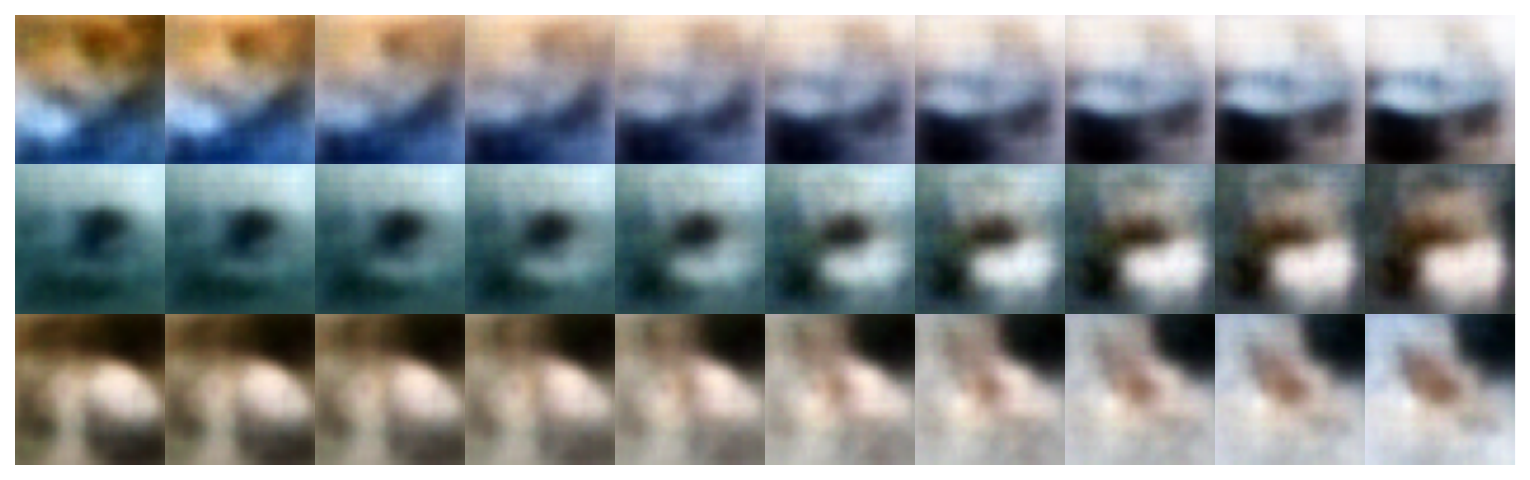}}
    \subfloat[CIFAR10 (256)]{\includegraphics[width=2.3in]{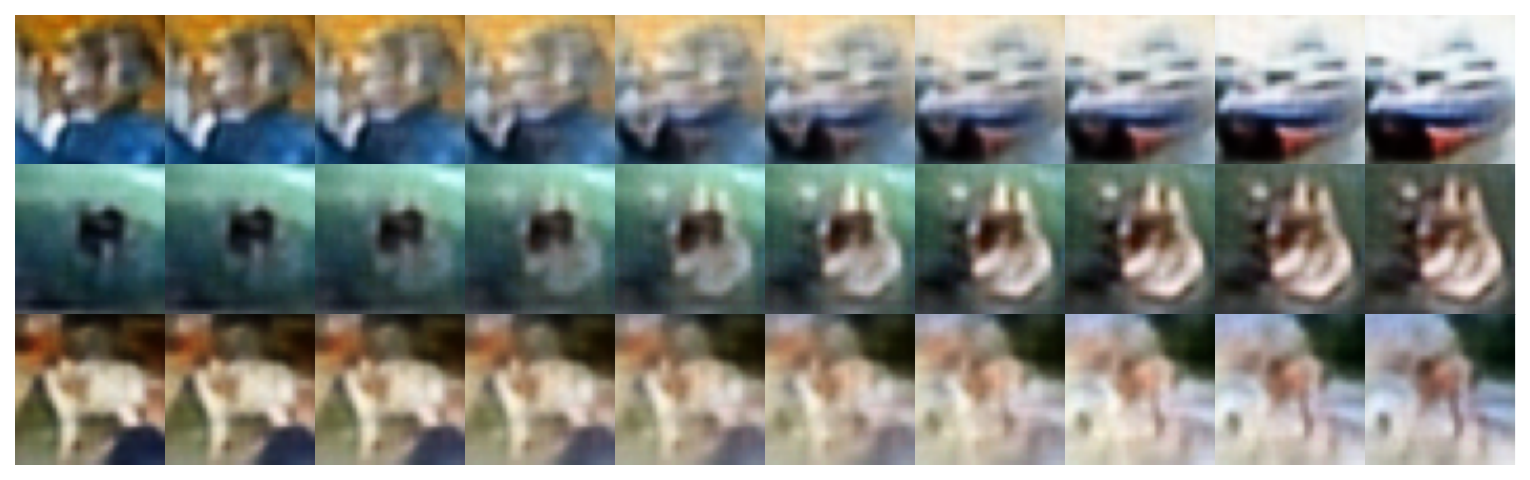}}\\\vspace{-1.2em}
    \adjustbox{minipage=6em,raise=\dimexpr -5.\height}{\small MSSSIM-VAE}
    \subfloat{\includegraphics[width=2.3in]{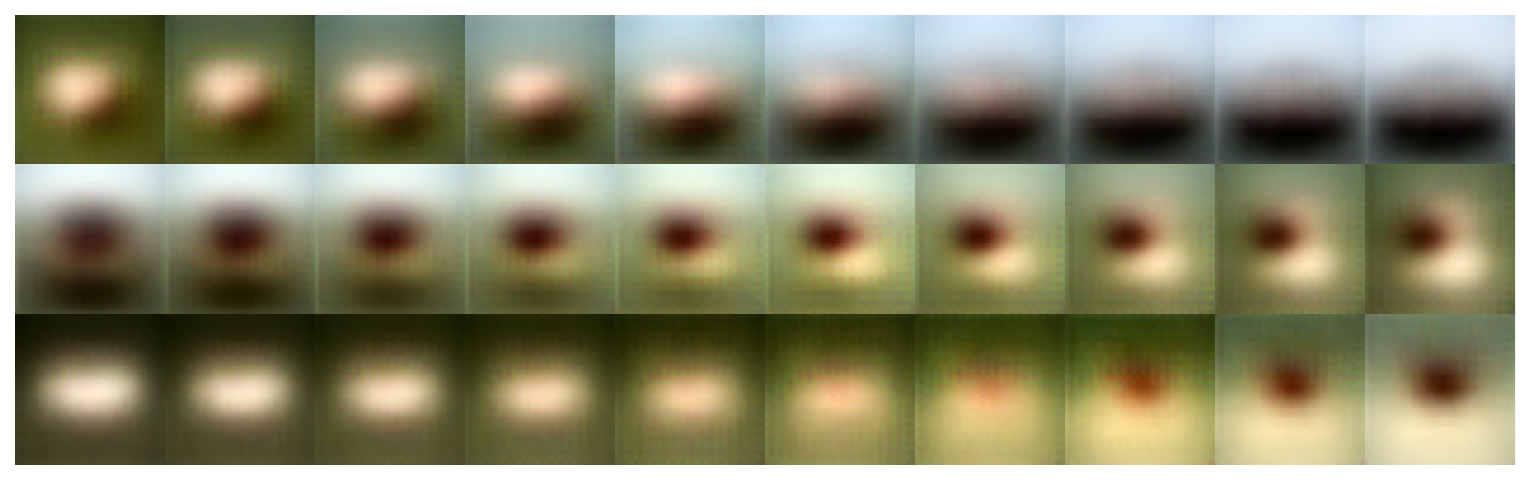}}
    \subfloat{\includegraphics[width=2.3in]{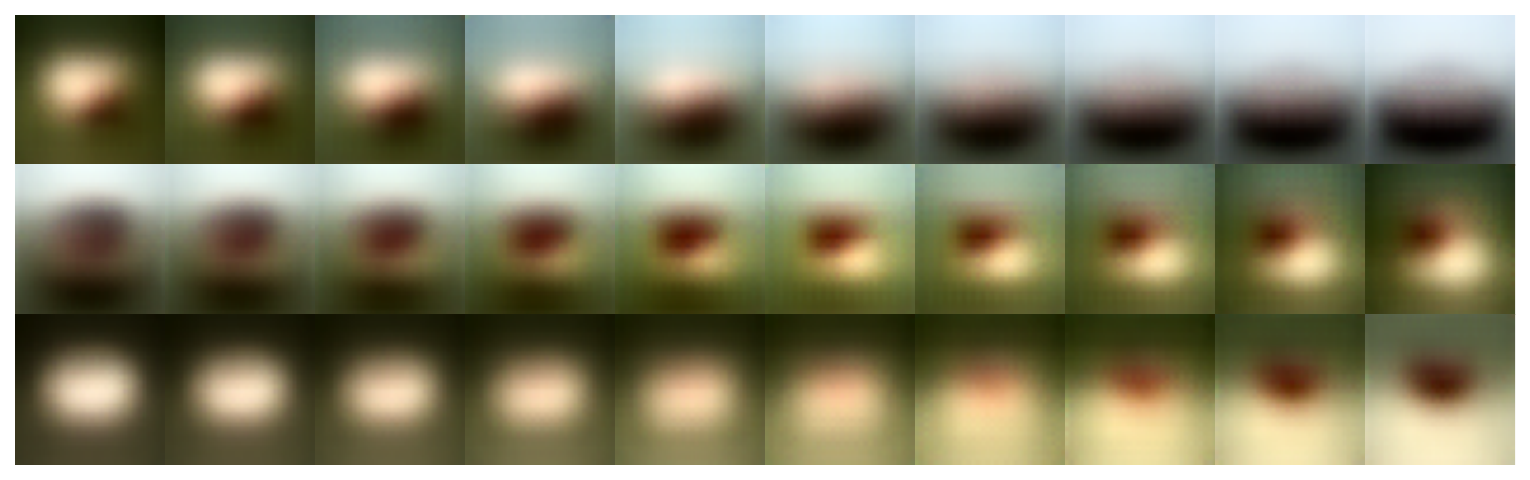}}\\\vspace{-1.2em}
    \adjustbox{minipage=6em,raise=\dimexpr -5.\height}{\small VAEGAN}
    \subfloat{\includegraphics[width=2.3in]{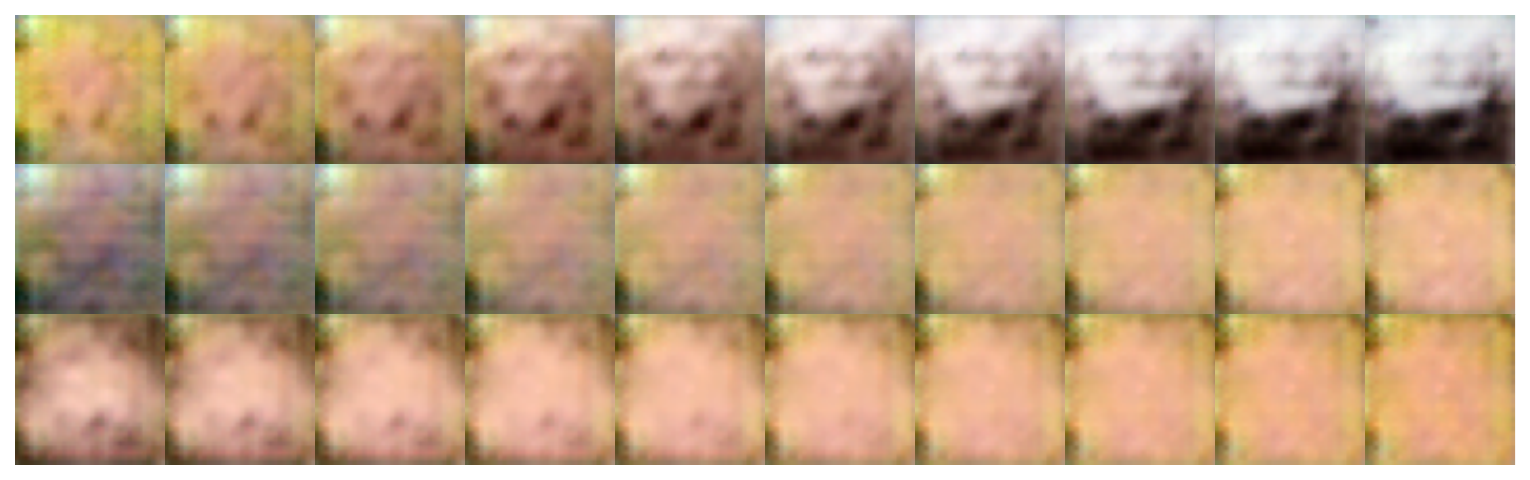}}
    \subfloat{\includegraphics[width=2.3in]{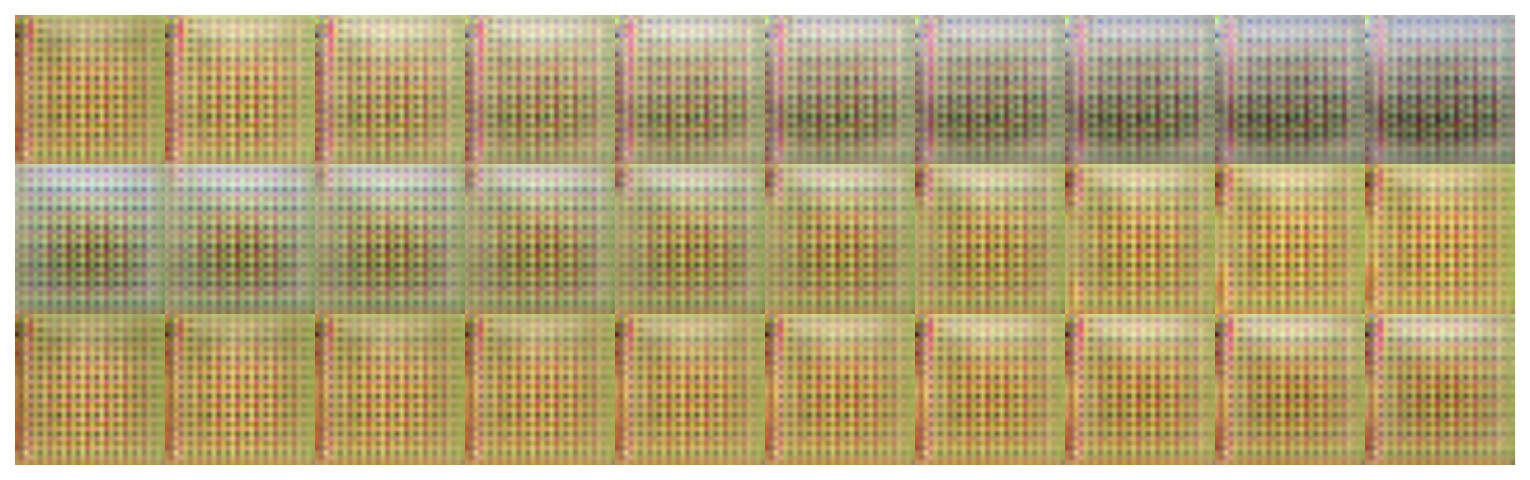}}\\\vspace{-1.2em}
    \adjustbox{minipage=6em,raise=\dimexpr -5.\height}{\small AE}
    \subfloat{\includegraphics[width=2.3in]{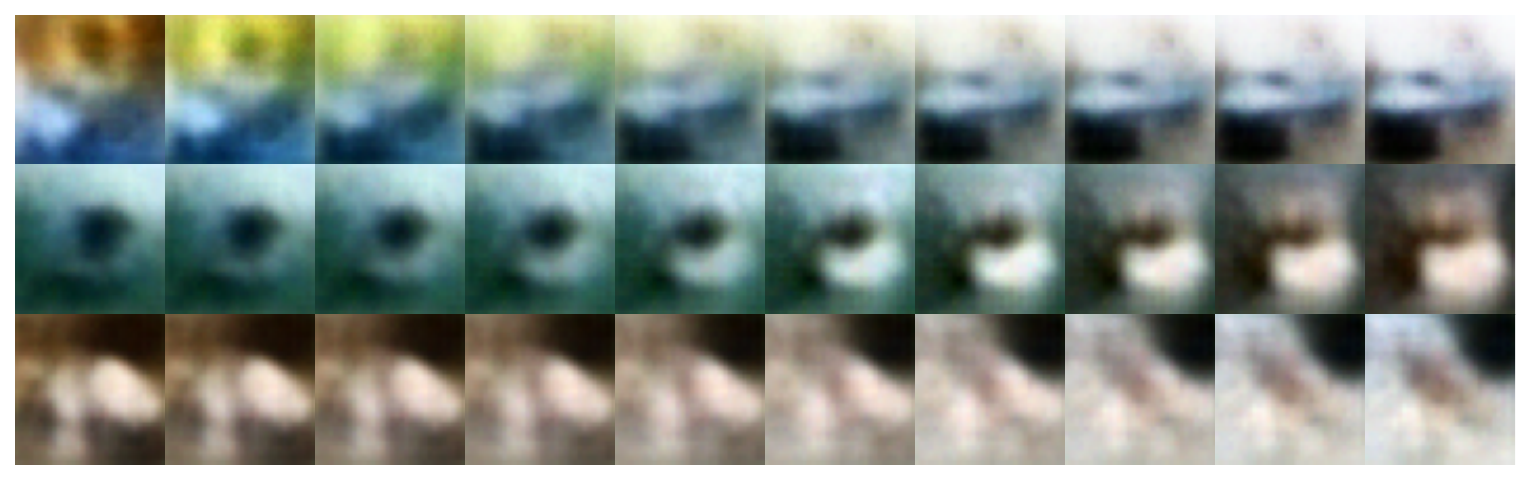}}
    \subfloat{\includegraphics[width=2.3in]{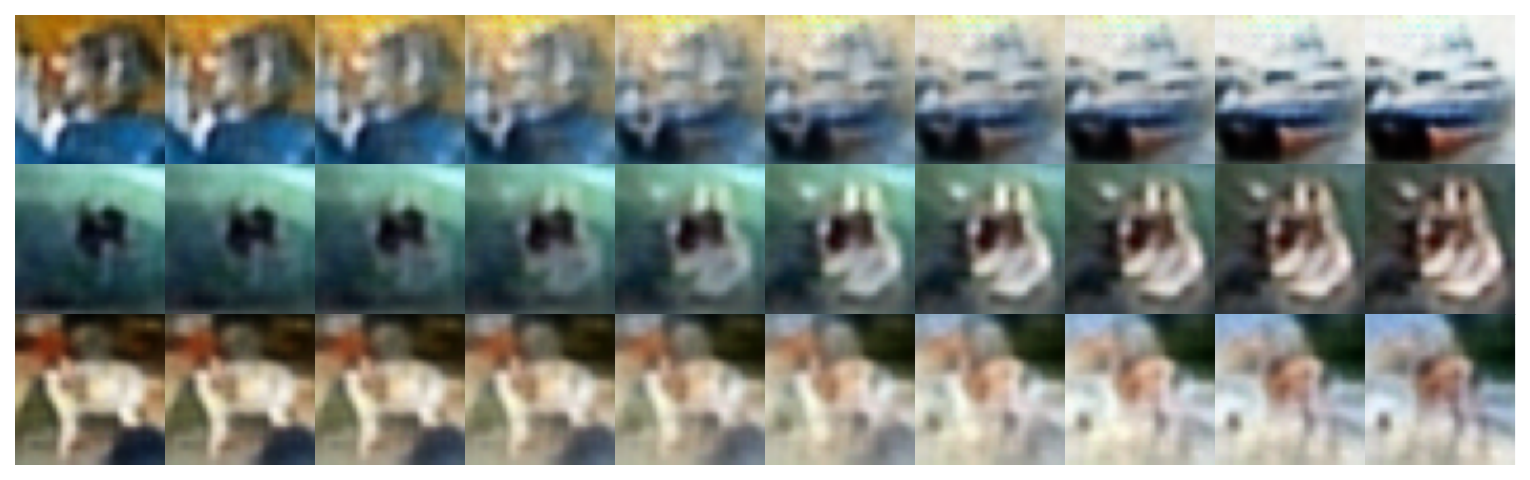}}\\\vspace{-1.2em}
    \adjustbox{minipage=6em,raise=\dimexpr -5.\height}{\small WAE-IMQ}
    \subfloat{\includegraphics[width=2.3in]{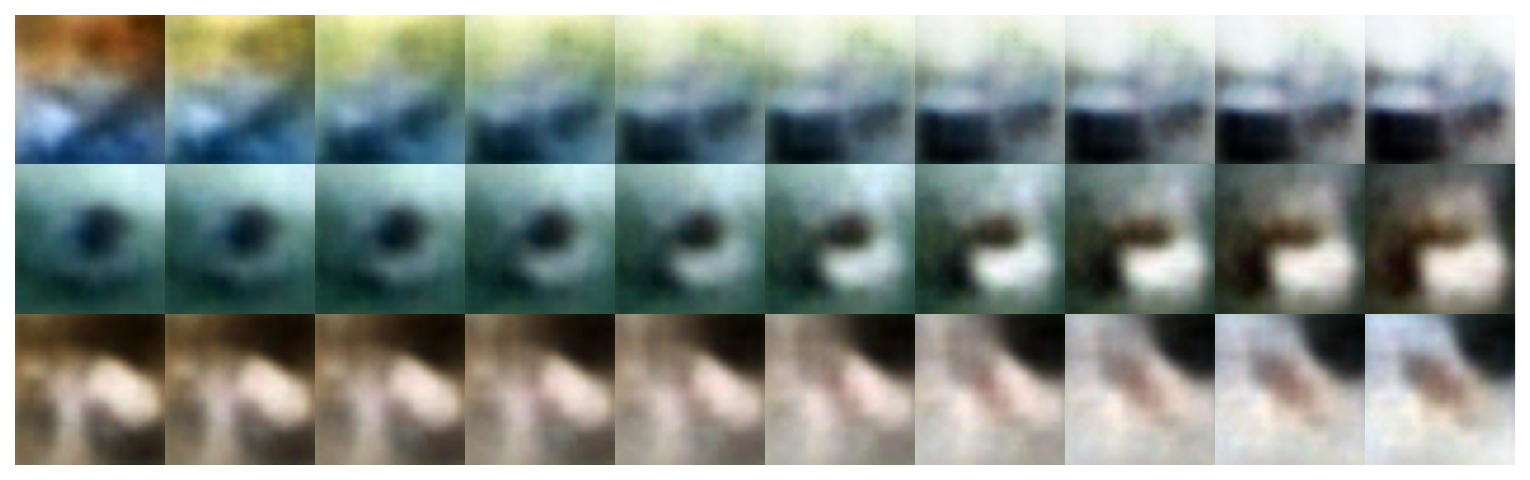}}
    \subfloat{\includegraphics[width=2.3in]{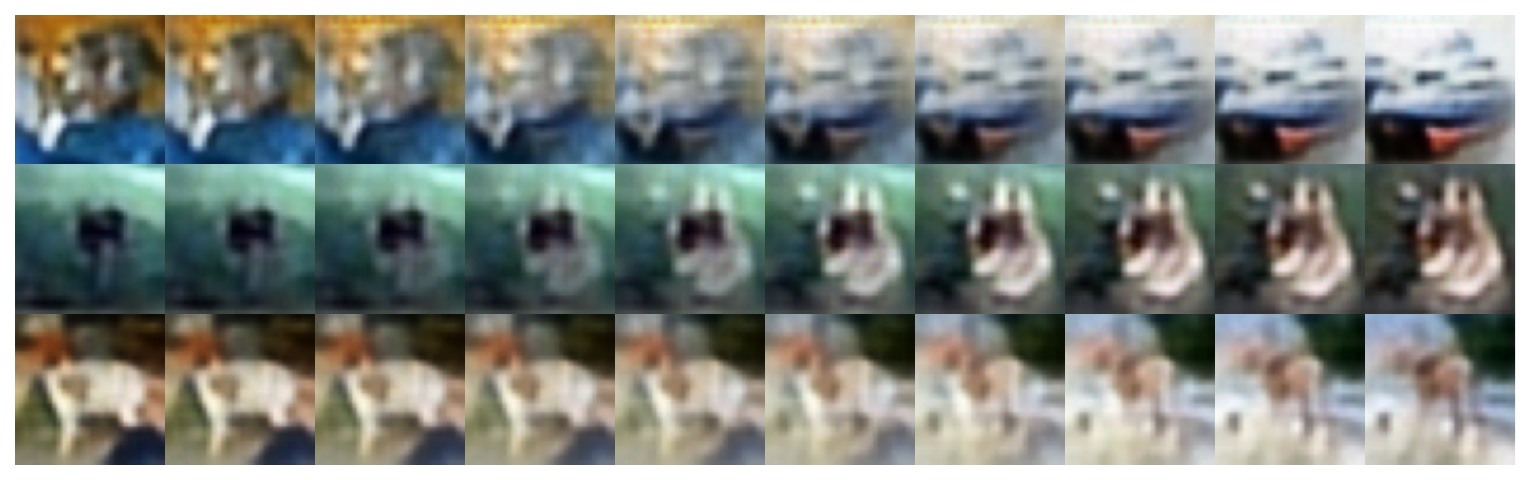}}\\\vspace{-1.2em}
    \adjustbox{minipage=6em,raise=\dimexpr -5.\height}{\small WAE-RBF}
    \subfloat{\includegraphics[width=2.3in]{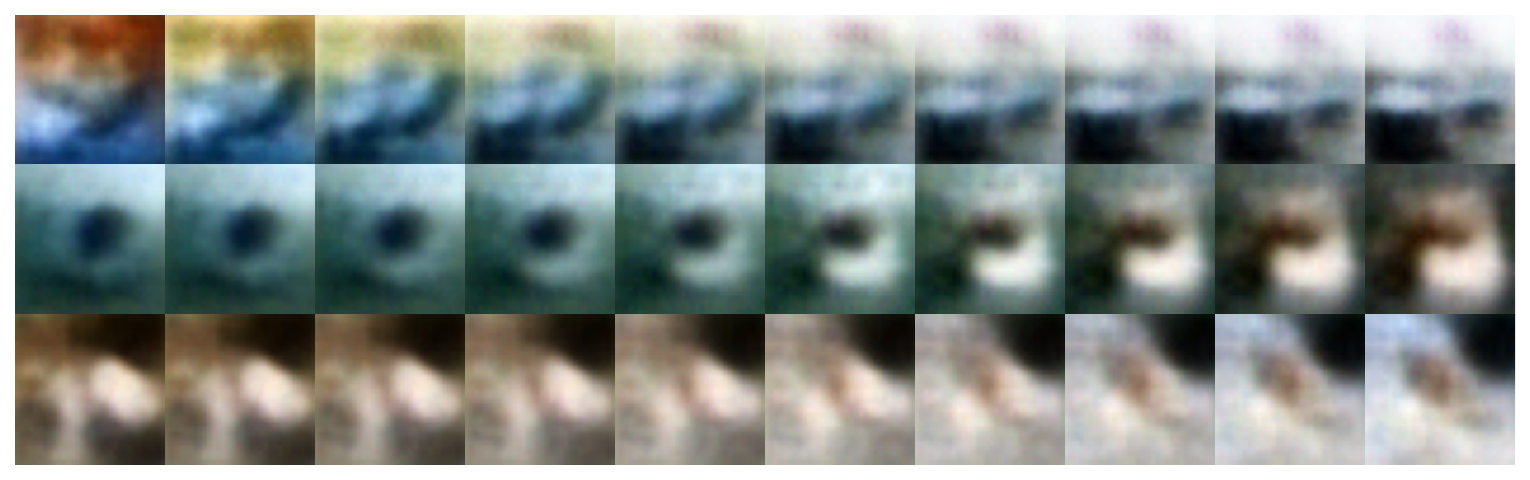}}
    \subfloat{\includegraphics[width=2.3in]{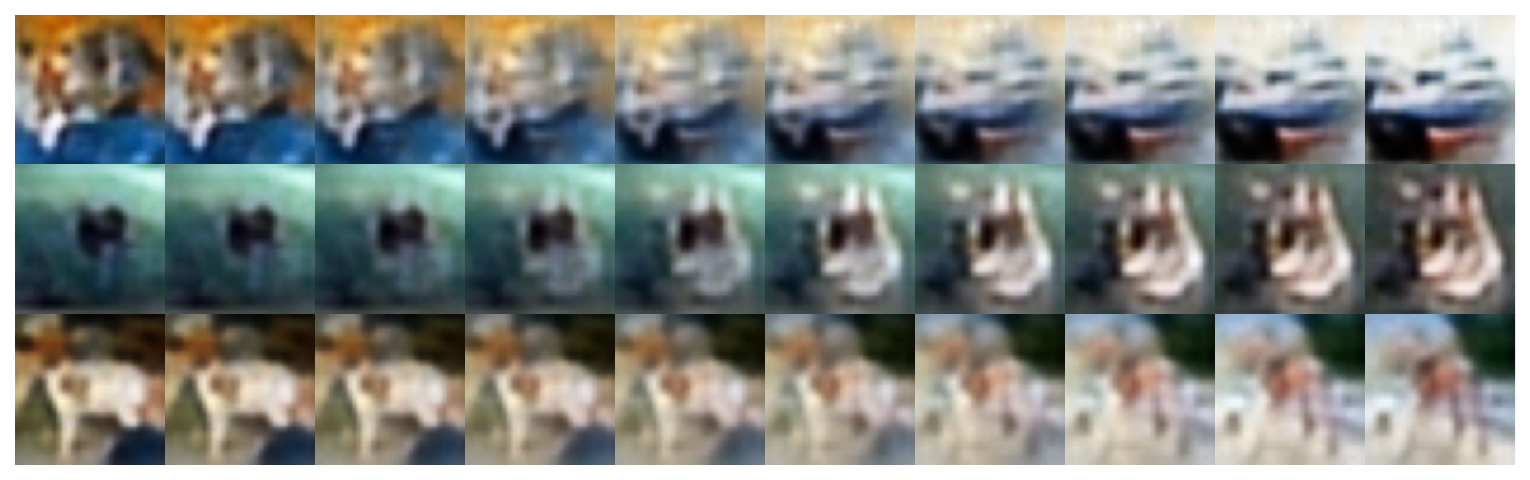}}\\\vspace{-1.2em}
    \adjustbox{minipage=6em,raise=\dimexpr -5.\height}{\small VQVAE}
    \subfloat{\includegraphics[width=2.3in]{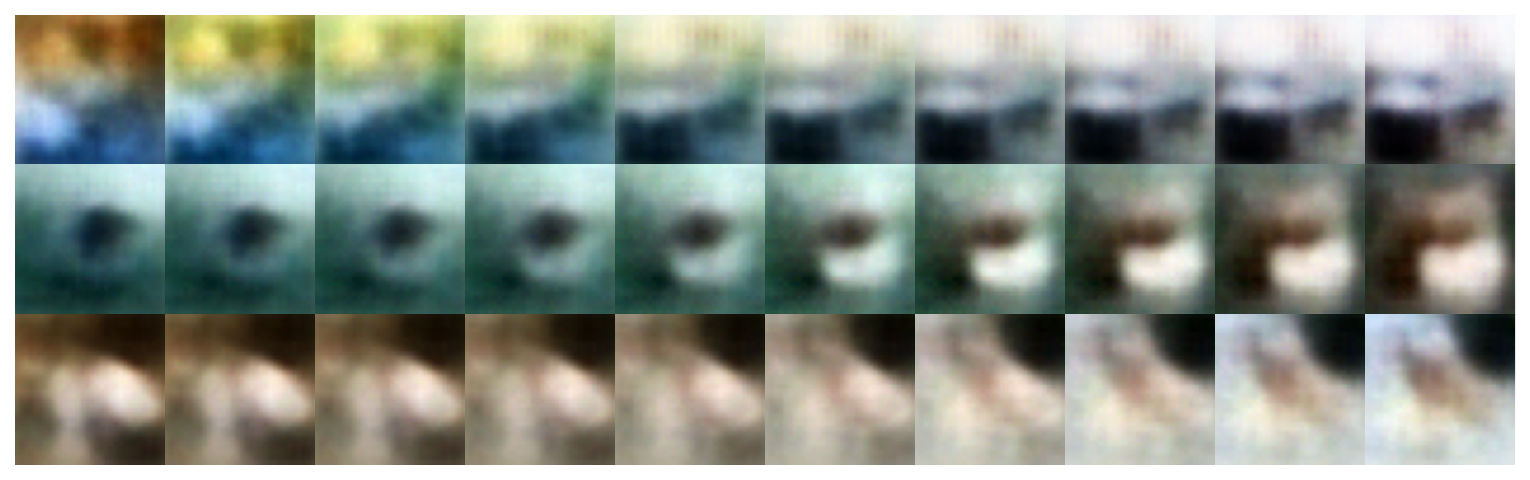}}
    \subfloat{\includegraphics[width=2.3in]{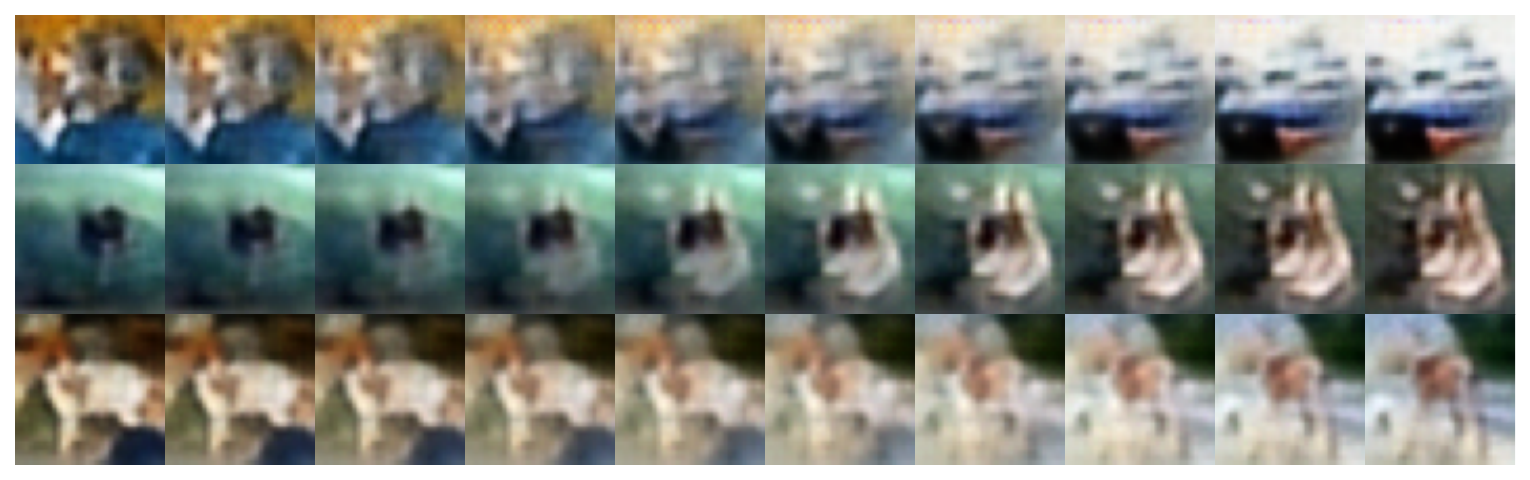}}\\\vspace{-1.2em}
    \adjustbox{minipage=6em,raise=\dimexpr -5.\height}{\small RAE-l2}
    \subfloat{\includegraphics[width=2.3in]{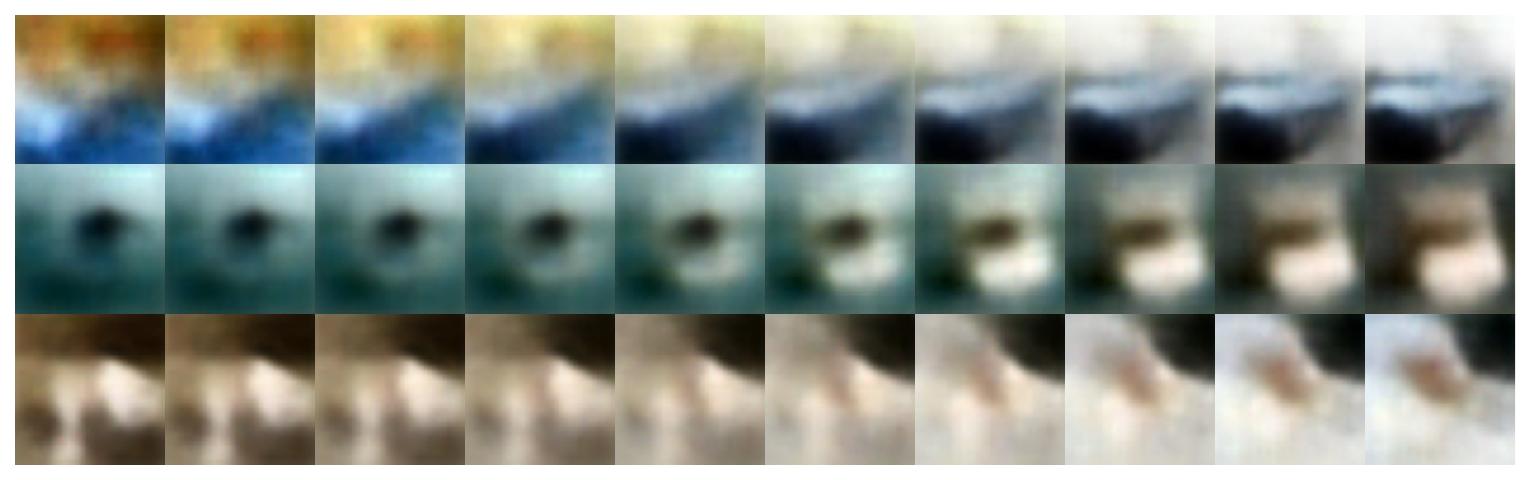}}
    \subfloat{\includegraphics[width=2.3in]{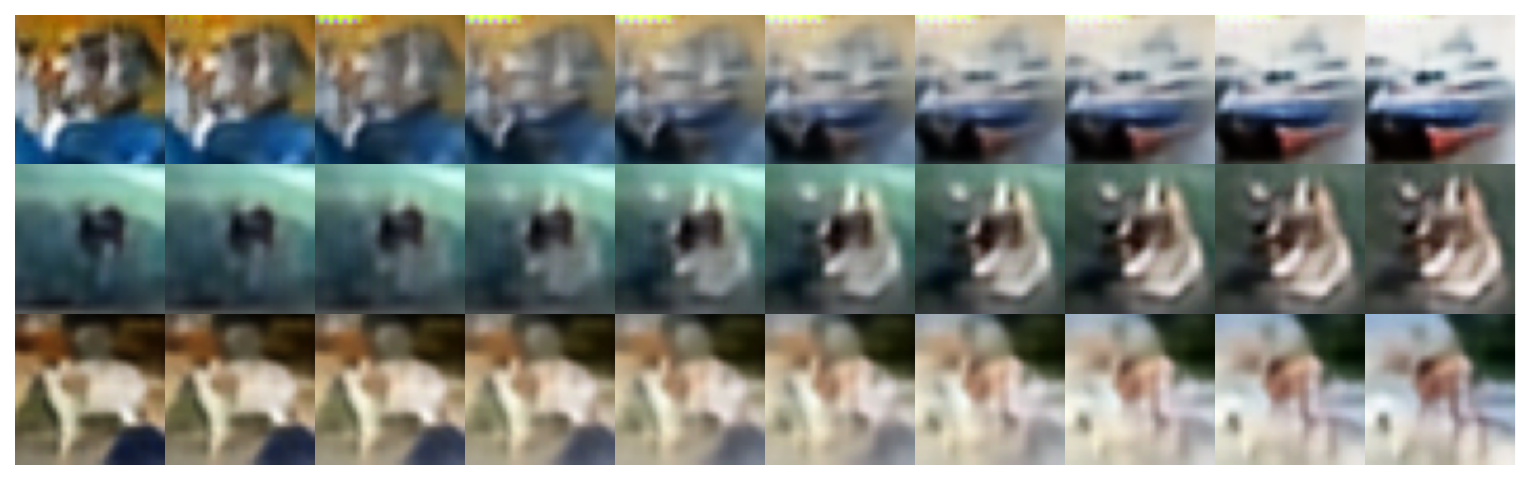}}\\\vspace{-1.2em}
    \adjustbox{minipage=6em,raise=\dimexpr -5.\height}{\small RAE-GP}
    \subfloat{\includegraphics[width=2.3in]{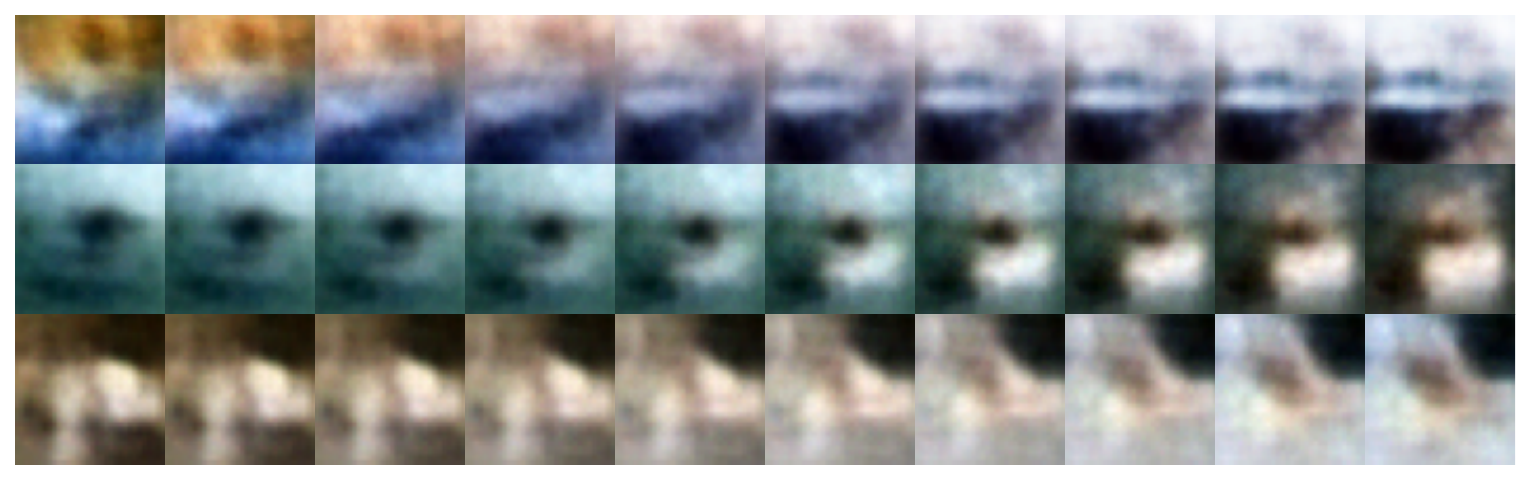}}
    \subfloat{\includegraphics[width=2.3in]{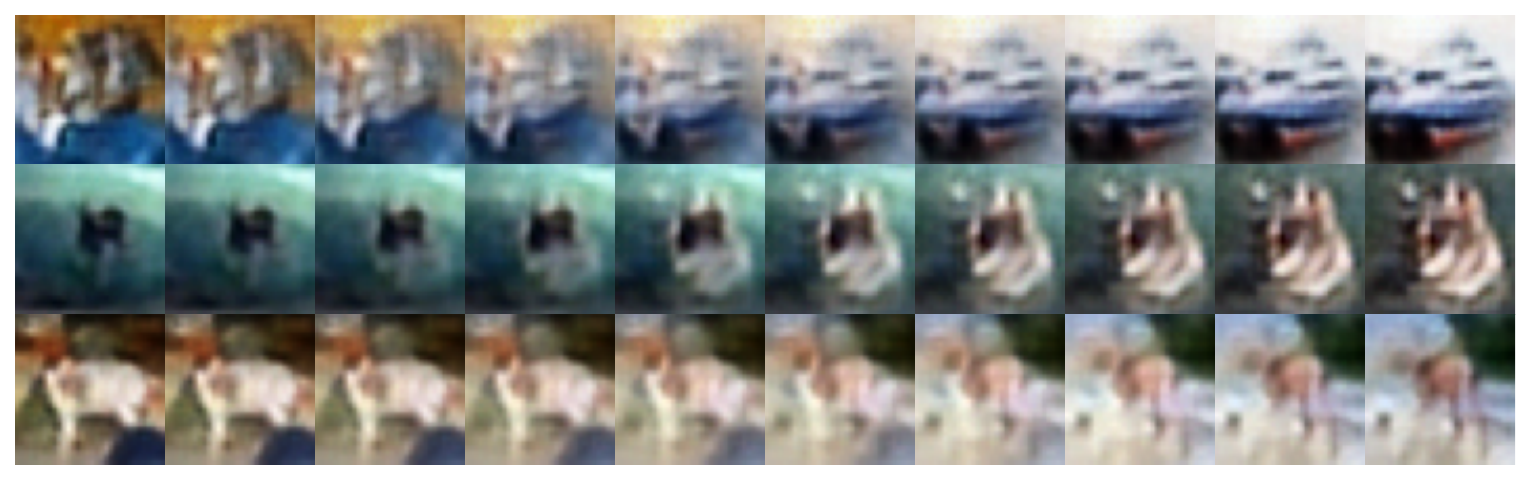}}
    \caption{Interpolations on CIFAR10 with the same starting and ending images for latent spaces of dimension 32 and 256. For each model we select the configuration achieving the lowest FID on the generation task on the validation set with a GMM sampler.}
    \label{fig:interpolations cifar 2}
    \end{figure}

%%%%%%%%%%%%%%%%%%%%%%%%%%%%%%%%%%%

\begin{figure}[ht]
    \centering
    \captionsetup[subfigure]{position=above, labelformat = empty}
     \adjustbox{minipage=6em,raise=\dimexpr -5.\height}{\small VAE}
    \subfloat[CELEBA (64)]{\includegraphics[width=2.3in]{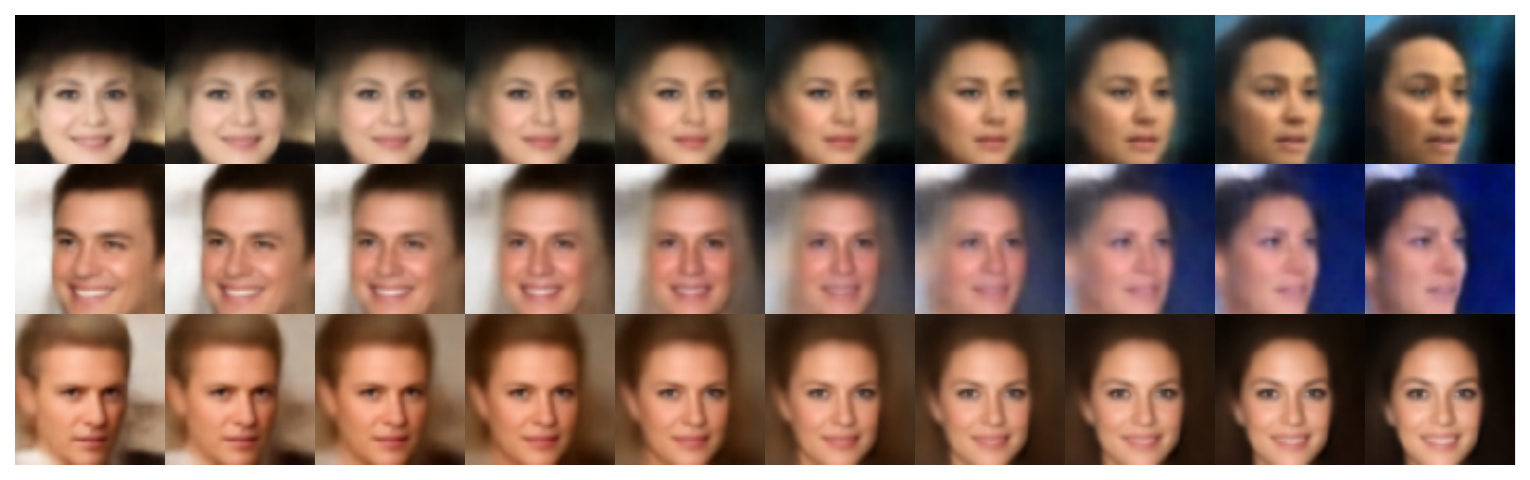}}
    \subfloat[CELEBA (64)]{\includegraphics[width=2.3in]{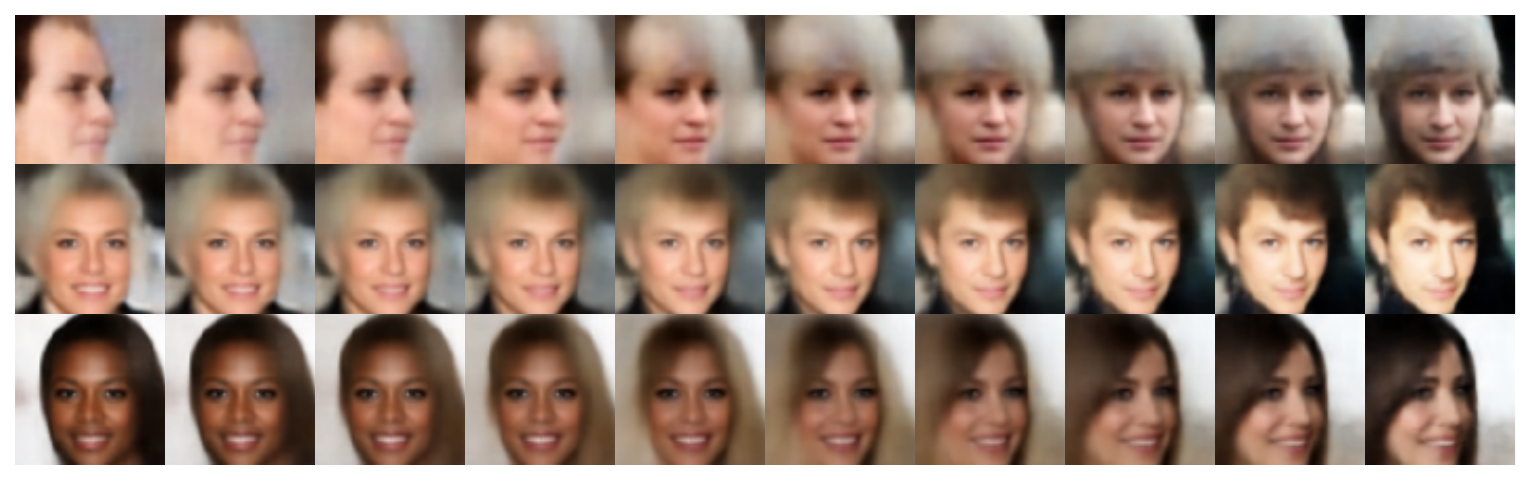}}\\\vspace{-1.2em}
    \adjustbox{minipage=6em,raise=\dimexpr -5.\height}{\small VAMP}
    \subfloat{\includegraphics[width=2.3in]{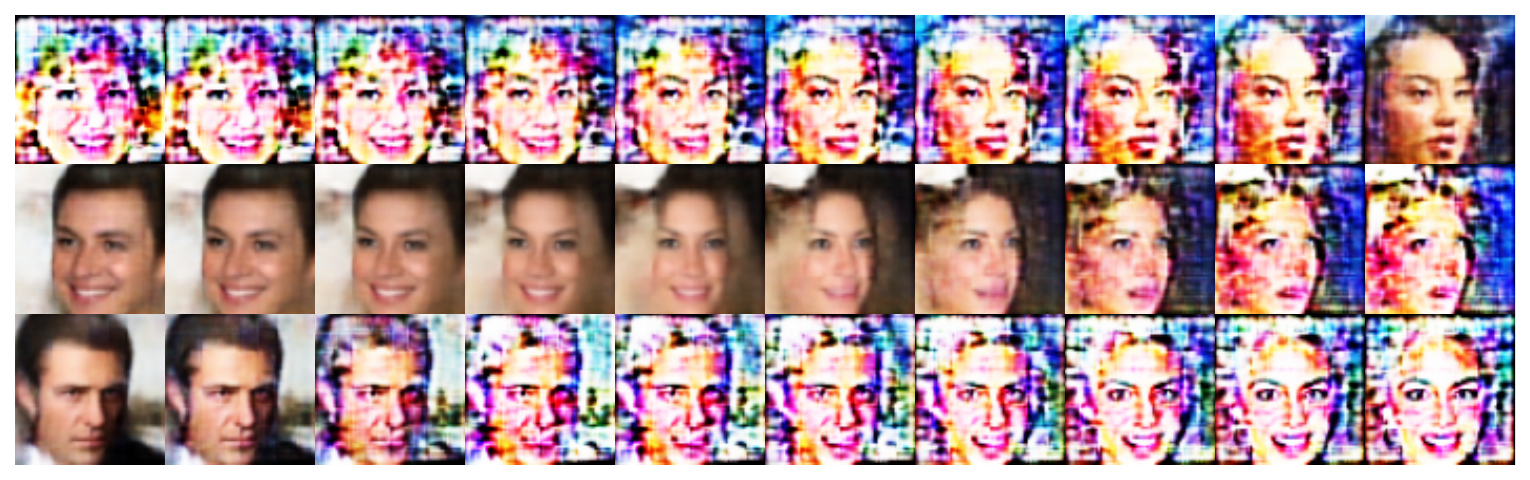}}
    \subfloat{\includegraphics[width=2.3in]{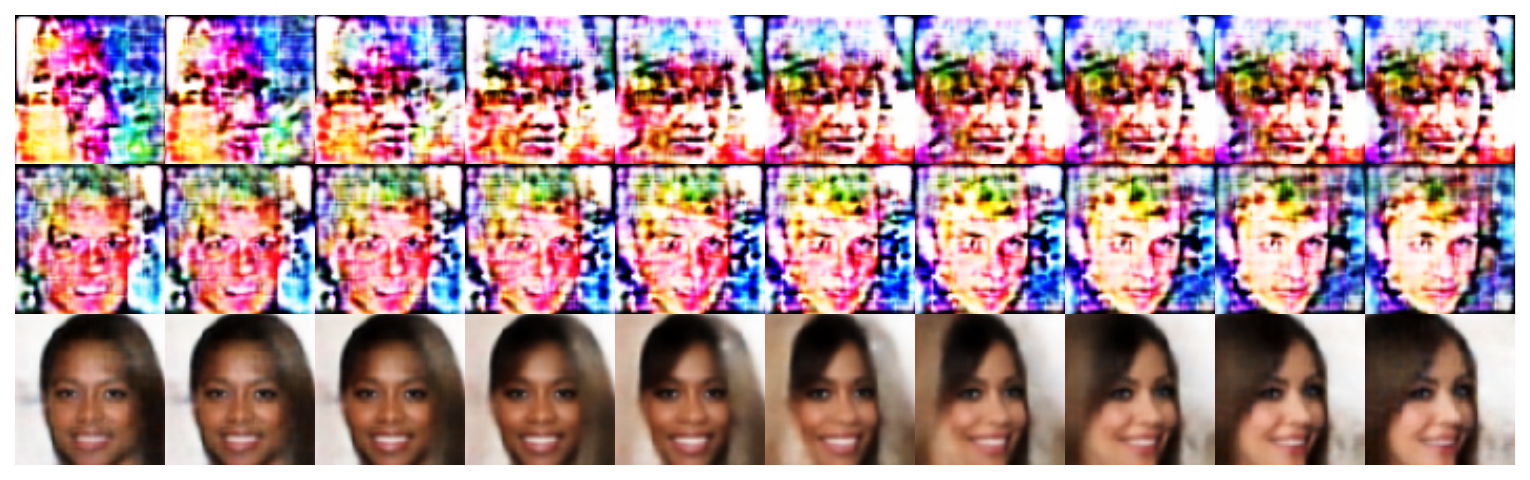}}\\\vspace{-1.2em}
    \adjustbox{minipage=6em,raise=\dimexpr -5.\height}{\small IWAE}
    \subfloat{\includegraphics[width=2.3in]{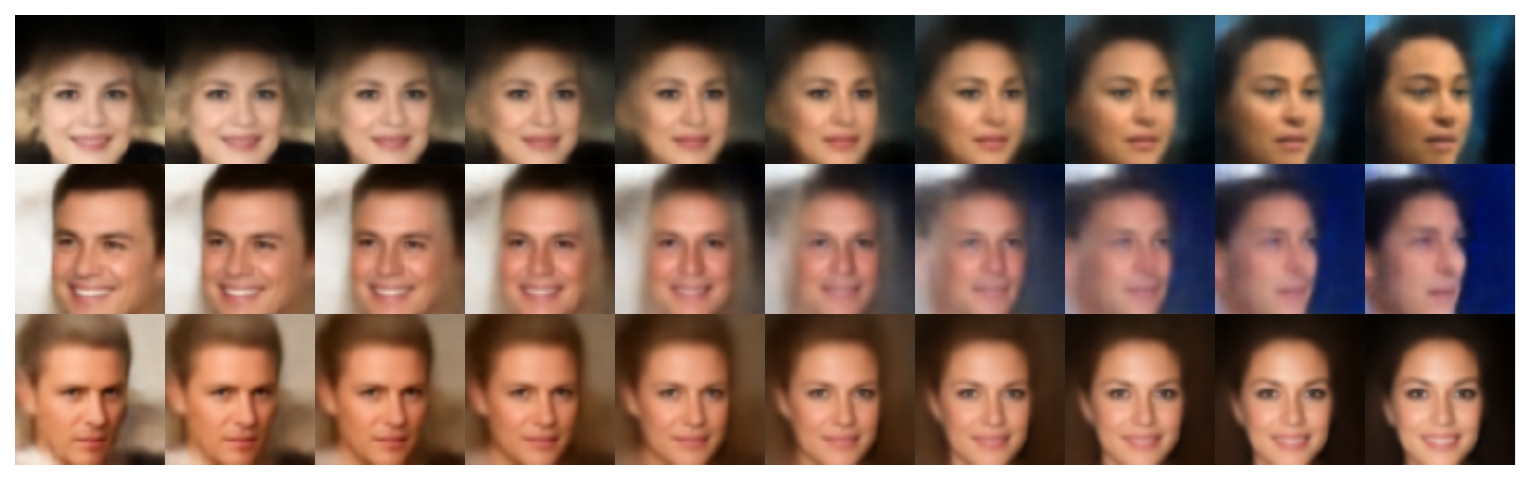}}
    \subfloat{\includegraphics[width=2.3in]{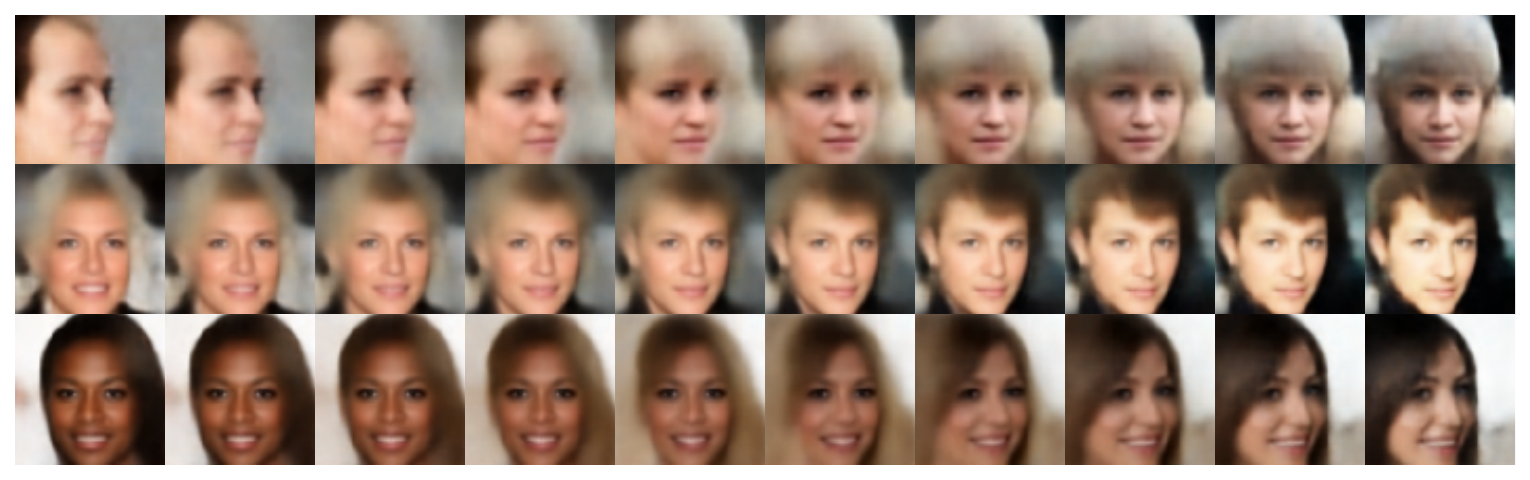}}\\\vspace{-1.2em}
    \adjustbox{minipage=6em,raise=\dimexpr -5.\height}{\small VAE-lin-NF}
    \subfloat{\includegraphics[width=2.3in]{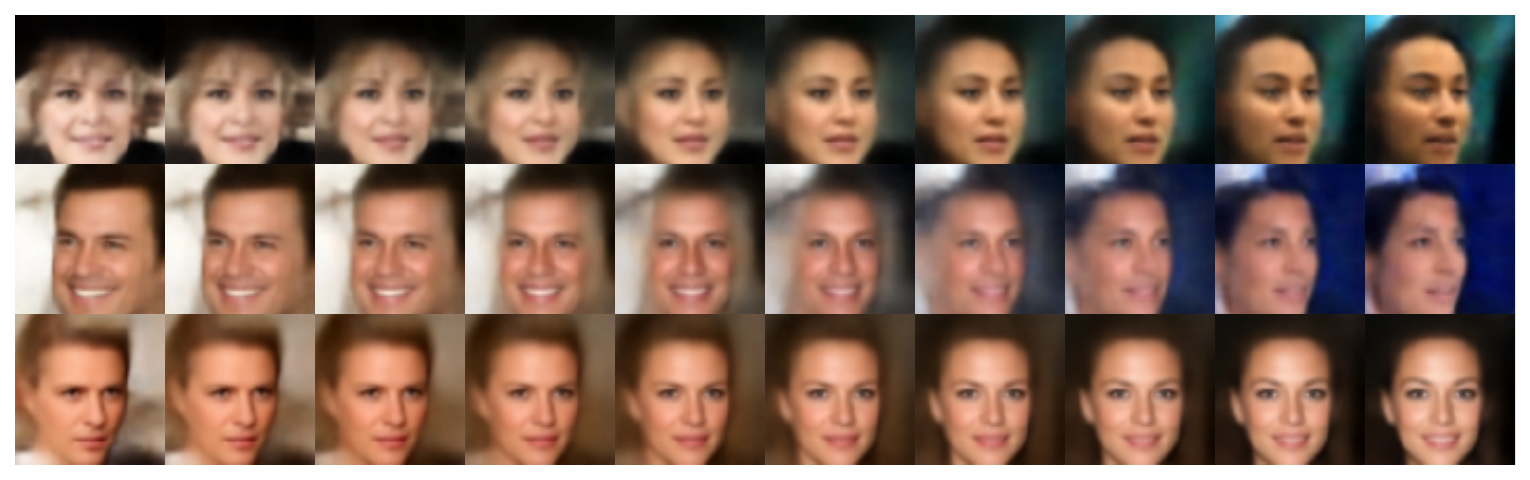}}
    \subfloat{\includegraphics[width=2.3in]{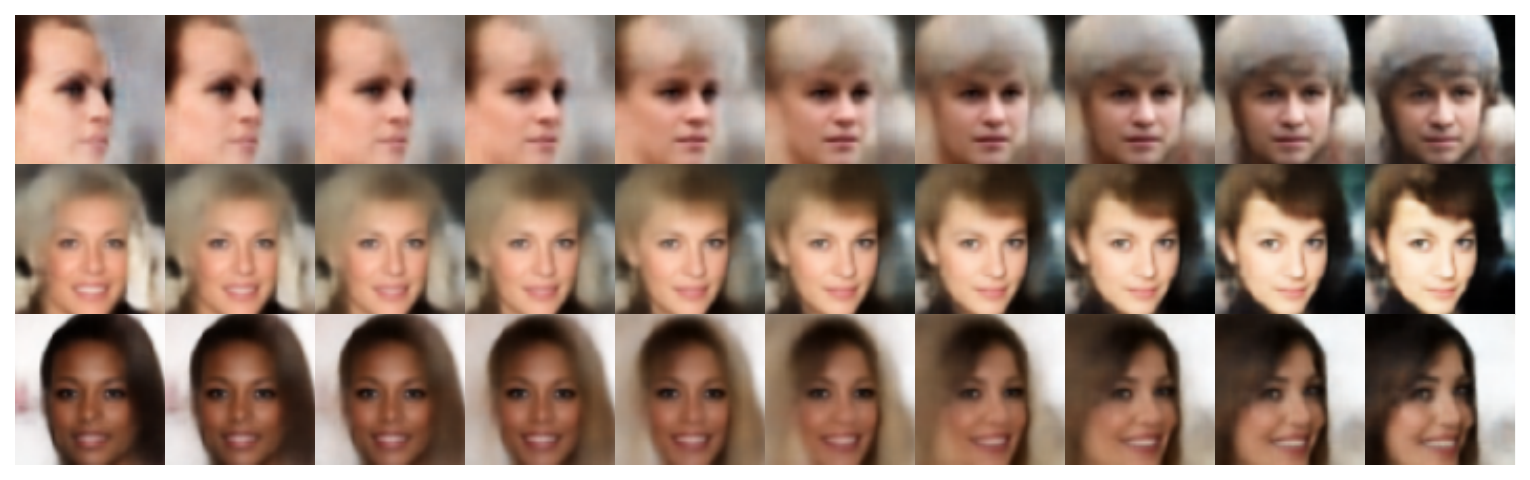}}\\\vspace{-1.2em}
    \adjustbox{minipage=6em,raise=\dimexpr -5.\height}{\small VAE-IAF}
    \subfloat{\includegraphics[width=2.3in]{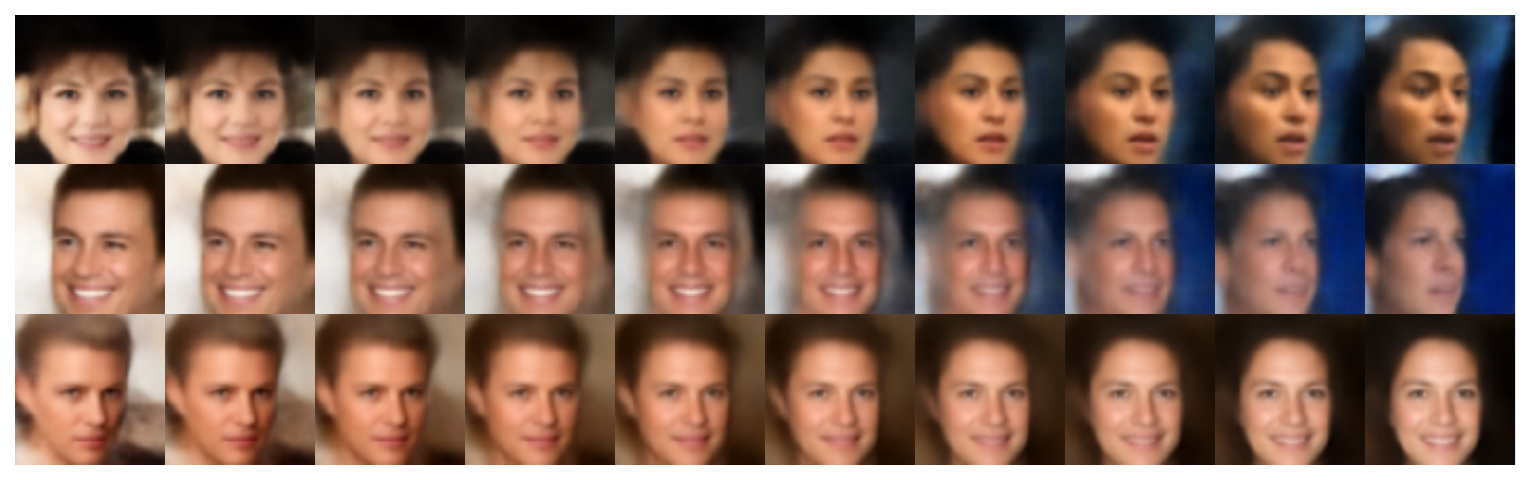}}
    \subfloat{\includegraphics[width=2.3in]{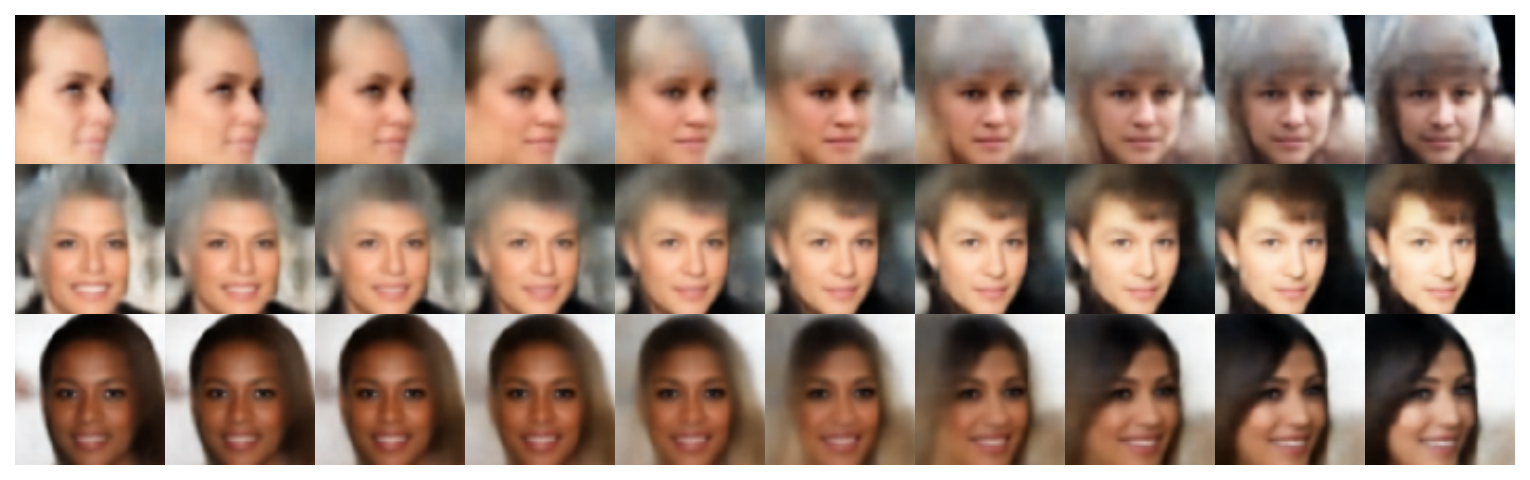}}\\\vspace{-1.2em}
    \adjustbox{minipage=6em,raise=\dimexpr -5.\height}{\small $\beta$-VAE}
    \subfloat{\includegraphics[width=2.3in]{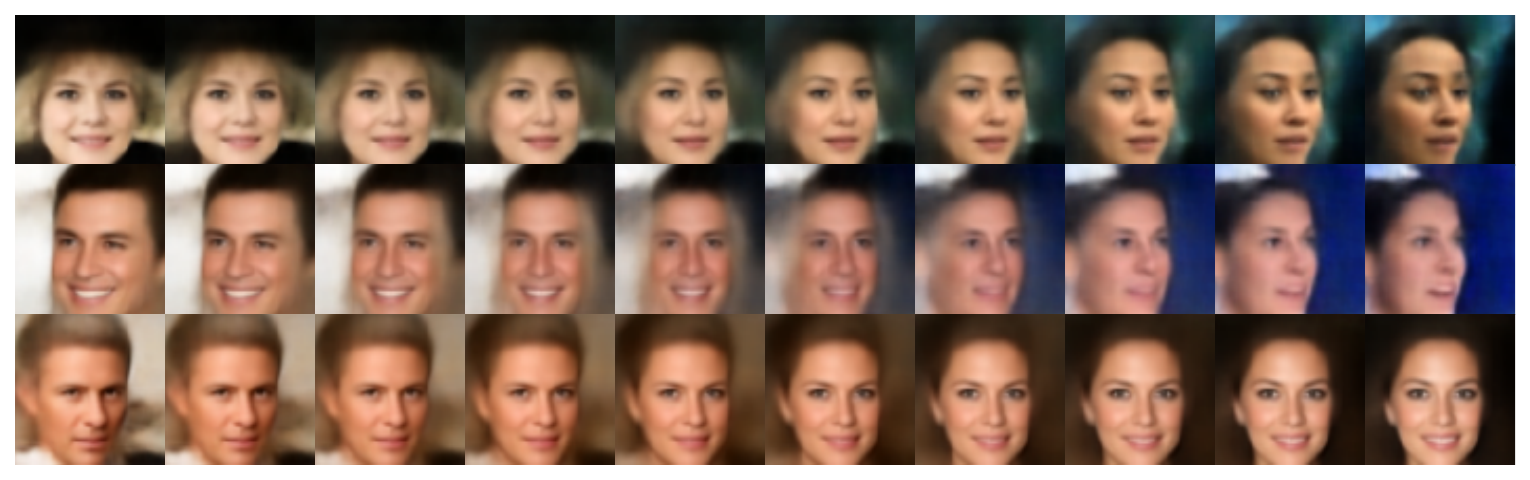}}
    \subfloat{\includegraphics[width=2.3in]{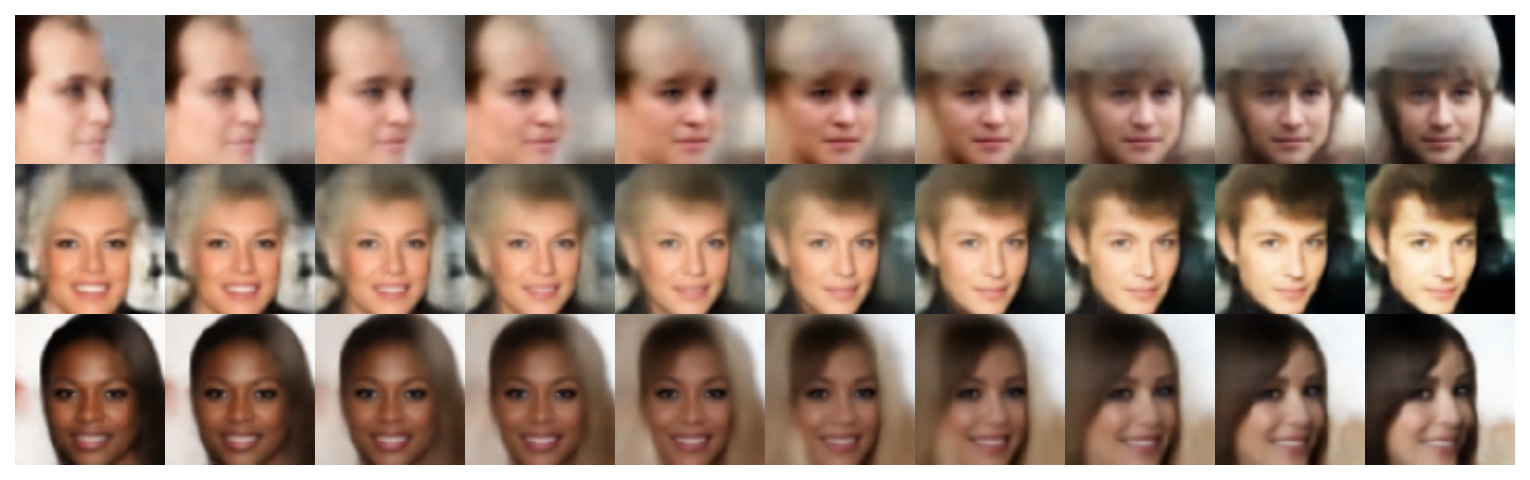}}\\\vspace{-1.2em}
    \adjustbox{minipage=6em,raise=\dimexpr -5.\height}{\small $\beta$-TC-VAE}
    \subfloat{\includegraphics[width=2.3in]{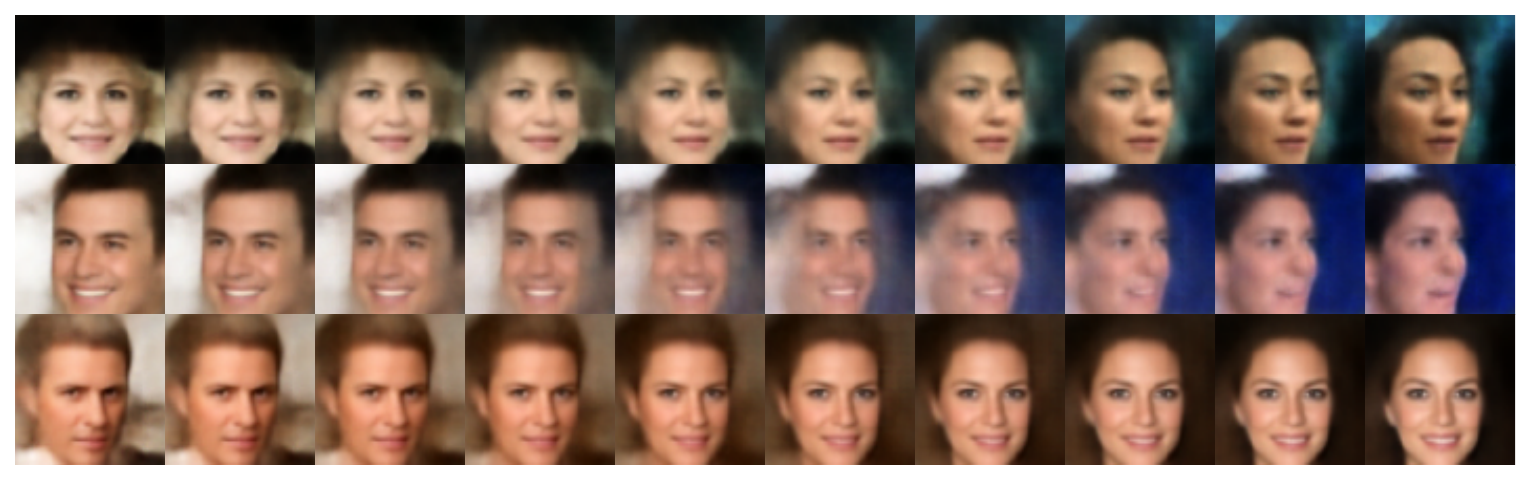}}
    \subfloat{\includegraphics[width=2.3in]{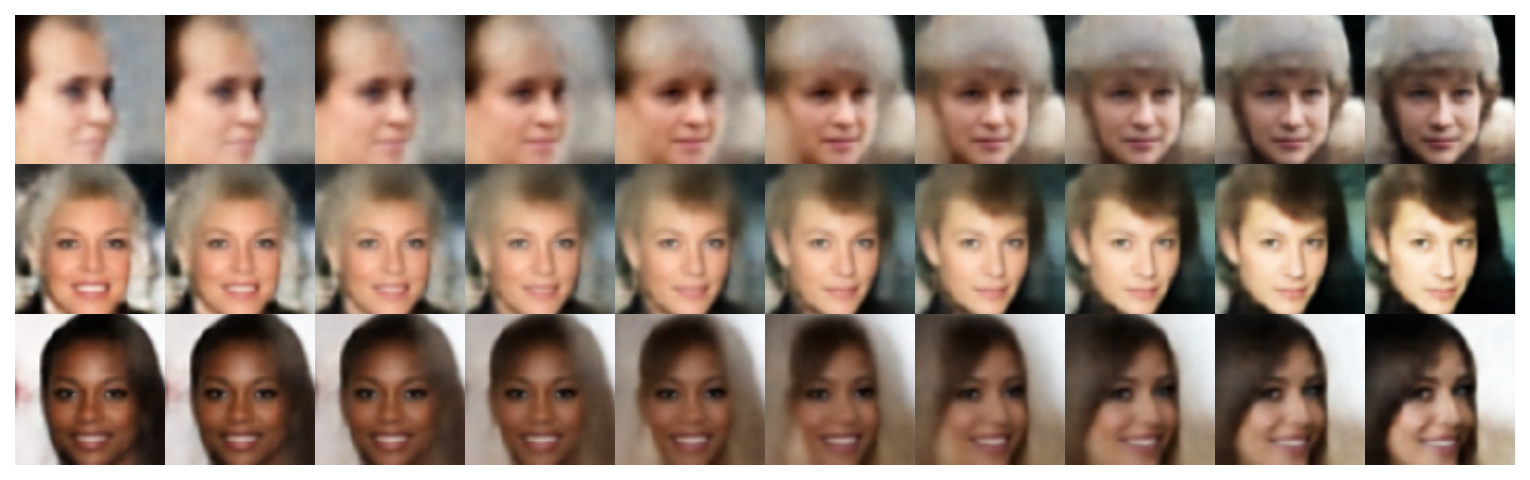}}\\\vspace{-1.2em}
    \adjustbox{minipage=6em,raise=\dimexpr -5.\height}{\small Factor-VAE}
    \subfloat{\includegraphics[width=2.3in]{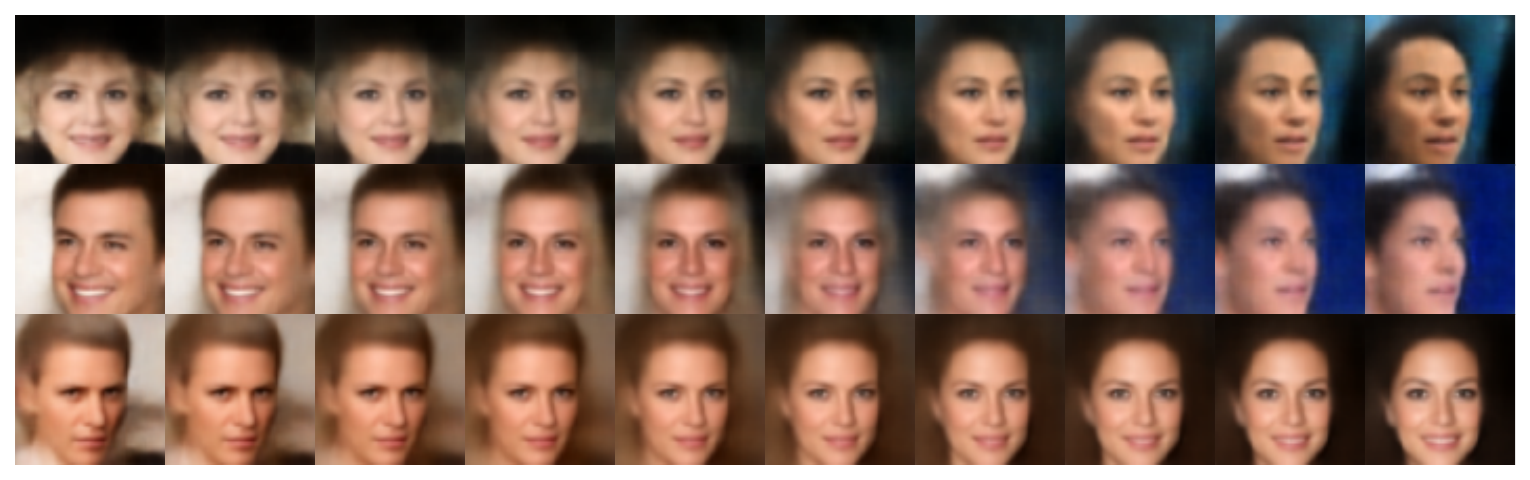}}
    \subfloat{\includegraphics[width=2.3in]{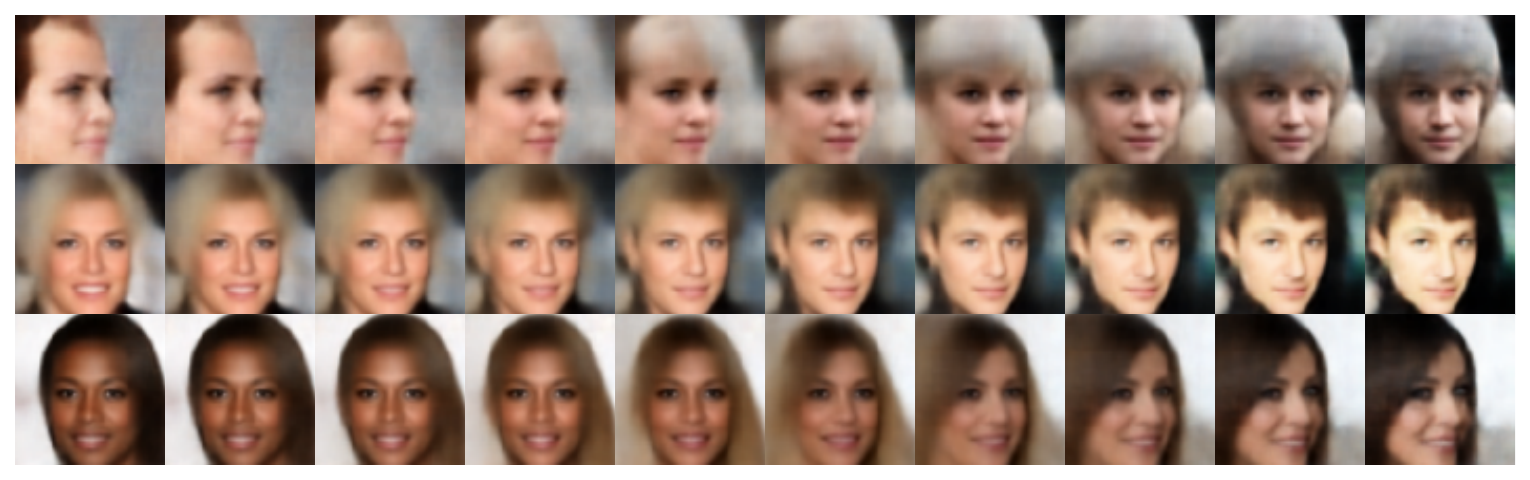}}\\\vspace{-1.2em}
    \adjustbox{minipage=6em,raise=\dimexpr -5.\height}{\small InfoVAE - IMQ}
    \subfloat{\includegraphics[width=2.3in]{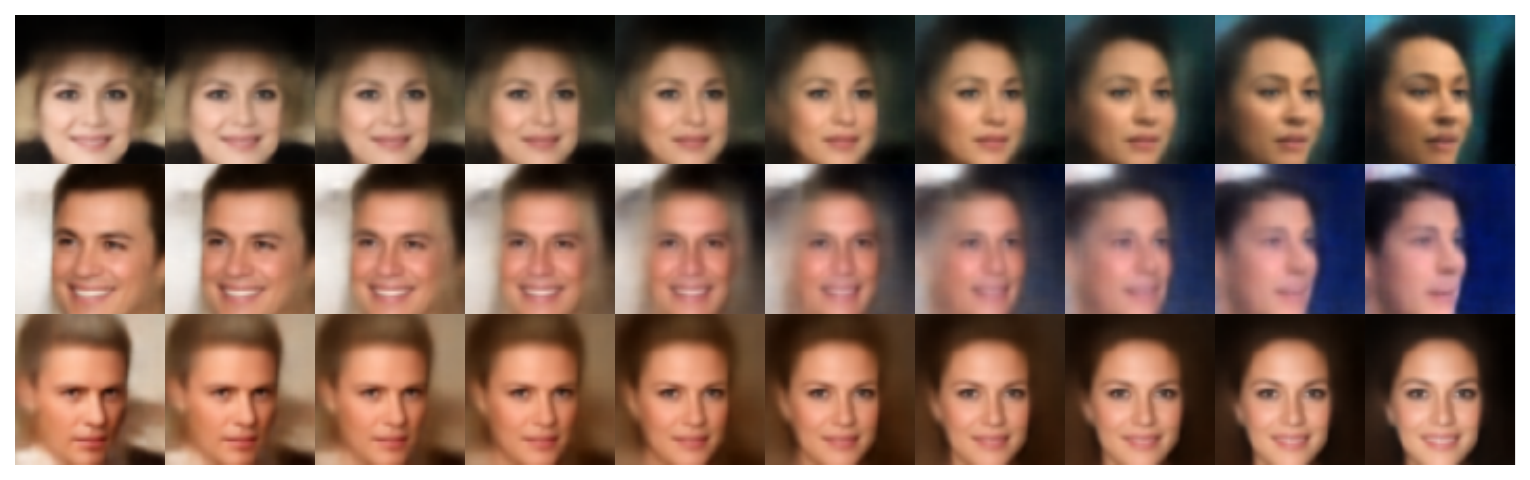}}
    \subfloat{\includegraphics[width=2.3in]{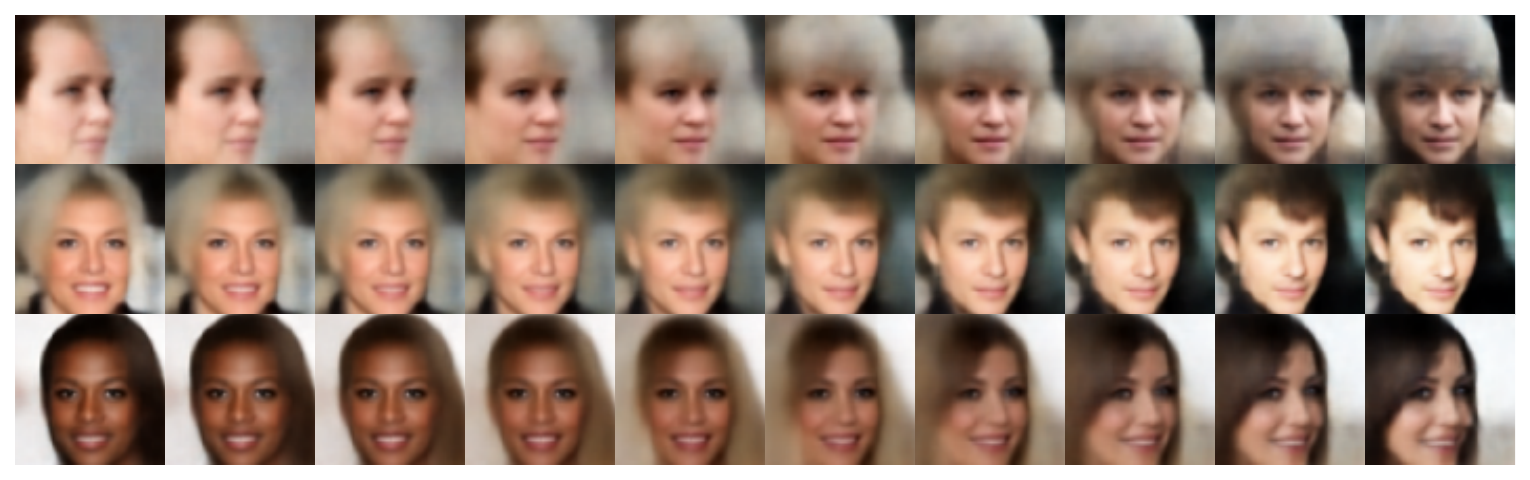}}\\\vspace{-1.2em}
    \adjustbox{minipage=6em,raise=\dimexpr -5.\height}{\small InfoVAE - RBF}
    \subfloat{\includegraphics[width=2.3in]{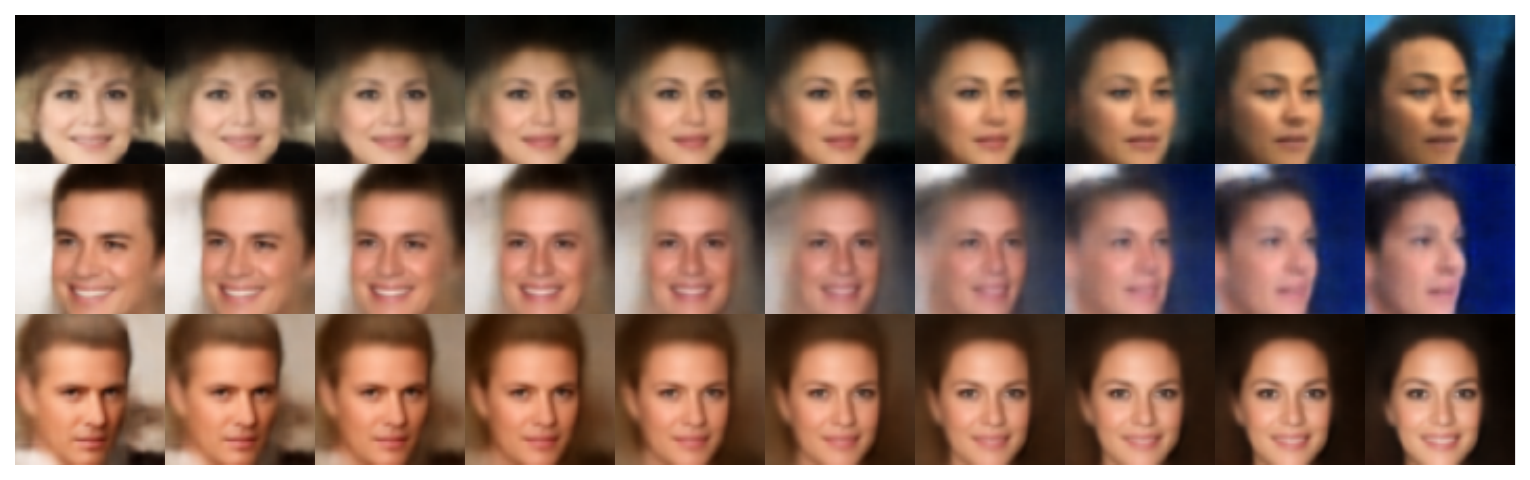}}
    \subfloat{\includegraphics[width=2.3in]{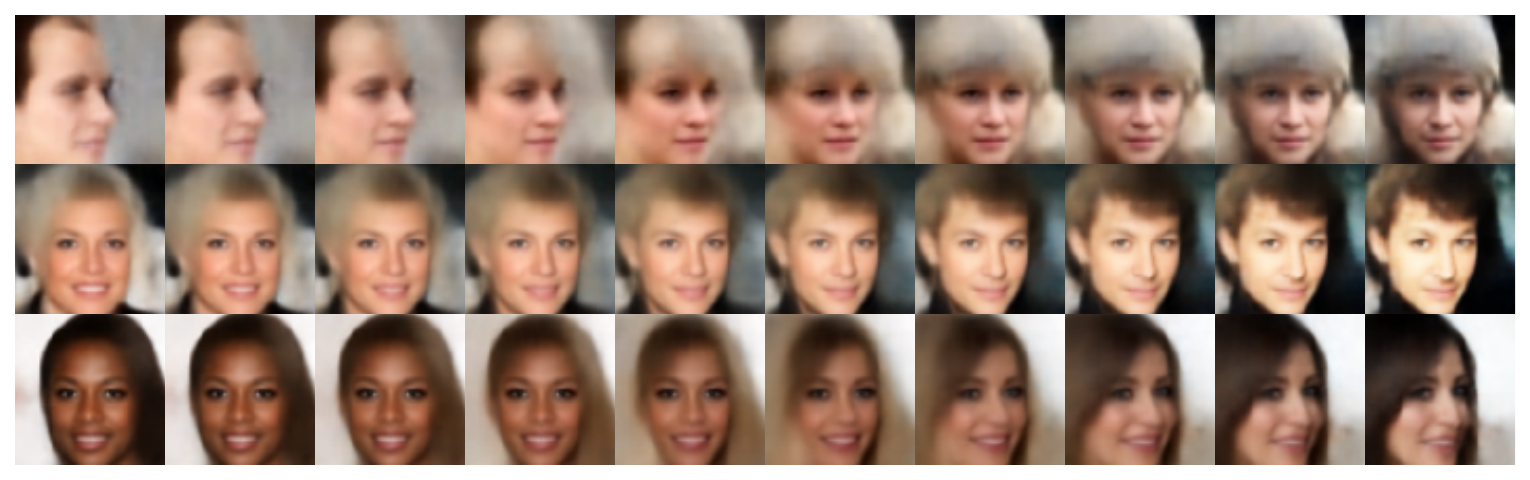}}
    \caption{Interpolations on CELEBA with the same starting and ending images for a latent space of dimension 64. For each model we select the configuration achieving the lowest FID on the generation task on the validation set with a GMM sampler.}
    \label{fig:interpolations celeba 1}
    \end{figure}

\begin{figure}[ht]
    \centering
    \captionsetup[subfigure]{position=above, labelformat = empty}
    \adjustbox{minipage=6em,raise=\dimexpr -5.\height}{\small AAE}
    \subfloat[CELEBA (64)]{\includegraphics[width=2.3in]{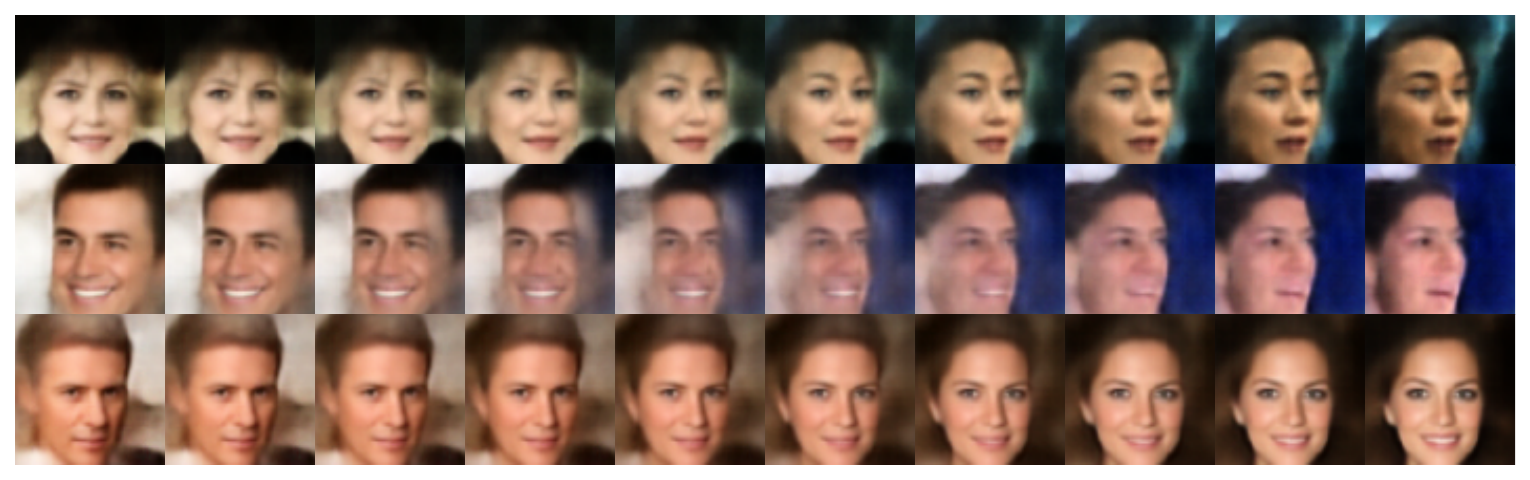}}
    \subfloat[CELEBA (64)]{\includegraphics[width=2.3in]{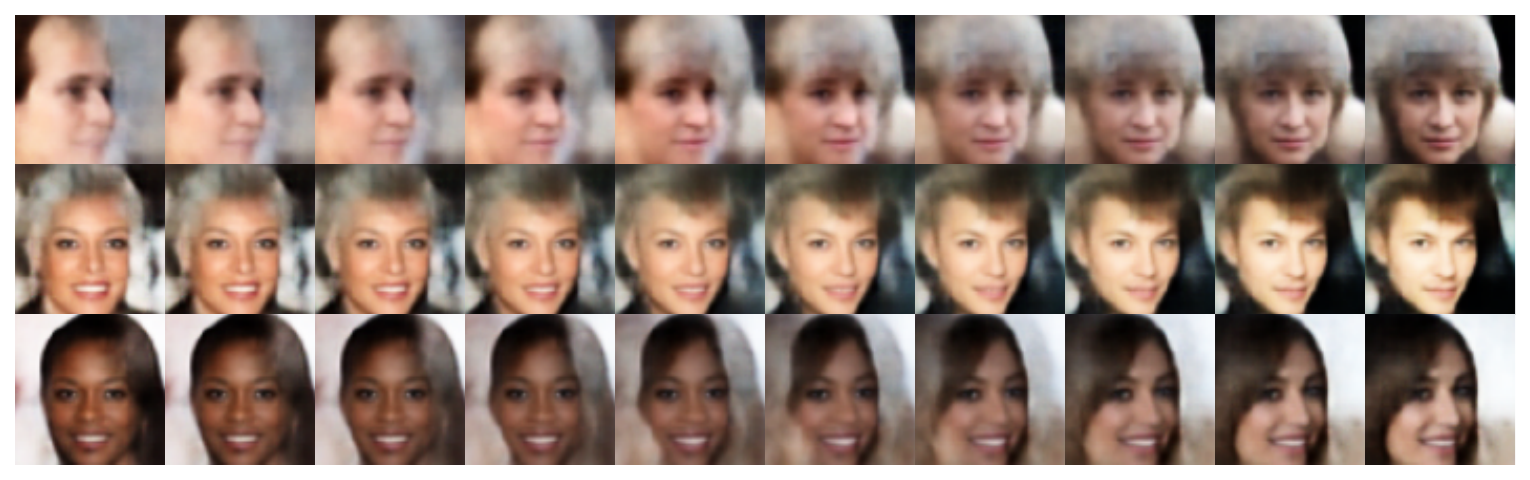}}\\\vspace{-1.2em}
    \adjustbox{minipage=6em,raise=\dimexpr -5.\height}{\small MSSSIM-VAE}
    \subfloat{\includegraphics[width=2.3in]{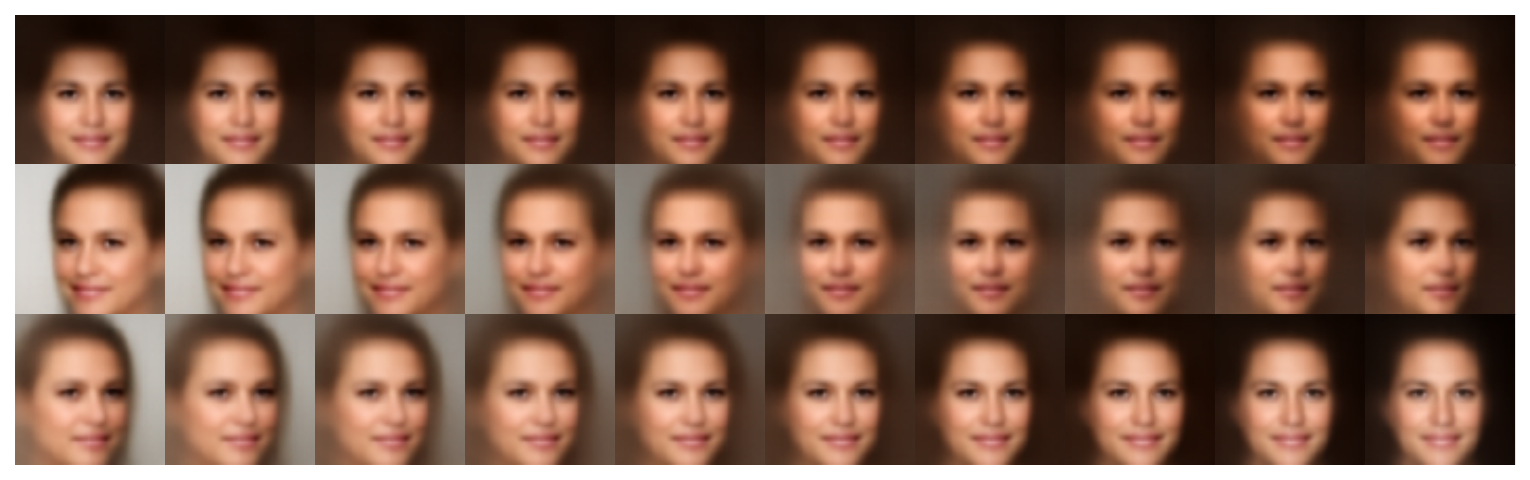}}
    \subfloat{\includegraphics[width=2.3in]{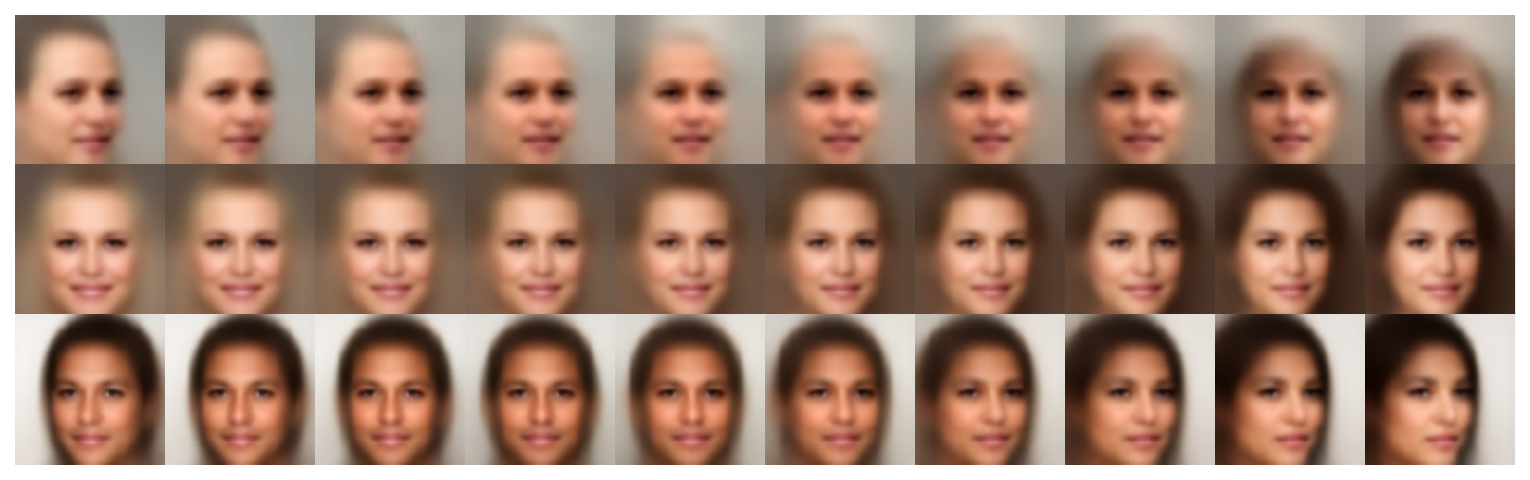}}\\\vspace{-1.2em}
    \adjustbox{minipage=6em,raise=\dimexpr -5.\height}{\small VAEGAN}
    \subfloat{\includegraphics[width=2.3in]{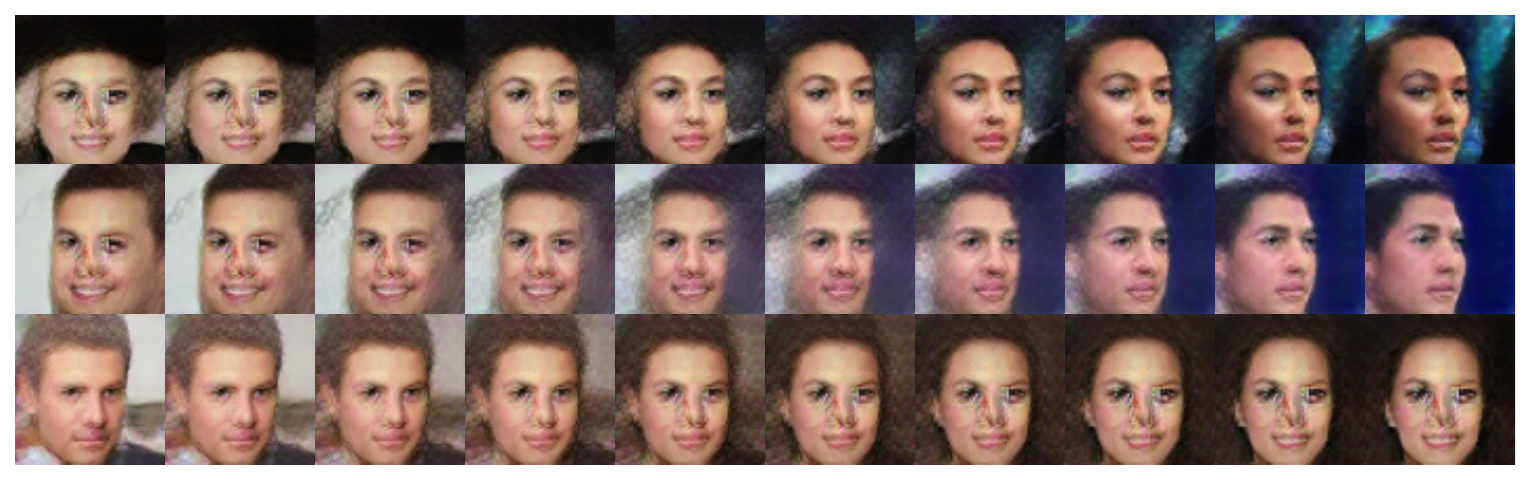}}
    \subfloat{\includegraphics[width=2.3in]{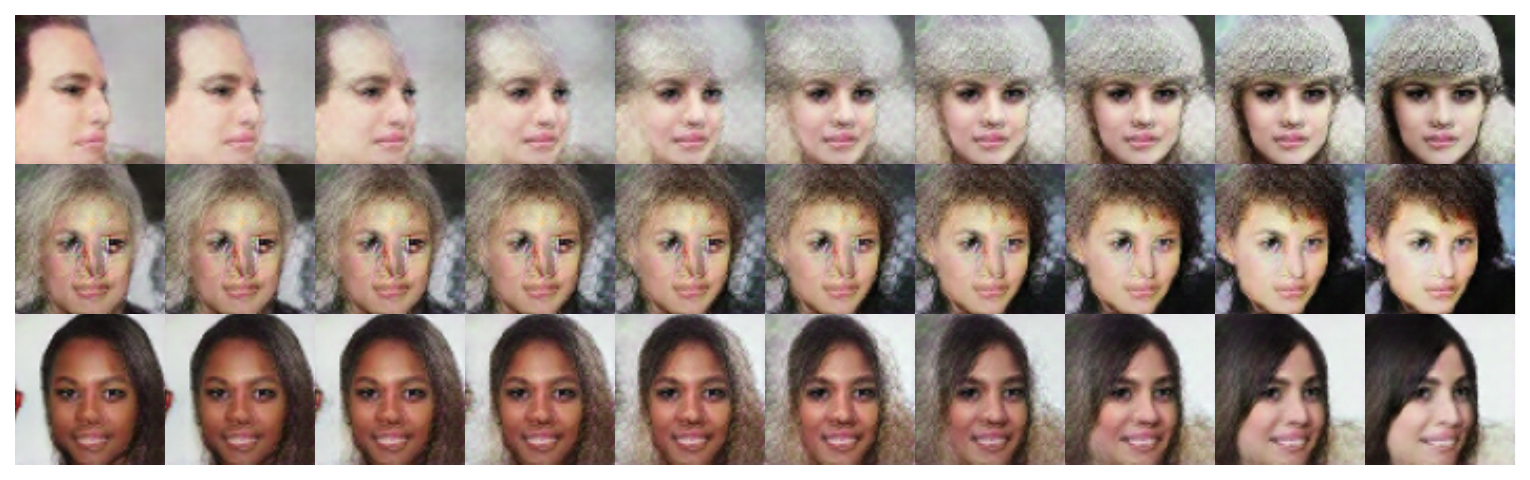}}\\\vspace{-1.2em}
    \adjustbox{minipage=6em,raise=\dimexpr -5.\height}{\small AE}
    \subfloat{\includegraphics[width=2.3in]{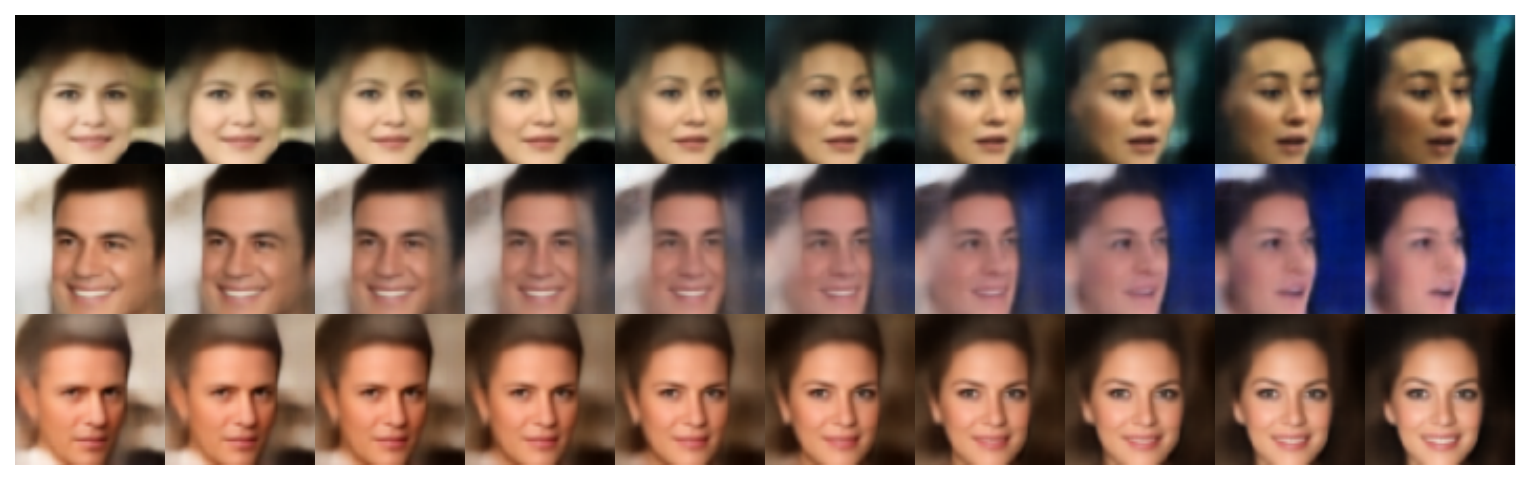}}
    \subfloat{\includegraphics[width=2.3in]{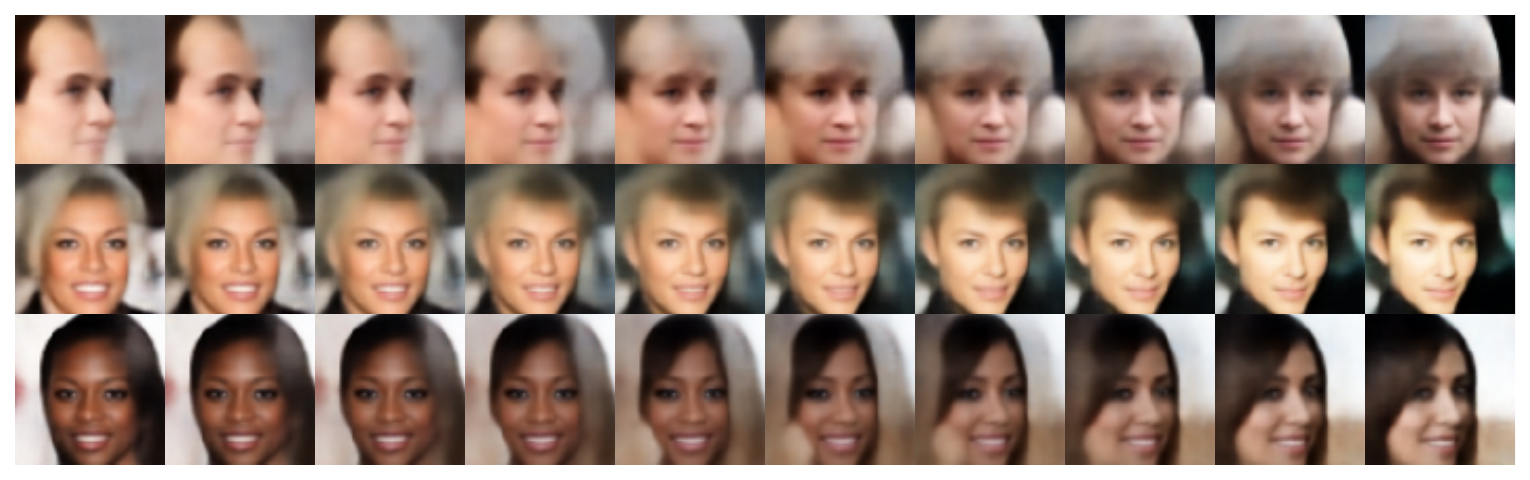}}\\\vspace{-1.2em}
    \adjustbox{minipage=6em,raise=\dimexpr -5.\height}{\small WAE-IMQ}
    \subfloat{\includegraphics[width=2.3in]{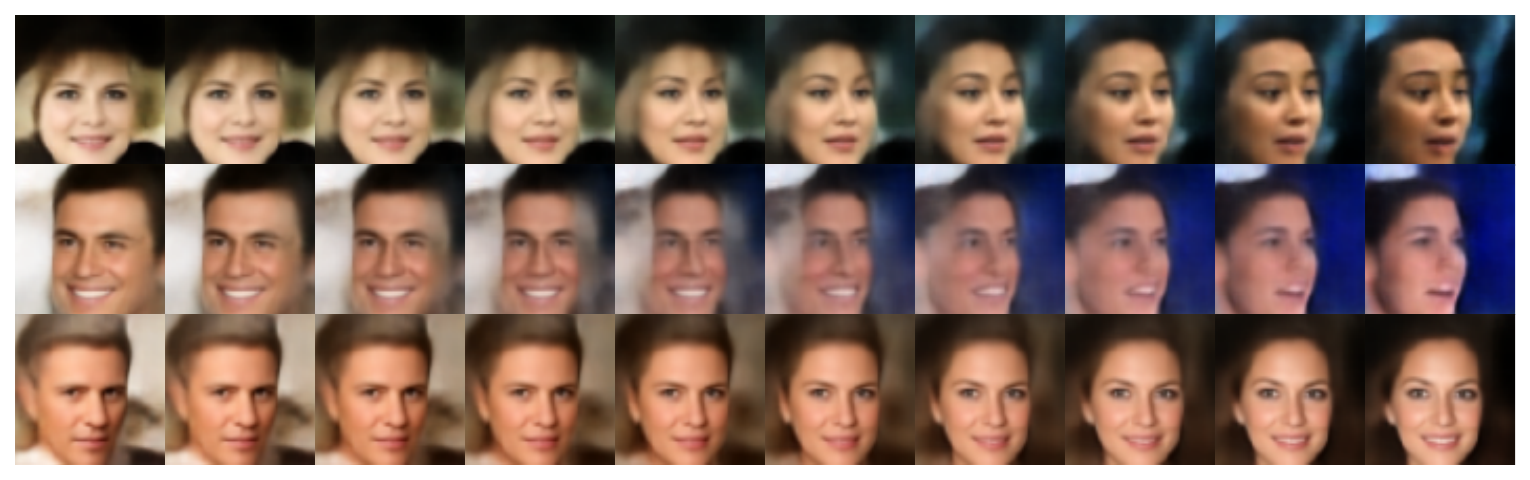}}
    \subfloat{\includegraphics[width=2.3in]{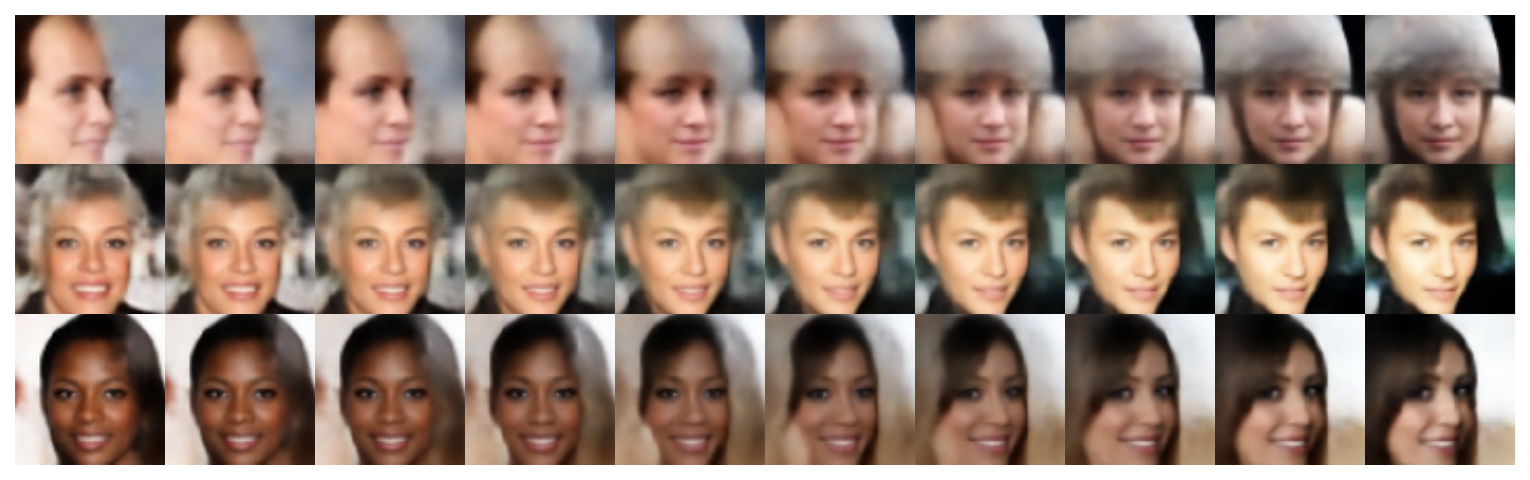}}\\\vspace{-1.2em}
    \adjustbox{minipage=6em,raise=\dimexpr -5.\height}{\small WAE-RBF}
    \subfloat{\includegraphics[width=2.3in]{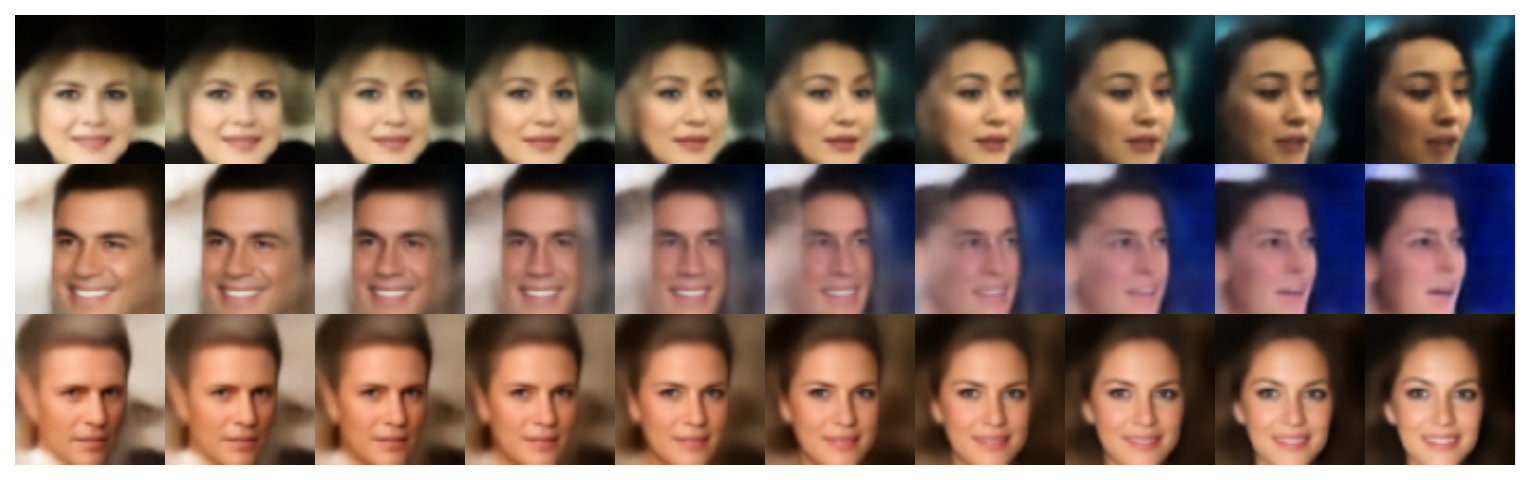}}
    \subfloat{\includegraphics[width=2.3in]{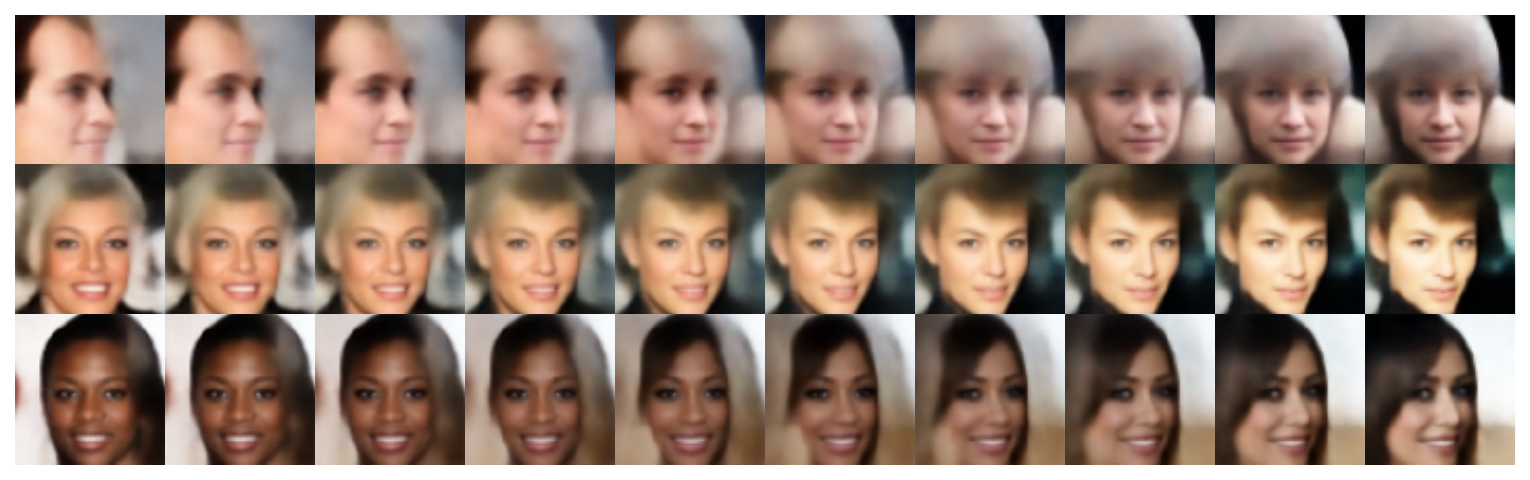}}\\\vspace{-1.2em}
    \adjustbox{minipage=6em,raise=\dimexpr -5.\height}{\small VQVAE}
    \subfloat{\includegraphics[width=2.3in]{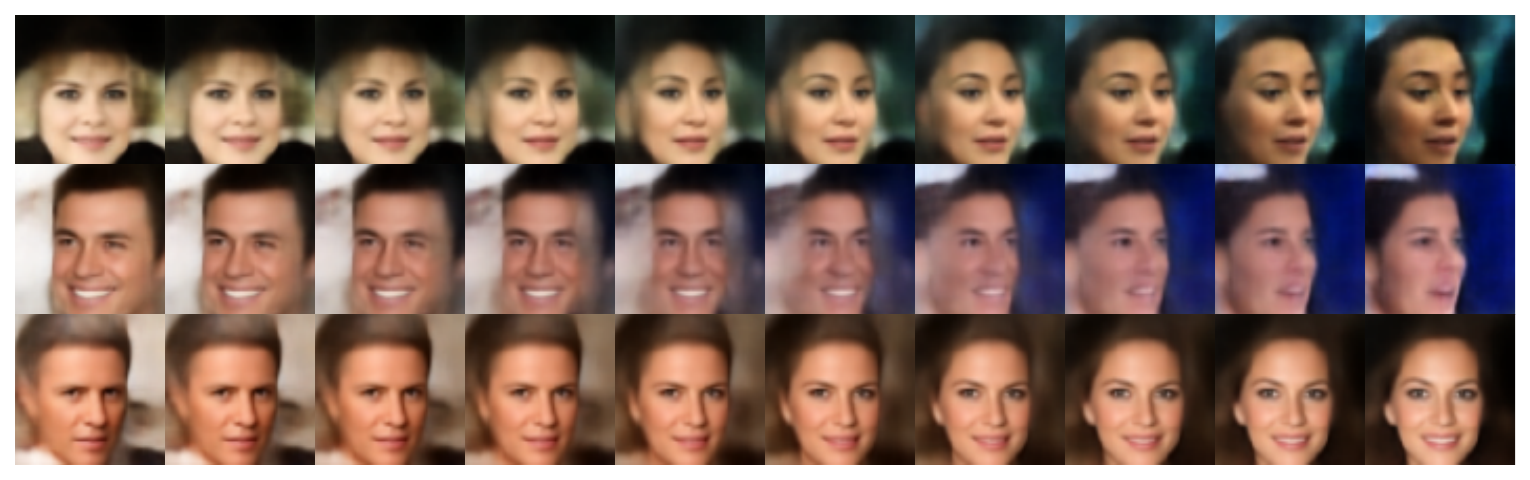}}
    \subfloat{\includegraphics[width=2.3in]{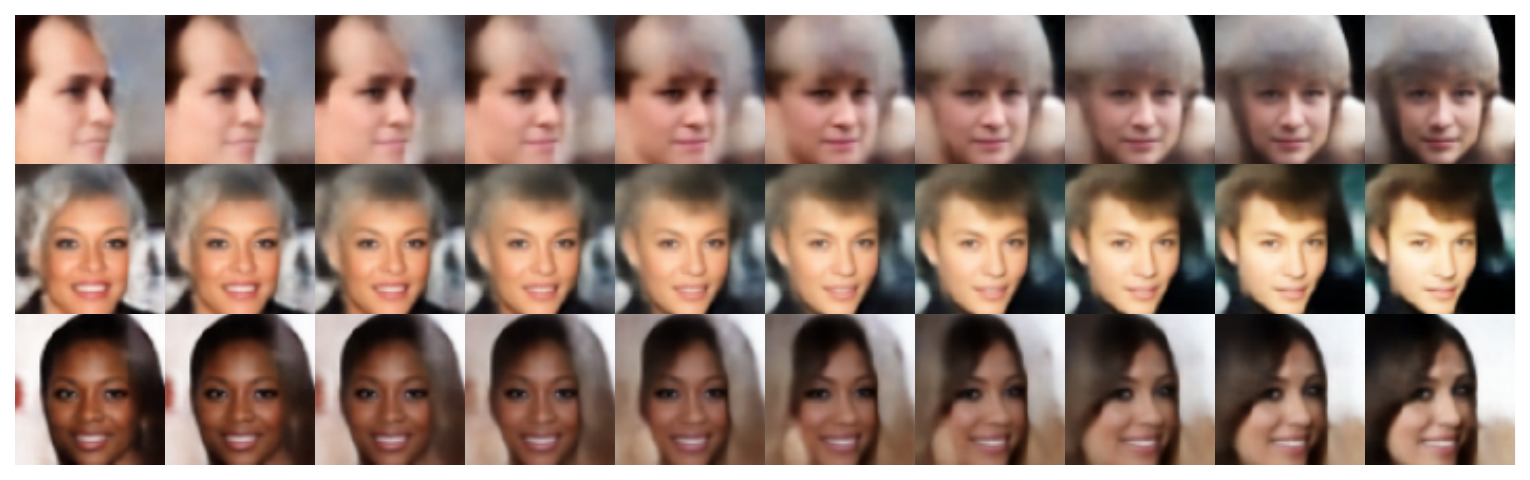}}\\\vspace{-1.2em}
    \adjustbox{minipage=6em,raise=\dimexpr -5.\height}{\small RAE-l2}
    \subfloat{\includegraphics[width=2.3in]{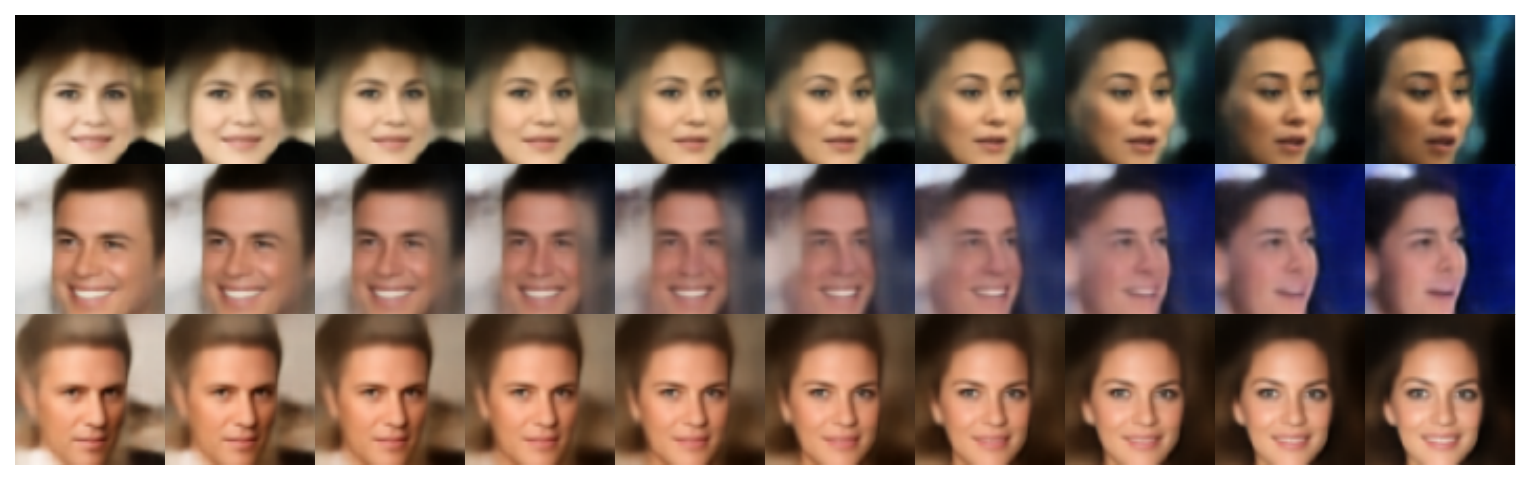}}
    \subfloat{\includegraphics[width=2.3in]{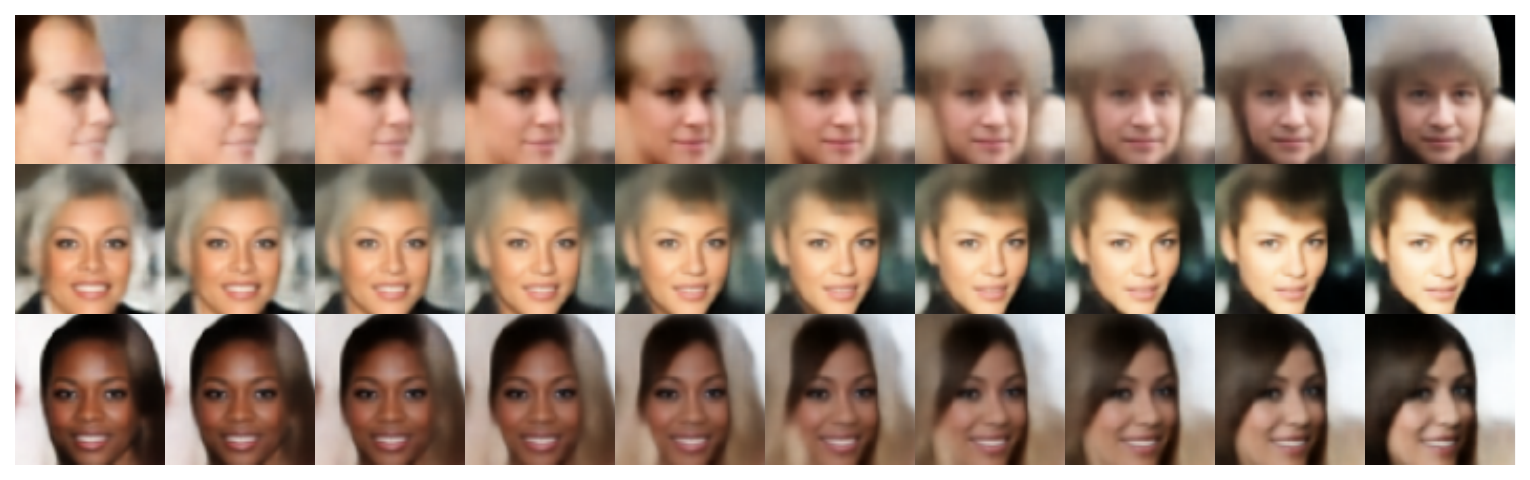}}\\\vspace{-1.2em}
    \adjustbox{minipage=6em,raise=\dimexpr -5.\height}{\small RAE-GP}
    \subfloat{\includegraphics[width=2.3in]{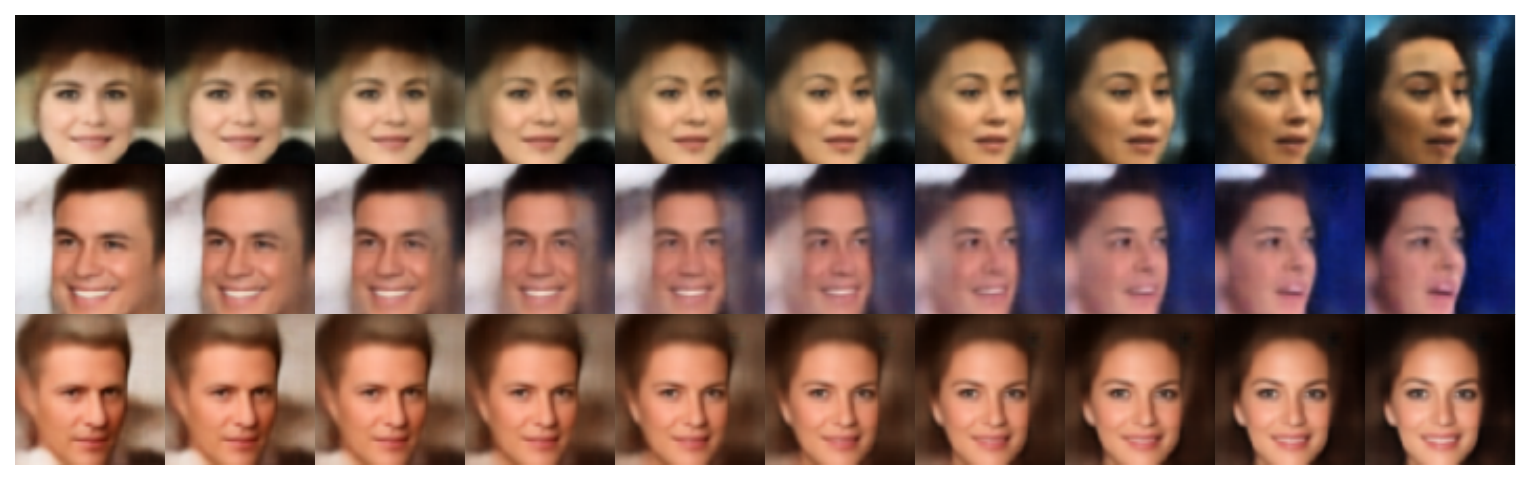}}
    \subfloat{\includegraphics[width=2.3in]{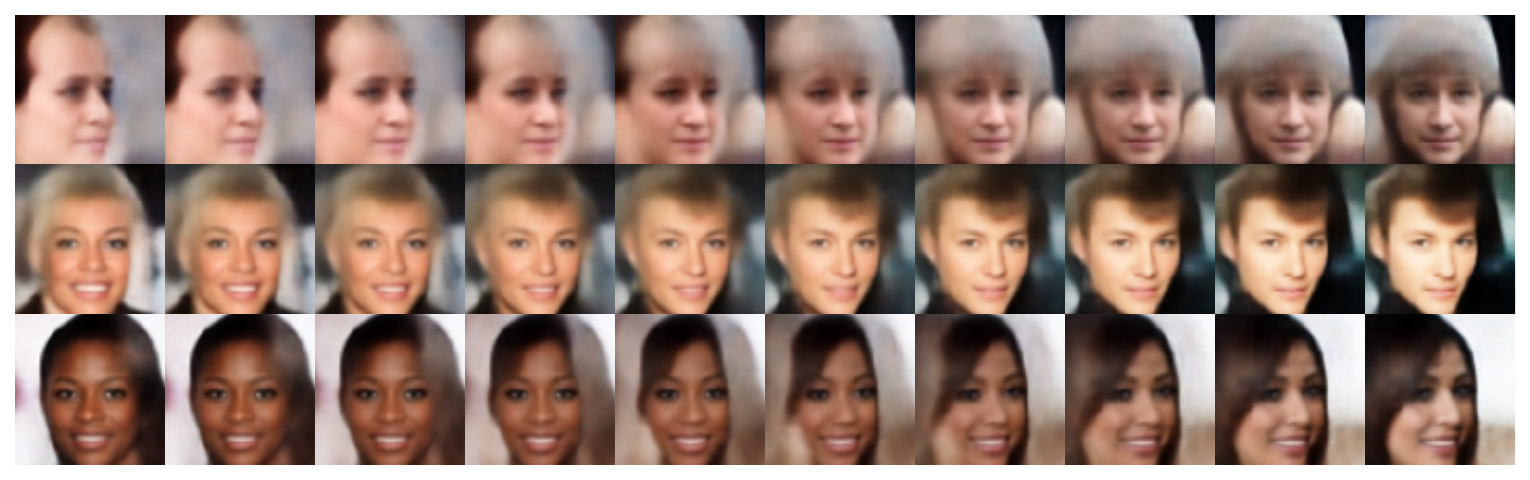}}
    \caption{Interpolations on CELEBA with the same starting and ending images for a latent space of dimension 64. For each model we select the configuration achieving the lowest FID on the generation task on the validation set with a GMM sampler.}
    
    \clearpage
    \label{fig:interpolations celeba 2}
    \end{figure}

\clearpage

\section{Detailed experiments set-up}\label{appC}

We detail here the main experimental set-up and implementation choices made in the benchmark. We let the reader refer to the code available online for specific implementation aspects

\paragraph{The data}
To perform the benchmaks presented in the paper, we select 3 classical \emph{free-to-use} image datasets: MNIST \citep{lecun_mnist_1998}, CIFAR10 \citep{krizhevsky2009learning} and CELEBA \citep{liu2015faceattributes}. These datasets are publicly available, widely used for generative model related papers and have well known associated metrics in the literature. Each dataset is split into a train set, a validation set and a test set. For MNIST and CIFAR10 the validation set is composed of the last 10k images extracted from the official train set and the test set corresponds to the official one. For CELEBA, we use the official train/val/test split. %For all the tasks presented in this benchmark, these splits remains unchanged.

\paragraph{Training paradigm} We equip each model used in the benchmark with the same neural network architecture for both the encoder and decoder, taken as a ConvNet and ResNet (architectures given in Tables.~\ref{tab:convNet archi}  and \ref{tab:resNet archi}) leading to a comparable number of parameters \footnote{Some models may actually have additional parameters in their intrinsic structure \emph{e.g.} a VQVAE learns a dictionary of embeddings, a VAMP learns the pseudo-inputs, a VAE-IAF learns the auto-regressive flows. Nonetheless, since we work on images, the number of parameters remains in the same order of magnitude.}. For the 19 considered models, due to computational limitations, 10 different configurations are considered, allowing a simple exploration of the models' hyper-parameters. The sets of hyper-parameters explored are detailed in Appendix.~\ref{appD} for each model. The models are then trained on MNIST and CIFAR10 for 100 epochs, a starting learning rate of $1e^{-4}$ and batch size of 100 with Adam optimizer \citep{kingma_adam_2014}. A scheduler reducing the learning rate by half if the validation loss does not improve for 10 epochs is also used. For CELEBA, we use the same setting but we train the models for 50 epochs  with a starting learning rate of $1e^{-3}$. Models with unstable training (NaN, huge training spikes...) are iteratively retrained with a starting learning rate divided by 10 until training stabilises. All 19 models are trained on a single 32GB V100 GPU. This leads to 10 trained models for each method, each dataset (MNIST, CIFAR10 or CELEBA) and each neural network (ConvNet or ResNet) leading to a total of 1140 models. The training setting (curves, configs ...) can be found at \url{https://wandb.ai/benchmark_team/trainings}.

\paragraph{Sampling paradigm for the MAF and VAE samplers} 

For the Masked Autoregressive Flow sampler used for sampling we use a 3-layer MADE \citep{germain2015made} with 128 hidden units and ReLU activation for each layer and stack 2 blocks of MAF to create the flow. For the masked layers, the mask is made sequentially and the ordering is reversed between each MADE. For this normalising flow we consider a starting distribution given by a standard Gaussian. For the auxiliary VAE sampling method proposed in \citep{dai_diagnosing_2018}, we consider a simple VAE with a Multi Layer Perceptron (MLP) encoder and decoder, with 2 hidden layers composed of 1024 units and ReLU activation. Both samplers are fitted with 200 epochs using the train and evaluation embeddings coming from the trained autoencoder models. A learning rate of $1e^{-4}$, a scheduler decreasing the learning rate by half if the validation loss does not improve for 10 epochs and a batch size of 100 are used for these samplers.

\begin{table}[ht]
    \centering
    \scriptsize
    \caption{Neural network architecture used for the convolutional networks.}
    \label{tab:convNet archi}
    \begin{tabular}{cccc}
    \toprule
         & MNIST & CIFAR10 & CELEBA \\
         \midrule
         Encoder & (1, 28, 28)& (3, 32, 32) & (3, 64, 64) \\
        \multirow{1}{*}{Layer 1} & Conv(128, 4, 2), BN, ReLU & Conv(128, 4, 2), BN, ReLU & Conv(128, 4, 2), BN, ReLU\\
        \multirow{1}{*}{Layer 2} & Conv(256, 4, 2), BN, ReLU & Conv(256, 4, 2), BN, ReLU & Conv(256, 4, 2), BN, ReLU\\
        \multirow{1}{*}{Layer 3} & Conv(512, 4, 2), BN, ReLU & Conv(512, 4, 2), BN, ReLU & Conv(512, 4, 2), BN, ReLU \\
        \multirow{1}{*}{Layer 4} & Conv(1024, 4, 2), BN, ReLU & Conv(1024, 4, 2), BN, ReLU & Conv(1024, 4, 2), BN, ReLU \\
        \multirow{1}{*}{Layer 5}& Linear(1024, latent\_dim)* & Linear(4096, latent\_dim)* & Linear(16384, latent\_dim)*\\
        \midrule
        Decoder & \\
        Layer 1 & Linear(latent\_dim, 16384) & Linear(latent\_dim, 65536) & Linear(latent\_dim, 65536) \\
        \multirow{1}{*}{Layer 2} & ConvT(512, 3, 2), BN, ReLU & ConvT(512, 4, 2), BN, ReLU & ConvT(512, 5, 2), BN, ReLU \\
        \multirow{1}{*}{Layer 3} & ConvT(256, 3, 2), BN, ReLU & ConvT(256, 4, 2), BN, ReLU & ConvT(256, 5, 2), BN, ReLU \\
        \multirow{1}{*}{Layer 4} & Conv(1, 3, 2), Sigmoid & Conv(3, 4, 1), Sigmoid & ConvT(128, 5, 2), BN, ReLU \\
        \multirow{1}{*}{Layer 5} & - & - & ConvT(3, 5, 1), Sigmoid \\
    \bottomrule
    \multicolumn{3}{l}{*Doubled for VAE-based models}
    \end{tabular}
\end{table}

\begin{table}[ht]
    \centering
    \scriptsize
    \caption{Neural network architecture used for the residual networks.}
    \label{tab:resNet archi}
    \begin{tabular}{cccc}
    \toprule
         & MNIST & CIFAR10 & CELEBA \\
         \midrule
         Encoder & (1, 28, 28)& (3, 32, 32) & (3, 64, 64) \\
        \multirow{1}{*}{Layer 1} & Conv(64, 4, 2) & Conv(64, 4, 2) & Conv(64, 4, 2) \\
        \multirow{1}{*}{Layer 2} & Conv(128, 4, 2) & Conv(128, 4, 2) & Conv(128, 4, 2)\\
        \multirow{1}{*}{Layer 3} & Conv(128, 3, 2) & Conv(128, 3, 1) & Conv(128, 3, 2)\\
        \multirow{1}{*}{Layer 4} & ResBlock** & ResBlock** & Conv(128, 3, 2)\\
        \multirow{1}{*}{Layer 5} & ResBlock** & ResBlock** & ResBlock** \\
         \multirow{1}{*}{Layer 6} & Linear(2048, latent\_dim)* & Linear(8192, latent\_dim)* & ResBlock** \\
        \multirow{1}{*}{Layer 7}& - & - & Linear(2048, latent\_dim)*\\
        \midrule
        Decoder & \\
        Layer 1 & Linear(latent\_dim, 2048) & Linear(latent\_dim, 8192) & Linear(latent\_dim, 2048) \\
        \multirow{1}{*}{Layer 2} & ConvT(128, 3, 2) & ResBlock** & ConvT(128, 3, 2)\\ 
        \multirow{1}{*}{Layer 3} & ResBlock** & ResBlock** & ResBlock**\\
        \multirow{1}{*}{Layer 4} & ResBlock**, ReLU & ConvT(64, 4, 2) & ResBlock** \\
        \multirow{1}{*}{Layer 5} & ConvT(64, 3, 2), ReLU &  ConvT(3, 4, 2), Sigmoid & ConvT(128, 5, 2), Sigmoid \\
        \multirow{1}{*}{Layer 6} & ConvT(1, 3, 2), Sigmoid & - & ConvT(64, 5, 2), Sigmoid \\
        \multirow{1}{*}{Layer 6} & - & - & ConvT(3, 4, 2), Sigmoid \\
    \bottomrule
    \multicolumn{3}{l}{*Doubled for VAE-based models}\\
    \multicolumn{4}{l}{**The ResBlocks are composed of one Conv(32, 3, 1) followed by Conv(128, 1, 1) with ReLU.}\\

    \end{tabular}
\end{table}

\clearpage

\section{Additional results}\label{appD}

\subsection{Effect of the latent dimension on the 4 tasks with the CIFAR10 database}

\begin{figure}[ht]
    \centering
    \includegraphics[width=\linewidth]{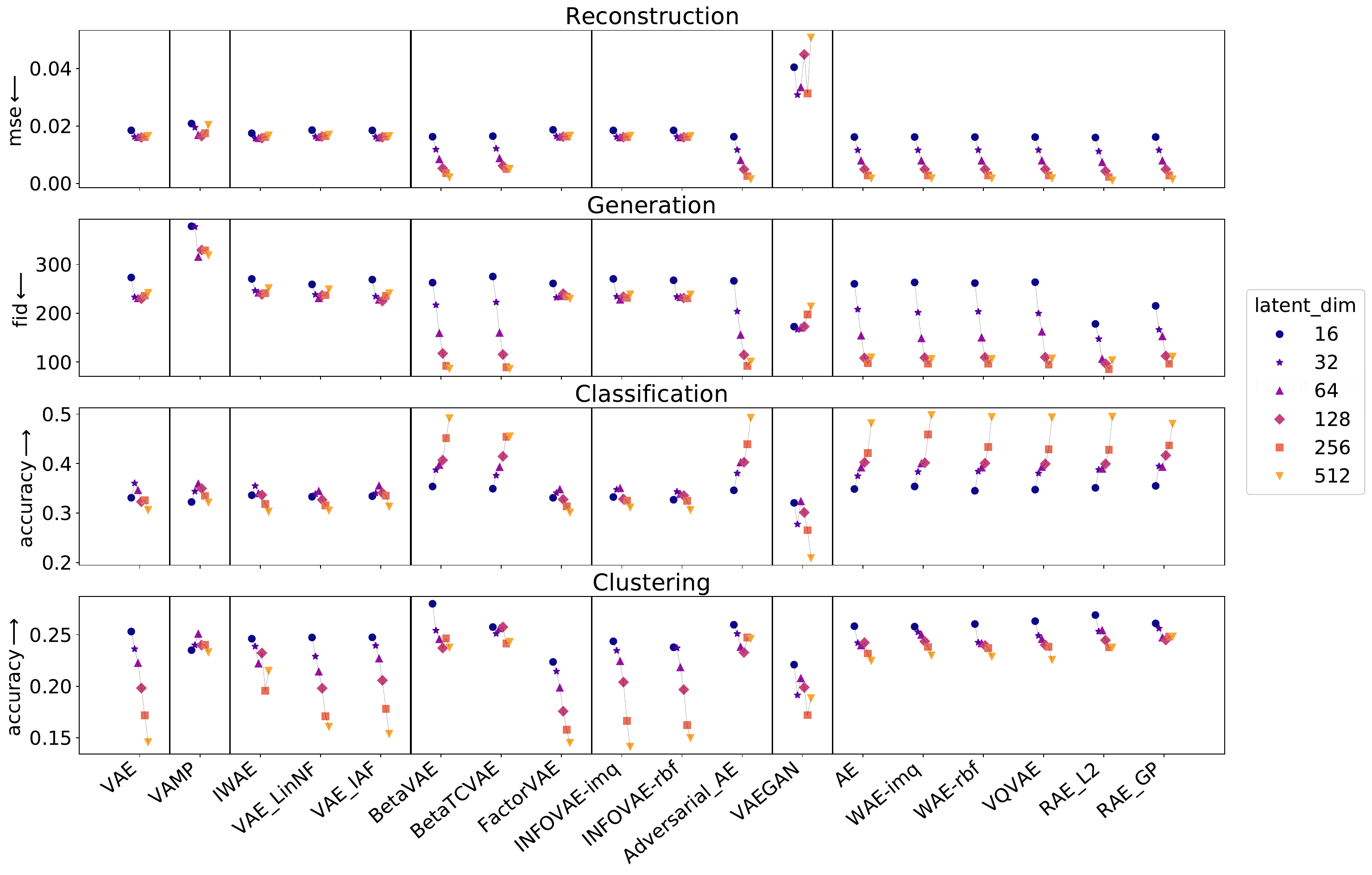}
    \caption{\emph{From top to bottom:} Evolution of the reconstruction MSE, generation FID, classification accuracy and clustering accuracy with respect to the latent space dimension on the CIFAR dataset.}
    \label{fig:benchmark_cifar}
    \end{figure}

Analogously to the results shown in the paper on the MNIST dataset for the 4 chosen tasks (reconstruction, generation, classification and clustering), Fig.~\ref{fig:benchmark_cifar} shows the impact the choice of the latent space dimension has on the performances of the models on the CIFAR10 dataset, whose image arguably have a greater intrinsic latent dimension than images of the MNIST dataset. Similarly to MNIST, two distinct groups appear: the AE-based methods and variational methods. Again, for all tasks but clustering, variational based methods demonstrate good robustness properties with respect to the dimension of the latent space when compared to AE approaches.

\subsection{Complete generation table}
In Table.~\ref{tab:generation_full} are presented the full results obtained for generation \emph{i.e.} including the MAF and 2-stage VAE sampler \citep{dai_diagnosing_2018}. As mentioned in the paper, it is interesting to note that fitting a GMM instead of using the prior for the variational-based approaches seems to often allow a better image generation since it allows a better prospecting of the learned latent space of each model. Interestingly, it seems that fitting more complex density estimators such as a normalising flow (MAF sampler) or another VAE (2-stage sampler) does not improve the generation results when compared to the GMM for those datasets.

\begin{table}[ht]
    \centering
    \tiny
    \caption{Inception Score (higher is better) and FID (lower is better) computed with 10k samples on the test set. For each model and sampler we report the results obtained by the model achieving the lowest FID score on the validation set.}
    \label{tab:generation_full}
    \begin{tabular}{c|c|cccccc|cccccc}
    \toprule
        \multirow{3}{*}{Model} & \multirow{3}{*}{Sampler} & \multicolumn{6}{c|}{ConvNet}& \multicolumn{6}{c}{ResNet} \\
        %\cmidrule(r){4-7}
        &&\multicolumn{2}{c}{MNIST} & \multicolumn{2}{c}{CIFAR10}& \multicolumn{2}{c|}{CELEBA} & \multicolumn{2}{c}{MNIST}& \multicolumn{2}{c}{CIFAR10}& \multicolumn{2}{c}{CELEBA} \\
        & & FID $\downarrow$  & IS $\uparrow$  & FID & IS  & FID  & IS $\uparrow$ & FID $\downarrow$  & IS $\uparrow$  & FID & IS  & FID  & IS  \\
        \midrule
        \gc & \gc $\mathcal{N}$ & \gc 28.5 & \gc 2.1 & \gc 241.0 & \gc 2.2 & \gc 54.8 & \gc 1.9 & \gc 31.3 & \gc 2.0 & \gc 181.7 & \gc 2.5 & \gc 66.6 & \gc 1.6  \\
                           \gc & \gc  GMM & \gc  26.9 & \gc  2.1 & \gc  235.9 & \gc  2.3 & \gc  52.4 & \gc  1.9 & \gc  32.3 & \gc  2.1 & \gc  179.7 & \gc  2.5 & \gc  63.0 & \gc  1.7  \\
                           \gc & \gc VAE & \gc 40.3 & \gc 2.0 & \gc 337.5 & \gc 1.7 & \gc 70.9 & \gc 1.6 & \gc 48.7 & \gc  1.8 & \gc 358.0 & \gc 1.3 & \gc 76.4 & \gc 1.4 \\
                            \multirow{-4}{*}{\gc VAE} & \gc  MAF & \gc  26.8 & \gc  2.1 & \gc  239.5 & \gc  2.2 & \gc  52.5 & \gc  2.0 & \gc  31.0 & \gc   2.1 & \gc  181.5 & \gc  2.5 & \gc  62.9 & \gc  1.7 \\
                            % & IAF Sampler \\
        \midrule
        \multirow{1}{*}{VAMP} & VAMP & 64.2 & 2.0 & 329.0 & 1.5 & 56.0 & 1.9 & 34.5 &  2.1 & 181.9 & 2.5 & 67.2 & 1.6  \\
        \midrule
        \gc & \gc $\mathcal{N}$ & \gc29.0 & \gc  2.1 & \gc  245.3 & \gc  2.1 & \gc  55.7 & \gc  1.9 & \gc  32.4 & \gc   2.0 & \gc  191.2 & \gc  2.4 & \gc  67.6 & \gc  1.6  \\
                            \gc & \gc GMM & \gc 28.4 & \gc 2.1 & \gc 241.2 & \gc 2.1 & \gc 52.7 & \gc 1.9 & \gc 34.4 & \gc  2.1 & \gc 188.8 & \gc 2.4 & \gc 64.1 & \gc 1.7 \\
                            \gc & \gc  VAE & \gc  42.4 & \gc  2.0 & \gc  346.6 & \gc  1.5 & \gc  74.3 & \gc  1.5 & \gc  50.1 & \gc   1.9 & \gc  364.8 & \gc  1.2 & \gc  76.4 & \gc  1.4  \\
                            \multirow{-4}{*}{\gc IWAE} & \gc MAF & \gc 28.1 & \gc 2.1 & \gc 243.4 & \gc 2.1 & \gc 52.7 & \gc 1.9 & \gc 32.5 & \gc  2.1 & \gc 190.4 & \gc 2.4 & \gc 64.3 & \gc 1.7  \\
                            % & IAF Sampler \\
        %\midrule
        \multirow{4}{*}{VAE-lin NF} & $\mathcal{N}$ &  29.3 & 2.1 & 240.3 & 2.1 & 56.5 & 1.9 & 32.5 &  2.0 & 185.5 & 2.4 & 67.1 & 1.6 \\
                             &  GMM &  28.4 &  2.1 &  237.0 &  2.2 &  53.3 &  1.9 &  33.1 &   2.1 &  183.1 &  2.5 &  62.8 &  1.7 \\
                             & VAE & 40.1 & 2.0 & 311.0 & 1.6 & 71.1 & 1.6 & 49.7 &  1.9 & 296.2 & 1.7 & 75.6 & 1.4  \\
                             &  MAF &  27.7 &  2.1 &  239.1 &  2.1 &  53.4 &  2.0 &  32.4 &   2.0 &  184.2 &  2.5 &  62.7 &  1.7 \\
                            % & IAF Sampler \\
        %\midrule
        \gc & \gc $\mathcal{N}$ & \gc 27.5 & \gc 2.1 & \gc 236.0 & \gc 2.2 & \gc 55.4 & \gc 1.9 & \gc 30.6 & \gc  2.0 & \gc 183.6 & \gc 2.5 & \gc 66.2 & \gc 1.6  \\
                            \gc & \gc GMM & \gc 27.0 & \gc 2.1 & \gc 235.4 & \gc 2.2 & \gc 53.6 & \gc 1.9 & \gc 32.2 & \gc  2.1 & \gc 180.8 & \gc 2.5 & \gc 62.7 & \gc 1.7  \\
                            \gc & \gc VAE & \gc 39.4 & \gc 2.0 & \gc 330.5 & \gc 1.1 & \gc 73.0 & \gc 1.5 & \gc 44.8 & \gc  1.9 & \gc 322.7 & \gc 1.5 & \gc 76.7 & \gc 1.4  \\
                            \multirow{-4}{*}{\gc VAE-IAF}  & \gc MAF & \gc 26.9 & \gc 2.1 & \gc 236.8 & \gc 2.2 & \gc 53.6 & \gc 1.9 & \gc 30.6 & \gc  2.1 & \gc 182.5 & \gc 2.5 & \gc 63.0 & \gc 1.7 \\
                            % & IAF Sampler \\
        \midrule
        \multirow{4}{*}{$\beta$-VAE} & $\mathcal{N}$ & 21.4 & 2.1 & 115.4 & 3.6 & 56.1 & 1.9 & 19.1 &  2.0 & 124.9 & 3.4 & 65.9 & 1.6  \\
                             &  GMM &  9.2 &  2.2 &  92.2 &  3.9 &  51.7 &  1.9 &  11.4 &   2.1 &  112.6 &  3.6 &  59.3 &  1.7 \\
                              & VAE & 14.0 & 2.2 & 139.6 & 3.6 & 55.0 & 1.9 & 20.3 &  2.1 & 152.5 & 3.5 & 61.5 & 1.7 \\
                             &  MAF &  9.5 &  2.2 &  100.9 &  3.5 &  51.5 &  2.0 &  12.0 &   2.1 &  120.0 &  3.6 &  59.7 &  1.8 \\
                            % & IAF Sampler \\ 
        %\midrule
        %  \multirow{4}{*}{Dis $\beta$-VAE} & $\mathcal{N}(0,1)$ & 96.5 & 2.3 & 219.4 & 3.6 & 130.8 & 1.6 & 109.4 &  2.7 & 209.8 & 3.2 & 110.6 & 1.5 \\
        %                       & GMM &  192.7 & 2.3 & 300.8 & 1.8 & 94.0 & 1.5 & 98.7 &  2.1 & 202.3 & 2.6 & 83.2 & 1.6 \\
        %                       & $2$-s sampler & 236.1 & 1.5 & 371.2 & 1.0 & 167.9 & 1.0 & 250.8 &  1.1 & 349.0 & 1.2 & 161.0 & 1.2  \\
        %                       & MAF sampler & 191.8 & 2.2 & 300.6 & 1.8 & 94.3 & 1.5 & 98.3 &  2.1 & 200.6 & 2.6 & 82.6 & 1.7  \\
        % %                     % & IAF Sampler \\
        %  \midrule
        \gc & \gc $\mathcal{N}$ & \gc 21.3 & \gc 2.1 & \gc 116.6 & \gc 2.8 & \gc 55.7 & \gc 1.8 & \gc 20.7 & \gc  2.0 & \gc 125.8 & \gc 3.4 & \gc 65.9 & \gc 1.6  \\
                            \gc & \gc GMM & \gc 11.6 & \gc 2.2 & \gc 89.3 & \gc 4.1 & \gc 51.8 & \gc 1.9 & \gc 13.3 & \gc  2.1 & \gc \textbf{106.5} & \gc \textbf{3.7} & \gc 59.3 & \gc 1.7  \\
                            \gc & \gc VAE & \gc 18.4 & \gc 2.2 & \gc 127.9 & \gc 4.2 & \gc 59.7 & \gc 1.8 & \gc 28.3 & \gc  2.0 & \gc 164.0 & \gc 3.3 & \gc 66.4 & \gc 1.5 \\
                            \multirow{-4}{*}{\gc $\beta$-TC VAE} & \gc MAF & \gc 12.0 & \gc 2.2 & \gc 95.6 & \gc 3.6 & \gc 52.2 & \gc 1.9 & \gc 13.7 & \gc  2.1 & \gc 116.6 & \gc 3.4 & \gc 60.1 & \gc 1.7  \\
                            % & IAF Sampler \\
        %\midrule
        \multirow{4}{*}{FactorVAE} & $\mathcal{N}$ & 27.0 & 2.1 & 236.5 & 2.2 & 53.8 & 1.9 & 31.0 &  2.0 & 185.4 & 2.5 & 66.4 & 1.7 \\
                             &  GMM &  26.9 &  2.1 &  234.0 &  2.2 &  52.4 &  2.0 &  32.7 &   2.1 &  184.4 &  2.5 &  63.3 &  1.7 \\
                             & VAE & 41.2 & 1.9 & 338.3 & 1.5 & 75.0 & 1.5 & 54.7 &  1.8 & 316.2 & 1.3 & 77.7 & 1.4 \\
                             &  MAF &  26.7 &  2.2 &  236.7 &  2.2 &  52.7 &  1.9 &  32.8 &   2.1 &  185.8 &  2.5 &  63.4 &  1.7 \\
                            % & IAF Sampler \\
        %\midrule
        \gc & \gc $\mathcal{N}$ & \gc 27.5 & \gc 2.1 & \gc 235.2 & \gc 2.1 & \gc 55.5 & \gc 1.9 & \gc 31.1 & \gc  2.0 & \gc 182.8 & \gc 2.5 & \gc 66.5 & \gc 1.6  \\
                            \gc & \gc GMM & \gc 26.7 & \gc 2.1 & \gc 230.4 & \gc 2.2 & \gc 52.7 & \gc 1.9 & \gc 32.3 & \gc  2.1 & \gc 179.5 & \gc 2.5 & \gc 62.8 & \gc 1.7 \\
                            \gc & \gc VAE & \gc 39.7 & \gc 2.0 & \gc 327.2 & \gc 1.5 & \gc 73.7 & \gc 1.5 & \gc 50.6 & \gc  1.9 & \gc 363.4 & \gc 1.2 & \gc 75.8 & \gc 1.4  \\
                            \multirow{-4}{*}{\gc InfoVAE - RBF} & \gc MAF & \gc 25.9 & \gc 2.1 & \gc 233.5 & \gc 2.2 & \gc 52.2 & \gc 2.0 & \gc 30.5 & \gc  2.1 & \gc 181.3 & \gc 2.5 & \gc 62.7 & \gc 1.7  \\
                            % & IAF Sampler \\
        \midrule
        \multirow{4}{*}{InfoVAE - IMQ} & $\mathcal{N}$ & 28.3 & 2.1 & 233.8 & 2.2 & 56.7 & 1.9 & 31.0 &  2.0 & 182.4 & 2.5 & 66.4 & 1.6  \\
                             &  GMM &  27.7 &  2.1 &  231.9 &  2.2 &  53.7 &  1.9 &  32.8 &   2.1 &  180.7 &  2.6 &  62.3 &  1.7 \\
                             & VAE & 40.4 & 1.9 & 323.8 & 1.6 & 73.7 & 1.5 & 49.9 &  1.9 & 341.8 & 1.8 & 75.7 & 1.4  \\
                             &  MAF &  27.2 &  2.1 &  232.3 &  2.1 &  53.8 &  2.0 &  30.6 &   2.1 &  182.5 &  2.5 &  62.6 &  1.7  \\
                            % & IAF Sampler \\
        %\midrule
        \gc & \gc $\mathcal{N}$ & \gc 16.8 & \gc 2.2 & \gc 139.9 & \gc 2.6 & \gc 59.9 & \gc 1.8 & \gc 19.1 & \gc  2.1 & \gc 164.9 & \gc 2.4 & \gc 64.8 & \gc 1.7  \\
                            \gc & \gc GMM & \gc 9.3 & \gc 2.2 & \gc 92.1 & \gc 3.8 & \gc 53.9 & \gc 2.0 & \gc 11.1 & \gc  2.1 & \gc 118.5 & \gc 3.5 & \gc 58.7 & \gc 1.8 \\
                            \gc & \gc VAE & \gc 13.4 & \gc 2.2 & \gc 144.0 & \gc 3.4 & \gc 58.2 & \gc 1.8 & \gc 15.1 & \gc  2.1 & \gc 145.2 & \gc 3.6 & \gc 59.0 & \gc 1.7 \\
                            \multirow{-4}{*}{\gc AAE} & \gc MAF & \gc 9.3 & \gc 2.2 & \gc 101.1 & \gc 3.2 & \gc 53.8 & \gc 2.0 & \gc 11.9 & \gc  2.1 & \gc 133.6 & \gc 3.1 & \gc 59.2 & \gc 1.8  \\
                            % & IAF Sampler \\
        \midrule
        \multirow{4}{*}{MSSSIM-VAE} & $\mathcal{N}$ & 26.7 & 2.2 & 279.9 & 1.7 & 124.3 & 1.3 & 28.0 &  2.1 & 254.2 & 1.7 & 119.0 & 1.3 \\
                             &  GMM &  27.2 &  2.2 &  279.7 &  1.7 &  124.3 &  1.3 &  28.8 &   2.1 &  253.1 &  1.7 &  119.2 &  1.3 \\
                             & VAE & 51.2 & 1.9 & 355.5 & 1.1 & 137.9 & 1.2 & 51.6 &  1.9 & 372.1 & 1.1 & 136.5 & 1.2  \\
                             &  MAF &  26.9 &  2.2 &  279.8 &  1.7 &  124.0 &  1.3 &  27.5 &   2.1 &  254.1 &  1.7 &  119.5 &  1.3  \\
                            % & IAF Sampler \\
        %\midrule
        \gc & \gc $\mathcal{N}$ & \gc 8.7 & \gc 2.2 & \gc 199.5 & \gc 2.2 & \gc 39.7 & \gc 1.9 & \gc 12.8 & \gc  2.2 & \gc 198.7 & \gc 2.2 & \gc 122.8 & \gc 2.0  \\
                            \gc & \gc GMM & \gc \textbf{6.3} & \gc \textbf{2.2} & \gc 197.5 & \gc 2.1 & \gc \textbf{35.6} & \gc 1.8 & \gc \textbf{6.5} & \gc  2.2 & \gc 188.2 & \gc 2.6 & \gc 84.3 & \gc 1.7  \\
                            \gc & \gc VAE & \gc 11.2 & \gc 2.1 & \gc 310.9 & \gc 2.0 & \gc 54.5 & \gc 1.6 & \gc 9.2 & \gc  2.1 & \gc 272.7 & \gc 2.0 & \gc 88.8 & \gc 1.6  \\
                            \multirow{-4}{*}{\gc VAEGAN} & \gc MAF & \gc 6.9 & \gc 2.3 & \gc 199.0 & \gc 2.1 & \gc 36.7 & \gc 1.8 & \gc 6.6 & \gc \textbf{ 2.2} & \gc 191.9 & \gc 2.5 & \gc 84.8 & \gc 1.7 \\
                            % & IAF Sampler \\
        \midrule
        \multirow{3}{*}{AE}  & $\mathcal{N}$ & 26.7 & 2.1 & 201.3 & 2.1 & 327.7 & 1.0 & 221.8 &  1.3 & 210.1 & 2.1 & 275.0 & 2.9  \\
&  GMM &  9.3 &  2.2 &  97.3 &  3.6 &  55.4 &  2.0 &  11.0 &   2.1 &  120.7 &  3.4 &  \textbf{57.4} &  1.8  \\
                             & MAF &  9.9 & 2.2 & 108.3 & 3.1 & 55.7 & 2.0 & 12.0 &  2.1 & 136.5 & 3.0 & 58.3 & 1.8 \\
                            % & IAF Sampler \\
        %\midrule
        \gc & \gc $\mathcal{N}$ & \gc 21.2 & \gc 2.2 & \gc 175.1 & \gc 2.0 & \gc 332.6 & \gc 1.0 & \gc 21.2 & \gc  2.1 & \gc 170.2 & \gc 2.3 & \gc 69.4 & \gc 1.6  \\
                            \gc & \gc GMM & \gc 9.2 & \gc 2.2 & \gc 97.1 & \gc 3.6 & \gc 55.0 & \gc 2.0 & \gc 11.2 & \gc  2.1 & \gc 120.3 & \gc 3.4 & \gc 58.3 & \gc 1.7 \\
                            \multirow{-3}{*}{\gc WAE - RBF} & \gc MAF & \gc 9.8 & \gc 2.2 & \gc 108.2 & \gc 3.1 & \gc 56.0 & \gc 2.0 & \gc 11.8 & \gc  2.2 & \gc 135.3 & \gc 3.0 & \gc 58.3 & \gc 1.8  \\
        %\midrule
        \multirow{3}{*}{WAE - IMQ} & $\mathcal{N}$ & 18.9 & 2.2 & 164.4 & 2.2 & 64.6 & 1.7 & 20.3 &  2.1 & 150.7 & 2.5 & 67.1 & 1.6 \\
                             &  GMM &  8.6 &  2.2 &  96.5 &  3.6 &  51.7 &  2.0 &  11.2 &   2.1 &  119.0 &  3.5 &  57.7 &  1.8  \\
                             & MAF &  9.5 & 2.2 & 107.8 & 3.1 & 51.6 & 2.0 & 11.8 &  2.1 & 130.2 & 3.0 & 58.7 & 1.7  \\
        %\midrule
        \gc  & \gc $\mathcal{N}(0,1)$ & \gc 28.2 & \gc 2.0 & \gc 152.2 & \gc 2.0 & \gc 306.9 & \gc 1.0 & \gc 170.7 & \gc  1.6 & \gc 195.7 & \gc 1.9 & \gc 140.3 & \gc \textbf{2.2}  \\
       \gc & \gc GMM & \gc 9.1 & \gc 2.2 & \gc 95.2 & \gc 3.7 & \gc 51.6 & \gc 2.0 & \gc 10.7 & \gc  2.1 & \gc 120.1 & \gc 3.4 & \gc 57.9 & \gc 1.8  \\
                            \multirow{-3}{*}{\gc VQVAE} & \gc MAF & \gc 9.6 & \gc 2.2 & \gc 104.7 & \gc 3.2 & \gc 52.3 & \gc 1.9 & \gc 11.7 & \gc  2.2 & \gc 136.8 & \gc 3.0 & \gc 57.9 & \gc 1.8 \\
        %\midrule
        \multirow{3}{*}{RAE-L2} & $\mathcal{N}$ & 25.0 & 2.0 & 156.1 & 2.6 & 86.1 & \textbf{2.8} & 63.3 &  2.2 & 170.9 & 2.2 & 168.7 & 3.1  \\
                             &  GMM &  9.1 &  2.2 &  \textbf{85.3} &  \textbf{3.9} &  55.2 &  1.9 &  11.5 &   2.1 &  122.5 &  3.4 &  58.3 &  1.8 \\
                             & MAF & 9.5 & 2.2 & 93.4 & 3.5 & 55.2 & 2.0 & 12.3 &  2.2 & 136.6 & 3.0 & 59.1 & 1.7  \\
                            % & IAF Sampler \\
        %\midrule
        \gc & \gc $\mathcal{N}$ & \gc  27.1 & \gc 2.1 & \gc 196.8 & \gc 2.1 & \gc 86.1 & \gc 2.4 & \gc 61.5 & \gc  2.2 & \gc 229.1 & \gc 2.0 & \gc 201.9 & \gc 3.1  \\
                            \gc & \gc GMM & \gc 9.7 & \gc 2.2 & \gc 96.3 & \gc 3.7 & \gc 52.5 & \gc 1.9 & \gc 11.4 & \gc  2.1 & \gc 123.3 & \gc 3.4 & \gc 59.0 & \gc 1.8 \\
                            \multirow{-3}{*}{\gc RAE - GP} & \gc MAF & \gc 9.7 & \gc 2.2 & \gc 106.3 & \gc 3.2 & \gc 52.5 & \gc 1.9 & \gc 12.2 & \gc  2.2 & \gc 139.4 & \gc 3.0 & \gc 59.5 & \gc 1.8  \\

    \bottomrule
    \end{tabular}
\end{table}

\clearpage

\subsection{Further interesting results}

\paragraph{Generated samples}

\begin{figure}[ht]
    \centering
    \captionsetup[subfigure]{position=above, labelformat = empty}
    \adjustbox{minipage=6em,raise=\dimexpr -2.2\height}{\small VAE}
    \subfloat[MNIST - $\mathcal{N}$]{\includegraphics[width=2.3in]{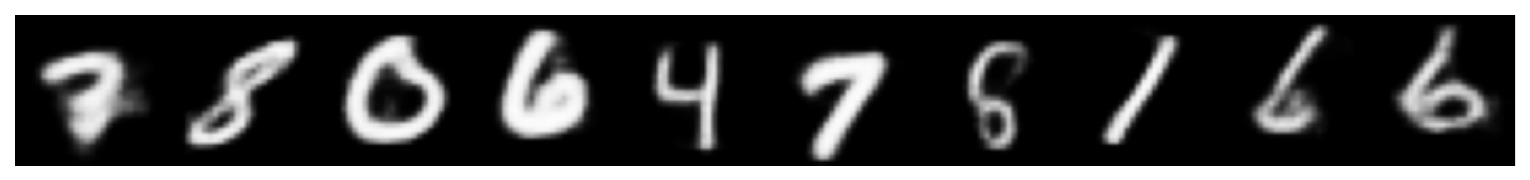}}
    \subfloat[MNIST - GMM]{\includegraphics[width=2.3in]{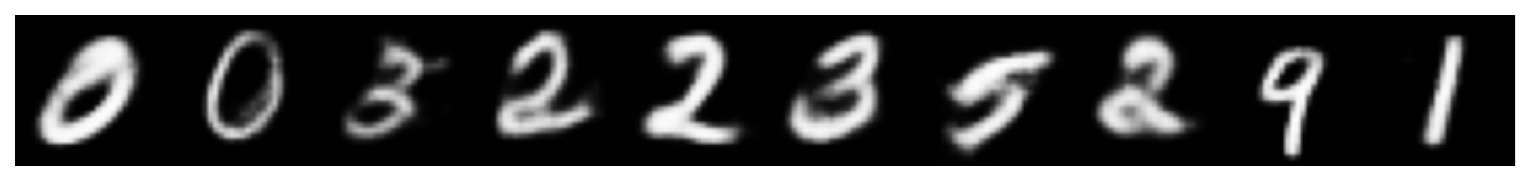}}\\\vspace{-1.3em}
    \adjustbox{minipage=6em,raise=\dimexpr -2.2\height}{\small IWAE}
    \subfloat{\includegraphics[width=2.3in]{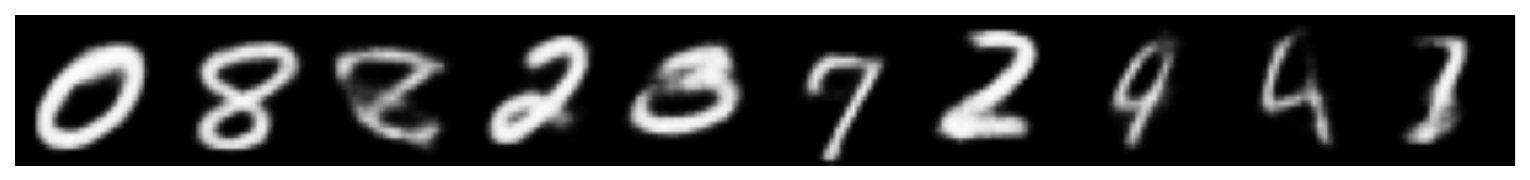}}
    \subfloat{\includegraphics[width=2.3in]{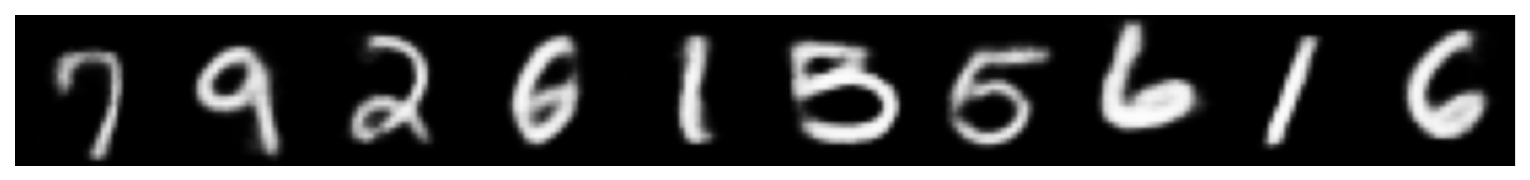}}\\\vspace{-1.3em}
    \adjustbox{minipage=6em,raise=\dimexpr -2.2\height}{\scriptsize VAE-lin-NF}
    \subfloat{\includegraphics[width=2.3in]{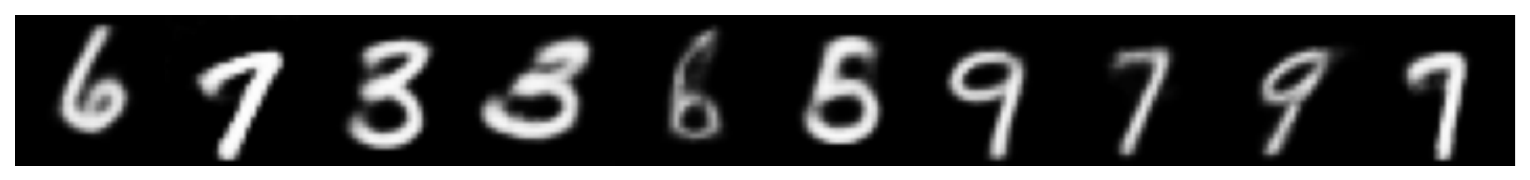}}
    \subfloat{\includegraphics[width=2.3in]{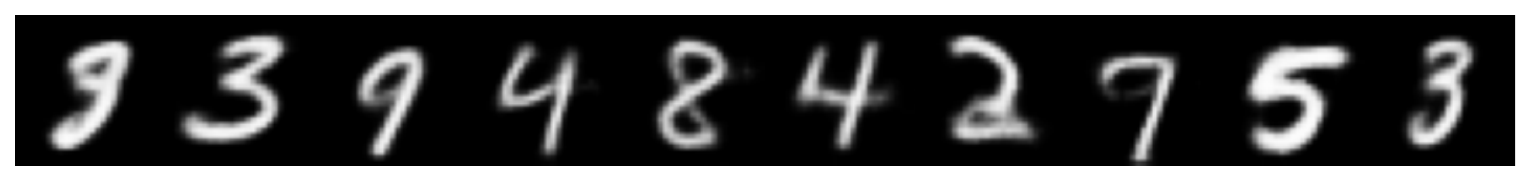}}\\\vspace{-1.3em}
     \adjustbox{minipage=6em,raise=\dimexpr -2.2\height}{\small VAE-IAF}
    \subfloat{\includegraphics[width=2.3in]{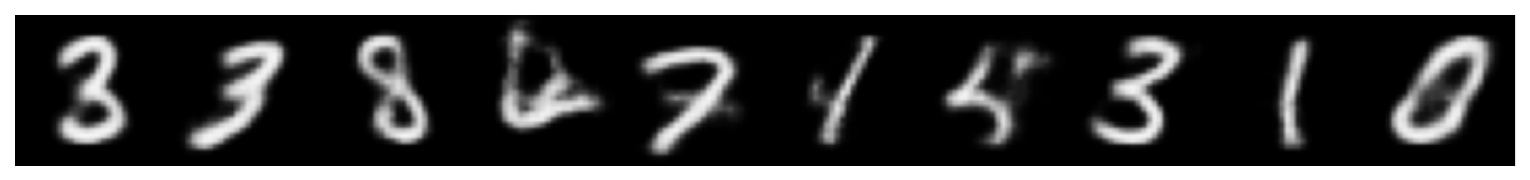}}
    \subfloat{\includegraphics[width=2.3in]{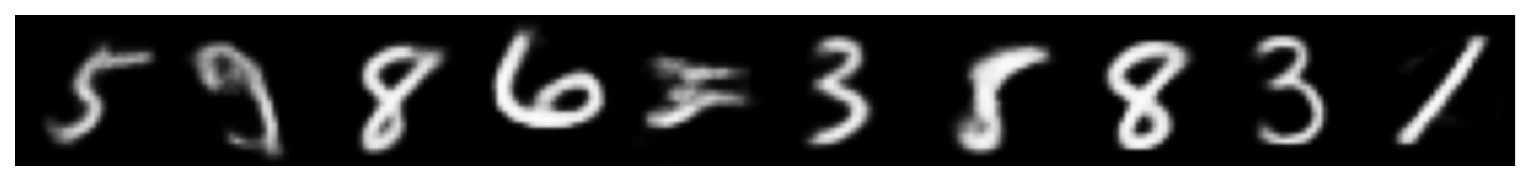}}\\\vspace{-1.3em}
    \adjustbox{minipage=6em,raise=\dimexpr -2.2\height}{\small $\beta$-VAE}
    \subfloat{\includegraphics[width=2.3in]{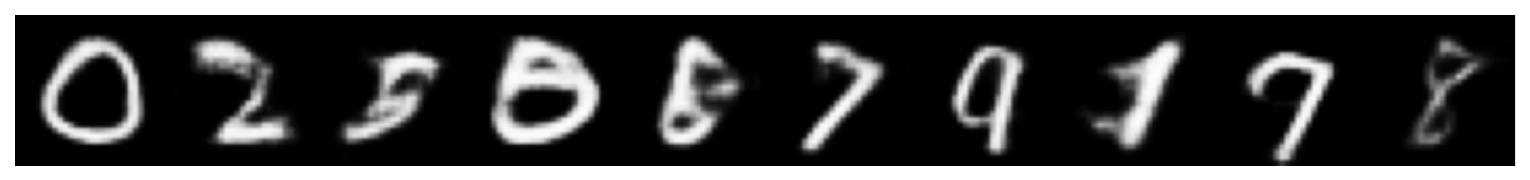}}
    \subfloat{\includegraphics[width=2.3in]{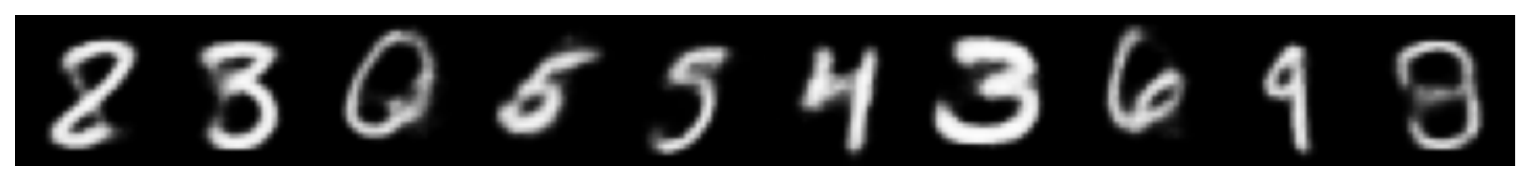}}\\\vspace{-1.3em}
    \adjustbox{minipage=6em,raise=\dimexpr -2.2\height}{\small $\beta$-TC-VAE}
    \subfloat{\includegraphics[width=2.3in]{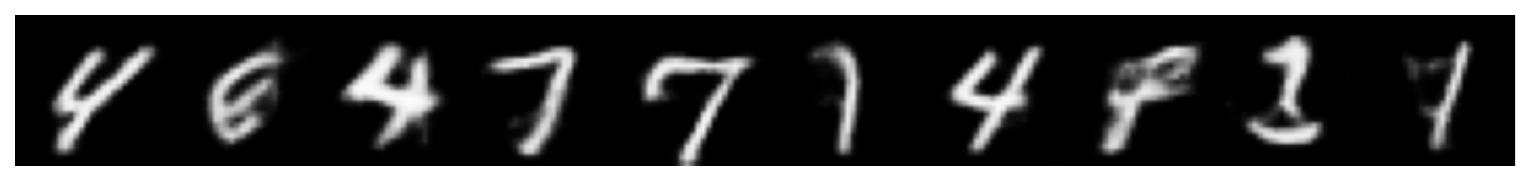}}
    \subfloat{\includegraphics[width=2.3in]{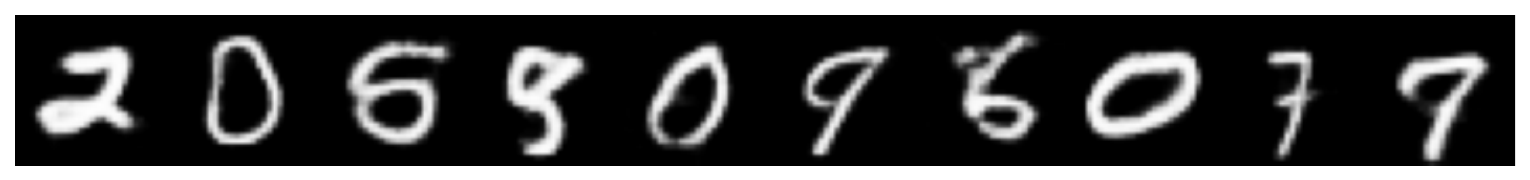}}\\\vspace{-1.3em}
    \adjustbox{minipage=6em,raise=\dimexpr -2.2\height}{\small Factor-VAE}
    \subfloat{\includegraphics[width=2.3in]{plots/generation/mnist/16/normal/factorvae.pdf}}
    \subfloat{\includegraphics[width=2.3in]{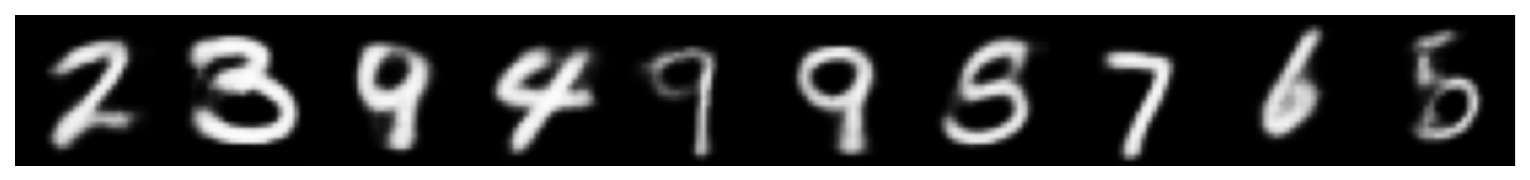}}\\\vspace{-1.3em}
    \adjustbox{minipage=6em,raise=\dimexpr -2.2\height}{\small InfoVAE - IMQ}
    \subfloat{\includegraphics[width=2.3in]{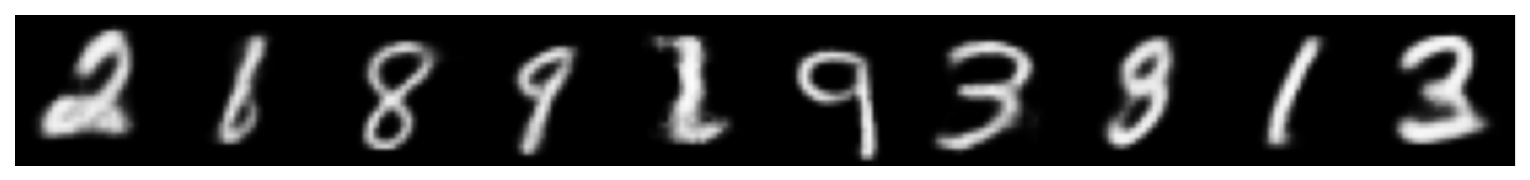}}
    \subfloat{\includegraphics[width=2.3in]{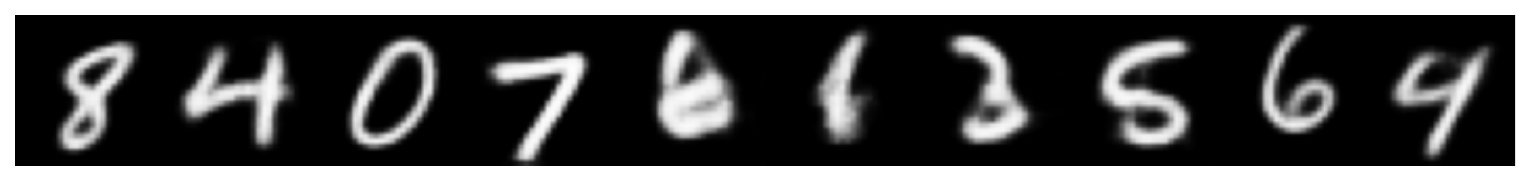}}\\\vspace{-1.3em}
    \adjustbox{minipage=6em,raise=\dimexpr -2.2\height}{\small InfoVAE - RBF}
    \subfloat{\includegraphics[width=2.3in]{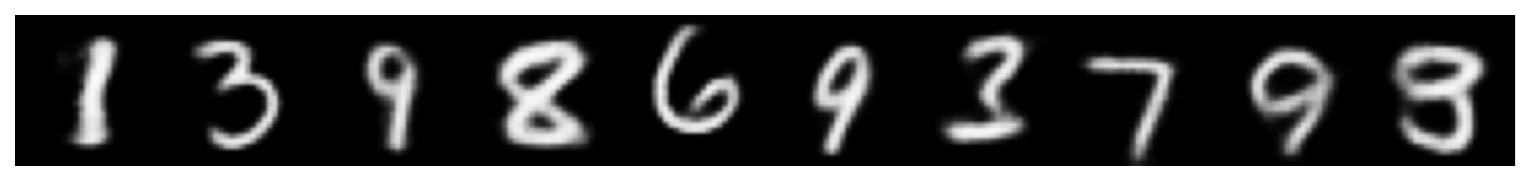}}
    \subfloat{\includegraphics[width=2.3in]{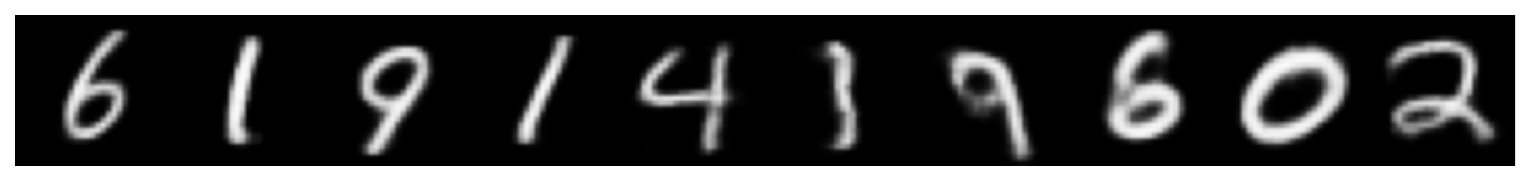}}\\\vspace{-1.3em}
    \adjustbox{minipage=6em,raise=\dimexpr -2.2\height}{\small AAE}
    \subfloat{\includegraphics[width=2.3in]{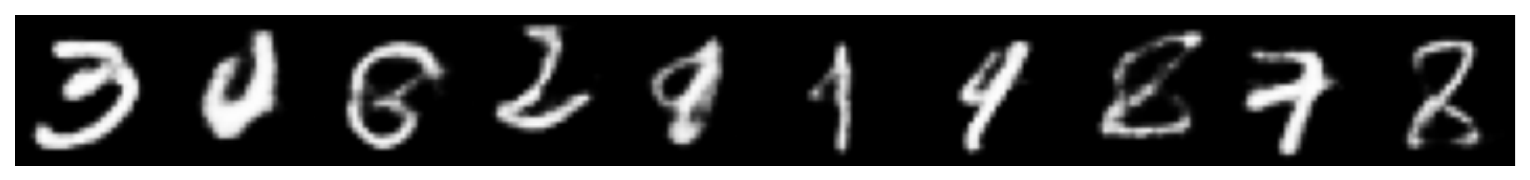}}
    \subfloat{\includegraphics[width=2.3in]{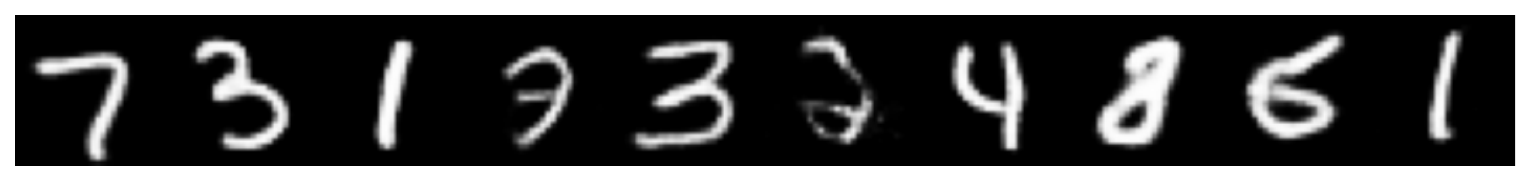}}\\\vspace{-1.3em}
    \adjustbox{minipage=6em,raise=\dimexpr -2.2\height}{\small MSSSIM-VAE}
    \subfloat{\includegraphics[width=2.3in]{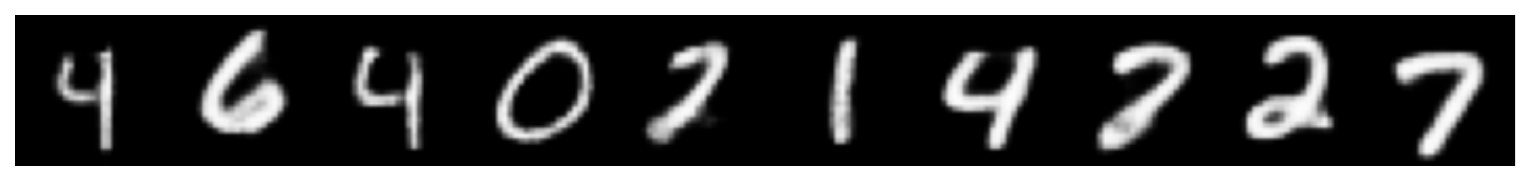}}
    \subfloat{\includegraphics[width=2.3in]{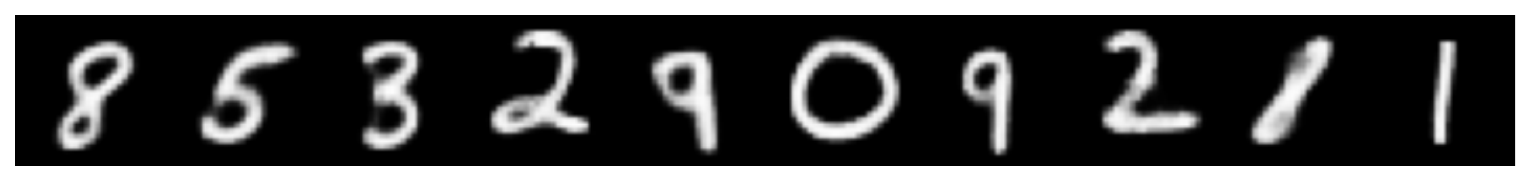}}\\\vspace{-1.3em}
     \adjustbox{minipage=6em,raise=\dimexpr -2.2\height}{\small VAEGAN}
    \subfloat{\includegraphics[width=2.3in]{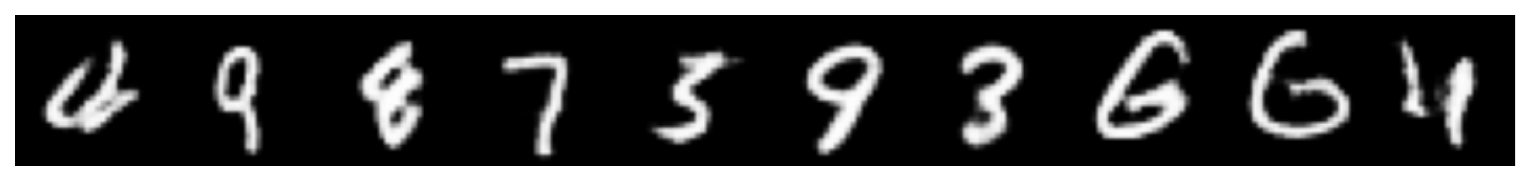}}
    \subfloat{\includegraphics[width=2.3in]{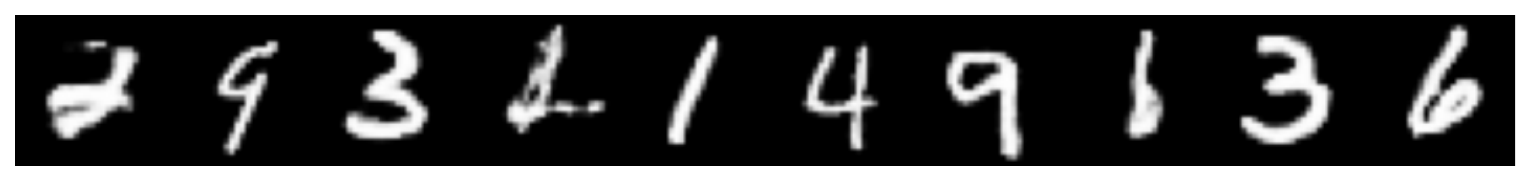}}\\\vspace{-1.3em}
    \adjustbox{minipage=6em,raise=\dimexpr -2.2\height}{\small AE}
    \subfloat{\includegraphics[width=2.3in]{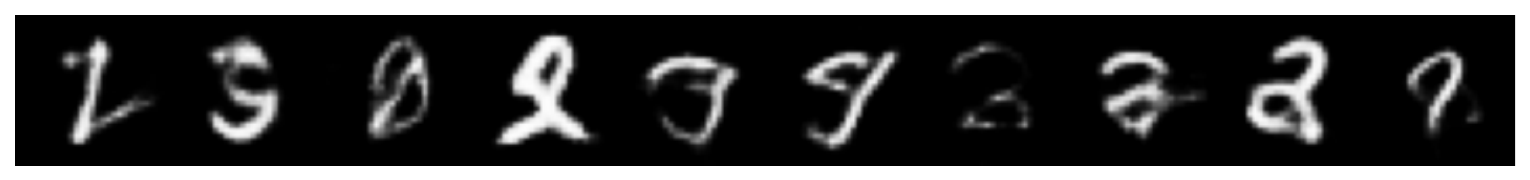}}
    \subfloat{\includegraphics[width=2.3in]{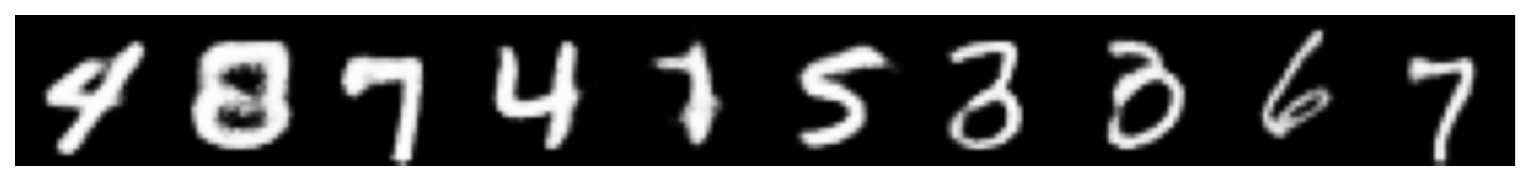}}\\\vspace{-1.3em}
    \adjustbox{minipage=6em,raise=\dimexpr -2.2\height}{\small WAE-IMQ}
    \subfloat{\includegraphics[width=2.3in]{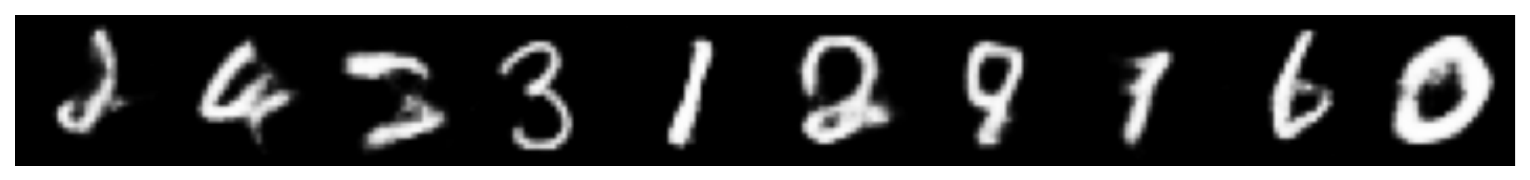}}
    \subfloat{\includegraphics[width=2.3in]{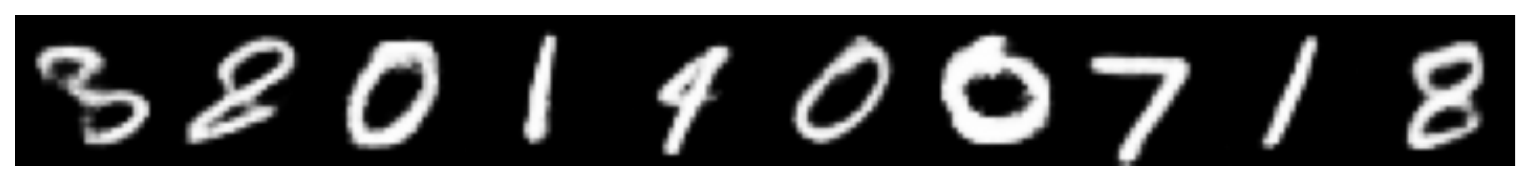}}\\\vspace{-1.3em}
     \adjustbox{minipage=6em,raise=\dimexpr -2.2\height}{\small WAE-RBF}
    \subfloat{\includegraphics[width=2.3in]{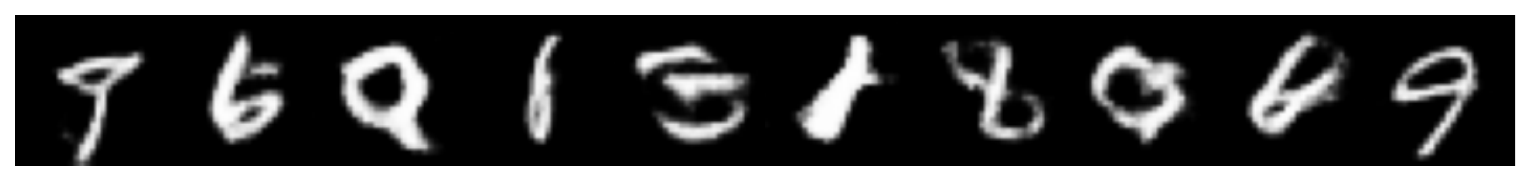}}
    \subfloat{\includegraphics[width=2.3in]{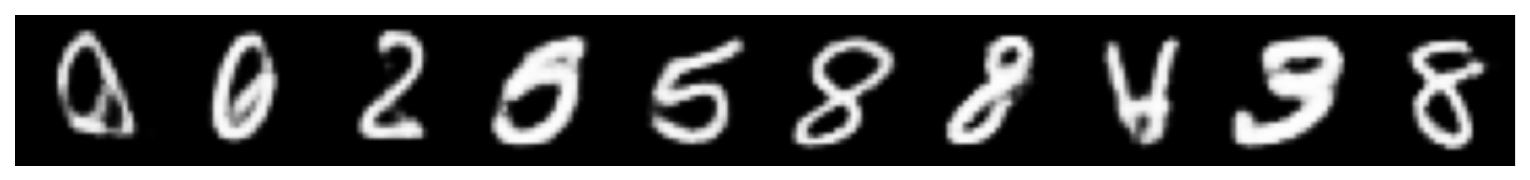}}\\\vspace{-1.3em}
    \adjustbox{minipage=6em,raise=\dimexpr -2.2\height}{\small VQVAE}
    \subfloat{\includegraphics[width=2.3in]{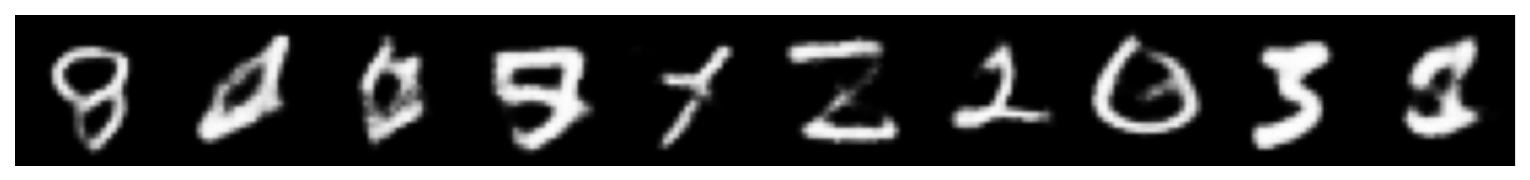}}
    \subfloat{\includegraphics[width=2.3in]{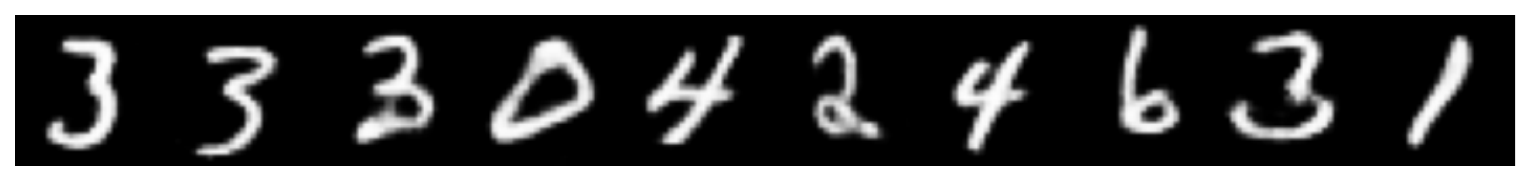}}\\\vspace{-1.3em}
    \adjustbox{minipage=6em,raise=\dimexpr -2.2\height}{\small RAE-L2}
    \subfloat{\includegraphics[width=2.3in]{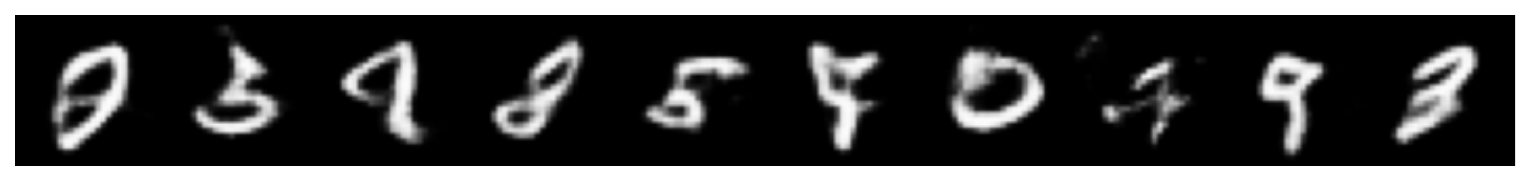}}
    \subfloat{\includegraphics[width=2.3in]{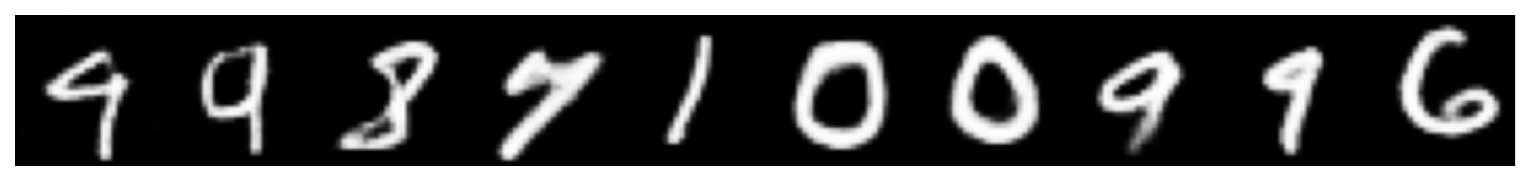}}\\\vspace{-1.3em}
    \adjustbox{minipage=6em,raise=\dimexpr -2.2\height}{\small RAE-GP}
    \subfloat{\includegraphics[width=2.3in]{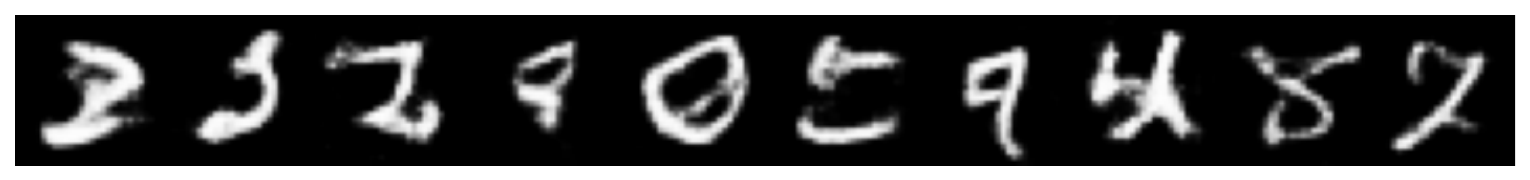}}
    \subfloat{\includegraphics[width=2.3in]{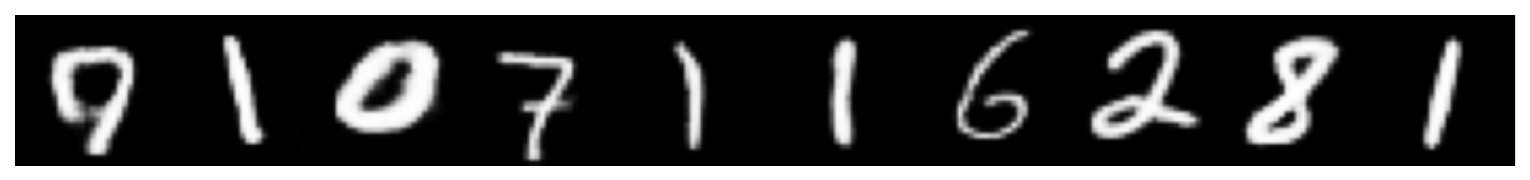}}
    \caption{Generated samples on MNIST for a latent space of dimension 16 and ConvNet architecture. For each model, we select the configuration achieving the lowest FID on the validation set.}
    \label{fig:generation mnist}
    \end{figure}

\begin{figure}[ht]
    \centering
    \captionsetup[subfigure]{position=above, labelformat = empty}
    \adjustbox{minipage=6em,raise=\dimexpr -2.2\height}{\small VAE}
    \subfloat[CELEBA - $\mathcal{N}$]{\includegraphics[width=2.3in]{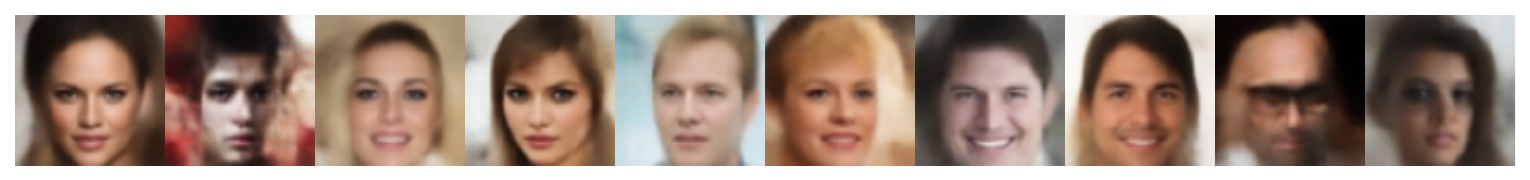}}
    \subfloat[CELEBA - GMM]{\includegraphics[width=2.3in]{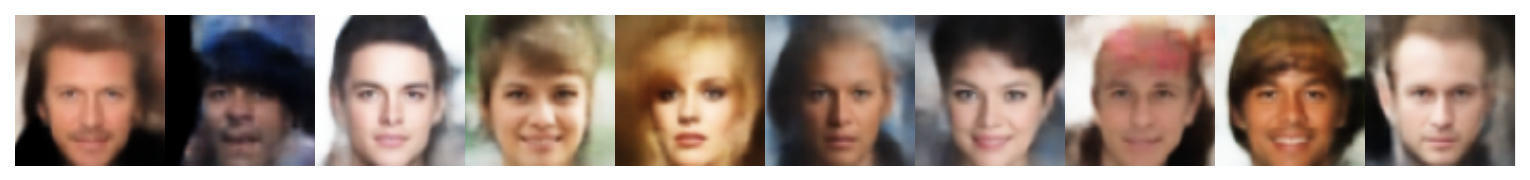}}\\\vspace{-1.3em}
    \adjustbox{minipage=6em,raise=\dimexpr -2.2\height}{\small IWAE}
    \subfloat{\includegraphics[width=2.3in]{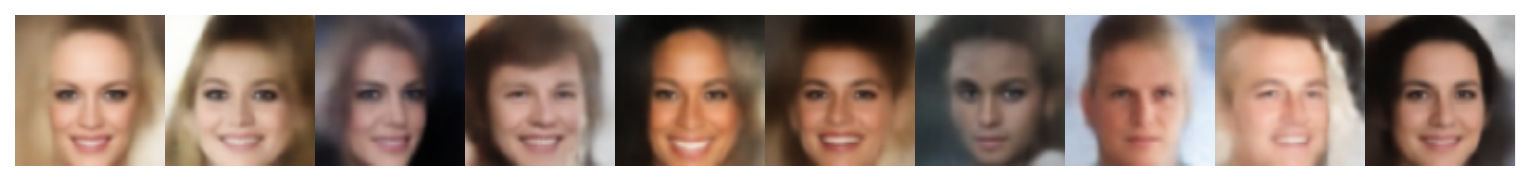}}
    \subfloat{\includegraphics[width=2.3in]{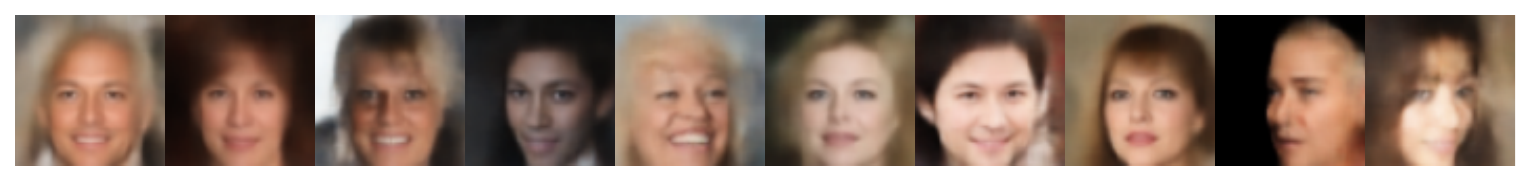}}\\\vspace{-1.3em}
    \adjustbox{minipage=6em,raise=\dimexpr -2.2\height}{\small VAE-lin-NF}
    \subfloat{\includegraphics[width=2.3in]{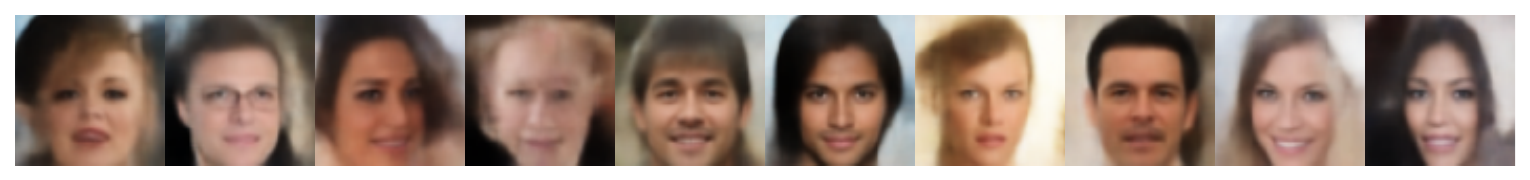}}
    \subfloat{\includegraphics[width=2.3in]{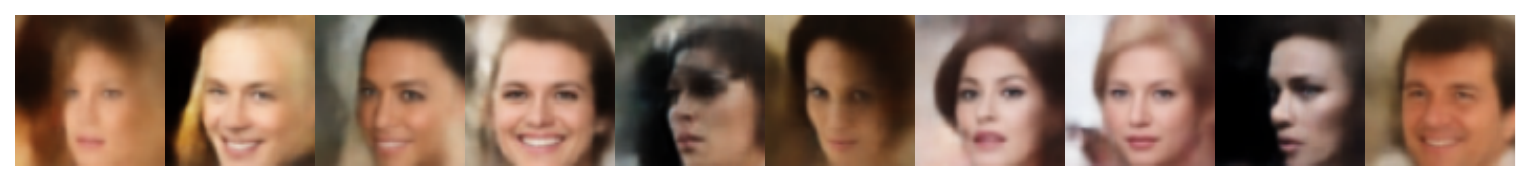}}\\\vspace{-1.3em}
     \adjustbox{minipage=6em,raise=\dimexpr -2.2\height}{\small VAE-IAF}
    \subfloat{\includegraphics[width=2.3in]{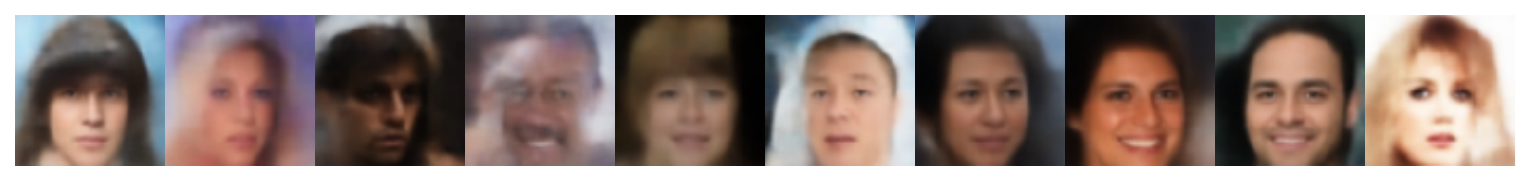}}
    \subfloat{\includegraphics[width=2.3in]{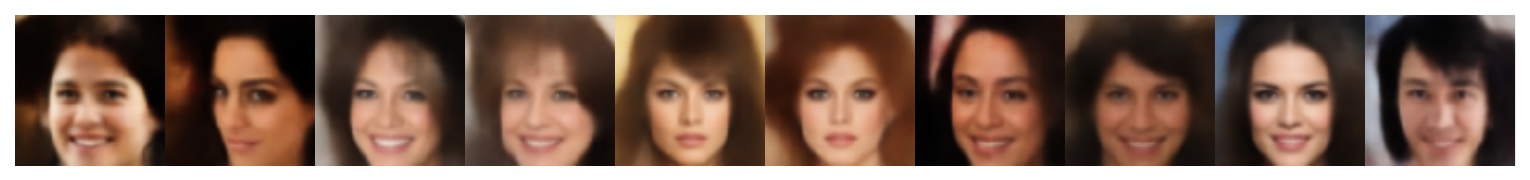}}\\\vspace{-1.3em}
    \adjustbox{minipage=6em,raise=\dimexpr -2.2\height}{\small $\beta$-VAE}
    \subfloat{\includegraphics[width=2.3in]{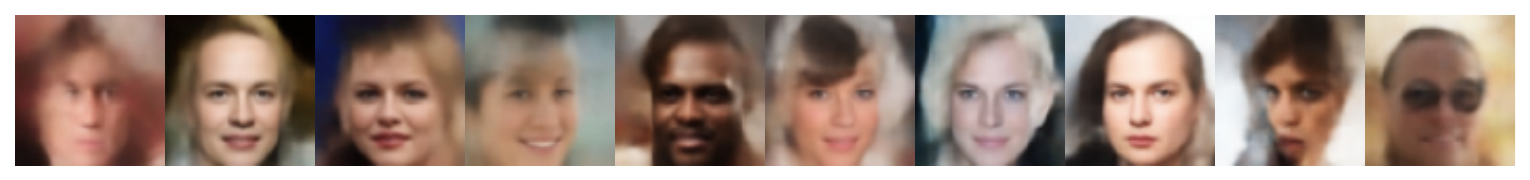}}
    \subfloat{\includegraphics[width=2.3in]{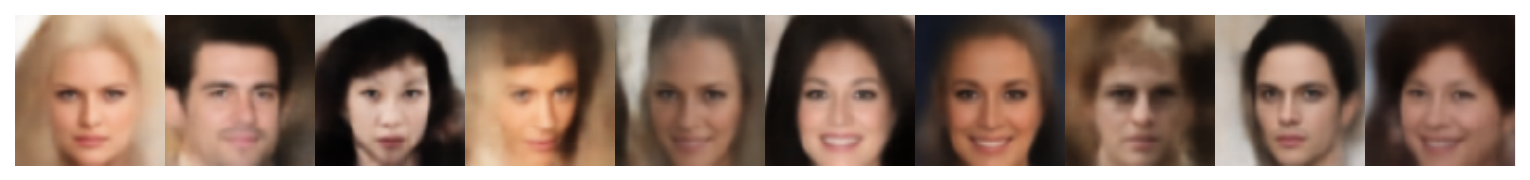}}\\\vspace{-1.3em}
    \adjustbox{minipage=6em,raise=\dimexpr -2.2\height}{\small $\beta$-TC-VAE}
    \subfloat{\includegraphics[width=2.3in]{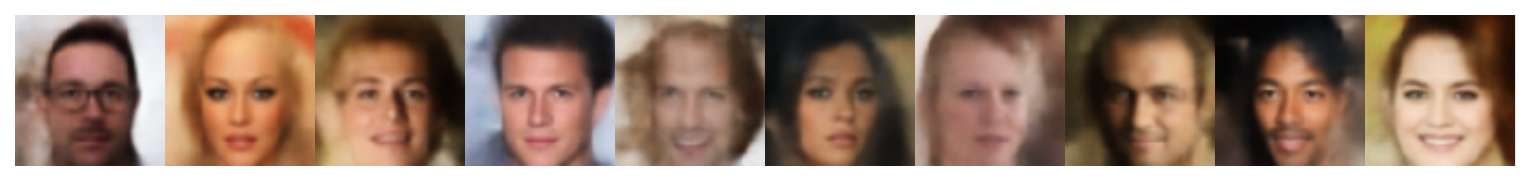}}
    \subfloat{\includegraphics[width=2.3in]{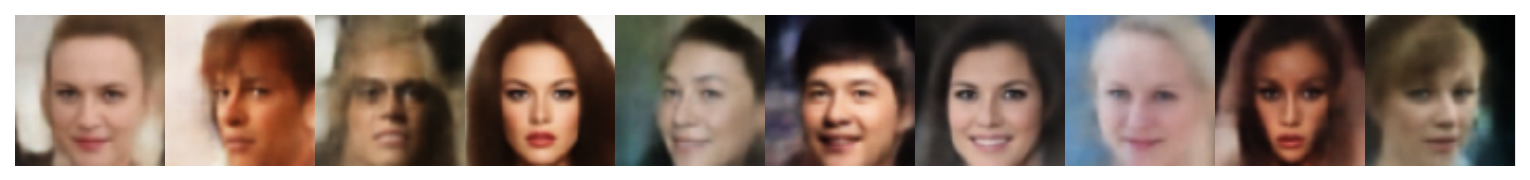}}\\\vspace{-1.3em}
    \adjustbox{minipage=6em,raise=\dimexpr -2.2\height}{\small Factor-VAE}
    \subfloat{\includegraphics[width=2.3in]{plots/generation/celeba/64/normal/factorvae.pdf}}
    \subfloat{\includegraphics[width=2.3in]{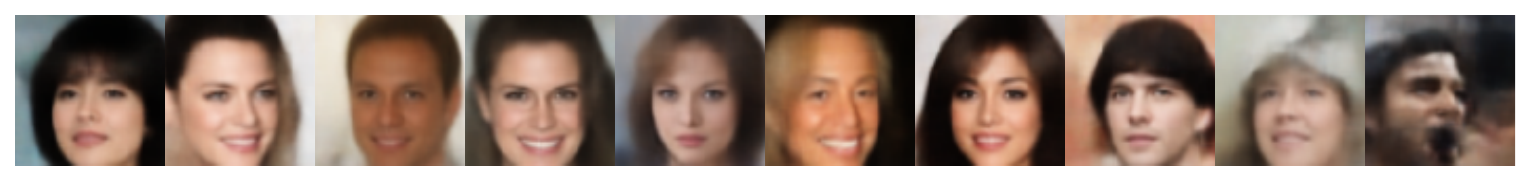}}\\\vspace{-1.3em}
    \adjustbox{minipage=6em,raise=\dimexpr -2.2\height}{\small InfoVAE - IMQ}
    \subfloat{\includegraphics[width=2.3in]{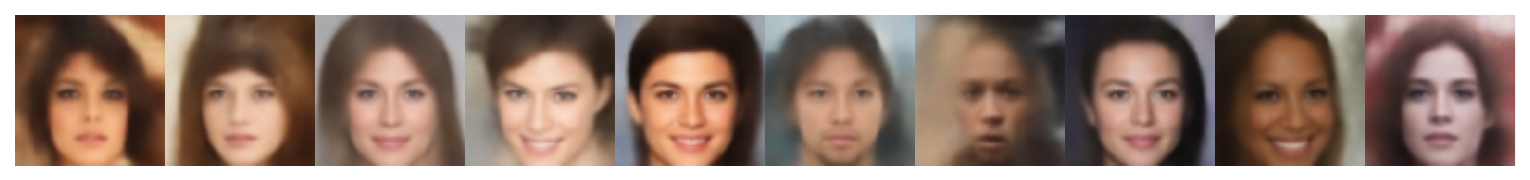}}
    \subfloat{\includegraphics[width=2.3in]{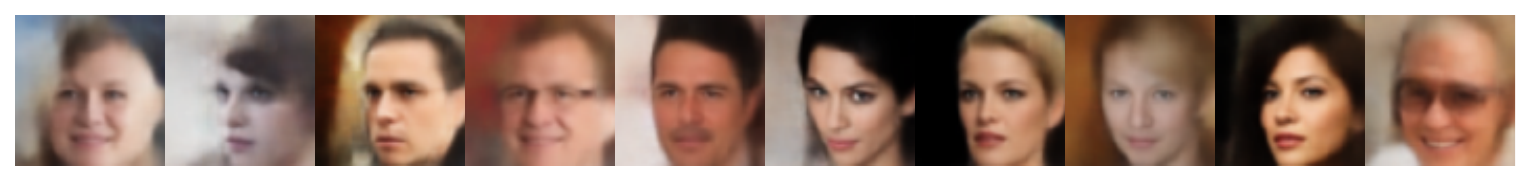}}\\\vspace{-1.3em}
    \adjustbox{minipage=6em,raise=\dimexpr -2.2\height}{\small InfoVAE - RBF}
    \subfloat{\includegraphics[width=2.3in]{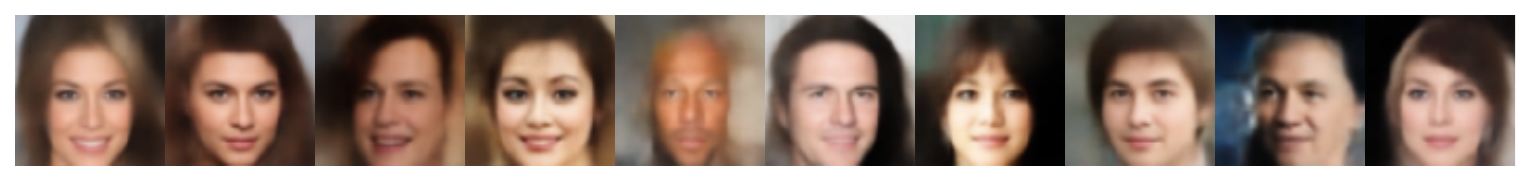}}
    \subfloat{\includegraphics[width=2.3in]{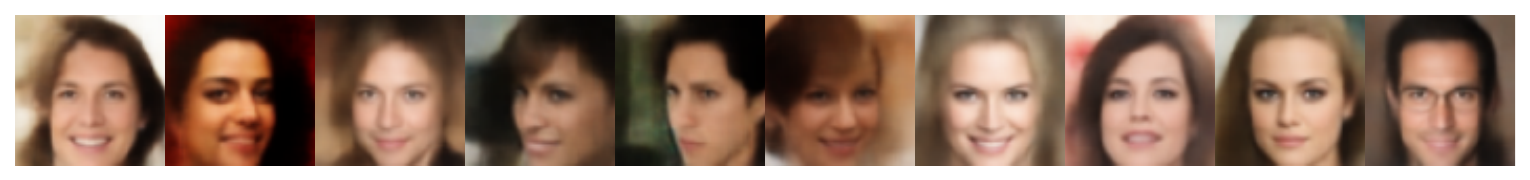}}\\\vspace{-1.3em}
    \adjustbox{minipage=6em,raise=\dimexpr -2.2\height}{\small AAE}
    \subfloat{\includegraphics[width=2.3in]{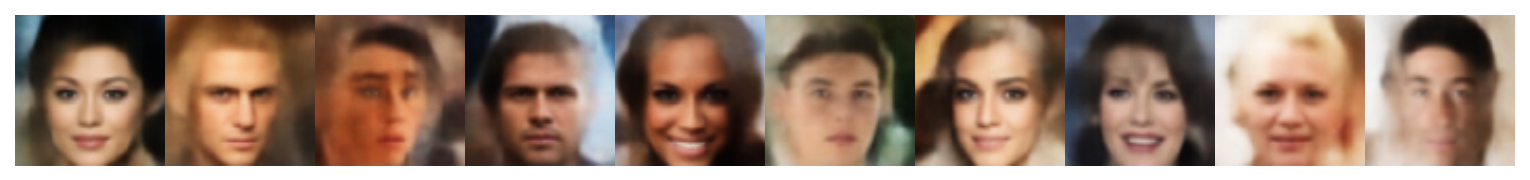}}
    \subfloat{\includegraphics[width=2.3in]{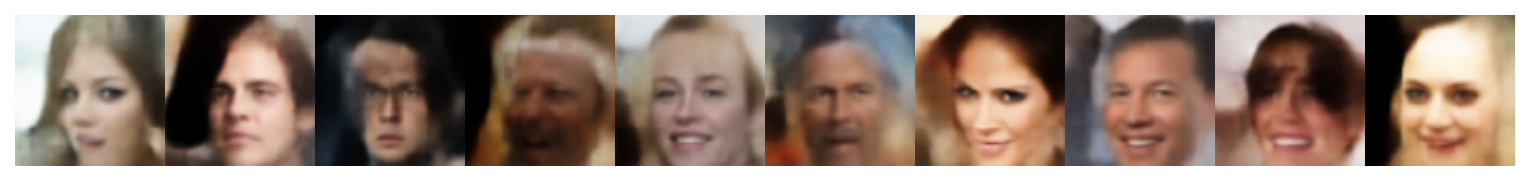}}\\\vspace{-1.3em}
    \adjustbox{minipage=6em,raise=\dimexpr -2.2\height}{\small MSSSIM-VAE}
    \subfloat{\includegraphics[width=2.3in]{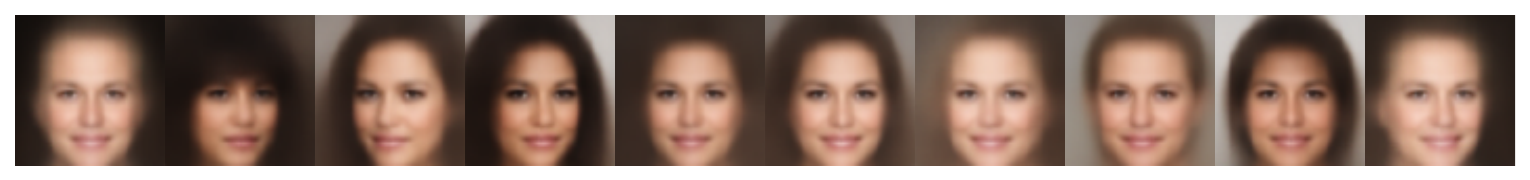}}
    \subfloat{\includegraphics[width=2.3in]{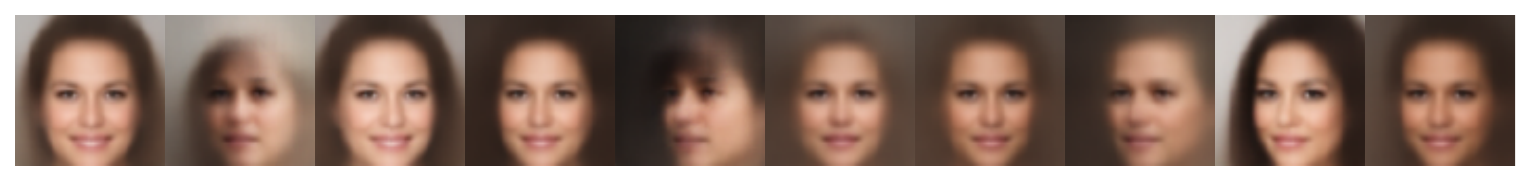}}\\\vspace{-1.3em}
    \adjustbox{minipage=6em,raise=\dimexpr -2.2\height}{\small VAEGAN}
    \subfloat{\includegraphics[width=2.3in]{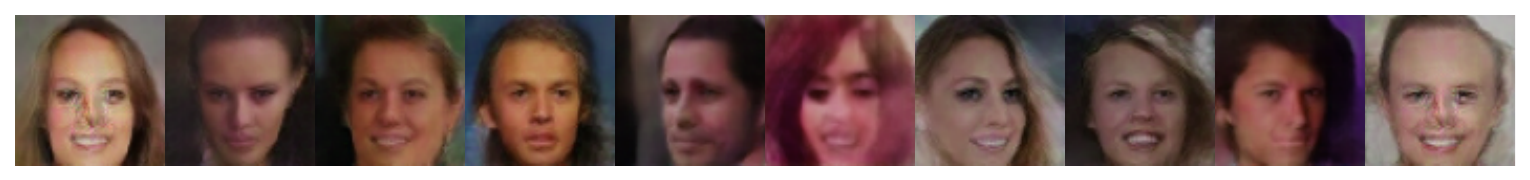}}
    \subfloat{\includegraphics[width=2.3in]{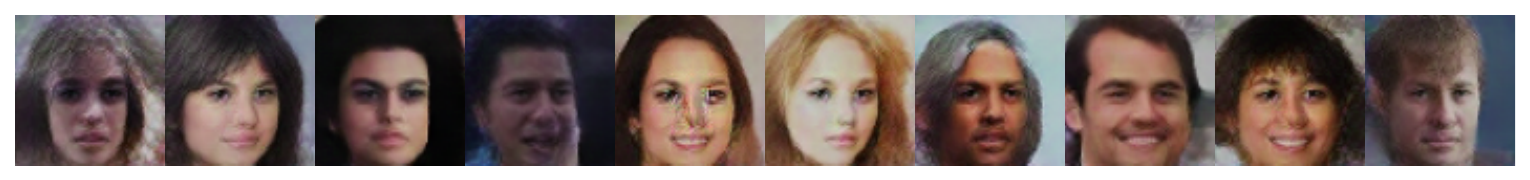}}\\\vspace{-1.3em}
    \adjustbox{minipage=6em,raise=\dimexpr -2.2\height}{\small AE}
    \subfloat{\includegraphics[width=2.3in]{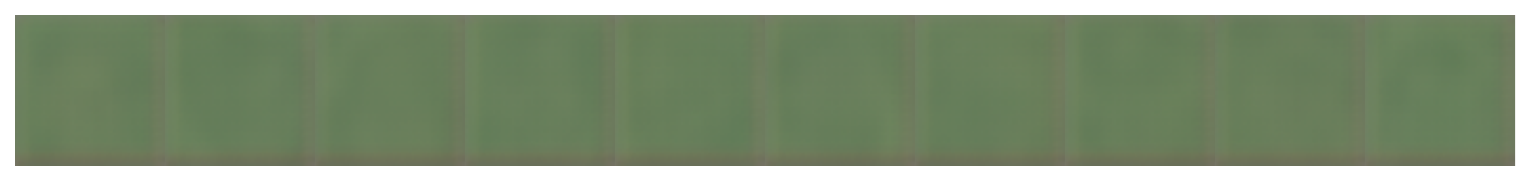}}
    \subfloat{\includegraphics[width=2.3in]{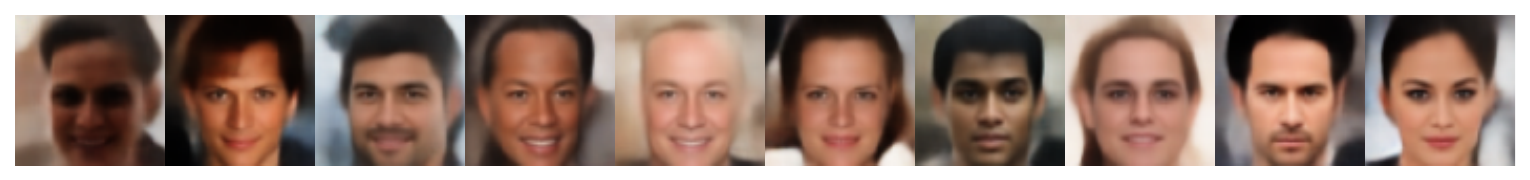}}\\\vspace{-1.3em}
    \adjustbox{minipage=6em,raise=\dimexpr -2.2\height}{\small WAE-IMQ}
    \subfloat{\includegraphics[width=2.3in]{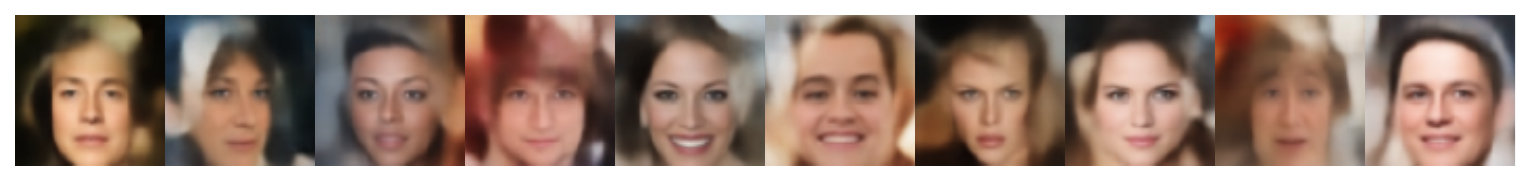}}
    \subfloat{\includegraphics[width=2.3in]{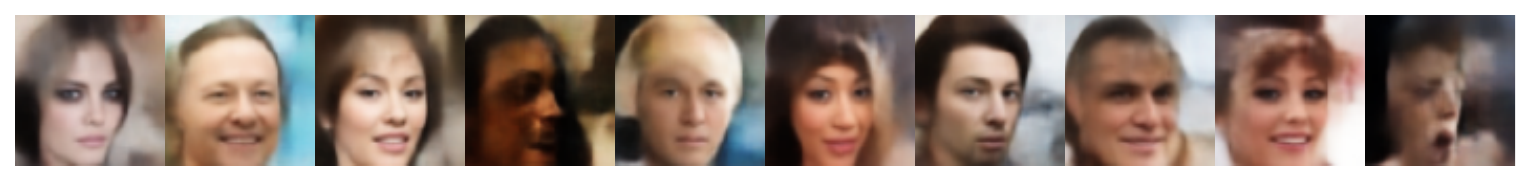}}\\\vspace{-1.3em}
     \adjustbox{minipage=6em,raise=\dimexpr -2.2\height}{\small WAE-RBF}
    \subfloat{\includegraphics[width=2.3in]{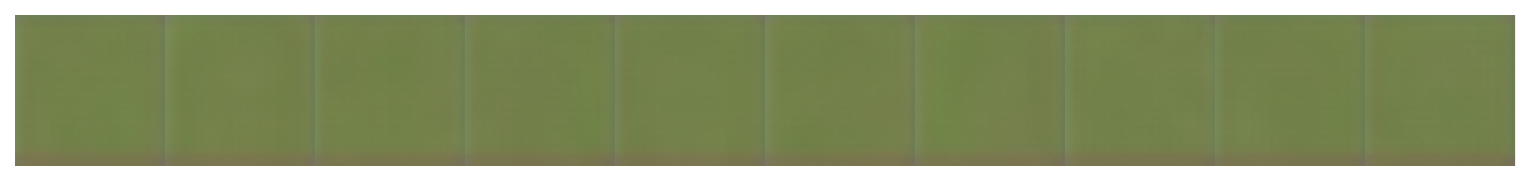}}
    \subfloat{\includegraphics[width=2.3in]{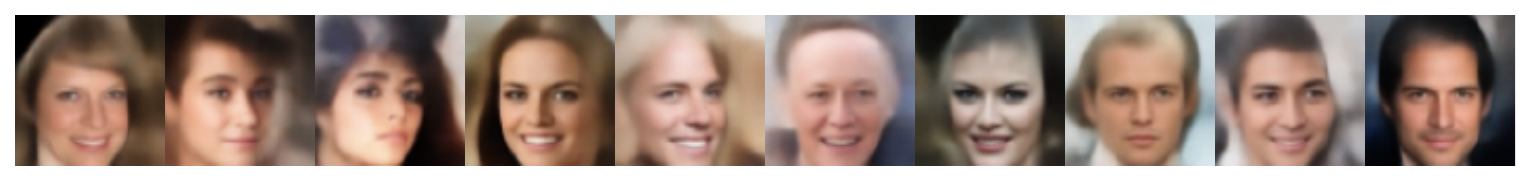}}\\\vspace{-1.3em}
    \adjustbox{minipage=6em,raise=\dimexpr -2.2\height}{\small VQVAE}
    \subfloat{\includegraphics[width=2.3in]{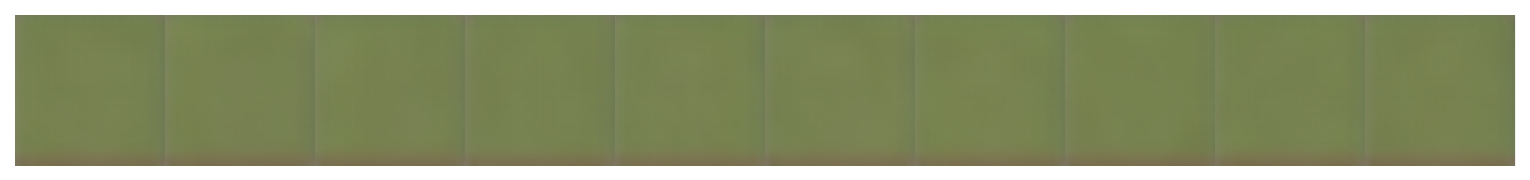}}
    \subfloat{\includegraphics[width=2.3in]{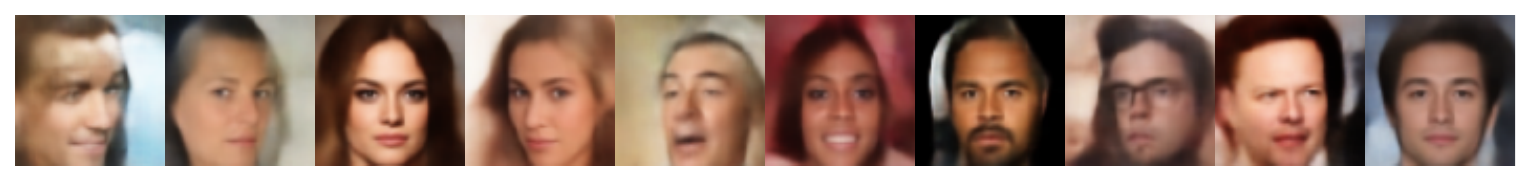}}\\\vspace{-1.3em}
    \adjustbox{minipage=6em,raise=\dimexpr -2.2\height}{\small RAE-L2}
    \subfloat{\includegraphics[width=2.3in]{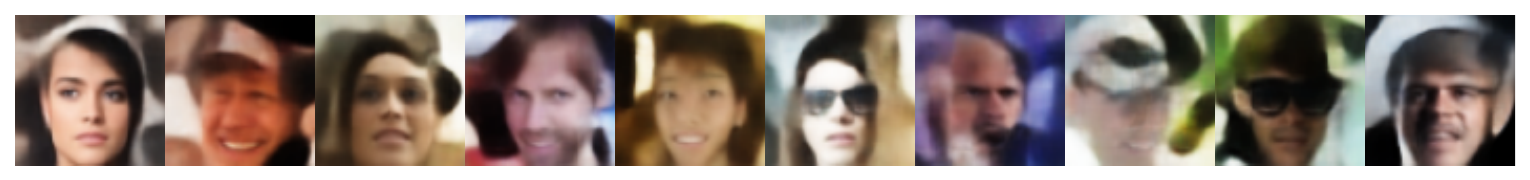}}
    \subfloat{\includegraphics[width=2.3in]{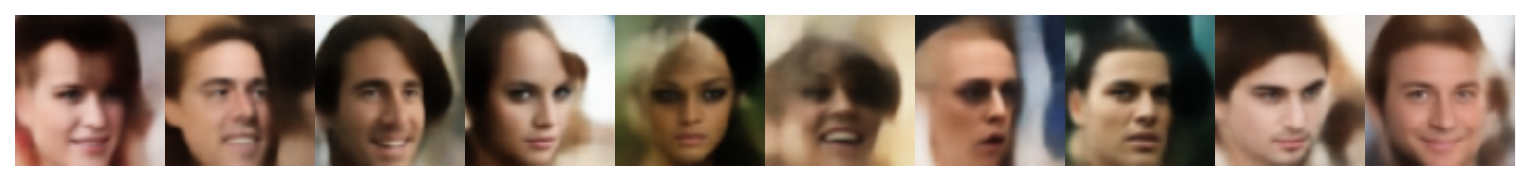}}\\\vspace{-1.3em}
    \adjustbox{minipage=6em,raise=\dimexpr -2.2\height}{\small RAE-GP}
    \subfloat{\includegraphics[width=2.3in]{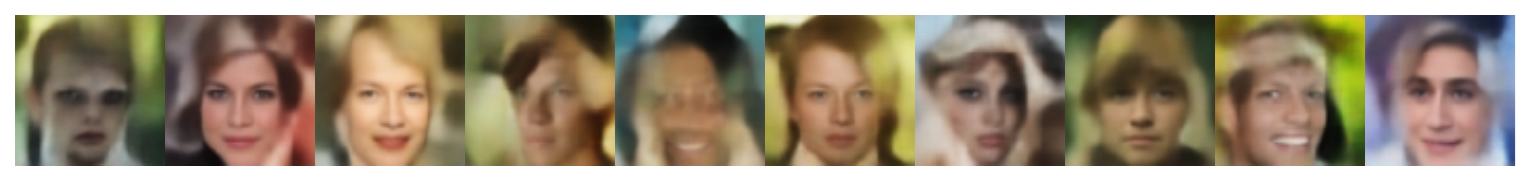}}
    \subfloat{\includegraphics[width=2.3in]{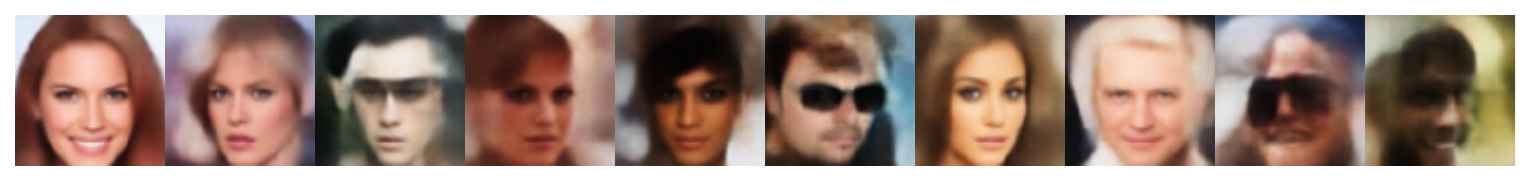}}
    \caption{Generated samples on CELEBA for a latent space of dimension 64 and ConvNet architecture. For each model, we select the configuration achieving the lowest FID on the validation set.}
    \label{fig:generation celeba}
    \end{figure}

In addition to quantitative metrics, we also provide in Fig.~\ref{fig:generation mnist} and Fig.~\ref{fig:generation celeba} some samples coming from the different models using either a $\mathcal{N}(0, I_d)$ or fitting a GMM with 10 components on MNIST and CELEBA. This allows to visually differentiate the quality of the different sampling methods.

\begin{figure}[ht]
    \centering
    \includegraphics[width=\linewidth]{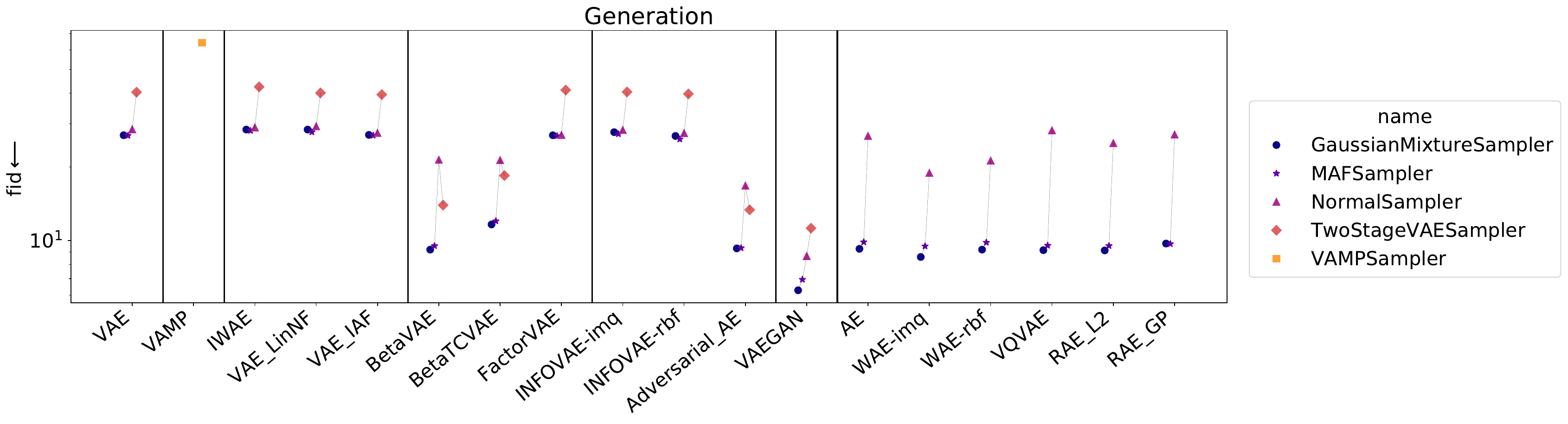}
    \caption{Evolution of the FID for the generation task depending on the sampler, for a ConvNet, the MNIST dataset and a latent dimension of 16. For each sampler and model, we select the configuration achieving the lowest FID on the validation set.}
    \label{fig:sensi_sampler}
    \end{figure}

\clearpage
\paragraph{Sampler ablation study}
Fig.~\ref{fig:sensi_sampler} shows the same results as Table.~\ref{tab:generation_full}
but under a different prism. In this plot, we show the influence each sampler has on the generation quality for all the models considered in this study. Note that sampling using a $\mathcal{N}(0, I_d)$ for an AE, RAE or VQVAE is far from being optimal since those models do not enforce explicitly the latent variables to follow this distribution. As mentioned in the paper, this experiment shows that using more complex density estimators such as a GMM or a normalising flow almost always improves the generation metric. 

\begin{figure}[ht]
    \centering
    \includegraphics[width=\linewidth]{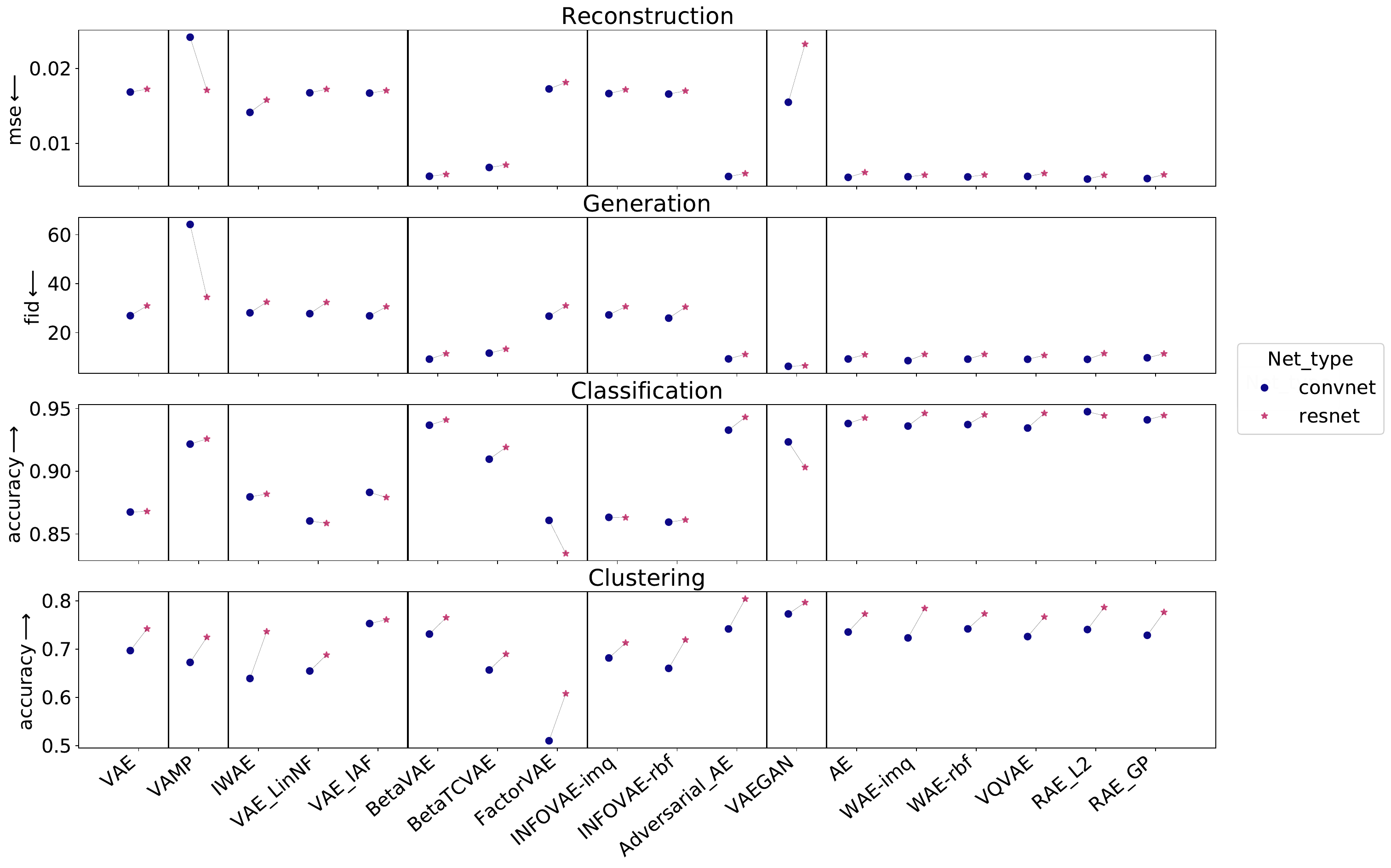}
    \caption{Evolution of the metrics for the 4 tasks depending on the network type on the MNIST dataset and a latent dimension of 16.}
    \label{fig:sensi_nn}
    \end{figure}

\paragraph{Neural network architecture ablation study} 
As explained in the paper and in Appendix.~\ref{appC}, we consider two different neural architectures for the encoder and decoder of each model: a ConvNet (convolutional neural network) and a ResNet (residual neural network). Fig.~\ref{fig:sensi_nn} shows the influence the choice of the neural architecture has on the ability of the model to perform the 4 tasks presented in the paper. The results are computed for each model on MNIST and a latent dimension of 16. The ConvNet architecture has approximately 20 times more parameters than the ResNet in such conditions. We select the best configuration for each model and each task on the validation set and report the results on the test set. 
Unsuprisingly, we see in Fig.~\ref{fig:sensi_nn} that the ConvNet architecture, more adapted to capture features intrinsic to images, leads to the best performances for reconstruction and generation.
Interestingly, the ResNet outperforms the ConvNet for the classification and clustering tasks, meaning that in addition to the network complexity, its structure can play a major role in the representation learned by the models. 
%On other noteworthy aspect is that the considered models appear quite robust to the autoencoding architecture since the ranking remains quite unchanged for each tasks regardless of the neural nets considered. 

\paragraph{Training time}

\begin{figure}[ht]
    \centering
    \includegraphics[width=\linewidth]{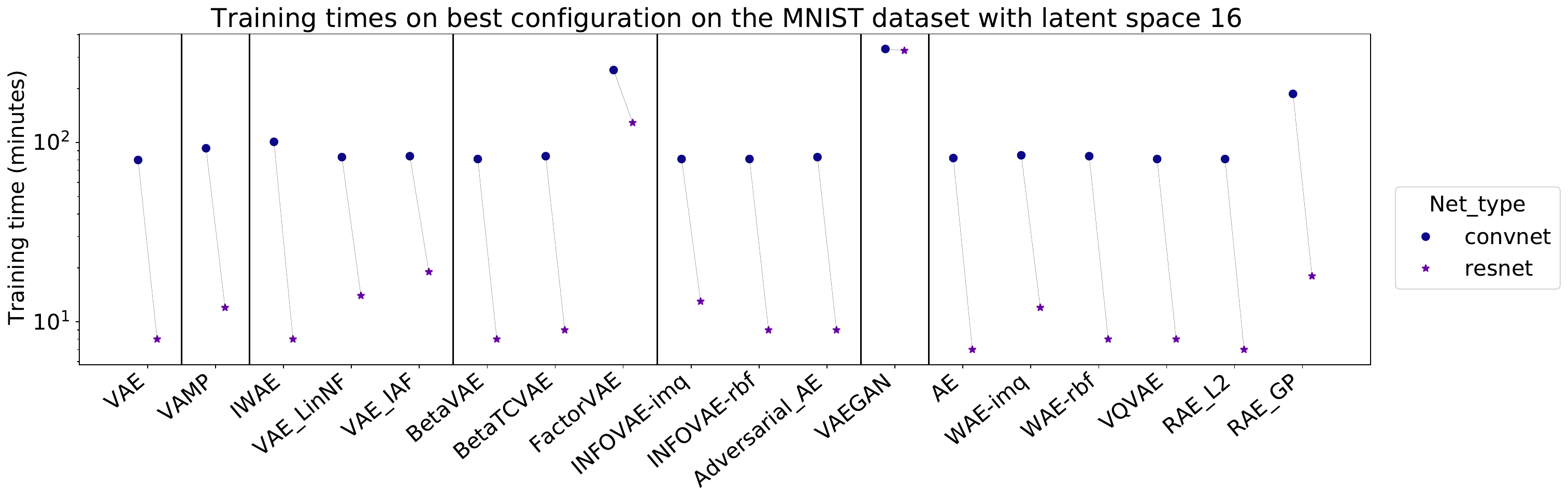}
    \caption{Total training time for the models trained on the MNIST dataset with latent dimension 16 with the best performance on the generation task.}
    \label{fig:training_time}
    \end{figure}

Fig.~\ref{fig:training_time} shows the training times required for each model for both network architectures on the MNIST dataset. For each model, we show the results obtained with the configuration giving the best performances on the generation task with fixed latent dimension 16.
It is interesting to note that although VAEGAN outperforms other models on the generation task, it is at the price of a higher computational time. This is due to the discriminator network (a convolutional neural net) that is called several times during training and takes images as inputs.
It should be noted that methods applying normalising flows to the posterior (VAE-lin-NF and VAE-IAF) maintain a reasonable training time, as the flows were chosen for their scalability.

\clearpage 

\subsection{Configurations and results by models}

    %\subsubsection{Presentation of VAEs}
    
    In this section we briefly explain each model considered in the benchmark, and show the evolution of performances on the 4 tasks and the training speed with respect to the choice of the hyper-parameters. For all 4 tasks we consider the MNIST dataset and a fixed latent space of dimension 16, as well as the Normal Gaussian sampler (if applicable) and the convolutional network architecture.
    For each model, 10 configuration runs with different hyper-parameters were tested. It should be noted that this configuration search was done empirically and is not exhaustive, therefore models with multiple hyper-parameters or that are sensitive to the choice of hyper-parameters will tend to have sub-optimal configuration choices. 
    Although hyper-parameter choices are dependant on both the auto-encoder architecture and the dataset, it is interesting to note the relative evolution of the performances on the different tasks and the training time induced by different hyper-parameter choices.
    
    \paragraph{Notations}
    
    In order to better underline the differences between different models and for clarity purposes, we set the following unified notations:
    
    \begin{itemize}
        
        \item $X = \{x_1, \dots, x_N\} \in \mathcal{X}^N$ the input dataset
        
        \item $x \in \mathcal{X}$ an observation from the dataset, and $z \in \mathcal{Z} = \mathbb{R}^d$ its corresponding latent vector
        
        %\item $E$ and $D$ respectively the encoder and decoder 
        
        \item $\hat{x}$ the reconstruction of $x$ by the auto-encoder model
        
        \item $p_z(z)$ the prior distribution, with $p_z \equiv \mathcal{N}(0,I_d)$ under standard VAE assumption
        
        \item $q_\phi(z|x)$ the approximate posterior distribution, modelled by the encoder. \citet{kingma_auto-encoding_2014} set $$q_\phi(z|x) \equiv \mathcal{N}\big( \mu_\phi(x),\Sigma_\phi(x)) \big)$$ where $\Sigma_\phi(x) = diag[\sigma_\phi(x)]$ and $\big(\mu_\phi(x),\sigma_\phi(x)\big) \in \mathbb{R}^{2 \times d}$ are outputs of the encoder network. The sampling process $z \sim q_\phi(z|x)$ is therefore performed by sampling  $\varepsilon \sim \mathcal{N}(0,I_d)$ and setting $z = \mu_\phi(x) + \Sigma_\phi(x)^{1/2} \cdot \varepsilon$ (re-parametrization trick).
        
        \item $p_\theta(x | z)$ the distribution of $x$ given $z$
        
        \item $p_\theta(x) = \int_\mathcal{Z} p_\theta(x | z) p_z(z) dz$ the marginal distribution of $x$ %derived from the prior distribution
        
        \item $q_\phi(z) = \dfrac{1}{N} \sum_{i=1}^N q_\phi(z | x_i) $ the aggregated posterior integrated over the training set.
        
        \item $\mathcal{D}_{KL}$ the Kullback-Leibler divergence
        
        %\item $q_\phi(z) = \int_\mathcal{X} q_\phi(z | x) p_theta(x) dx$ the marginal distribution of $x$ derived from the prior distribution
        
    \end{itemize}
    
    We further recall that \citet{kingma_auto-encoding_2014} use the unbiased estimate $\widehat{p}_{\theta}(x)$ of ${p}_{\theta}(x)$ defined as
    \begin{equation*}
        \widehat{p}_{\theta}(x) = \frac{p_{\theta}(x|z)p_z(z)}{q_{\phi}(z|x)}
    \end{equation*}
    to derive the standard Evidence Lower Bound (ELBO) of the log probability $\log p_\theta(x)$ which we wish to maximize:
    \begin{equation*}
        \mathcal{L}_{ELBO} = \mathbb{E}_{x \sim p_\theta(x)} \big[ \mathcal{L}_{ELBO}(x) \big]
    \end{equation*}
    with
    \begin{equation*}
        \mathcal{L}_{ELBO}(x) = \underbrace{\mathbb{E}_{z \sim q_\phi}[\log p_\theta(x | z)]}_\text{reconstruction} - \underbrace{\mathcal{D}_{KL}\big[ q_\phi(z|x) || p_z(z) \big]}_\text{regularisation}
    \end{equation*}
    
    The \emph{reconstruction} loss is maximized when $\hat{x}$ is close to $x$, thus encouraging a good reconstruction of the input $x$, while the \emph{regularisation} term is maximized when $q_\phi(z|x)$ is close to $p_z(z)$, encouraging the posterior distribution to follow the chosen prior distribution.
    
    The integration over $p_\theta(x)$ is approximated by the empirical distribution of the training dataset, and the negative ELBO function acts as a loss function to minimize for the encoder and decoder networks.

    %\subsubsection{Improving the prior }
    
    %It has been shown that the role of the prior distribution is crucial in the good performance of the VAE. Choosing a family of overly simplistic priors such as the standard normal distribution proposed in the original VAE implementation can lead to over-regularisation, and poor reconstruction performance. Several extensions have been proposed to overcome this issue. 
    
    \clearpage
    \paragraph{VAE with a VampPrior (VAMP)}
        Starting from the observation that a standard Gaussian prior may be too simplistic, \citet{tomczak_vae_2018} proposes a less restrictive prior: the Variational Mixture of Posteriors (VAMP). A VAE with a VAMP prior  aims at relaxing the posterior constraint by replacing the conventional normal prior with a multimodal aggregated posterior given by:
        \[
        p_z(z) = \frac{1}{K} \sum_{k=1}^K q_\phi(z | u_k)\,,
        \]
        where $u_k$ are pseudo-inputs living in the data space $\mathcal{X}$ learned through back-propagation and acting as anchor points for the prior distribution. For the VAMP VAE implementation, we use the same architecture as the authors' implementation for the network generating the pseudo-inputs: a MLP with a single layer and Tanh activation.

    \textbf{Results by configuration}

\begin{table}[ht]
    \centering
    \scriptsize
    \caption{VAMP configurations}
    \label{tab:vamp config}
    \begin{tabular}{lcccccccccc}
    \toprule
        Config & 1 & 2 & 3 & 4 & 5 & 6 & 7 & 8 & 9 & 10 \\ 
        \midrule
        Number of pseudo-inputs (K) & 10 & 20 &30 &500 &100 &150 &200 &250 &300 &500\\
    \bottomrule
    \end{tabular}
\end{table}

    \begin{figure}[ht]
    \centering
    \includegraphics[trim={0 4cm 0 0},clip,width=\linewidth]{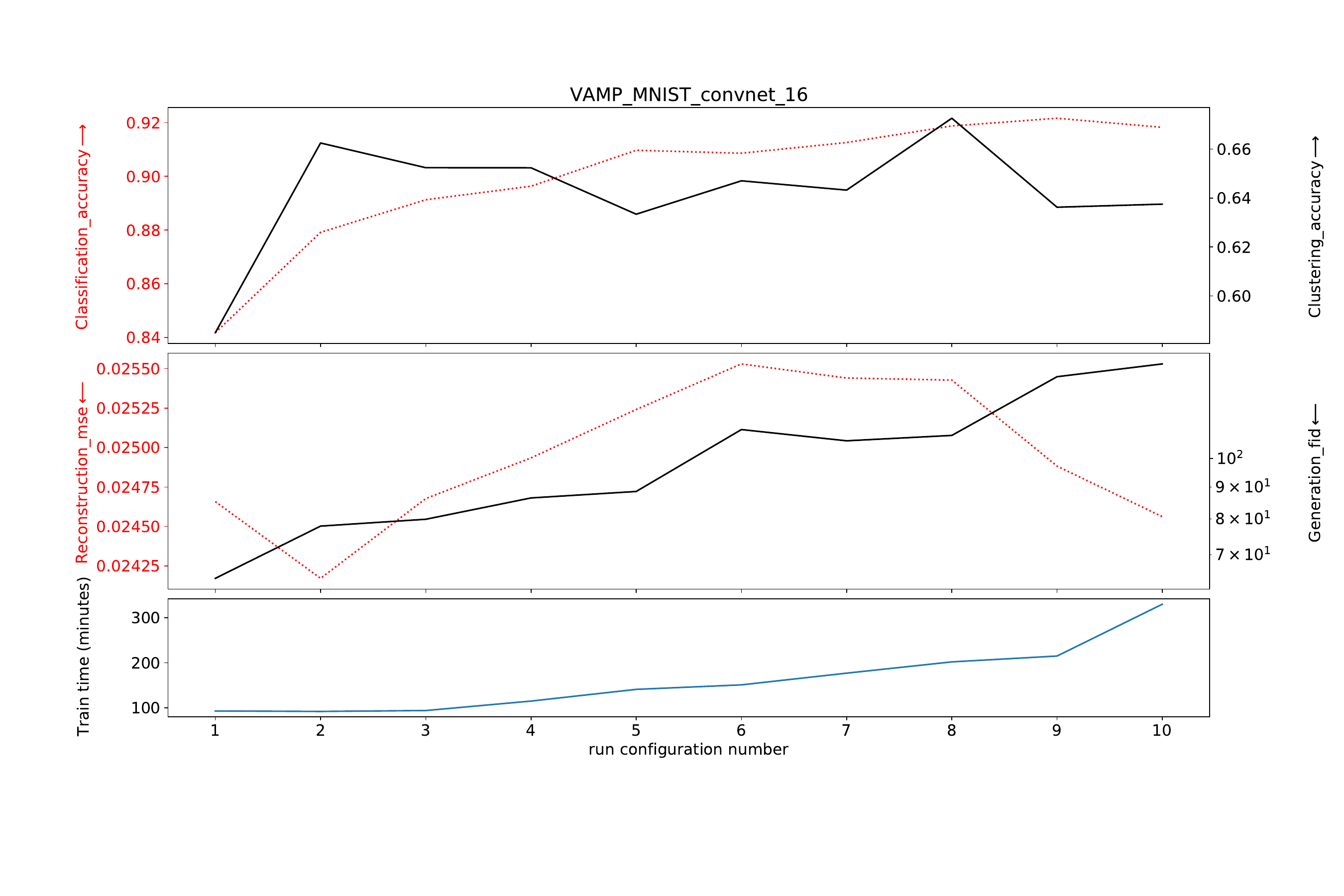}
    \caption{Results on VAMP}
    \label{fig:indiv_vamp}
    \end{figure}

%Some work has also been done on adapting the VAE scheme to a deterministic Auto Encoder
    \clearpage
    
    \paragraph{Importance Weighted Autoencoder (IWAE)}
    
    \citet{burda_importance_2016} introduce an alternative lower bound to maximize, derived from importance weighting where the new unbiased estimate $\widehat{p}_{\theta}(x)$ of the marginal distribution $p_{\theta}(x)$ is computed with $L$ samples $z_1, \dots, z_L \sim q_{\phi}(z|x)$ :
    \begin{equation*}
        \widehat{p}_{\theta}(x) = \dfrac{1}{L}\sum_{i=1}^L \frac{p_{\theta}(x|z_i)p_z(z_i)}{q_{\phi}(z_i|x)}\,.
    \end{equation*}
    This estimate induces a new lower bound of the true marginal distribution $p_{\theta}(x)$ using Jensen's inequality:
\begin{equation*}
    \mathcal{L}_\text{IWAE}(x) := \mathbb{E}_{z_1, \dots, z_L \sim q(z|x)}\bigg[\log\frac{1}{L}\sum \limits_{i=1}^L \frac{p_{\theta}(x|z_i)p_z(z_i)}{q_{\phi}(z_i|x)} \bigg] \leq \log \underbrace{\mathbb{E}_{z_1, \dots, z_L \sim q(z|x)}\big[\widehat{p}_{\theta}(x) \big]}_{p_{\theta}(x)}.
\end{equation*}
    As the number of samples $L$ increases, $\mathcal{L}_\text{IWAE}(x)$ becomes closer to $\log p_\theta(x)$, therefore providing a tighter bound on the true objective. Note that when $L=1$ we recover the original VAE framework. 
    
    As expected the reconstruction quality increases with the number of samples. Nonetheless, we note that increasing the number of samples has a significant impact on the computation of a single training step, therefore leading to a much slower training process.

    \textbf{Results by configuration}
    
    \begin{table}[ht]
    \centering
    \scriptsize
    \caption{IWAE configurations}
    \label{tab:iwae config}
    \begin{tabular}{lcccccccccc}
    \toprule
        Config & 1 & 2 & 3 & 4 & 5 & 6 & 7 & 8 & 9 & 10 \\ 
        \midrule
        Number of samples (L) & 2 & 3 & 4 &5 &6 &7 &8 &9 &10 &12\\
    \bottomrule
    \end{tabular}
\end{table}
    
    \begin{figure}[ht]
    \centering
    \includegraphics[trim={0 4cm 0 0},clip,width=\linewidth]{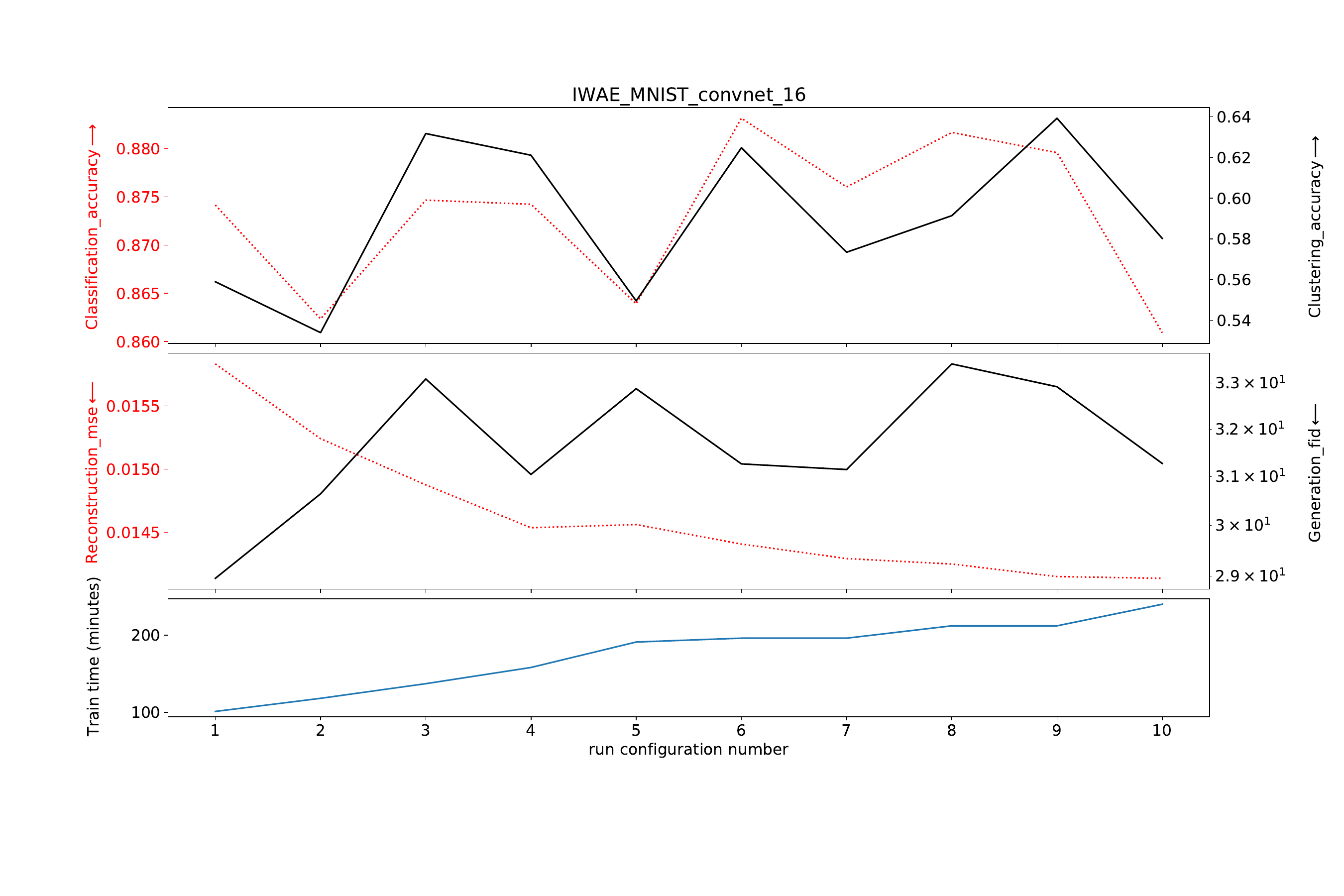}
    \caption{Results on IWAE}
    \label{fig:indiv_iwae}
    \end{figure}

    % \begin{table}[ht]
    % \CatchFileDef{\mytable}{plots/configs/configs_IWAE.tex}{}
    %     \centering
    %     \begin{tabular}{r|cccccccccc} 
    %         \mytable
    %     \end{tabular}
    %     %\caption{Caption}
    %     \label{tab:my_label}
    % \end{table}

    \clearpage
    
    \paragraph{Variational Inference with Normalizing Flows (VAE-lin-NF)}
    
    In order to model a more complex family of approximate posterior distributions, \citet{rezende_variational_2015} propose to use a succession of normalising flows to transform the simple distribution $q_{\phi}(z | x)$, allowing it to model more complex behaviours. In practice, after having sampled $z_0 \sim q_\phi(z | x)$, $z_0$ is passed through a chain of $K$ invertible smooth mappings from the latent space $\mathbb{R}^d$ to itself:
        \[
        z_K = f_K \circ \dots \circ f_2 \circ f_1(z_0)\,.
        \]
        The modified latent vector $z_K$ is then used as input $z$ for the decoder network. In their paper, the authors propose to use two types of transformations: planar and radial flows.  
    \[
       f_{\mathrm{planar}}(z) = z + uh(w^{\top} z + b) \hspace{5mm};\hspace{5mm} f_{\mathrm{radial}}(z) = z + \beta g(\alpha, r) (z-z_0)\,,
    \]
    where $h$ is a smooth non-linearity with tractable derivatives and $g(\alpha, r)=\frac{1}{\alpha + r}$. The parameters are such that $r= \|z - z_0 \|$, $u, w, z_0 \in \mathbb{R}^d$, $\alpha \in \mathbb{R}^+$ and $b, \beta \in \mathbb{R}$. We can easily compute the resulting density $q$ given by 
        \[
            \log q(z_K) = \log q_{\phi}(z_0|x) - \sum \limits _{k=1}^K \log \bigg | \det \frac{\partial f_k}{\partial z}\bigg |\,.
        \]
    This makes the ELBO tractable and optimisation possible. 
    
    %For this model, we decide to change the complexity of the flow using different length and combinations as shown in Table.~\ref{tab:vae-lin config}. Results acccording to the configurations considered are displyed in Fig.~\ref{fig:indiv_vae_lin_nf}.  
    
    \textbf{Results by configuration}
    
    \begin{table}[ht]
    \centering
    \scriptsize
    \caption{VAE-lin-NF configurations}
    \label{tab:vae-lin config}
    \begin{tabular}{lcccccccccc}
    \toprule
        Config & 1 & 2 & 3 & 4 & 5 & 6 & 7 & 8 & 9 & 10 \\ 
        \midrule
        Flow sequence & PPPPP & RRRRR & PRPRP & 10P & 15P &20P &30P &PRPRPRPRPR & PPP & PRPPRPPPPR \\
    \bottomrule
    \multicolumn{6}{l}{`P' stands for planar flow - `R' stands for radial flow}\\
    \end{tabular}
\end{table}
        
    \begin{figure}[ht]
    \centering
    \includegraphics[trim={0 4cm 0 0},clip,width=\linewidth]{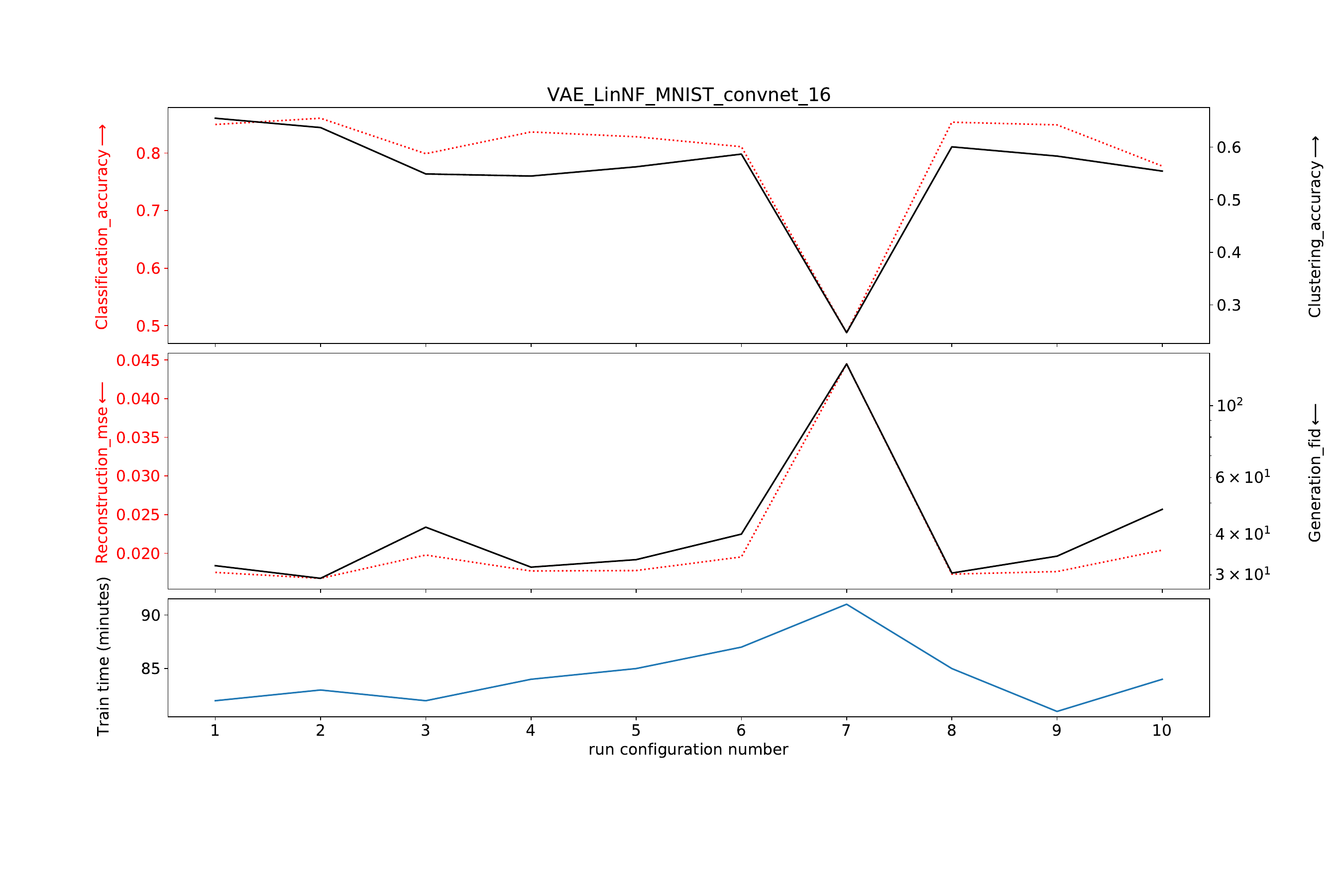}
    \caption{Results on VAE-lin-NF}
    \label{fig:indiv_vae_lin_nf}
    \end{figure}

% \begin{table}[ht]
% \CatchFileDef{\mytable}{plots/configs/configs_VAE_LinNF.tex}{}
%     \centering
%     \begin{tabular}{|||c||c|} 
%         \mytable
%     \end{tabular}
%     %\caption{Caption}
%     \label{tab:my_label}
% \end{table}
    
    \clearpage
    \paragraph{Variational Inference with Inverse Autoregressive Flow (VAE-IAF)}
    
    \citet{kingma_improved_2016} improve upon the works of \cite{rezende_variational_2015} with a new type of normalising flow that better scales to high-dimensional latent spaces. The main idea is again to apply several transformations to a sample from a simple distribution in order to model richer distributions. Starting from $z_0 \sim q_\phi(z | x)$, the proposed IAF flow consists in applying consecutively the following transformation
    \[
        z_k = \mu_k + \sigma_k \cdot z_{k-1}\,,
    \]
    where $\mu_k$ and $\sigma_k$ are the outputs of an autoregressive neural network taking $z_{k-1}$ as input. Inspired from the original paper, to implement one Inverse Autoregressive Flow we use MADE \citep{germain2015made} and stack multiple IAF together to create a richer flow. 
    The MADE mask is made sequentially for the masked autoencoders and the ordering is reversed after each MADE. 
    
    %The main parameter we wish to test in the benchmark is the influence of the complexity of the flow. Hence, the size and number of hidden units in the MADEs along with the number of IAF stacked blocks are part of the search.
    
    \textbf{Results by configuration}
    
    \begin{table}[ht]
    \centering
    \scriptsize
    \caption{VAE-IAF configurations}
    \label{tab:vae-iaf config}
    \begin{tabular}{lcccccccccc}
    \toprule
        Config & 1 & 2 & 3 & 4 & 5 & 6 & 7 & 8 & 9 & 10 \\ 
        \midrule
        hidden size in MADE & 32.0 & 32.0 & 32.0 & 32.0 & 32.0 & 32.0 & 32.0 & 32.0 & 64.0 & 128.0 \\ 
 number hidden units in MADE & 2.0 & 2.0 & 2.0 & 2.0 & 2.0 & 2.0 & 4.0 & 6.0 & 2.0 & 2.0 \\
number of IAF blocks & 1.0 & 2.0 & 5.0 & 10.0 & 20.0 & 4.0 & 4.0 & 4.0 & 4.0 & 4.0\\
    \bottomrule

    \end{tabular}
\end{table}

    \begin{figure}[ht]
    \centering
    \includegraphics[trim={0 4cm 0 0},clip,width=\linewidth]{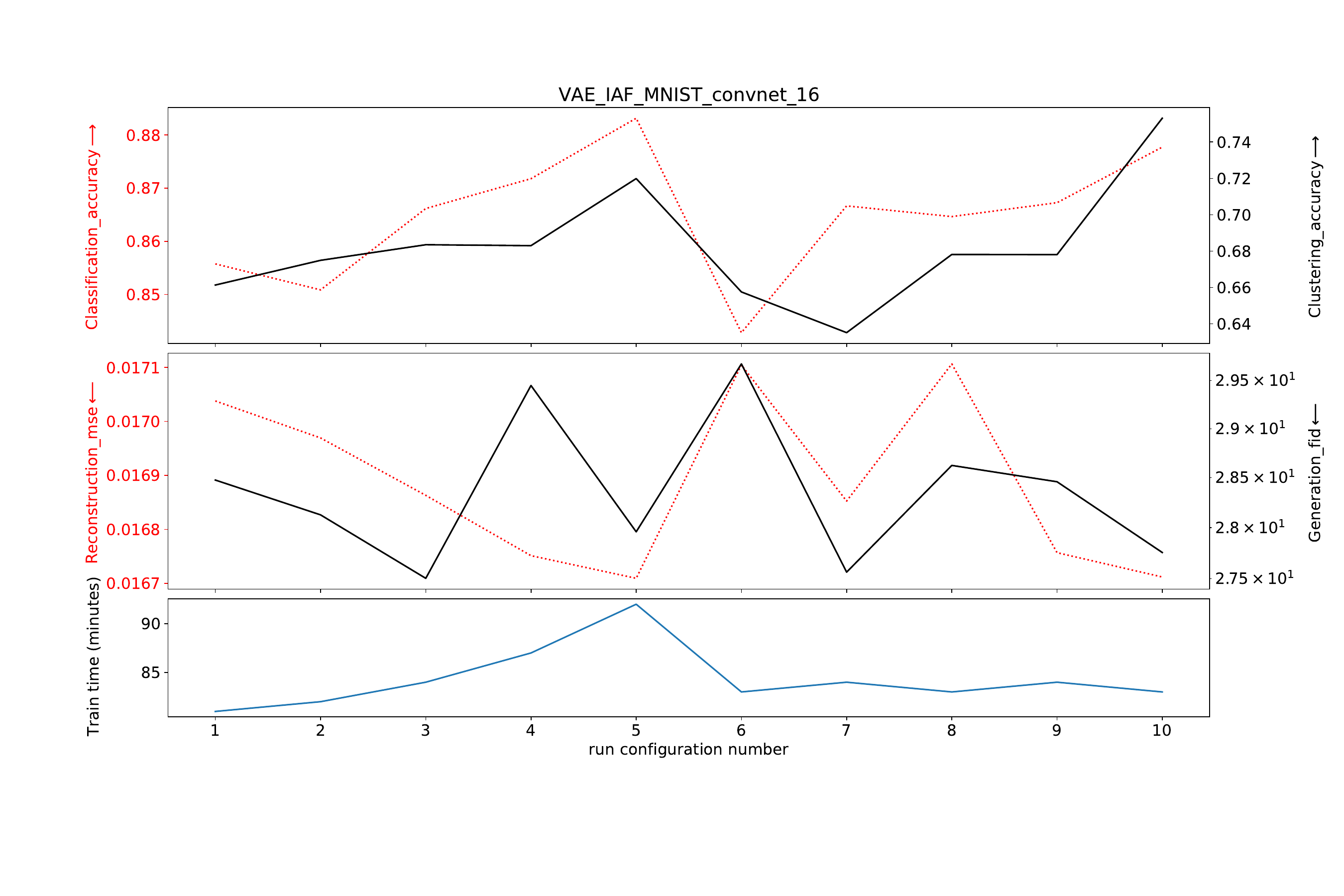}
    \caption{Results on VAE-IAF}
    \label{fig:indiv_vae_iaf}
    \end{figure}

% \begin{table}[ht]
% \CatchFileDef{\mytable}{plots/configs/configs_VAE_IAF.tex}{}
%     \centering
%     \begin{tabular}{|||c||c|} 
%         \mytable
%     \end{tabular}
%     %\caption{Caption}
%     \label{tab:my_label}
% \end{table}

    \clearpage
    
    \paragraph{$\beta$-VAE}
    
    \citet{higgins_beta-vae_2017} argue that increasing the weight of the KL divergence term in the ELBO loss enforces a stronger disentanglement of the latent features as the posterior probability is forced to match a multivariate standard Gaussian. They propose to add a hyper-parameter $\beta$ in the ELBO leading to the following objective to maximise:
        \[ 
        \mathcal{L}_{\beta\text{-VAE}}(x) = \mathbb{E}_{z \sim q_\phi}[\log p_\theta(x | z)] - \beta \mathcal{D}_{KL}\big[ q_\phi(z|x) || p_z(z) \big]\,.
        \]
    Although the original publication specifies $\beta > 1$ to encourage a better disentanglement, a smaller value of $\beta$ can be used to relax the regularisation constraint of the VAE. Therefore, for this model we consider a range of values for $\beta$ from $1e^{-3}$ to $1e^{3}$. 
    %See all the values we considered in the benchmark in Table.~\ref{tab:betavae config} and the obtained results in Fig.~\ref{fig:indiv_betavae}. 
    
\begin{figure}[ht]
    \centering
    \includegraphics[trim={0 4cm 0 0},clip,width=\linewidth]{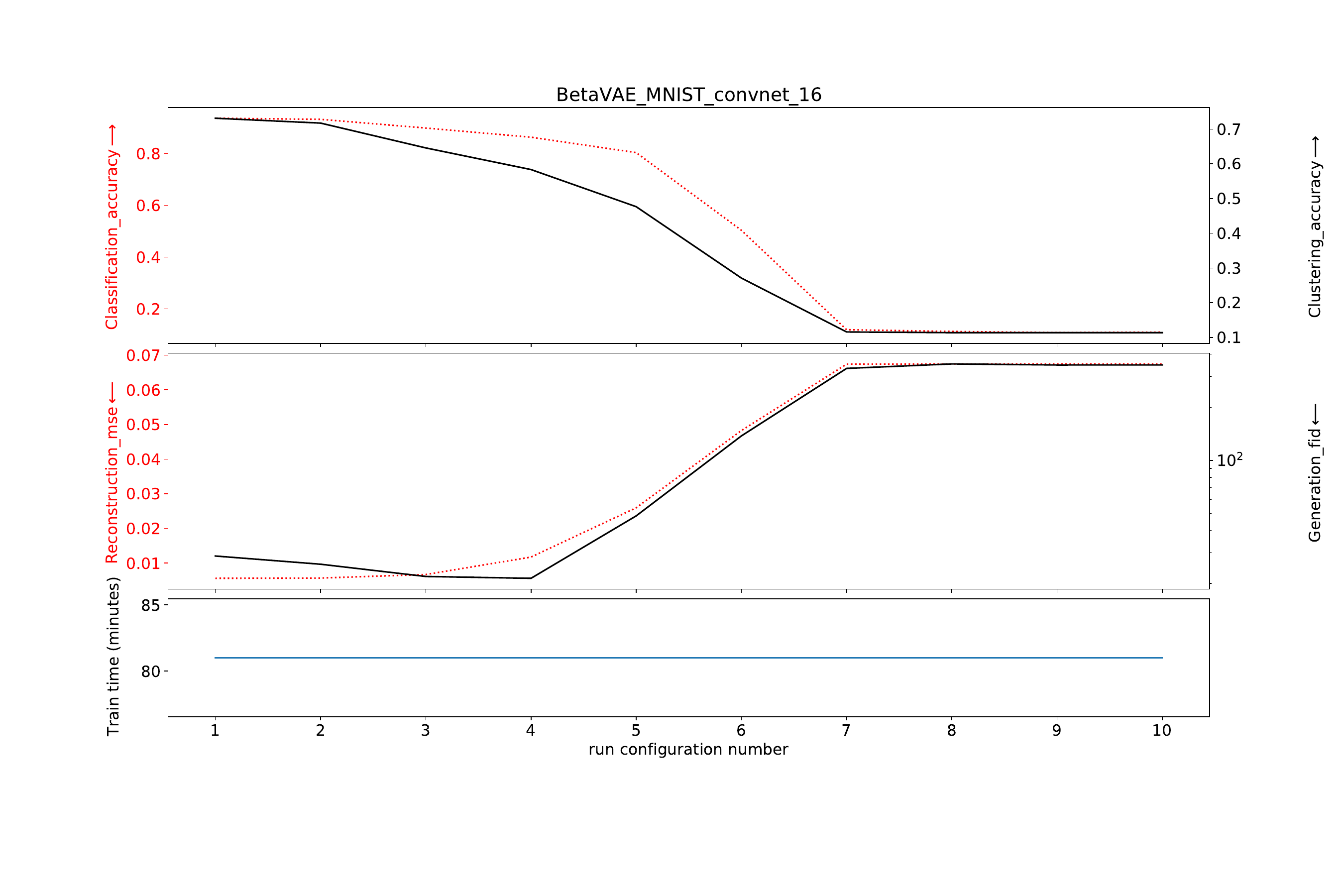}
    \caption{Results on $\beta$-VAE}
    \label{fig:indiv_betavae}
    \end{figure}

    As expected, we see in Fig.~\ref{fig:indiv_betavae} a trade-off appearing between reconstruction and generation. Indeed, a very small $\beta$ will tend to less regularise the model since the latent variables will no longer be driven to follow the prior, favouring a better reconstruction. On the other hand, a higher value for $\beta$ will constrain the model, leading to a better generation quality. Moreover, as can be seen in Fig.~\ref{fig:indiv_betavae}, too high a value of $\beta$ will lead to over-regularisation, resulting in poor performances on all evaluated tasks.
    
    \textbf{Results by configuration}
    
    \begin{table}[ht]
    \centering
    \scriptsize
    \caption{$\beta$-VAE configurations}
    \label{tab:betavae config}
    \begin{tabular}{lcccccccccc}
    \toprule
        Config & 1 & 2 & 3 & 4 & 5 & 6 & 7 & 8 & 9 & 10 \\ 
        \midrule
        $\beta$ & $1e^{-3}$ & $1e^{-2}$ & $1e^{-1}$ & $0.5$ & 2 & 5 & 10 & 20 & $1e^{2}$ & $1e^{3}$ \\ 
    \bottomrule
    \end{tabular}
\end{table}
    
    \clearpage
    
    \paragraph{$\beta$-TC-VAE}
    
    \citet{chen2019vae} extend on the ideas of \citet{higgins_beta-vae_2017} by rewriting and re-weighting specific terms in the ELBO loss with multiple hyperparameters. The authors note that the KL-divergence term of the ELBO loss can be rewritten as
        \[
        \begin{aligned}
            \mathbb{E}_{x \sim p_\theta}\bigg[
            \mathcal{D}_{KL}\big[ q_\phi(z | x) || p_z(z) \big] 
            \bigg] = \underbrace{I(x,z)}_\text{Mutual information} &+ \underbrace{\mathcal{D}_{KL}\big[ q_\phi(z) || \prod_{j=1}^d q_\phi(z_j) \big]}_\text{TC-loss}\\ &+ \underbrace{\sum_{j=1}^d \mathcal{D}_{KL}\big[ q_\phi(z_j) || p_z(z_j) \big]}_\text{Dimension-wise KL}
        \end{aligned}
        \]
        \begin{itemize}
            \item The mutual information term corresponds to the amount of information shared by $x$ and its latent representation $z$. It is claimed that maximising the mutual information encourages better disentanglement and a more compact representation of the data.
            
            \item The TC-loss corresponds to the total correlation between the latent distribution and its fully disentangled version, maximising it enforces the dimensions of the latent vector to be uncorrelated.
            
            \item Maximising the dimension-wise KL prevents the marginal distribution of each latent dimension from diverging too far from the prior Gaussian distribution
        \end{itemize}
        
\begin{figure}[ht]
    \centering
    \includegraphics[trim={0 4cm 0 0},clip,width=\linewidth]{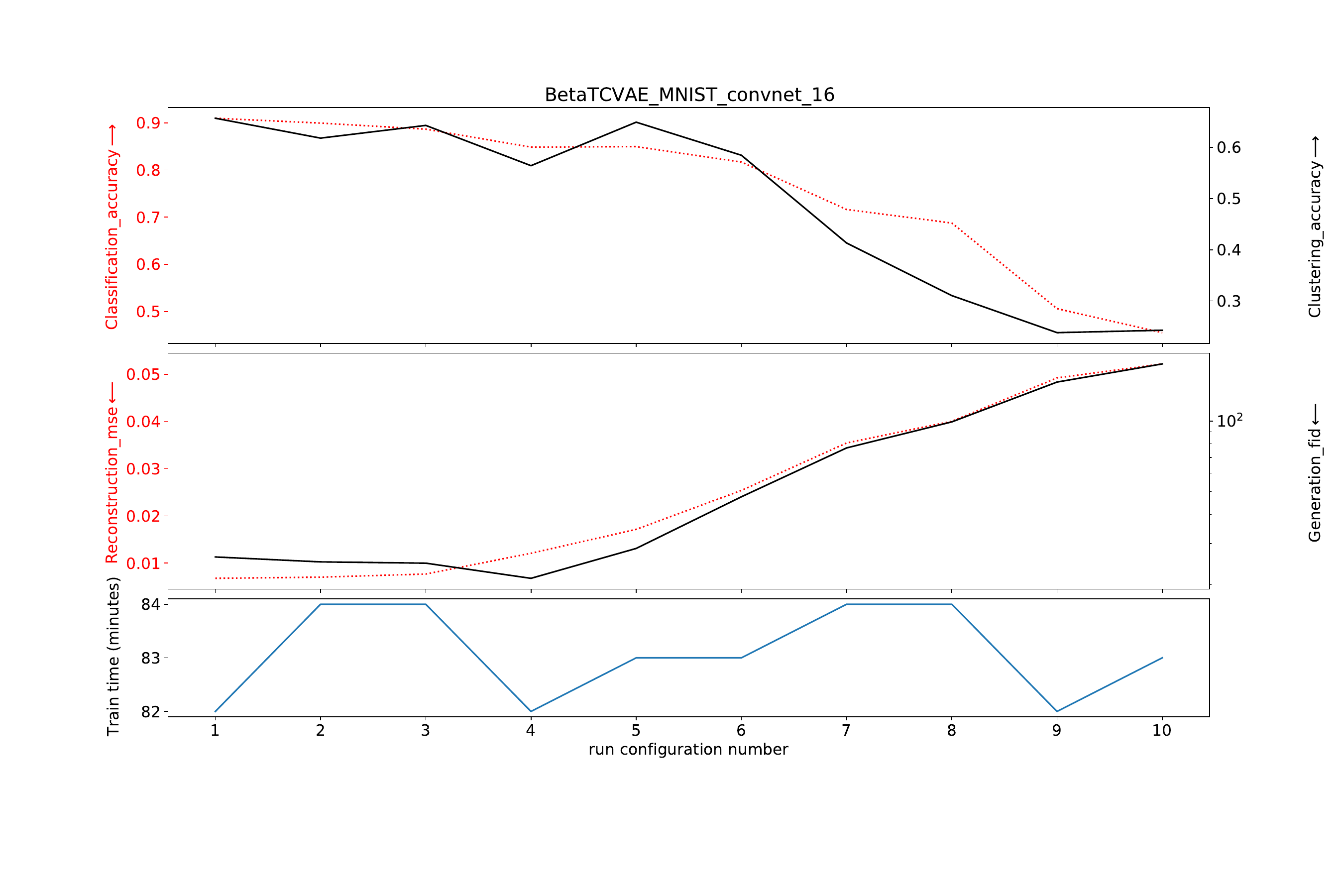}
    \caption{Results on $\beta$-TC-VAE}
    \label{fig:indiv_betatcvae}
    \end{figure}

        The authors therefore propose to replace the classical regularisation term with the more general term
    \[
    \mathcal{L}_\text{reg} := \alpha I(x,n) + \beta \mathcal{D}_{KL}\big[ q_\phi(z) || \prod_ j q_\phi(z_j) \big] + \gamma \sum_j \mathcal{D}_{KL}\big[ q_\phi(z_j) || p_z(z_j) \big]\,.
    \]
        Similarly to the authors, we set $\alpha=\gamma=1$ and only perform a search on the parameter $\beta$. 
        %See Table.~\ref{tab:betatcvae config} for details regarding the considered values.
        Fig.~\ref{fig:indiv_betatcvae} shows a reconstruction-generation trade-off similar to the $\beta$-VAE model
        
    \textbf{Results by configuration}
    
    \begin{table}[ht]
    \centering
    \scriptsize
    \caption{$\beta$-TC-VAE configurations}
    \label{tab:betatcvae config}
    \begin{tabular}{lcccccccccc}
    \toprule
        Config & 1 & 2 & 3 & 4 & 5 & 6 & 7 & 8 & 9 & 10 \\ 
        \midrule
        $\beta$ & $1e^{-3}$ & $1e^{-2}$ & $1e^{-1}$ & $0.5$ & 1 & 2 & 5 & 10 & 50 & $1e^{2}$ \\ 
    \bottomrule
    \end{tabular}
\end{table}
    
    \clearpage
    
    \paragraph{Factor VAE}
    
    \citet{kim2018factorvae} augment the VAE objective with a penalty that encourages factorial representation of the marginal distributions, enforcing a stronger disentangling of the latent space. Noting that a high $\beta$ value in $\beta$-VAE ELBO loss encourages disentanglement at the expense of reconstruction quality, FactorVAE proposes a new lower bound of the log likelihood with an added disentanglement term:
\begin{equation*} \mathcal{L}_\text{FactorVAE}(x) := \mathcal{L}_{ELBO}(x) - \gamma \mathcal{D}_{KL}\big( q_\phi(z) || \bar{q}_\phi(z) \big) \ \text{, with } \bar{q}_\phi(z) := \prod_{j=1}^d q_\phi(z_j)
\end{equation*}
The distribution of representations $q_\phi(z) = \frac{1}{N}\sum_{i=1}^N q_\phi(z | x_i)$ of the entire dataset is therefore forced to be close to its fully-disentangled equivalent $\bar{q}_\phi(z)$ while leaving the ELBO loss as it is. They further propose to approximate the KL divergence with a discriminator network $D$ that is trained jointly to the VAE:
\begin{equation*}
    \mathcal{D}_{KL}\big( q(z) || \bar{q}(z) \big) \approx \mathbb{E}_{q_z(z)}\bigg[ \log \frac {D(z)}{1 - D(z)}\bigg]
\end{equation*}
As suggested in the authors's paper, the discriminator is set as a MLP composed of 6 layers each with 1000 hidden units and LeakyReLU activation. 
            
    \textbf{Results by configuration}
    
    \begin{table}[ht]
    \centering
    \scriptsize
    \caption{FactorVAE configurations}
    \label{tab:factorvae config}
    \begin{tabular}{lcccccccccc}
    \toprule
        Config & 1 & 2 & 3 & 4 & 5 & 6 & 7 & 8 & 9 & 10 \\ 
        \midrule
        $\gamma$ & 1 & 2 & 5 & 10 & 15 & 20 & 30 & 40 & 50 & 100 \\ 
    \bottomrule
    \end{tabular}
\end{table}

    \begin{figure}[ht]
    \centering
    \includegraphics[trim={0 4cm 0 0},clip,width=\linewidth]{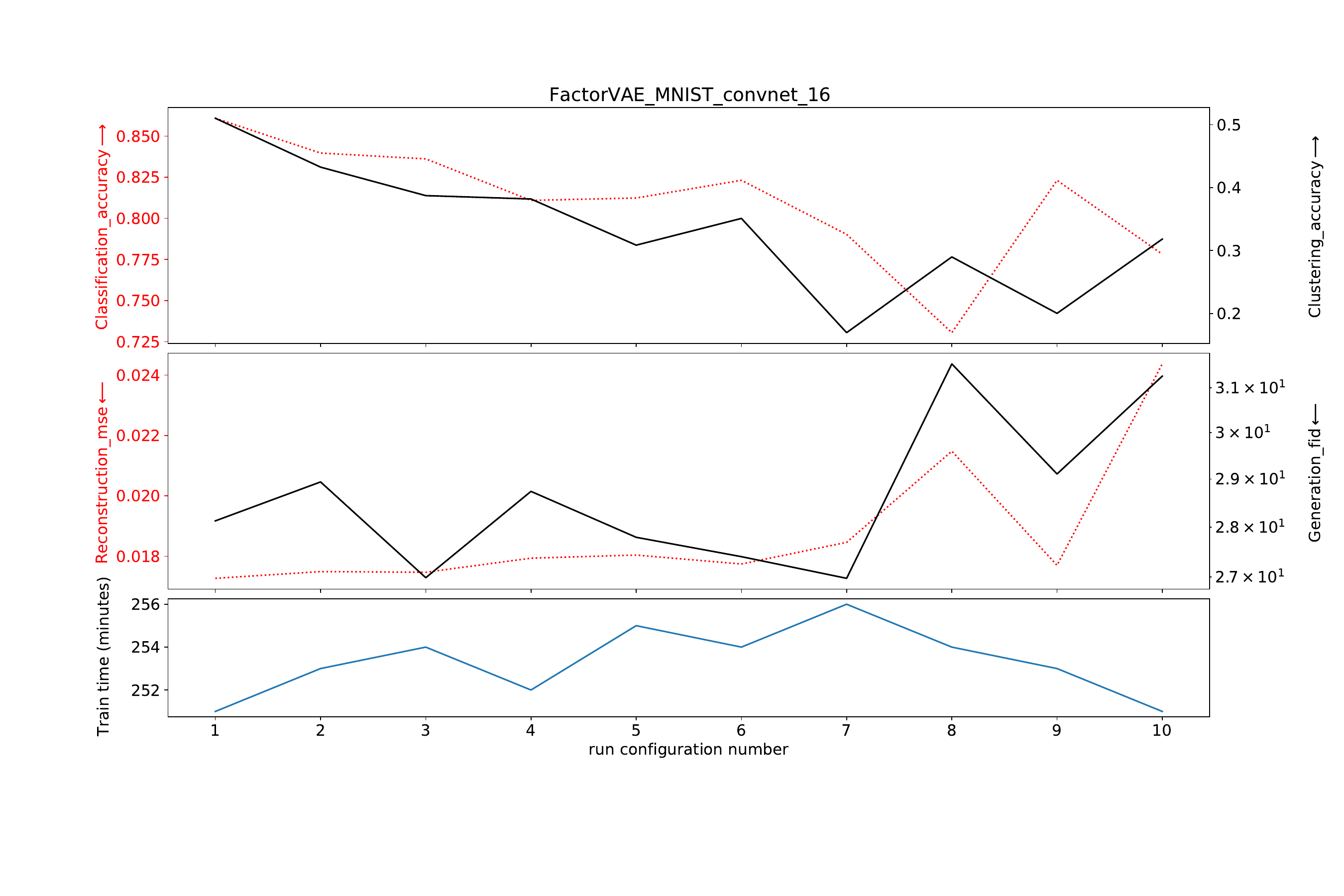}
    \caption{Results on FactorVAE}
    \label{fig:indiv_factorvae}
    \end{figure}

    \clearpage
    
    \paragraph{InfoVAE}
    
    \citet{zhao2018vae} note that the traditional VAE ELBO objective can lead to both inaccurate amortized inference and VAE models that tend to ignore most of the latent variables, therefore not fully taking advantage of the modelling capacities of the VAE scheme and learning less meaningful latent representations. In order to counteract these two issues, they propose to rewrite and re-weight the ELBO objective in order to counterbalance the imbalance between the distribution in the data space and the latent space, and add a mutual information term between $x$ and $z$ to encourage a stronger dependency between the two variables, preventing the model from ignoring the latent encoding. One can re-write the ELBO loss in order to explicit the KL divergence between the marginalised posterior and the prior
    \[
    \mathcal{L}_{\text{ELBO}}(x) := - \mathcal{D}_{KL}[q_\phi(z) || p_z(z) ] - \mathbb{E}_{z \sim p_z} \bigg[ \mathcal{D}_{KL}[q_\phi(x | z) || p_\theta(x | z) ] \bigg]\,.
    \]
    Introducing an additional mutual information term $I_q(x;z)$ and extending the objective function to use any given divergence $D$ between probability measures instead of the KL objective, the authors propose a new objective defined as 
    \[\mathcal{L}_{\text{InfoVAE}}(x) := - \lambda D[q_\phi(z) || p_z(z) ] - \mathbb{E}_{z \sim p_z} \bigg[ \mathcal{D}_{KL}[q_\phi(x | z) || p_\theta(x | z) ] \bigg] + \alpha I_q(x;z)
    \]
    where $\lambda$ and $\alpha$ are hyperparameters. In our experiments, $\alpha$ is set to 0 as recommended in the paper in the case where $p_\theta(x|z)$ is a simple distribution. 
    $D$ is chosen as the Maximum Mean Discrepancy (MMD) \citep{gretton2012kernel}, defined as
    \[\mbox{MMD}_k(p_\lambda(z),q_\phi(z)) = || \int_{\mathcal{Z}} k(z,.)d p_\lambda(z) - \int_{\mathcal{Z}} k(z,.)d q_\phi(z) ||_{\mathcal{H}_k}
    \]
    with $k: \mathcal{Z} \times \mathcal{Z} \rightarrow \mathbb{R}$ a positive-definite kernel and its associated RKHS $\mathcal{H}_k$. We choose to differentiate 2 cases in the benchmark: one with a Radial Basis Function (RBF) kernel, the other with the Inverse MultiQuadratic (IMQ) kernel as proposed in \citep{tolstikhin2018wasserstein} where the kernel is given by $k(x, y) = \sum \limits _{s \in \mathcal{S}} \frac{s \cdot C}{s \cdot C + \|x-y \|_2^2}$ with $s \in [0.1, 0.2, 0.5, 1, 2, 5, 10]$ and $C=2\cdot d\cdot \sigma^2$, $d$ being the dimension of the latent space and $\sigma$ a parameter part of the hyper-parameter search. 
    
    The authors underline that choosing $\lambda > 0$, $\alpha = 1 - \lambda$ and $D = \mathcal{D}_{KL}$, we recover the $\beta$-VAE model \citep{higgins_beta-vae_2017}, while choosing $\alpha=\lambda=1$ and setting $D$ as the Jensen Shannon divergence we recover the Adversarial AE model \citep{makhzani2016vae}.
            
\textbf{Results by configuration}
        
\begin{table}[ht]
    \centering
    \scriptsize
    \caption{InfoVAE configurations}
    \label{tab:infovae config}
    \begin{tabular}{lcccccccccc}
    \toprule
        Config & 1 & 2 & 3 & 4 & 5 & 6 & 7 & 8 & 9 & 10 \\ 
        \midrule
        kernel bandwidth - $\sigma$ & $1e^{-2}$ & $1e^{-1}$ & 0.5 & 1 & 1 & 1 & 1 & 1 & 2 & 5 \\
        $\lambda$ & 10 & 10 & 10 & $1e^{-2}$ & $1e^{-1}$ & 10 & 100 & 100 & 10 & 10 \\
    \bottomrule
    \end{tabular}
\end{table}

    \begin{figure}[ht]
    \centering
    \includegraphics[trim={0 4cm 0 0},clip,width=\linewidth]{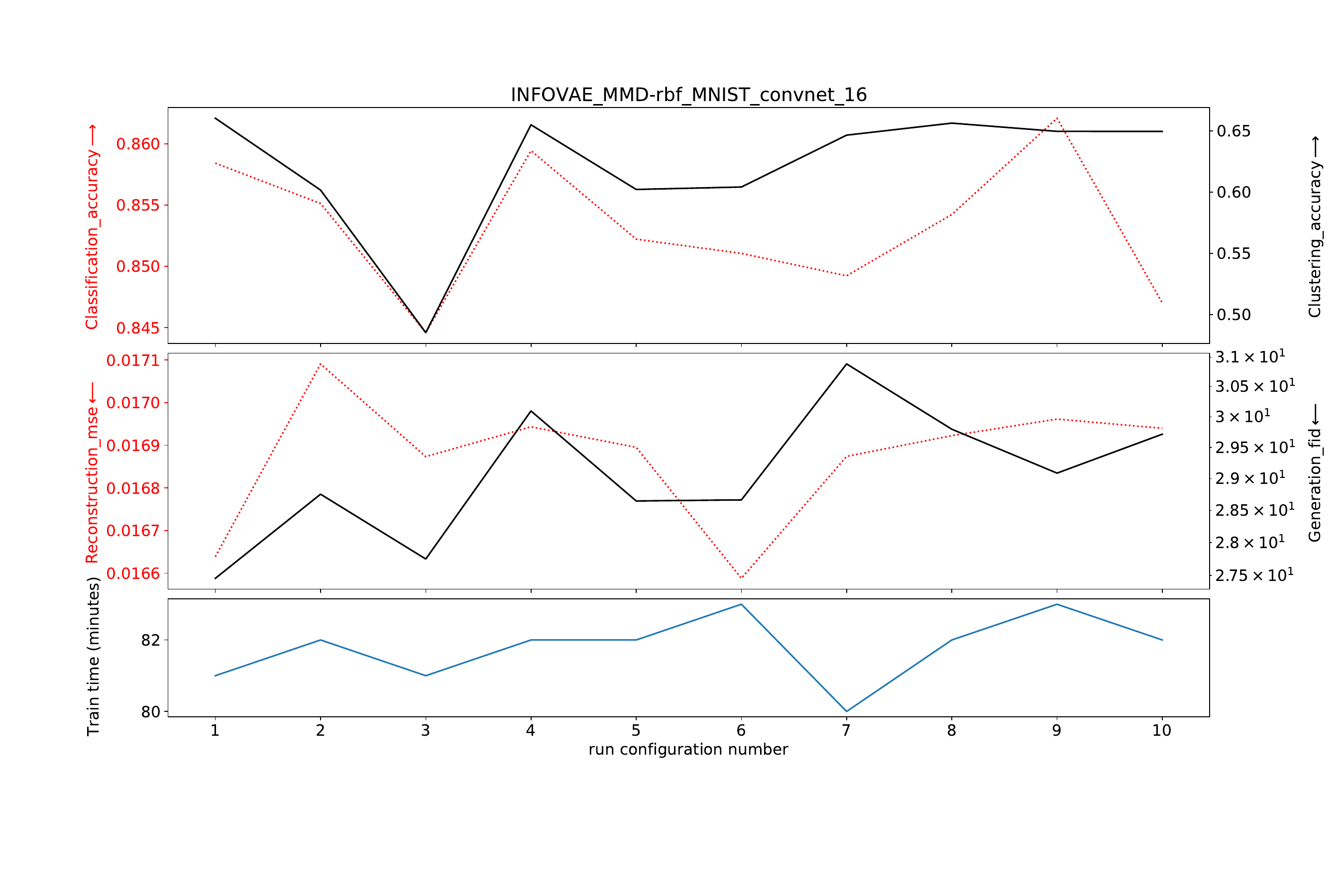}
    \caption{Results on InfoVAE-RBF}
    \label{fig:indiv_infovae_imq}
    \end{figure}
    
    \begin{figure}[ht]
    \centering
    \includegraphics[trim={0 4cm 0 0},clip,width=\linewidth]{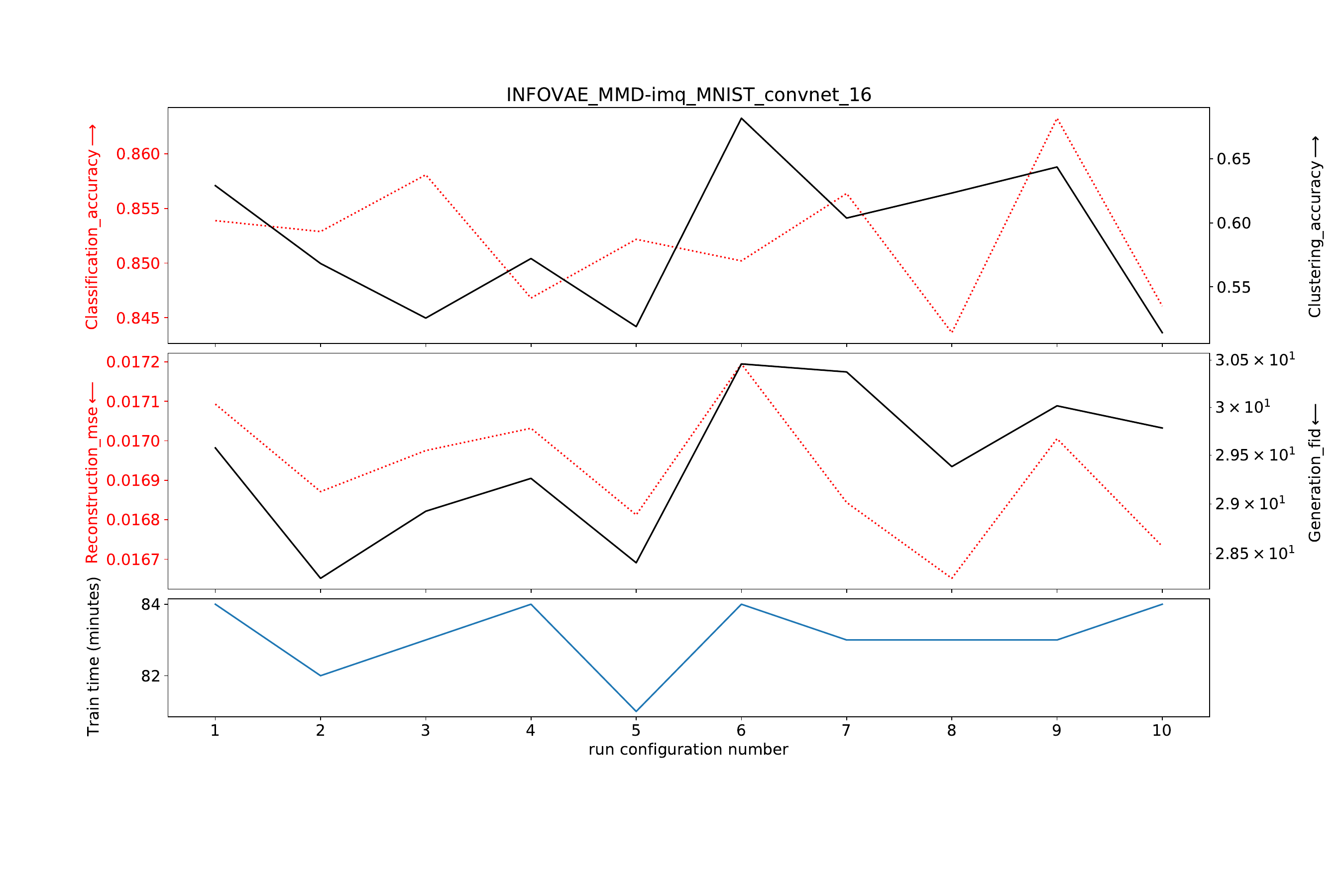}
    \caption{Results on InfoVAE-IMQ}
    \label{fig:indiv_infovae_rbf}
    \end{figure}

    \clearpage
    
    \paragraph{Adversarial AE (AAE)}
    
    \citet{makhzani2016vae} propose to use a GAN-like approach by replacing the regularisation induced by the KL divergence with a discriminator network $D$ trained to differentiate between samples from the prior and samples from the posterior distribution. The encoder network therefore acts as a generator network, leading to the following objective
    \begin{equation*}
        \mathcal{L}_{\text{AAE}}(x) = \mathbb{E}_{z \sim q_\phi(z | x)}[\log p_\theta(x | z)] + \alpha \mathcal{L}_{\mathrm{GAN}}\,, %\bigg)\,,
    \end{equation*}
    with $\mathcal{L}_{\mathrm{GAN}}$ the standard GAN loss defined by
    \begin{equation*}
        \mathcal{L}_{\mathrm{GAN}} = \mathbb{E}_{\tilde{z} \sim p_z(z)} \bigg[\log (1 - D(\tilde{z})) ) \bigg] +  \mathbb{E}_{x \sim p_\theta} \bigg[ \mathbb{E}_{z \sim q_\phi(z | x)}[\log D(z)] \bigg]\,.
    \end{equation*}
    For the Adversarial Autoencoder implementation, we use a  MLP neural network for the discriminator composed of a single hidden layer with 256 units and ReLU activation. 
    
    We observe a similar trade-off between reconstruction and generation quality as observed with $\beta$-VAE type models, as the $\alpha$ term acts like the $\beta$ term, balancing between regularisation and reconstruction.
    
   % The encoder network acts as the generator
    
    %Keeping the general encoder-decoder scheme, we add a latent discriminator network $\mathcal{D}$
    
    %At each iteration, we sample as usual $z \sim q_\phi (z | x)$, and also sample $z_{prior} \sim p_z$. The discriminator is trained to differentiate the prior distribution $p_z$ from the posterior $q_\phi (z | x)$.
    
    %The auto encoder is trained with the following loss
    
    %$$ \mathbb{E}_{x \sim p_\theta} \bigg[ \mathbb{E}_{z \sim q_\phi(z | x)}[\log p_\theta(x | z)] + \alpha \mathcal{D}(z) \bigg]$$
    
    %where $\alpha$ is a chosen hyperparameter.
    
    %The discriminator is trained to minimise the following loss
    
    %$$ \mathbb{E}_{z_{\mathrm{prior}} \sim p_z} \bigg[1 - \mathcal{D}(z_{\mathrm{prior}}) \bigg] +  \mathbb{E}_{x \sim p_\theta} \bigg[ \mathbb{E}_{z \sim q_\phi(z | x)}[\mathcal{D}(z)] \bigg] $$

    \textbf{Results by configuration}

    \begin{table}[ht]
    \centering
    \scriptsize
    \caption{AAE configurations}
    \label{tab:aae config}
    \begin{tabular}{lcccccccccc}
    \toprule
        Config & 1 & 2 & 3 & 4 & 5 & 6 & 7 & 8 & 9 & 10 \\ 
        \midrule
        $\alpha$ & $1e^{-3}$ & $1e^{-2}$ & $1e^{-1}$ & 0.25 & 0.5 & 0.75 & 0.9 & 0.95 & 0.99 & 0.999 \\
    \bottomrule
    \end{tabular}
\end{table}
    
    \begin{figure}[ht]
    \centering
    \includegraphics[trim={0 4cm 0 0},clip,width=\linewidth]{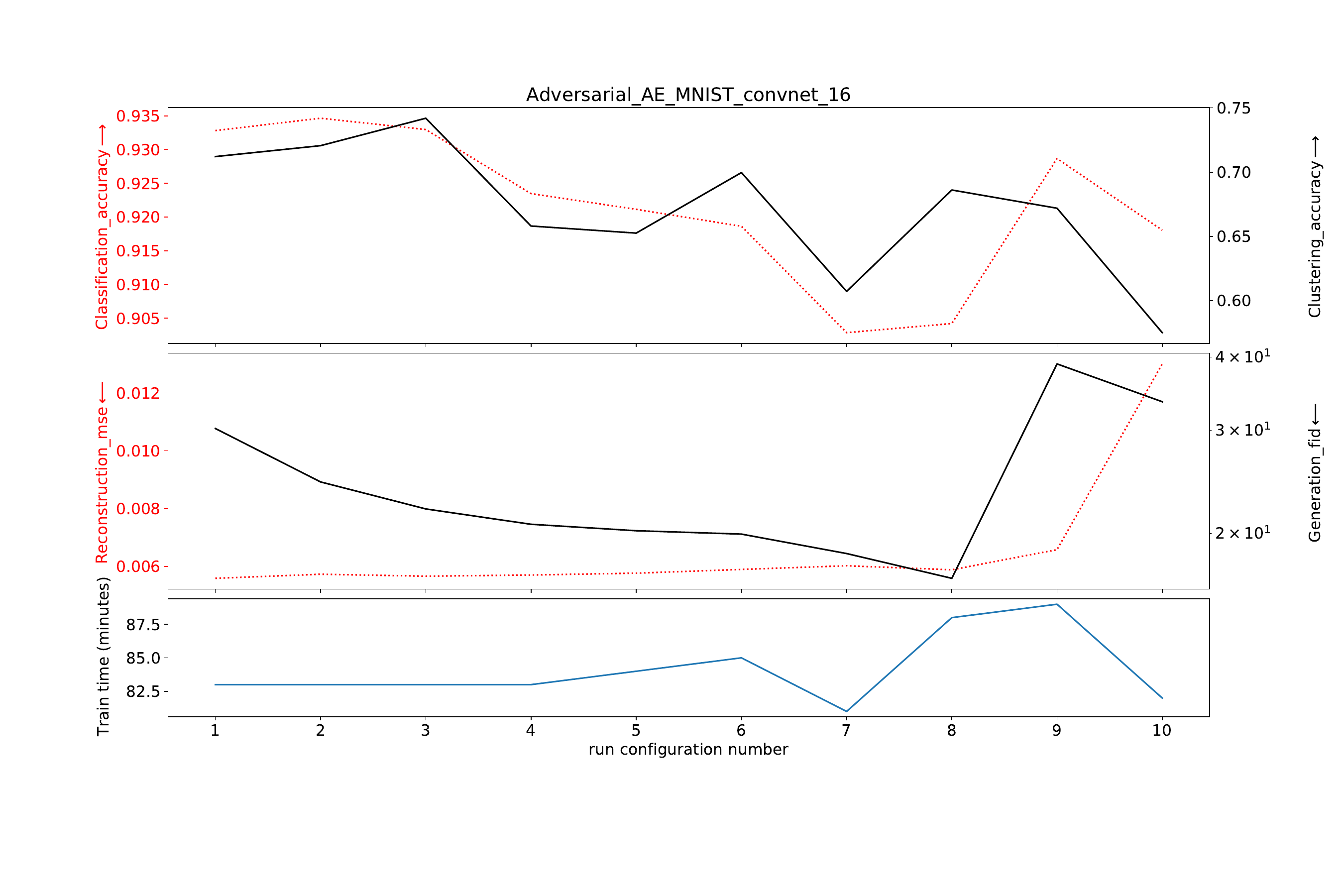}
    \caption{Results on Adversarial AE}
    \label{fig:indiv_aae}
    \end{figure}
    
    \clearpage
    
    \paragraph{EL-VAE (MSSSIM-VAE)}

    \citet{snell2017vae} propose an extension of the ELBO loss to a more general case where any deterministic reconstruction loss $\Delta(x,\hat{x})$ can be used by replacing the probabilistic decoder $p_\theta$ with a deterministic equivalent $f_\theta$ such that the reconstruction $\hat{x}$ of $x$ given $z \sim q_\phi(z|x)$ is defined as $\hat{x} = f_\theta(z)$. 
    The modified ELBO objective is thus defined as
    \[ \mathcal{L}_{\mathrm{EL-VAE}}(x) = \Delta(x,\hat{x}) - \beta \mathcal{D}_{KL}\big( q_\phi(z | x) || p(z) \big)\,,
    \]
    with $\beta \leq 1$. As suggested in the original paper we use a multi scale variant of the single scale SSIM \citep{wang2004image}: the Multi Scale Structural Similarity Metric (MS-SSIM) \citep{wang2003multiscale}. 
    
    \textbf{Results by configuration}

    \begin{table}[ht]
    \centering
    \scriptsize
    \caption{MSSSIM-VAE configurations}
    \label{tab:msssimvae config}
    \begin{tabular}{lcccccccccc}
    \toprule
        Config & 1 & 2 & 3 & 4 & 5 & 6 & 7 & 8 & 9 & 10 \\ 
        \midrule
        $\beta$ & $1e^{-2}$ & $1e^{-2}$ & $1e^{-2}$ & $1e^{-1}$ & $1e^{-1}$ & $1e^{-1}$ & 1 & 1 & 1 & 1 \\
        window size in MSSSIM & 3 & 5 & 11 & 5 &  3 & 11 & 11&  5 & 3 & 15\\ 
    \bottomrule
    \end{tabular}
\end{table}

    \begin{figure}[ht]
    \centering
    \includegraphics[trim={0 4cm 0 0},clip,width=\linewidth]{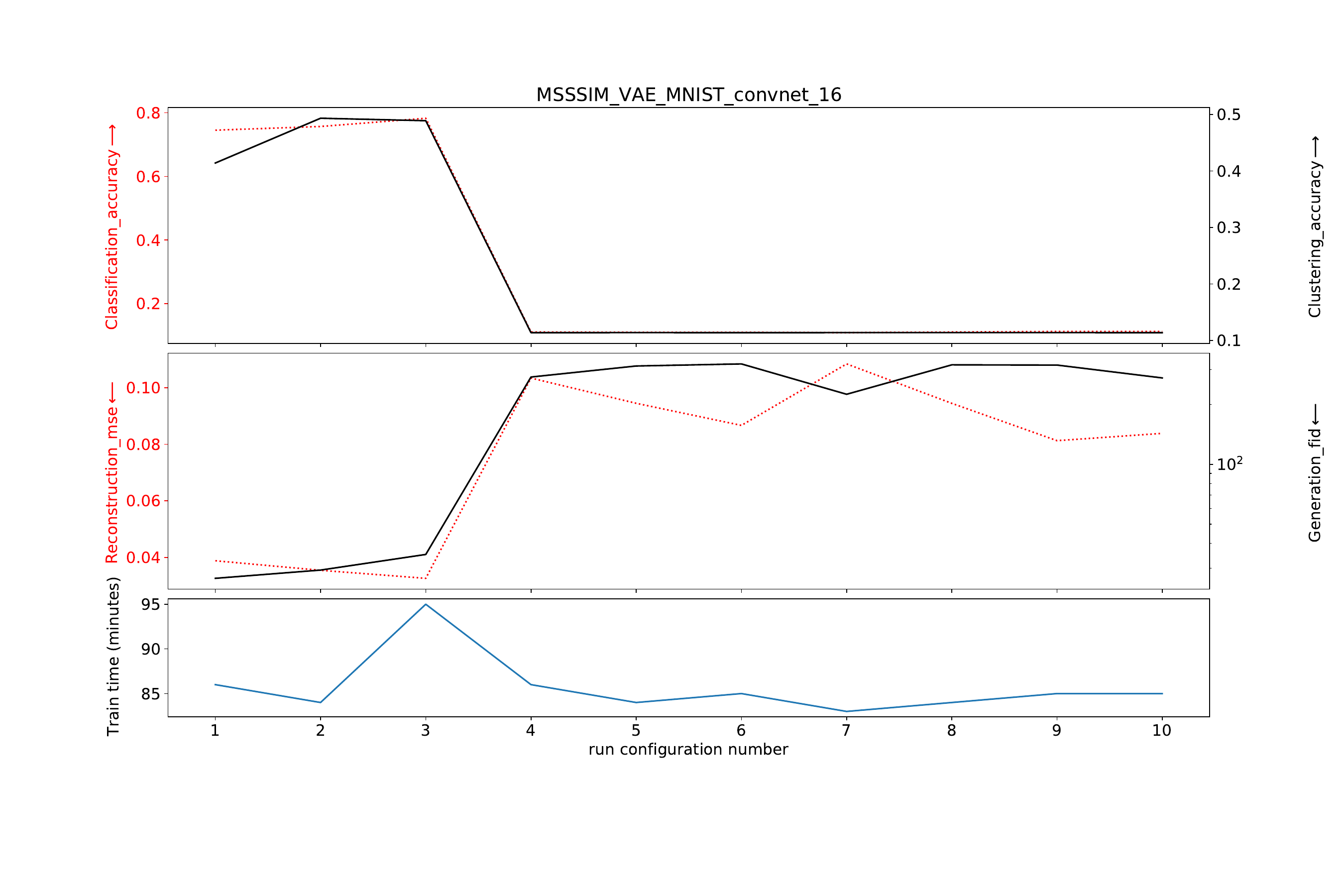}
    \caption{Results on MSSSIM-VAE}
    \label{fig:indiv_msssimvae}
    \end{figure}
    
    \clearpage
    
    \paragraph{VAE-GAN}

    \citet{larsen_autoencoding_2015} use a GAN like approach by training a discriminator to distinguish real data from reconstructed data. In addition, the discriminator learns to distinguish between real data and data generated by sampling from the prior distribution in the latent space.
    
    Noting that intermediate layers of a discriminative network trained to differentiate real from generated data can act as data-specific features, the authors propose to replace the reconstruction loss of the ELBO with a Gaussian log-likelihood between outputs of intermediate layers of a discriminative network $D$:
    \begin{equation*}
        \mathcal{L}_\text{VAE-GAN} = \underbrace{\mathbb{E}_{z \sim q_\theta(z|x)}\bigg[ \log \mathcal{N}(D_l(x)|D_l(\hat{x}),I) \bigg]}_\text{reconstruction} - \underbrace{\mathcal{D}_{KL}\big[ q_\phi(z|x) || p_z(z) \big]}_\text{regularisation} - \mathcal{L}_{GAN}\,,
    \end{equation*}
        where $D_l$ is the output of the $l^{\mathrm{th}}$ layer of the discriminator $D$, chosen to be representative of abstract intermediate features learned by the discriminator, and $\mathcal{L}_{\mathrm{GAN}}$ is the standard GAN objective defined as
        \begin{equation*}
            \mathcal{L}_{\mathrm{GAN}} = \log \bigg( \frac{D(x)}{1-D(x_{\mathrm{gen}})} \bigg)\,,
        \end{equation*}
        where $x_{\mathrm{gen}}$ is generated using $z \sim p_z(z)$. As encouraged by the authors, we add a hyper-parameter $\alpha$ to the reconstruction loss for the decoder only, such that a higher value of $\alpha$ will encourage better reconstruction abilities with respect to the features extracted at the $l^{\mathrm{th}}$ layer of the discriminator network, whereas a smaller value will encourage fooling the discriminator, therefore favouring regularisation toward the prior distribution. For the VAEGAN implementation, we use a discriminator whose architecture is similar to the model's encoder given in Table.~\ref{tab:convNet archi}. For MNIST and CIFAR we remove the BatchNorm layer and change the activation of layer 2  to Tanh instead of ReLU. For CELEBA, the BatchNorm layer is kept and the activation of layer 2 is also changed to Tanh. For all datasets, the output size of the last linear layer is set to 1 instead of $d$ and followed by a Sigmoid activation.
        
        % In particular, we see that even though the VAEGAN was performing the best by far with regards to the metrics, when looking at the generated images we see some artefacts arising. This is particularly visible on CELEBA and these aspects are well known drawbacks of the GAN-based approaches.
        
        %Discriminateur entrainé ou pas, ++
    
    %{\color{red} If observed we should state something on the VAEGAN sensitiveness to the hyper-parameters}
    
    \textbf{Results by configuration}
    
    \begin{table}[ht]
    \centering
    \scriptsize
    \caption{VAEGAN configurations}
    \label{tab:vaegan config}
    \begin{tabular}{lcccccccccc}
    \toprule
        Config & 1 & 2 & 3 & 4 & 5 & 6 & 7 & 8 & 9 & 10 \\ 
        \midrule
        $\alpha$ & 0.3 & 0.5 & 0.7 & 0.8 & 0.8 & 0.8 & 0.9 & 0.9 & 0.99 & 0.999 \\
        reconstruction layer (l) & 3 & 3 & 3 & 3 &  2 & 4 & 3& 3 & 3 & 3\\ 
    \bottomrule
    \end{tabular}
\end{table}

    \begin{figure}[ht]
    \centering
    \includegraphics[trim={0 4cm 0 0},clip,width=\linewidth]{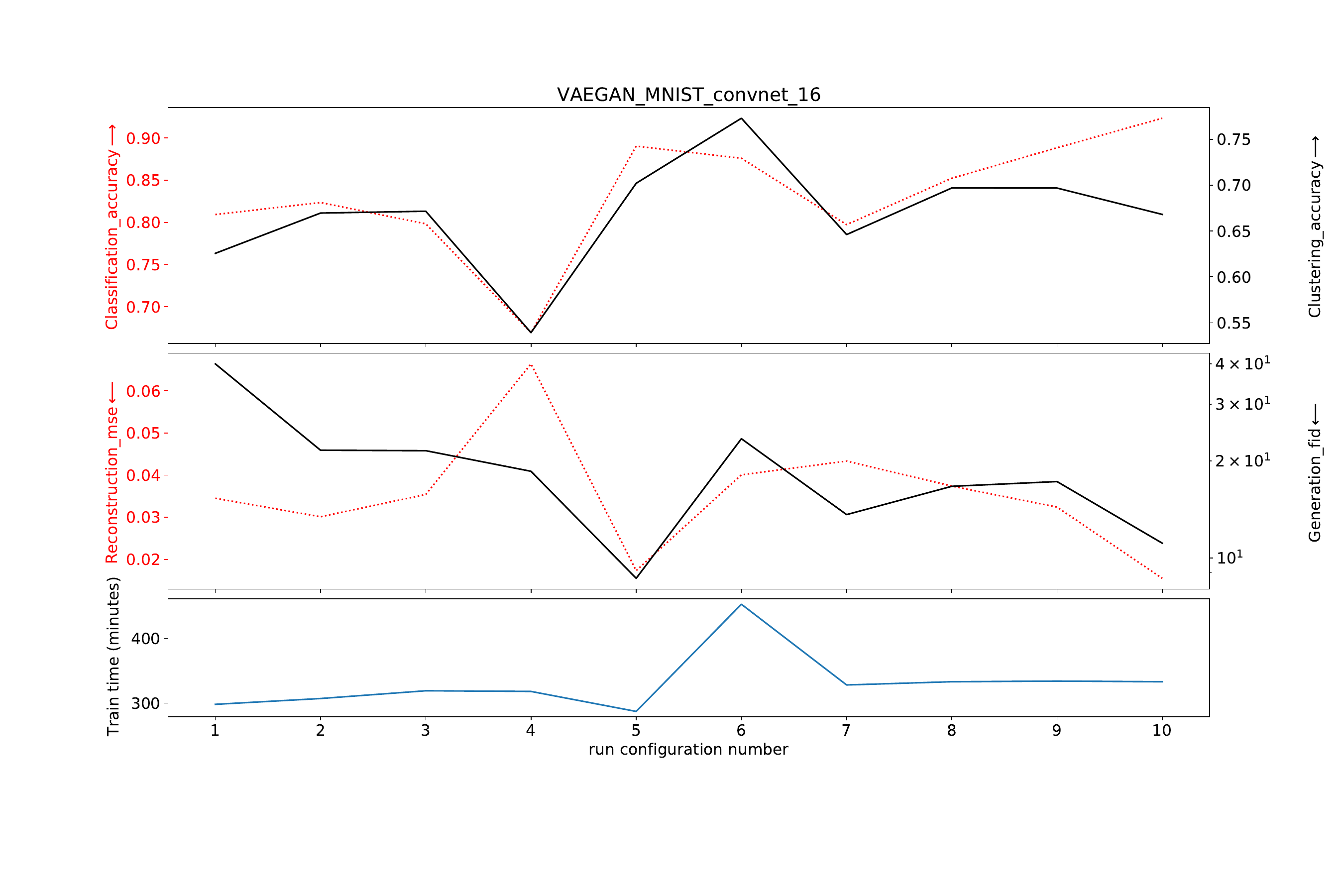}
    \caption{Results on VAEGAN}
    \label{fig:indiv_vegan}
    \end{figure}

    % \textbf{Improve disentanglement}

    % One key interest of VAEs is to learn a {more representative and interpretable} latent representation of the input data. In that regard, several publications have been proposed encouraging a better disentanglement of the features in the latent space.

    \clearpage
    
    \paragraph{Wasserstein Autoencoder (WAE)}
    
    \citet{tolstikhin2018wasserstein} generalise the VAE objective by replacing both terms in the ELBO: similarly to \cite{snell2017vae} (EL-VAE), the reconstruction loss is replaced by any measurable cost function $\Delta$, and the standard KL divergence is substituted with any arbitrary divergence $\mathcal{D}$ between two distributions, leading to the following objective function
    \begin{equation*}
        E_{q_\phi(z | x)}[\Delta(x,\hat{x})]+ \lambda \mathcal{D}_z(p_z(z),q_\phi(z))\,,
    \end{equation*}
    with $\lambda$ a hyper-parameter. The authors propose two different penalties for $\mathcal{D}_z$:
    \begin{enumerate}
        \item \textbf{GAN-based: WAE-GAN} %{\color{red} to remove or specify since this is equivalent to AAE}
        
        An adversarial discriminatory network $D(z,z')$ is trained jointly to separate the "true" points sampled from the prior $p_z(z)$ from the "fake" ones sampled from $q_\phi(z|x)$, similarly to \cite{makhzani2016vae} (Adversarial AE). %We have $\mathbb{E}_{x \sim p_\theta}\mathbb{E}_{p_\lambda}[D(z,Q(x))]:= \mathcal{D}_Z(p_\lambda(z),q_\phi(z))$ 
        
        \item \textbf{MMD-based} 
        
        The Maximum Mean Discrepancy is used as a distance between the prior and the posterior distribution. This is the case considered in the benchmark. We choose to differentiate 2 cases in the benchmark: one with a Radial Basis Function (RBF) kernel, the other with the Inverse MultiQuadratic (IMQ) kernel as proposed in \citep{tolstikhin2018wasserstein} where the kernel is given by $k(x, y) = \sum \limits _{s \in \mathcal{S}} \frac{s \cdot C}{s \cdot C + \|x-y \|_2^2}$ with $s \in [0.1, 0.2, 0.5, 1, 2, 5, 10]$ and $C=2\cdot d\cdot \sigma^2$, $d$ being the dimension of the latent space and $\sigma$ a parameter part of the hyper-parameter search. As proposed by the authors, for this model we choose to use a deterministic encoder meaning that $q_{\phi}(z|x) = \delta_{\mu_{\phi}(x)}$. 
        %{\color{red} Difficile de le formuler autrement que comme dans l'article}
        
        %Given a positive-definite kernel $k: \mathcal{Z} \times \mathcal{Z} \rightarrow \mathbb{R}^+$ and its associated RKHS $\mathcal{H}_k$, the maximum mean discrepancy (MMD) is defined as:
        
        %$$\mbox{MMD}_k(p_\lambda(z),q_\phi(z)) = || \int_{\mathcal{Z}} k(z,.)d p_\lambda(z) - \int_{\mathcal{Z}} k(z,.)d q_\phi(z) ||_{\mathcal{H}_k}$$

    \end{enumerate}
    
    %cite{ovinnikov_poincare_2020} propose to consider the representation learning problem as an Optimal Transport problem. We assign a distribution $p_\theta$ to our input space
    %This results in the following loss
    
    %Observing that the KL divergence is 
    %The standard KL divergence is substituted with any arbitrary divergence $\mathcal{D}$ between two distributions, which amounts to a penalized form of the Wasserstein distance between two distributions derived from optimal transport.
    %Additionally, similarly to \cite{snell2017vae} (EL-VAE), the reconstruction loss is replaced by any measurable cost function $\Delta$, leading to the following cost function  

    %$$\mathbb{E}_{x \sim p_\theta}\bigg[E_{q_\phi(z | x)}[c(x,p_\theta(x|z))]\bigg] + \lambda \mathcal{D}_z(p_\lambda(z),q_\phi(z))$$ with $q_\phi(z):=\mathbb{E}_{x \sim p_\theta}[q_\phi(z|x)]$ the marginal distribution of the latent space generated by the encoder, $c$ is any measurable cost function, $\mathcal{D}_z$ is an arbitrary divergence between 2 distributions (The KL divergence for example) and $\lambda>0$ is a hyper-parameter.
    
    %The first term $\mathbb{E}_{x \sim p_\theta}\bigg[E_{q_\phi(z | x)}[c(x,p_\theta(x|z))]\bigg]$ forces the decoded point $p_\theta(x|z)$ to be close to $x$, while the second term $\lambda \mathcal{D}_Z(p_\lambda(z),q_\phi(z))$ acts as a regularisation term.

    \textbf{Results by configuration}
    
    \begin{table}[ht]
    \centering
    \scriptsize
    \caption{WAE configurations}
    \label{tab:wae config}
    \begin{tabular}{lcccccccccc}
    \toprule
        Config & 1 & 2 & 3 & 4 & 5 & 6 & 7 & 8 & 9 & 10 \\ 
        \midrule
        kernel bandwidth - $\sigma$ & $1e^{-2}$ & $1e^{-1}$ & 0.5 & 1 & 1 & 1 & 1 & 1 & 2 & 5 \\
        $\lambda$ & 1 & 1 & 1 & $1e^{-2}$ & $1e^{-1}$ & 1 & 10 & 100 & 1 & 1 \\
    \bottomrule
    \end{tabular}
\end{table}
    
    \begin{figure}[ht]
    \centering
    \includegraphics[trim={0 4cm 0 0},clip,width=\linewidth]{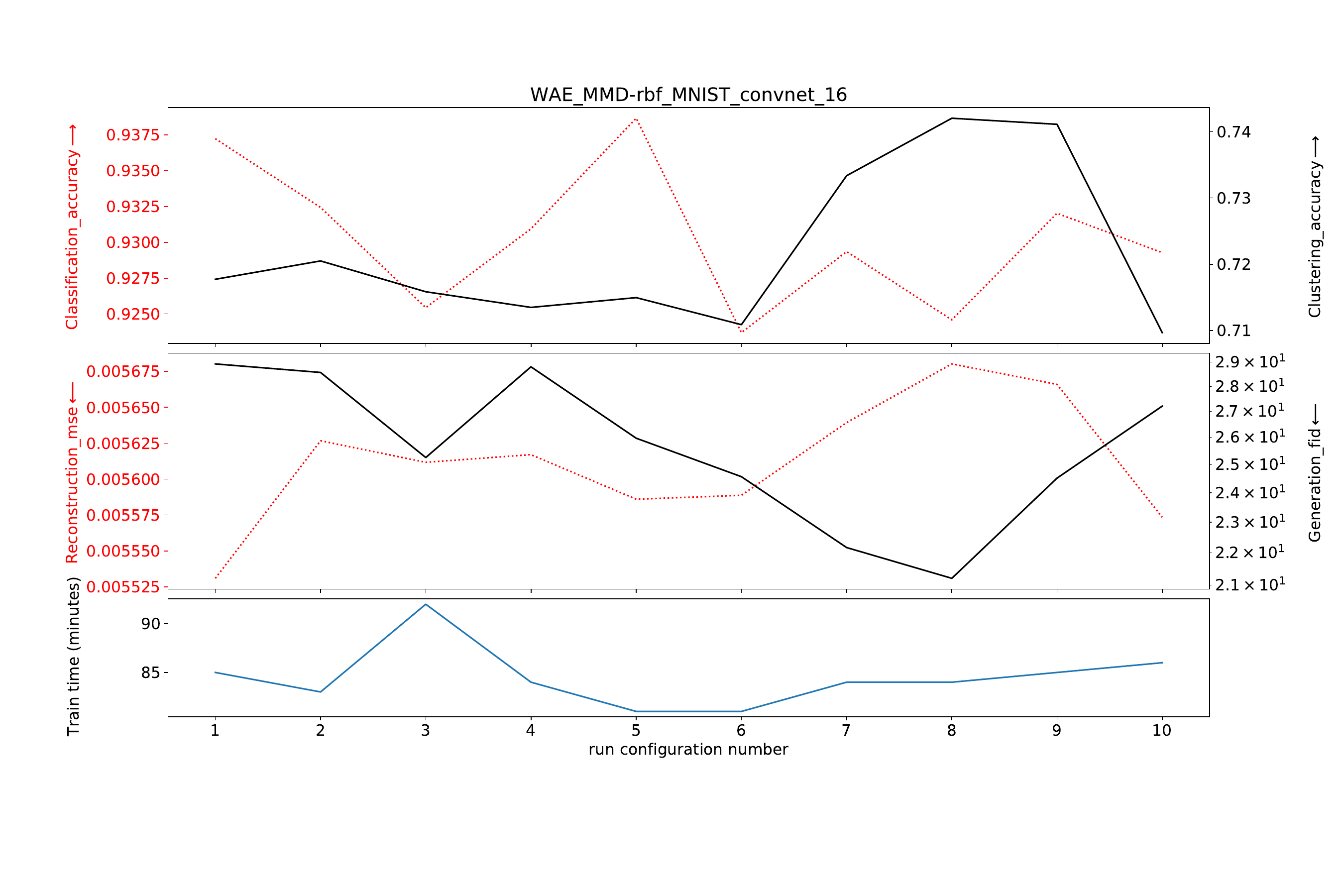}
    \caption{Results on WAE-RBF}
    \label{fig:indiv_wae_rbf}
    \end{figure}
    
    \begin{figure}[ht]
    \centering
    \includegraphics[trim={0 4cm 0 0},clip,width=\linewidth]{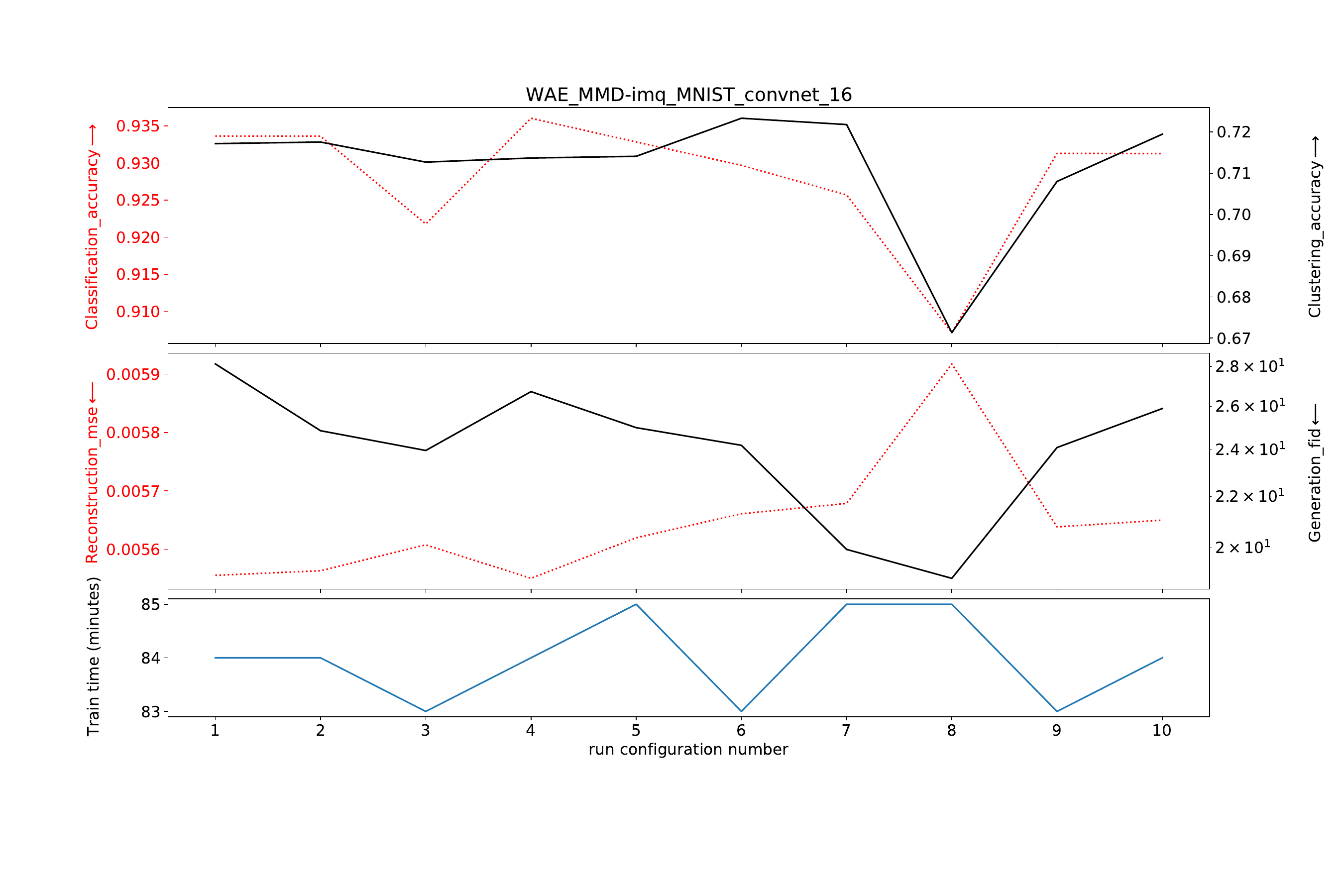}
    \caption{Results on WAE-IMQ}
    \label{fig:indiv_wae_imq}
    \end{figure}
    
    \clearpage
    \clearpage
    
    \paragraph{Vector Quantized VAE (VQ-VAE)}
    
    \citet{oord2018vae} propose to use a discrete space. Therefore, the latent embedding space is defined as a $\mathbb{R}^{K \times D}$ vector space of $K$ different $D$ dimensional embedding vectors $\mathcal{E} = \{e_1, \dots, e_K \}$ which are learned and updated at each iteration. 
    
    %(In general, D will vary from 1 to 10)..
    
    Given an embedding size $d$ and an input $x$, the output of the encoder $z_e(x)$ is of size $\mathbb{R}^{d \times D}$. Each of its $d$ elements is then assigned to the closest embedding vector resulting in an embedded encoding $z_q(x) \in \mathcal{E}^d$ such that $\big(z_q(x)\big)_j = e_l$ where  $l = \text{argmin}_{1 \leq l \leq d} ||(z_e(x))_j - e_l||_2 $ for $j \in [1, d]$. Since the argmin operation is not differentiable, learning of the embeddings and regularisation of the latent space is done by introducing the stopgradient operator $sg$ in the training objective:
    \begin{equation*}
        \mathcal{L}_\text{VQ-VAE}(x) := \log p(x|z_q(x)) + \alpha ||sg[z_e(x)] - e||_2^2 + \beta||z_e(x) - sg[e]||_2^2\,.
    \end{equation*}
    For the VQVAE implementation we use the Exponential Moving Average update as proposed in \citep{oord2018vae} to replace the term $||sg[z_e(x)] - e||_2^2$ in the loss. Thus, we consider only two hyper-parameters in the search: the size of the dictionary of embeddings $K$ and the regularisation factor $\beta$. 
    
    \textbf{Results by configuration}

    \begin{table}[ht]
    \centering
    \scriptsize
    \caption{VQVAE configurations}
    \label{tab:vqvae config}
    \begin{tabular}{lcccccccccc}
    \toprule
        Config & 1 & 2 & 3 & 4 & 5 & 6 & 7 & 8 & 9 & 10 \\ 
        \midrule
        $K$ & 128 & 256 & 512 & 512 & 512 & 512 & 512 & 512 & 1024 & 2948 \\
        $\beta$ & 0.25 & 0.25 & 0.9 & 0.1 & 0.5 & 0.25 & 0.75 & 0.25 & 0.25 & 0.25 \\
    \bottomrule
    \end{tabular}
\end{table}

    \begin{figure}[ht]
    \centering
    \includegraphics[trim={0 4cm 0 0},clip,width=\linewidth]{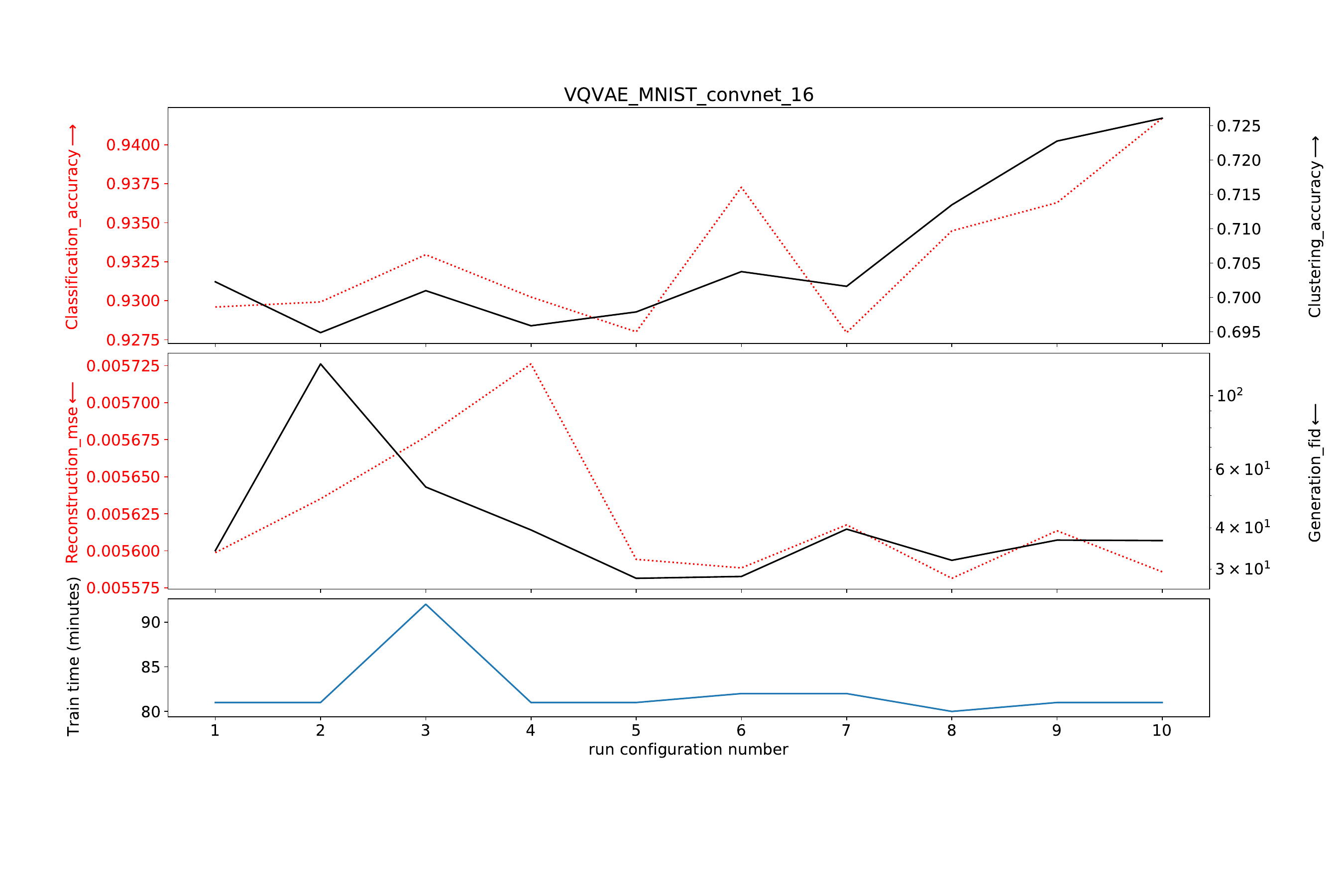}
    \caption{Results on VQVAE}
    \label{fig:indiv_vqvae}
    \end{figure}
    
    %$z \in \{1, \dots, K \}^l$ where $l$ is the embedding size, such that $z_j$ is equal to $\textit{argmin}_{1 \leq i \leq K}||z_j - e_i||$ {\color{red} can't say this}.
    
    \newpage
    
    \paragraph{RAE L2 and RAE GP}
    
    \citet{ghosh_variational_2020} propose to replace the stochastic VAE with a deterministic autoencoder by adapting the ELBO objective to a deterministic case. Under standard VAE assumption with Gaussian decoder, both the reconstruction and the regularisation terms in the ELBO loss can be written in closed form as
    \begin{equation*}
        \left\{\begin{aligned}
          \mathcal{L}_\text{reconstruction}(x) &= ||x - \hat{x}||_2^2\,, \\
          \mathcal{L}_\text{regularisation}(x) &= \frac{1}{2}\Big [ ||z||^2_2 - d + \sum_{i=1}^d (\sigma_\phi(x)_i - \log \sigma_\phi(x)_i) \Big ] \,.
        \end{aligned}\right.
    \end{equation*} 
    Arguing that the regularisation of the VAE model is done through a noise injection mechanism by sampling from the approximate posterior distribution $z \sim \mathcal{N}(\mu_\phi, diag( \sigma_\phi ))$, the authors propose to replace this stochastic regularisation with an explicit regularisation term, leading to the following deterministic objective:
    \begin{equation}\label{eq:rae_loss}
        \mathcal{L}_\text{RAE} = ||x - \hat{x}||_2^2 + \frac{\beta}{2}||z||^2_2 + \lambda \mathcal{L}_\text{REG}\,,
    \end{equation} where $\mathcal{L}_\text{REG}$ is an explicit regularisation. They propose to use either 
    \begin{itemize}
    \item a L2 loss on the weights of the decoder (RAE-L2), which amounts to applying weight decay on the parameters of the decoder.
    \item a gradient penalty on the output of the decoder (RAE-GP), which amounts to applying a L2 norm on the gradient of the output of the decoder.
    \end{itemize}

    %Hence, the parameter we decide to change in the configurations are the hyper-parameters introduced in the loss function in Eq.~\eqref{eq:rae_loss}. See table.~\ref{tab:rae config} for the values considered and Fig.~\ref{fig:indiv_rae_gp} and Fig.~\ref{fig:indiv_rae_l2} to assess their influence on the 4 considered tasks for the RAE-GP and RAE-L2 respectively. 
    
    \textbf{Results by configuration}
    
    \begin{table}[ht]
    \centering
    \scriptsize
    \caption{RAE configurations}
    \label{tab:rae config}
    \begin{tabular}{lcccccccccc}
    \toprule
        Config & 1 & 2 & 3 & 4 & 5 & 6 & 7 & 8 & 9 & 10 \\ 
        \midrule
        $\beta$ & $1e^{-6}$ & $1e^{-4}$ & $1e^{-3}$ & $1e^{-3}$ & $1e^{-3}$ & $1e^{-3}$ & $1e^{-3}$ & $1e^{-2}$ & $1e^{-1}$ & 1 \\
 $\lambda$ & $1e^{-3}$ & $1e^{-3}$ & $1e^{-6}$ & $1e^{-4}$ & $1e^{-2}$ & $1e^{-1}$ & 1 & $1e^{-3}$ & $1e^{-3}$ & $1e^{-3}$\\
    \bottomrule
    \end{tabular}
\end{table}

    \begin{figure}[ht]
    \centering
    \includegraphics[trim={0 4cm 0 0},clip,width=\linewidth]{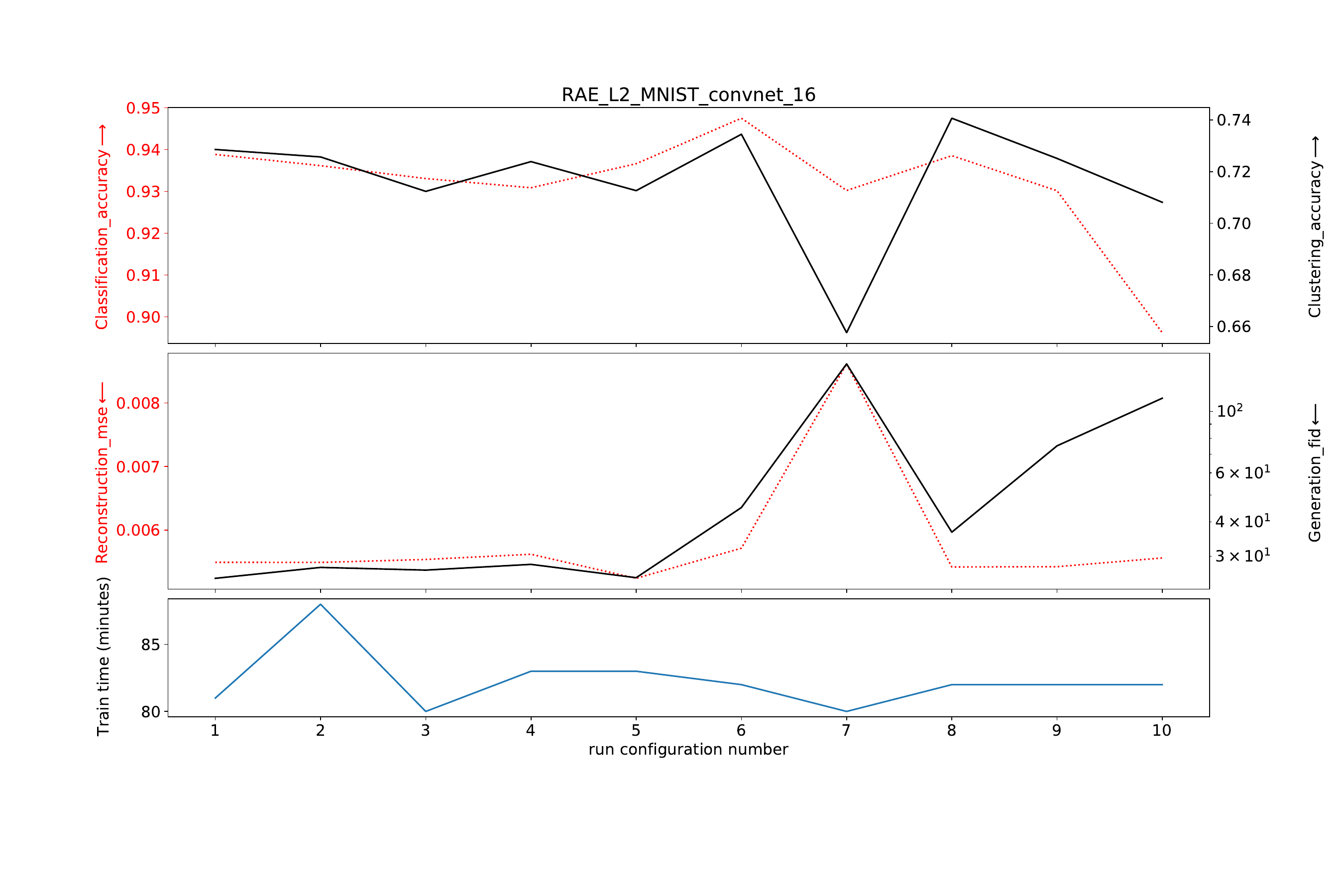}
    \caption{Results on RAE-L2}
    \label{fig:indiv_rae_l2}
    \end{figure}
        
    \begin{figure}[ht]
    \centering
    \includegraphics[trim={0 4cm 0 0},clip,width=\linewidth]{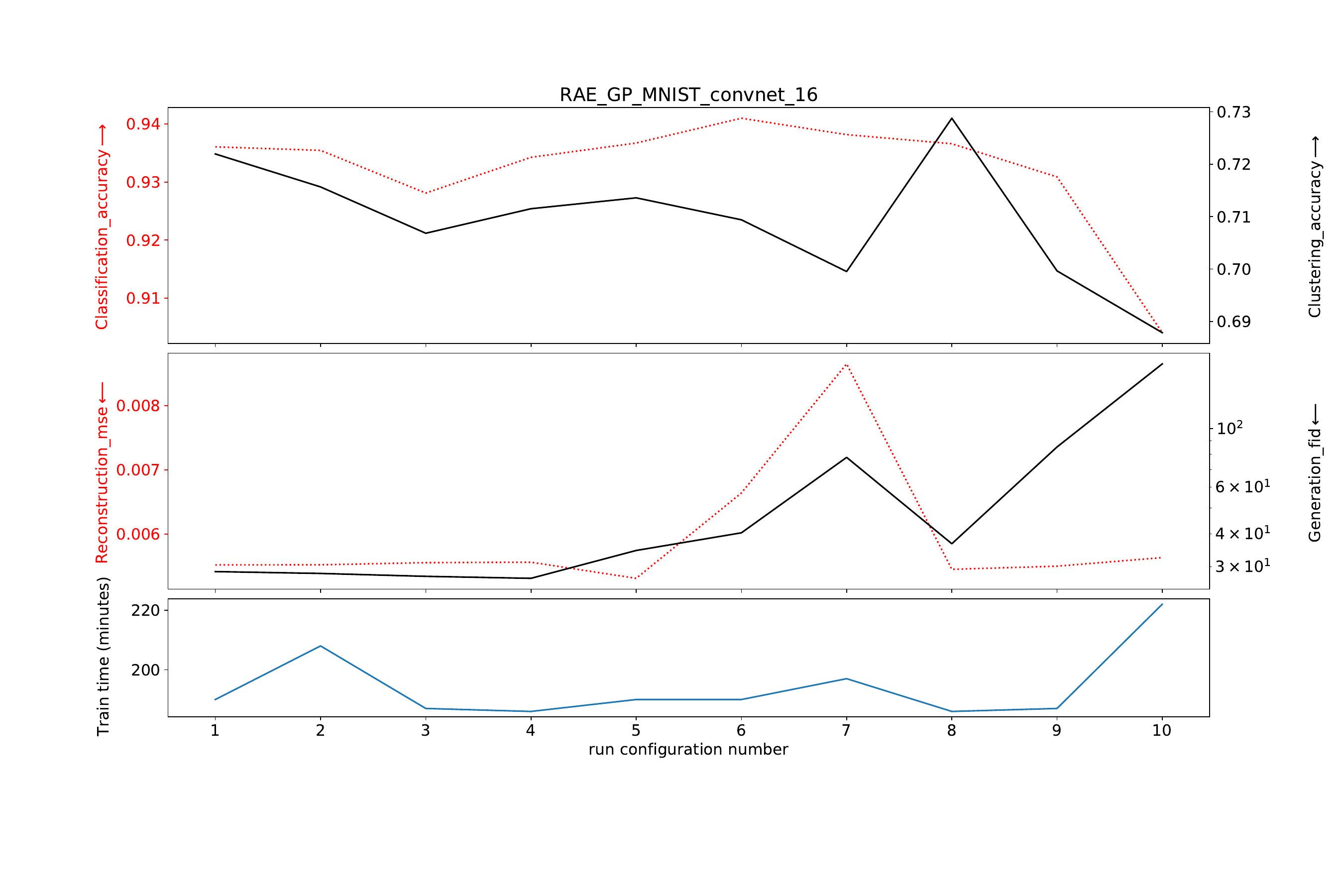}
    \caption{Results on RAE-GP}
    \label{fig:indiv_rae_gp}
    \end{figure}

% \begin{table}[ht]
% \CatchFileDef{\mytable}{plots/configs/configs_RAE_L2.tex}{}
%     \centering
%     \begin{tabular}{|||c||c|} 
%         \mytable
%     \end{tabular}
%     %\caption{Caption}
%     \label{tab:my_label}
% \end{table}

    \clearpage

\end{document}